\documentclass[11pt, oneside, b5paper, openleft]{book} 
\usepackage{KUstyle}
\usepackage[Lenny]{fncychap}
\ChTitleVar{\LARGE}
\usepackage[top = 1.2in, bottom = 1.2in]{geometry} %
\usepackage{layouts} %

\usepackage{xpatch}
\usepackage{tocloft}%
\usepackage{CJKutf8}

\makeatletter
\xpatchcmd{\@part}{%
  \addcontentsline{toc}{part}{\thepart\hspace{1em}#1}%
}{%
  \addtocontents{toc}{\begingroup\Large \mdseries\protect\centering\partname\ \thepart\par\large\protect\centering#1\par\endgroup}
}{}{}

\makeatother

\usepackage[round]{natbib}
\usepackage{bibentry}
\nobibliography*

 \graphicspath{ {images/} }

\usepackage{amsthm} %

\newtheorem*{prop*}{Proposition}

\newtheorem*{lemma*}{Lemma}

\usepackage{fourier-orns}
\usepackage{fancyhdr}
\usepackage{tabularx} %

\usepackage[T1]{fontenc}    %
\usepackage{booktabs}       %
\usepackage{amsfonts}       %

\usepackage{xcolor}         %
\usepackage{amsmath}
\usepackage{amssymb}
\usepackage{backref}
\usepackage{hyperref}
\usepackage[ruled,vlined,linesnumbered]{algorithm2e}
\usepackage[noabbrev,capitalize,nameinlink]{cleveref}
\crefformat{section}{#2Section~#1#3}
\crefformat{subsection}{#2Subsection~#1#3}
\crefformat{figure}{#2Figure~#1#3}
\crefformat{table}{#2Table~#1#3}

\usepackage[noend]{algpseudocode}
\usepackage{adjustbox}
\usepackage[draft]{minted}
\usepackage{soul, colortbl}
\usepackage{array, multirow, graphicx}
\usepackage{tikz}
\usepackage{tikz-dependency}
\usepackage{lscape, float} 
\usepackage{changes}
\usepackage[shortlabels]{enumitem}
\usepackage{fontawesome}
\usepackage{amssymb}%
\usepackage{pifont}%
\usepackage{ragged2e}
\usepackage{bbm}
\usepackage{todonotes}
\usepackage{emptypage}
\usepackage{xurl}
\usepackage{microtype}

\usepackage{titlesec}
\usepackage{fancyhdr}
\usepackage{CJKutf8}

\fancypagestyle{plain}{         
\fancyhf{}
\fancyfoot[C]{\thepage}
}%

\newcommand\mymainpagestyle{%
\fancyhf{}
\fancyhead[LE]{\nouppercase{\footnotesize{{\rightmark\hfill\leftmark}}}}
\fancyhead[RO]{\nouppercase{\footnotesize{{\leftmark\hfill\rightmark}}}}
\fancyfoot[C]{\thepage}
}

\usepackage[T1]{fontenc}
\usepackage{charter}

\usepackage{setspace}
\begin{document}

\begin{titlepage}
    \begin{center}
 {\Large \textbf{Computational Job Market Analysis}\\}
 {\normalsize \textbf{with Natural Language Processing}\\}
 \vspace{10em}
 Mike Zhang\\
 \begin{CJK*}{UTF8}{gbsn}
 张军军\\
 \end{CJK*}
    \vspace{20em}
This thesis has been submitted to the Ph.D. School of the \\IT University of Copenhagen on 14 January 2024
 \end{center}
\end{titlepage}

\frontmatter %
\pagestyle{plain} %
\thispagestyle{empty} 
\begin{center}
The research for this doctoral thesis has received funding from the Independent Research Fund Denmark (DFF) grant 9131-00019B and in parts by European Research Council (ERC) Consolidator grant DIALECT 101043235. 
\end{center}

\cleardoublepage

\thispagestyle{empty} 
\begin{center}
\small
\begin{tabular}{lll}
\textbf{Committee} & & \\\\
\textbf{Advisor}     &  prof. dr. B. Plank & Ludwig-Maximilians-Universität München\\
                     &                     & IT Universitetet i København\\\\
\textbf{Co-Advisor}  &  dr. R.M. van der Goot & IT Universitetet i København \\\\
\textbf{Members}     & prof. dr. C. Biemann & Universität Hamburg \\
            & dr. T. Bogers (chair) & IT Universitetet i København \\
            & dr. D. Maynard & University of Sheffield \\
\end{tabular}
\end{center}

\cleardoublepage

\section*{Abstract}
\thispagestyle{empty} 
\label{sec:abstract}
\addcontentsline{toc}{section}{Abstract} 
Recent technological advances underscore the dynamic nature of the labor market. These transformative shifts yield significant consequences for employment prospects, resulting in the increase of job vacancy data across platforms and languages. The aggregation of such data holds the potential to gain valuable insights into labor market demands, the emergence of new skills, and the overall facilitation of job matching. These benefits extend to various parties, including job platforms, recruitment agencies, applicants, and other stakeholders within the ecosystem. However, despite the prevalence of such insights in the private sector, we lack transparent language technology systems and data for this domain.

The primary objective of this thesis is to investigate the use of Natural Language Processing (NLP) technology for the extraction of relevant information from job descriptions. We identify several general challenges within this domain. These encompass a scarcity of available training and evaluation data, a lack of standardized guidelines to annotate data, and a shortage of effective methods for extracting information from job ads.

Therefore, we embark on a comprehensive study of the entire process: First, framing the problem and getting annotated data for training NLP models. Here, our contributions encompass job description datasets, including a de-identification dataset, and a novel active learning algorithm designed for efficient model training. Second, we introduce several extraction methodologies to tackle the task of information extraction from job advertisement data: A skill extraction approach using weak supervision, a taxonomy-aware pre-training methodology adapting a multilingual language model to the job market domain, and a retrieval-augmented model leveraging multiple skill extraction datasets to enhance overall extraction performance. Lastly, given the extracted information, we delve into the grounding of this data within a designated taxonomy.

\newpage
\section*{Resumé}
\thispagestyle{empty} 
\label{sec:resume}
\addcontentsline{toc}{section}{Resumé}
De seneste teknologiske fremskridt understreger arbejdsmarkedets dynamiske natur. Disse transformative skift har betydelige konsekvenser for beskæftigelsesudsigterne, hvilket resulterer i en stigning i data om ledige stillinger på tværs af platforme og sprog. Aggregering af sådan data har potentiale til at give værdifuld indsigt i arbejdsmarkedets efterspørgsler, fremkomst af nye færdigheder, og general facilitering af jobmatching. Disse fordele kommer forskellige parter til gode, herunder jobplatforme, rekrutteringsbureauer, ansøgere og andre interessenter i økosystemet. På trods af udbredelsen af sådanne indsigter i den private sektor, mangler vi gennemsigtige sprogteknologiske systemer og data til dette domæne.

Det primære formål med denne afhandling er at undersøge brugen af sprogteknologi (\textit{en}: Natural Language Processing) til at udtrække relevant information fra jobbeskrivelser. Vi identificerer flere generelle udfordringer inden for dette domæne. Disse omfatter en mangel på tilgængelig trænings- og evalueringsdata, en mangel på standardiserede retningslinjer for annotering af data samt en mangel på effektive metoder til at udtrække information fra jobannoncer.

Derfor går vi i gang med en omfattende undersøgelse af hele processen: Indledningsvis formuleres problemet og data annoteres med henblik på træning af NLP-modeller. Vores bidrag omfatter datasæt med jobbeskrivelser, herunder et de-identifikationsdatasæt, og en ny aktiv læringsalgoritme designet til effektiv modeltræning. For det andet introducerer vi flere ekstraktionsmetoder til at tackle opgaven med at udtrække information fra jobannoncedata: En metode til udtrækning af færdigheder ved hjælp af svag supervision, en taksonomi-bevidst prætræningsmetode, der tilpasser en flersproget sprogmodel til jobmarkedsdomænet, samt en informationssøgning-baseret model, der udnytter flere datasæt til at udtrække færdigheder for at forbedre den samlede effektivitet. Til sidst, givet den ekstraherede information dykker vi ned i forankringen af disse data inden for en udpeget taksonomi.

\newpage
\section*{Acknowledgements}
\thispagestyle{empty} 
\label{sec:acks}
\addcontentsline{toc}{section}{Acknowledgements}
Placeholder for Acknowledgments.

\newpage
\section*{Declaration of Work}
\thispagestyle{empty}
\addcontentsline{toc}{section}{Declaration of Work}
I, Mike Zhang, declare that this thesis - submitted in partial fulfillment of the requirements for the conferral of PhD, from the IT University of Copenhagen - is solely my own work unless otherwise referenced or attributed. Neither the thesis nor its content have been submitted (or published) for qualifications at another academic institution.
\\\\
\noindent
-- Mike Zhang

\newpage
\tableofcontents
\newpage

\mainmatter 
\mymainpagestyle{} %

\chapter{Introduction}
\label{chap:intro}
\newcommand{\jobstack}{\textsc{JobStack}}
\newcommand{\xlmr}{\texttt{XLM-R}\textsubscript{large}}
\newcommand{\escolmr}{\texttt{ESCOXLM-R}}
\newcommand{\mbert}{\texttt{mBERT}\textsubscript{large}}
\newcommand{\cls}{\texttt{[CLS]}}
\newcommand{\sos}{\texttt{<s>}}
\newcommand{\tss}[1]{\textsuperscript{#1}}
\newcommand{\lr}[2]{{#1}{$e^{-\text{#2}}$}}
\sloppy

\newcommand{\rqone}{How do Transformer-based models compare against recurrent neural network models for de-identification of privacy-related entities in job postings?}
\newcommand{\rqtwo}{Which auxiliary tasks and datasets improve de-identification performance of privacy-related entities from job postings?}
\newcommand{\rqthree}{Can training dynamic-based active learning lead to more efficient model training than previous methods for text classification while maintaining performance?}
\newcommand{\rqfour}{What are skills and how challenging is it to manually identify them in job postings?}
\newcommand{\rqfive}{To what extent does continuous pre-training on unlabeled job posting data improve skill extraction in English?}
\newcommand{\rqsix}{How much does continuous pre-training on large-scale unlabeled Danish job posting data improve skill classification in Danish?}
\newcommand{\rqseven}{How effective is distant supervision with ESCO for skill categorization?}
\newcommand{\rqeight}{Can we apply existing skill labels from the ESCO taxonomy as a weak supervision signal for skill extraction?}
\newcommand{\rqnine}{Can we use a graph-guided pre-training method with ESCO to improve a multilingual language model's performance on skill extraction and classification?}
\newcommand{\rqten}{How does a retrieval-augmented language model as a dataset-unifying method affect robustness for skill extraction?}
\newcommand{\rqeleven}{How effective are entity linking methodologies for linking skills to knowledge bases (e.g., ESCO)?}

From the 2010s until the 2020s, we were experiencing the Fourth Industrial Revolution~\citep{lasi2014industry,schwab2017fourth}, characterized by a digital revolution and more mobile internet. The cost of the technologies used is also decreasing as they develop. At the time of writing, 1TB of disk storage costs around 14 EUR, whereas when the author of this thesis was born, the price was 128,000 EUR. Such low-cost access to technologies drove innovation. Reasonably affordable graphical processing units (GPUs) were one of the reasons for the advances in Artificial Intelligence (AI) and Deep Learning (DL;~\citealp{krizhevsky2012imagenet,lecun2015deep,goodfellow2016deep}). Since 2021, AI has been seen as one of the main drivers for Industry 5.0~\citep{eu2021industry}. Such adoptions of technological changes have implications for the ever-changing labor market \citep{brynjolfsson2011race,brynjolfsson2014second}. One concrete example is Large Language Models (LLMs) and especially ChatGPT~\citep{openai-chat} in December 2022.\footnote{We would like to note that most of the work done in this thesis was before the emergence of LLMs in December 2022.} Investigations show the impact of the technology on the labor market~\citep{eloundou2023gpts}:

\begin{quote}
    ``80\% of the U.S. workforce could have at least 10\% of their work tasks affected by LLMs and 19\% of the U.S. workforce may see at least 50\% of their work tasks affected by LLMs.''
\end{quote}

Therefore, we can expect a surge of job vacancies emerging on a variety of platforms and in big quantities. Investigating these job vacancies can provide insights into labor market dynamics, such as skill demands or job matching~\citep{holt1966concept, balog2012expertise, azar2022labor}. To infer occupational skill demands automatically, one can apply natural language processing (NLP) technology to the job vacancy data to extract relevant skills from the job ad, e.g., ``Python programming'' or ``communication skills''. Once these skills are extracted, we could deduce skill demand and get a snapshot of the labor market.

In this thesis, we investigate how we can leverage NLP to extract relevant information from job vacancy data, and in part how to further standardize the extracted information. 
More specifically, in this thesis, we concern ourselves with the task of Skill Extraction (SE): The task of extracting relevant spans from a text that resembles a skill. SE is one of the core applications within a field we depict as \emph{Computational Job Market Analysis}.

The bottleneck of this task is multi-fold, we refer to these challenges in~\cref{sub:challenges}. There are also other general challenges surrounding this task, not only limited to SE, such as data annotation and how we can standardize these extracted skills. The necessary background for this domain is presented in~\cref{chap:chap0} (\textbf{Part I}). In the following sections, we introduce several research questions, which are also the larger overall themes of this thesis (\textbf{Part II}, \textbf{III}, and \textbf{IV}).

\section{Annotating Data (Part II)}
Job postings can contain privacy-related entities, such as phone numbers, names, and other personally identifiable information. To comply with, e.g., the European General Data Protection Regulation (GDPR;~\citealp{regulation2016regulation}), we need to remove such information from the text before we continue extracting other types of information (e.g., skills). Manually masking or removing this information is time-consuming and labor-intensive. To the best of our knowledge, there exists no de-identification work in the job posting domain. There are several works on de-identifying parts in other sources of text, such as the medical domain (e.g.,~\citealp{szarvas2007state, meystre2010automatic, liu2015automatic, jiang2017identification, friedrich-etal-2019-adversarial, trienes2020comparing}) or SMS messages (e.g.,~\citealp{treurniet2012collection, patel2013approaches, chen2013creating}). The task at the time was frequently approached with a bidirectional Long-Short Term Memory~\citep{hochreiter1997long} model (BiLSTM). We explore how Transformer-based models compare against recurrent neural networks (e.g., BiLSTMs). 

Additionally, as there are de-identification datasets available in other domains, we investigate how a model can benefit by learning from these other auxiliary, related tasks via multi-task learning~\citep{caruana1997multitask}. In this thesis, we annotate for these privacy-related entities in job postings and seek to answer the following research questions:

\noindent\rule{\linewidth}{2pt}
\begin{center}
    \textbf{\cref{chap:chap1}}
\end{center}
\begin{itemize}
    \item[\textbf{RQ1}] \rqone{}
    \item[\textbf{RQ2}] \rqtwo{}
\end{itemize}
\noindent\rule{\linewidth}{2pt}\\\\

\noindent
Generally, annotation is a labor-intensive and expensive process. A potential remedy involves the use of active learning~\citep{cohn1994active, lewis1994heterogeneous, settles2009active}. The fundamental concept behind active learning is that a Machine Learning (ML) model can enhance its accuracy using fewer labeled training instances when it can \emph{select} the data for learning. This approach is extensively used to address the challenges associated with the time-consuming and costly tasks of manually labeling data. 

The most straightforward and widely employed approach is \textit{uncertainty sampling}~\citep{lewis1994sequential, lewis1994heterogeneous}, wherein the learner queries instances about which it is most uncertain. However, uncertainty sampling is myopic, as it solely measures the information content of an individual data instance. Alternative AL algorithms focus on assembling a \textit{diverse} batch for annotation~\citep{geifman2017deep, sener2017active, gissin2019discriminative, zhdanov2019diverse} or estimating the \textit{uncertainty} distribution of the model~\citep{houlsby2011bayesian, gal2016dropout}. Nevertheless, these methods are typically constrained in their definition of informativeness, which is associated with post-training model uncertainty and batch diversity. Instead, it could be more useful if the behavior of the model on individual instances during training could be exploited, which is usually called training dynamics~\citep{swayamdipta2020dataset}, we investigate:

\noindent\rule{\linewidth}{2pt}
\begin{center}
    \textbf{\cref{chap:chap2}}
\end{center}
\begin{itemize}
    \item[\textbf{RQ3}] \rqthree{}\footnote{Here, we do not apply it to the job market domain or in the context of sequence labeling, but instead to text classification tasks in general domains (e.g., news and question types).}
\end{itemize}
\noindent\rule{\linewidth}{2pt}\\\\

\noindent
In the context of SE, annotation guidelines are lacking, it is unclear when a skill span starts and when it ends. Consequentially, there exists a lacuna of high-quality annotated data to train ML models for this domain. There are, however, existing alternative resources such as the European Skills, Competences, Qualifications, and Occupations (ESCO;~\citealp{le2014esco}). This is a large-scale taxonomy consisting of textual descriptions of occupations and skills (3K and 14K respectively). For the skills part of the taxonomy, there are descriptions and other possible denominations. For example, a skill is ``plan teamwork'' and an alternative to this is ``plan to work as a team''.\footnote{See this specific example here: \url{http://data.europa.eu/esco/skill/75e48971-3242-4ea5-a1ba-4deec48ad41c}.} However, these lists of alternative names for skills are non-exhaustive and therefore, are hard to match directly to a job posting. Additional reasons, for example, are not having a string match, due to small variations of wording (e.g., driver's license vs. driving license). There could also be misspellings. Therefore, we instead develop our annotation guidelines with inspiration from the definitions and spans in ESCO (\cref{chap:chap3}).

Alternatively, existing resources are also the unannotated job postings themselves. Previous work has shown that further training language models (domain-adaptive pre-training or continuous pre-training;~\citealp{gururangan2020don}) on large corpora of a specific domain (e.g., the medical domain) shows a significant performance improvement on the downstream tasks of that domain. We explore two questions:

\noindent\rule{\linewidth}{2pt}
\begin{center}
    \textbf{\cref{chap:chap3}}
\end{center}
\begin{itemize}
    \item[\textbf{RQ4}] \rqfour{}
    \item[\textbf{RQ5}] \rqfive{}
\end{itemize}
\noindent\rule{\linewidth}{2pt}\\\\

\noindent
Resources such as ESCO are also multilingual, it contains 28 different European languages. Previous works on skill identification focus on the English language~\citep{jia2018representation,sayfullina2018learning,tamburri2020dataops}, which could hinder local job seekers from finding an occupation suitable to their specific skills within their language community via online job platforms. In~\cref{chap:chap4}), we focus on the Danish labor market and annotate the Danish job postings in a similar approach using our own developed guidelines from~\cref{chap:chap3}.

Further categorization of skills can help us gain insights into talent demands. Skill classification is the task of further categorizing extracted skill spans. Existing label categorization could include ESCO, beyond these, there are alternative taxonomies, such as the Russian professional standard in \citet{Botov.2019} or the Chinese Occupation Classification Grand Dictionary used in \citet{Cao.2021}. However, in this thesis, we make use of ESCO, due to its comprehensiveness and coverage of the English and Danish language. Instead of manually annotating the skill category for each span, one can leverage alternative approaches to induce labels automatically. This is generally known as distant supervision~\citep{mintz2009distant}, further defined in~\cref{def:weaksupervision}. We seek to answer:

\noindent\rule{\linewidth}{2pt}
\begin{center}
    \textbf{\cref{chap:chap4}}
\end{center}
\begin{itemize}
    \item[\textbf{RQ6}] \rqsix{}
    \item[\textbf{RQ7}] \rqseven{}
\end{itemize}
\noindent\rule{\linewidth}{2pt}\\\\

\section{Modeling Occupational Skills (Part III)}

It is practically infeasible to annotate all the job posting data for skills. Most SE approaches are supervised and need costly and time-consuming annotation. There are other ways to obtain labeled data, such as distant supervision, as mentioned in the previous section. We can leverage knowledge bases like ESCO to find matching skill spans in text. This can be approached using exact or fuzzy string matching, or more powerful methods that find similar skill spans in embedding space.
\noindent\rule{\linewidth}{2pt}
\begin{center}
    \textbf{\cref{chap:chap5}}
\end{center}
\begin{itemize}
    \item[\textbf{RQ8}] \rqeight{}
\end{itemize}
\noindent\rule{\linewidth}{2pt}\\\\

\noindent
Instead of using ESCO as an external knowledge base to match skills, there are several investigations on integrating factual knowledge or rare/infrequent words into a language model~\citep{peters-etal-2019-knowledge,zhang-etal-2019-ernie,he-etal-2020-bert,wang-etal-2021-kepler,wang2021k,yu-etal-2022-dict}. We could instead integrate ESCO directly into a language model. As previous research has been focusing on English and Danish, we take it a step further and integrate all 28 languages of ESCO into a multilingual language model (e.g.,~\citealp{conneau2020unsupervised}) and improve performance for skill extraction and classification.
\clearpage
\noindent\rule{\linewidth}{2pt}
\begin{center}
    \textbf{\cref{chap:chap6}}
\end{center}
\begin{itemize}
    \item[\textbf{RQ9}] \rqnine{}
\end{itemize}
\noindent\rule{\linewidth}{2pt}\\\\

\noindent
Re-training or even further pre-training language models can be costly with the large compute necessary, especially in the context of large language models. One effective strategy to circumvent this is retrieval augmentation~\citep{grave2017improving, lee-etal-2019-latent, guu2020retrieval, lewis2020retrieval, Khandelwal2020Generalization, izacard2021distilling, izacard2022few}. Here, language models can use external modules to enhance their context-processing ability. Two approaches are commonly used: First, using a separately trained model to retrieve relevant documents from a collection. This approach is employed in open-domain question answering tasks~\citep{petroni-etal-2021-kilt}.

Second, previous work on explicit memorization showed promising results with a cache~\citep{grave2017improving}, which serves as a type of datastore. 
The cache contains past hidden states of the model as keys and the next word as tokens in key-value pairs. Memorization of hidden states in a datastore, involves using the $k$NN algorithm as the retriever. 
The first work of the $k$-nearest neighbors ($k$NN) algorithm as the retrieval component was by \citet{Khandelwal2020Generalization}, leading to several LM decoder-based works. Regardless, it is unclear how we can leverage multiple datasets for retrieval augmentation with encoder-based models.
\clearpage
\noindent\rule{\linewidth}{2pt}
\begin{center}
    \textbf{\cref{chap:chap7}}
\end{center}
\begin{itemize}[align=left]
    \item[\textbf{RQ10}] \rqten{}
\end{itemize}
\noindent\rule{\linewidth}{2pt}\\\\

\section{Linking Skills to Existing Resources (Part IV)}\label{sec:intro:linking}

Finally, once we extracted the skills, it is important to link surface form skills to a taxonomy entry, allowing us to quantify the current labor market dynamics and determine the demands and needs. This task can be framed as an entity linking problem~\citep{he-etal-2013-learning, logeswaran-etal-2019-zero, wu-etal-2020-scalable}. Specifically, we aim at the linking of fine-grained span-level skill mentions to a specific taxonomy entry. 

Generally, entity linking is the task of linking mentions of entities in unstructured text documents to their respective entities in a knowledge base, most commonly Wikipedia~\citep{he-etal-2013-learning}. Recent models address this problem by producing entity representations from a (sub)set of knowledge base information, e.g., entity descriptions~\citep{logeswaran-etal-2019-zero,wu-etal-2020-scalable}, fine-grained entity types~\citep{raiman2018deeptype,onoe2020fine,ayoola-etal-2022-refined}, or generation of the input text autoregressively~\citep{cao2021autoregressive,de2022multilingual}.

For linking skills to a taxonomy, we can use the European Skills, Competences, Qualifications, and Occupations (ESCO;~\citealp{le2014esco}) taxonomy. Previous work classified spans to its taxonomy code via multi-class classification~\citep{zhang-etal-2022-kompetencer} without surrounding context and neither the full breadth of ESCO.~\citet{gnehm-etal-2022-fine} approaches this as a coarse-grained sequence labeling task using a subset of ESCO, and others classify on the sentence level~\citep{decorte2022design,decorte2023extreme,clavie2023large}. Therefore, we investigate:

\noindent\rule{\linewidth}{2pt}
\begin{center}
\textbf{\cref{chap:chap8}}
\end{center}
\begin{itemize}[align=left]
    \item[\textbf{RQ11}] \rqeleven{}
\end{itemize}
\noindent\rule{\linewidth}{2pt}\\\\

\section{Chapter Guide \& Contributions}

\textbf{Part II}, \textbf{III}, and \textbf{IV} contains the collection of peer-reviewed main contributions (detailed in~\cref{sec:publications}), consisting of:

\cref{chap:chap1}, where we present \jobstack{}, a corpus for de-identification of personal data in job vacancies on Stackoverflow for ethical reasons and to comply with data regulations~\citep{regulation2016regulation}. We introduce baselines, comparing BiLSTMs and Transformer models (\textbf{RQ1}). To improve upon these baselines, we experiment with LSTM models, Transformer models, and distantly related auxiliary data via multi-task learning. Our results show that auxiliary data from both generic named entity recognition data and medical data improves de-identification performance from job postings (\textbf{RQ2}).

In~\cref{chap:chap2}, we introduce a novel Active Learning algorithm: Cartography Active Learning (CAL). It exploits the behavior of the model on individual instances \textit{during training} as a proxy to find the most informative instances for labeling. CAL is inspired by data maps, which were proposed to derive insights into dataset quality~\citep{swayamdipta-etal-2020-dataset} (\textbf{RQ3}).

We introduce \textsc{SkillSpan} in~\cref{chap:chap3}, a novel skill extraction dataset consisting of 14.5K sentences and over 12.5K annotated spans. We release its respective guidelines created over three different sources annotated for hard \emph{and} soft skills by domain experts (\textbf{RQ4}). Additionally, we continuously pre-train English-based language models on unlabeled job posting data and adapt them to the job market domain (\textbf{RQ5}).

In~\cref{chap:chap4}, we release the first Danish job posting dataset: \textsc{Kompetencer} (\emph{en}: competences), annotated for nested spans of competences. To improve upon coarse-grained annotations, we make use of the ESCO taxonomy API to obtain fine-grained labels via distant supervision. Additionally, we also further pre-train a Danish language model on unlabeled Danish job posting data and adapt it to the job market domain (\textbf{RQ6}, \textbf{RQ7}). 

In~\cref{chap:chap5}, we propose Skill Extraction with Weak Supervision. We leverage the ESCO taxonomy to find similar skills in job ads via latent representations. The method shows a strong positive signal, outperforming baselines based on token-level and syntactic patterns (\textbf{RQ8}).

There is a lack of generalized, multilingual models and benchmarks for skill extraction and classification tasks. In~\cref{chap:chap6}, we introduce a language model called \escolmr{}, based on \xlmr{}, which uses domain-adaptive taxonomy-driven pre-training objective on the ESCO taxonomy, covering 27 languages. The pre-training objectives for \escolmr{} include dynamic masked language modeling and a novel additional objective for inducing multilingual taxonomical ESCO relations (\textbf{RQ9}). We benchmark the model on nine job-related datasets in four different languages and shows that \escolmr{} outperforms the existing state-of-the-art on six out of nine datasets.\footnote{During the thesis, there were several contemporaneous datasets created and released.}

In~\cref{chap:chap7}, we tackle the complexity in occupational skill datasets tasks---combining and leveraging multiple datasets for skill extraction---to identify rarely observed skills within a dataset, and overcoming the scarcity of skills across datasets. In particular, we investigate the retrieval-augmentation of language models, employing an external datastore for retrieving similar skills in a dataset-unifying manner. Our proposed method, \textbf{N}earest \textbf{N}eighbor \textbf{O}ccupational \textbf{S}kill \textbf{E}xtraction (NNOSE) effectively leverages multiple datasets by retrieving neighboring skills from other datasets in the datastore (\textbf{RQ10}). This improves skill extraction without additional fine-tuning.

Standardizing skill mentions can help us get insight into the current labor market demands by getting a distribution over the skill requirements in job ads. In~\cref{chap:chap8}, we explore Entity Linking (EL) in this domain, specifically targeting the linkage of occupational skills to the ESCO taxonomy~\citep{le2014esco} (\textbf{RQ11}). Previous efforts linked full sentences to a corresponding ESCO skill. Instead, in this work, we link fine-grained span-level mentions of skills.

\section{Publications}\label{sec:publications}

The peer-reviewed work presented in this thesis consists of the following publications:

\begin{enumerate}
    \item \bibentry{jensen-etal-2021-de}.\footnote{Shared first authorship.} (\cref{chap:chap1}, \textbf{RQ1}, \textbf{RQ2})
    \item \bibentry{zhang-plank-2021-cartography-active}. (\cref{chap:chap2}, \textbf{RQ3})
    \item \bibentry{zhang-etal-2022-skillspan}. (\cref{chap:chap3}, \textbf{RQ4}, \textbf{RQ5})
    \item \bibentry{zhang-etal-2022-kompetencer}. (\cref{chap:chap4}, \textbf{RQ6}, \textbf{RQ7})
    \item \bibentry{zhang2022skill}. (\cref{chap:chap5}, \textbf{RQ8})
    \item \bibentry{zhang2023escoxlm}. (\cref{chap:chap6}, \textbf{RQ9})
    \item \bibentry{zhang2024nnose}. (\cref{chap:chap7}, \textbf{RQ10})
    \item \bibentry{zhang2024eljob}. (\cref{chap:chap8}, \textbf{RQ11})
\end{enumerate}

\noindent
All the papers above are covered in the thesis. I primarily led these articles in terms of framing the problem, conducting the experiments, and writing the papers with the help and supervision of my co-authors and supervisors. 

\subsection*{Other (Non-published) Articles}
I was also part of several other works (some not published) that are not in this thesis.

\begin{enumerate}
    \item[9.] \bibentry{bassignana-etal-2022-evidence}
    \item[10.] \bibentry{ulmer-etal-2022-experimental}
    \item[11.] \bibentry{barrett2024domain}
    \item[12.] \bibentry{senger2024survey}
    \item[13.] \bibentry{magron2024jobskape}
    \item[14.] \bibentry{nguyen2024rethinking}
\end{enumerate}

\newpage
\part{Background}
\newpage

\chapter{Challenges with Skill Extraction in the Job Market Domain}
\chaptermark{{Challenges}}
\label{chap:chap0}
\newcommand{\cmark}{\ding{51}}%
\newcommand{\xmark}{\ding{55}}%

This chapter provides the general background for the thesis. We discuss how skills are defined, what the task of skill extraction entails, how we categorize skills, evaluate and analyze the predictions of skill extraction models, considerations for data annotation (e.g., privacy-related information and faster annotation), and how we can transfer general-purpose models to the job market domain.

\section{Defining Skills}

\subsection{The European Skills, Competences, Qualifications and Occupations}\label{int:esco}

The notion of skill or competence encompasses diverse definitions within various disciplinary and research spheres (e.g., \citealp{rauner2008handbook,zhao2014areas}). For example, \citet{rauner2008handbook} defines competences as possessing requisite abilities, authority, skills, and knowledge, reflecting an individual's psychological attributes. The concept of competence involves considerations of various influencing factors, including educational systems, cultural contexts, economic structures, vocational traditions, and labor market dynamics.

Another definition of competences by~\citet{peterssen2001kleines} intends to capture specific reference points of human action. The differentiation follows the assumption that a person can be confronted with four types of challenges:

\begin{enumerate}
    \item \textbf{Factual competence}: The focus is on dealing with material or symbolic objects, i.e., objects of nature or culture. Such objects include computers, office furniture, tools, or even texts, formulas, and programming languages.
    \item \textbf{Social competence} is about dealing with other people in different communication situations (dyad, group/team, community).
    \item \textbf{Self-competence} focuses on dealing with facets of one's own person. For example, it is about dealing with one's own emotions (e.g., fear, and aggression) or with one's own learning behavior.
    \item \textbf{Methodological competence} is the usage of strategies and techniques of learned knowledge (i.e., factual competences).
\end{enumerate}

\begin{figure*}[t]
    \centering
    \includegraphics[width=\linewidth]{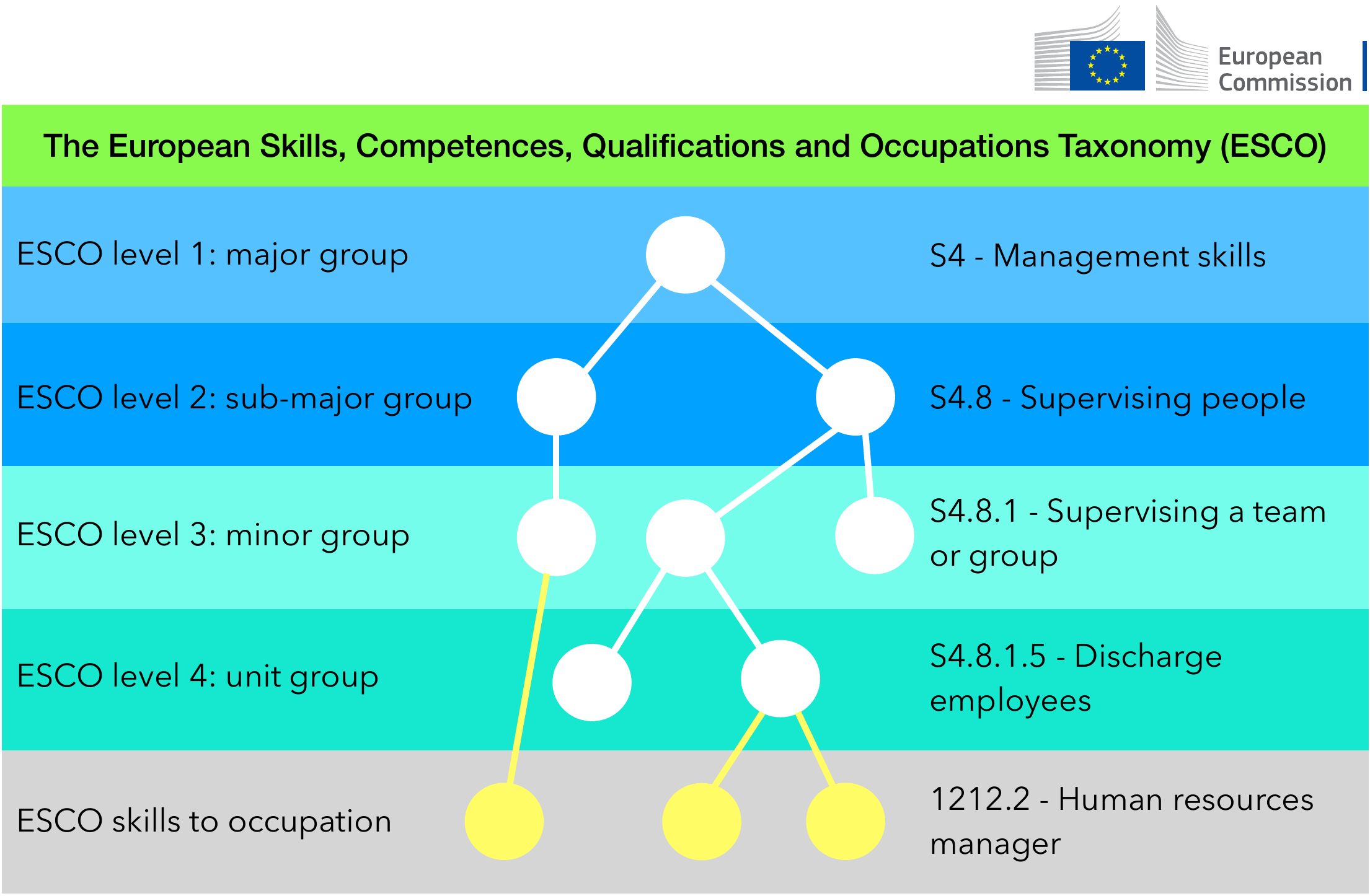}
    \caption{\textbf{A Graphical Illustration of ESCO.} We show the simplified structure of the hierarchical ESCO taxonomy. It contains four levels of skills, where each skill can be connected to a specific occupation.}
    \label{fig:esco-illustration}
\end{figure*}

As shown, competences are diverse, spurring substantial efforts toward categorization. The International Standard Classification of Occupations (ISCO; \citealp{elias1997occupational}) stands as a key global framework categorizing occupations and skills, operating within the international economic and social classification family. Similarly, the European Skills, Competences, Qualifications, and Occupations (ESCO; \citealp{le2014esco}) taxonomy, derived from ISCO, constitutes the European standard terminology connecting skills, competences, and qualifications to occupations. ESCO encompasses 13,890 competences and 3,008 occupations.\footnote{Per version 1.1.0.} We show a graphical illustration of the hierarchy in ESCO in~\cref{fig:esco-illustration}. Here, we only focused on a specific skill group (namely S4: management skills). The taxonomy is four levels deep, where skills in each level can be connected to an occupation. At the time of the work done in this thesis divided competences into three categories: \emph{Knowledge}, \emph{Skill}, and \emph{Attitudes}.

Knowledge, as per ESCO, is defined as the assimilation of information through learning, encompassing facts, principles, theories, and practices related to a particular field of work or study.\footnote{\url{https://ec.europa.eu/esco/portal/escopedia/Knowledge}} For instance, the acquisition of the Python programming language through learning represents a \emph{knowledge} component, often referred to as a \emph{hard skill}. Next, the application of this knowledge to specific tasks is regarded as a \emph{skill} component, defined by ESCO as the ability to apply knowledge and utilize know-how to accomplish tasks and solve problems.\footnote{\url{https://ec.europa.eu/esco/portal/escopedia/Skill}} ESCO further designates \emph{soft skills} as \emph{attitudes}, considering them integral to skill components:

\begin{quote}
``The ability to use knowledge, skills, and personal, social and/or methodological abilities, in work or study situations and professional and personal development''\footnote{\url{https://data.europa.eu/esco/skill/A}}
\end{quote}

In summary, \emph{hard skills} are commonly identified as \emph{knowledge} components, and the application of these skills is construed as a \emph{skill} component. On the other hand, \emph{soft skills}, termed as \emph{attitudes} in ESCO, are integral to skill components.

A concrete example of a \emph{knowledge} component is ``K8: engineering, manufacturing and construction
'', with one of the leaf nodes being ``
K8.3.1: architectural conservation
'', this has the description ``The practice of recreating forms, features, shapes, compositions, and architectural techniques of past constructions to preserve them''.\footnote{See the specific knowledge component here \url{https://esco.ec.europa.eu/en/classification/skills?uri=http://data.europa.eu/esco/isced-f/0831}.}

A specific example of a \emph{skill} component is ``S6: handling and moving'' with the description ``Sorting, arranging, moving, transforming, fabricating and cleaning goods and materials by hand or using handheld tools and equipment. Tending plants, crops, and animals''. Then a sibling of this specific skill is ``S6.2: moving and lifting'' with the description ``Performing physical activities to move, load, unload or store objects, position workpieces or equipment for assembly, or to climb structures, by hand or with the aid of equipment''.\footnote{See the specific skill example here \url{https://esco.ec.europa.eu/en/classification/skills?uri=http://data.europa.eu/esco/isced-f/08}.}

\subsection{Challenges}\label{sub:challenges}

The extraction of skills presents several distinctive nuances and challenges, which contribute to the complexity of this task and make it an intriguing area for exploration within the broader scope of NLP research.

\begin{enumerate}
    \item There can be an implicitness in the representation of skills in textual content.\footnote{In this context, an implicit skill refers to a standardized skill form that is not explicitly mentioned in the text.} For instance, the skill of ``teamwork'' may manifest in various formats such as ``being able to work together'', ``working in a team'', or ``collaborating efficiently''. The diverse expressions of these skills pose significant implications for the modeling approaches for SE.
    \item Skills, by their nature, can exhibit considerable length, posing a challenge in terms of extraction with NLP models.
    \item Furthermore, the distribution of skills follows an inherent long-tail pattern, signifying that certain skills are less frequently mentioned in a comprehensive set of job postings.
    \item Another intriguing problem pertains to the discontinuity of skill spans, analogous to what is referred to as \emph{coordinated conjunctions} in linguistics. This aspect introduces a layer of complexity in accurately delineating the boundaries of skills within the text.
    \item Last, if we make use of a taxonomy like ESCO as the labels, this results in a substantial output space during the prediction phase, adding a layer of complexity to the modeling process.
\end{enumerate}

We investigate and address several of these challenges in this thesis. 
First, in \cref{chap:chap8} covers how well a model can link implicit skills to a taxonomical counterpart (Challenge 1). 
Second, in~\cref{chap:chap3} and \cref{chap:chap5}, we investigate how well encoder-based language models can predict lengthy skills (Challenge 2).
Then, in~\cref{chap:chap7}, we address how we can leverage multiple datasets to tackle the inherent long-tail frequency pattern of skills (Challenge 3).
In~\cref{chap:chap4} and \cref{chap:chap8}, we look into the large label space of ESCO. We approach this challenge as both a classification and linking task (Challenge 5).
Last, the fourth challenge of discontinuous skill extraction is not addressed in this thesis and is left for future work.

\begin{figure}
    \centering
    \includegraphics[width=\linewidth]{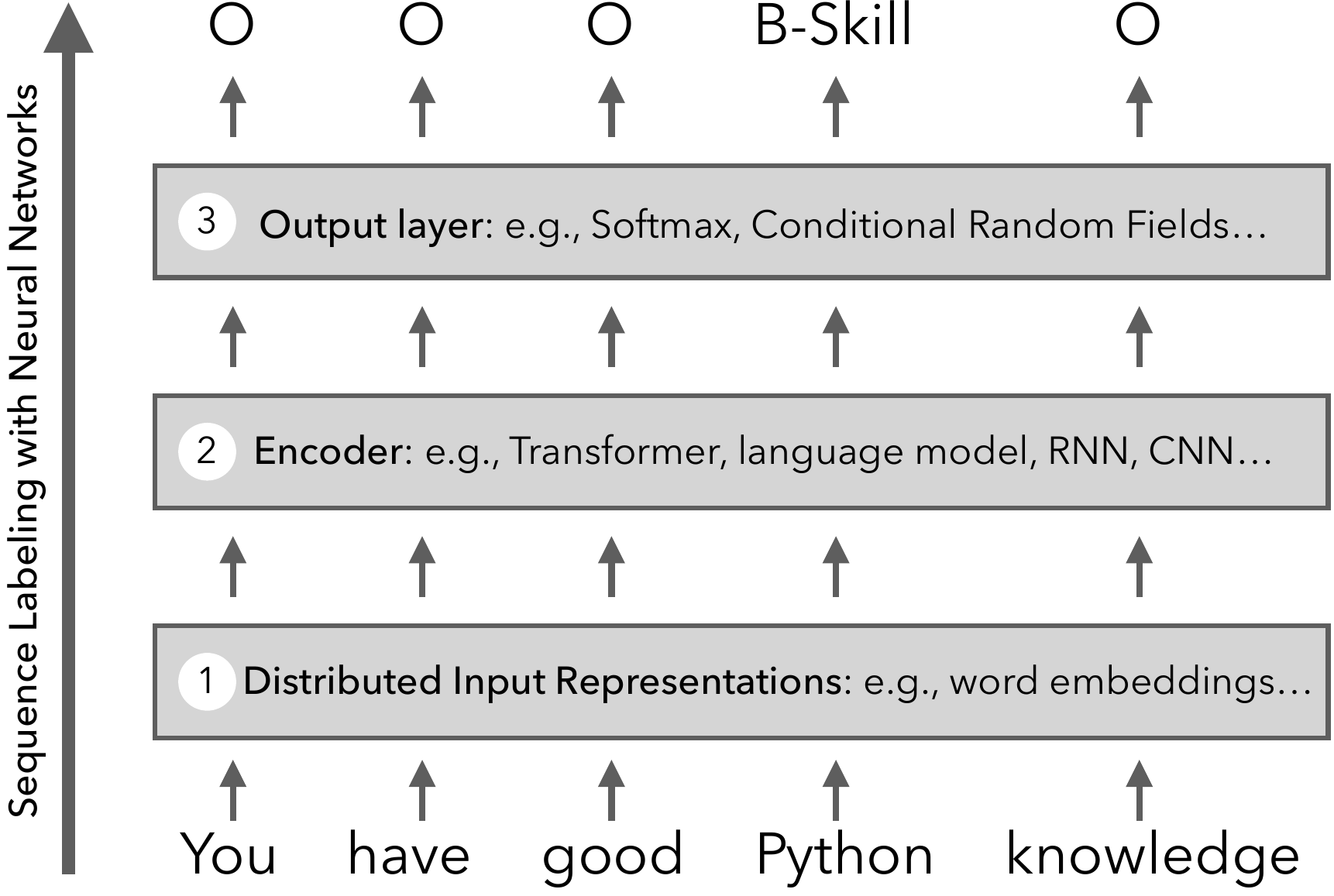}
    \caption{\textbf{Sequence Labeling with Neural Networks.} The flow for DL-based sequence labeling. The input sequence goes through 1) a layer that tokenizes the input, 2) an encoder that transforms the representations into meaningful vectors, 3) an output layer where the tags (\texttt{B-Skill}, \texttt{I-Skill}, 
    \texttt{O}) get predicted from each token vector.}
    \label{fig:seq-label-nn}
\end{figure}

\section{Skill Extraction}

\subsection{Skill Extraction Definition}
In this thesis, the concept of skill extraction is operationalized by framing it as a sequence labeling problem as we want to extract the exact subspan from the text. Specifically, we use the \texttt{BIO}-labels~\citep{ramshaw-marcus-1995-text} for token-level predictions. Formally, let $\mathcal{D}$ represent a set of Job Descriptions (JDs), where each $d \in \mathcal{D}$ constitutes a set of sequences, denoted as $\mathcal{X}^i_{d} = \{x_1, x_2, ..., x_T\}$ for the $i^\text{th}$ input sequence, and a corresponding target sequence of \texttt{BIO}-labels, $\mathcal{Y}^i_{d} = \{y_1, y_2, ..., y_T\}$ (e.g., \texttt{B-Skill}'', \texttt{I-Skill}'', ``\texttt{O}''). The primary objective is to train a sequence labeling algorithm, denoted as $h: \mathcal{X} \mapsto \mathcal{Y}$, utilizing $\mathcal{D}$ to accurately predict entity spans by assigning an output label $y_t$ to each token $x_t$.

\paragraph{Other Definitions} Another frequently used task formulation of SE is multi-label classification~\citep{bhola-etal-2020-retrieving, goyal-etal-2023-jobxmlc}. Formally, for a given input text \(t \in T\), we seek a mapping \(f: T \rightarrow [0, 1]^S\), where \(S = |S|\) represents the cardinality of the global skill set (i.e., predefined label space). The function \(f\) provides a probability score for each label \(s \in S\) given the input text \(t\):
\[ f(t) = P(r_i = 1 | t) \]
where \(i \in \{1, \ldots, S\}\), and \(r_i\) denotes the label corresponding to the \(i\)-th skill. $P$ is usually the result of a sigmoid function and ensures that the output of \(f(t)\) is constrained to the interval \([0, 1]\). To encapsulate the essence of the task, it entails assigning labels to textual descriptions from an extensive set of skills. However, in this thesis, we are mostly concerned with the sequence labeling task for skill extraction, as we can extract the exact subspan a skill label refers to from the text.

\subsection{Span Extraction with Neural Networks}

The task of sequence labeling is to assign a label to every token in a sentence. A single span could consist of multiple tokens within a sentence. The tokens in the sentence are usually represented in the BIO format (Beginning, Inside, and Outside) where every token is labeled as a B-label if the token is the beginning of the span, and I-label if it is the inside of the span, but not the first token~\citep{ramshaw-marcus-1995-text,tjong2003introduction}. 

In~\cref{fig:seq-label-nn}, we show the common framework of a sequence labeling problem in context of neural networks. The input sentence ``You have good Python knowledge'' goes through three stages: First, the sentence gets transformed into some form of input representations, these are frequently lookup tables assigning vectors to corresponding tokens. Second, the input representations go through an encoder, where the goal is to learn a representation of the input sequence. Last, the encoded representations are transformed in the output layer, consisting usually of a feedforward layer and a softmax or Conditional Random Fields (CRF;~\citealp{lafferty2001conditional}) layer to predict the final labels.

\subsection{Evaluation}
In this work, the evaluation of sequence labeling involves two primary evaluation schemes: \emph{strict} and \emph{loose} span-F1. We always use strict span-F1~\citep{tjong-kim-sang-de-meulder-2003-introduction,seqeval}, same as standard span-F1, unless otherwise stated. The emphasis on span-F1, as opposed to regular F1, is driven by the correctness and annotation at the full span level. The evaluation criteria for strict and loose spans are different based on the matching boundaries between predicted and gold spans, i.e., true positives (TP).

\paragraph{Strict}
We count towards TP if the exact boundaries of predicted and gold spans match, including the correct skill type.

\paragraph{Loose}
We count a TP when there is an overlap between the boundaries of predicted and gold spans, including the correct skill type.

False positives (FP) are the predicted spans that are not in the gold evaluation set and false negatives (FN) are the gold spans that are not predicted. Then, to calculate precision and recall, we do the following:

\paragraph{Precision}

One can interpret this as ``out of the predicted spans, how many are correct?''

\begin{equation}
    \text{Precision} = \frac{\text{TP}}{\text{TP} + \text{FP}}\text{, or}
\end{equation}
\begin{equation*}
    \text{Precision} = \frac{\text{Number of correctly predicted spans}}{\text{Total number of predicted spans}},
\end{equation*}

\paragraph{Recall}

One can interpret this as ``out of all gold spans, how many are correct?''

\begin{equation}
    \text{Recall} = \frac{\text{TP}}{\text{TP} + \text{FN}}\text{, or}
\end{equation}
\begin{equation*}
    \text{Recall} = \frac{\text{Number of correctly predicted spans}}{\text{Total number of true spans}}
\end{equation*}

\noindent
Then, the final Span-F1 score is calculated as the harmonic mean between precision and recall~\citep{van1979information}:

\begin{equation}\label{eq:f1}
    \text{Span-F1} = \frac{2 \times \text{Precision} \times \text{Recall}}{\text{Precision} + \text{Recall}}.
\end{equation}

\noindent
One could also consider other $\beta$ weighting for the F-score, instead of a balanced weighting as in span-F1. For example, two common values for $\beta$ are 2, which weighs recall higher than precision, and 0.5, which weighs recall lower than precision~\citep{van1979information}.

\section{Categorization of Skills}
Once skill spans are extracted from the text, we can further categorize them. In turn, this allows us to quantify skill demand more precisely (\cref{sec:intro:linking}). For categorization, we use the ESCO taxonomy labels as the gold standard. We outline two types of skill categorization: 1) \textit{Coarse-grained skill categorization} adopts the top-level major group of ESCO, yielding 23 labels (\cref{subsec:coarse}).
2) \textit{Fine-grained skill categorization} involves the entire breadth of the taxonomy, encompassing around 14,000 skills, each with its unique taxonomy code (\cref{subsec:fine}).

\subsection{Coarse-grained Skill Categorization}\label{subsec:coarse}

Coarse-grained skill classification refers to the classification of already extracted skill spans from text, where we define coarse as 50 labels or less. Considering a set of job descriptions $\mathcal{D}$, where $d \in \mathcal{D}$ is a collection of extracted spans (not full sentences), with the $i^\text{th}$ span denoted as $\mathcal{X}^i_{d} = \{x_1, x_2, ..., x_T\}$ and a target class $c \in \mathcal{C}$, where $\mathcal{C} = \{\text{\texttt{S*}}, \text{\texttt{K*}}\}$. \texttt{S*} and \texttt{K*} are \emph{skill} and \emph{knowledge} components from the ESCO taxonomy (\cref{int:esco}). The goal is to train an algorithm $h: \mathcal{X} \mapsto \mathcal{C}$ to predict skill tags accurately by assigning an output label $c$ for input $\mathcal{X}^i_{d}$.

\paragraph{Evaluation of Skill Categorization}
Given that some skills may be more prevalent than others~\citep{autor2003skill,autor2013growth}, an inherent class imbalance exists in the set of ground-truth skills. To take this imbalance into account, one can use the weighted macro-F1, a metric to evaluate a classification model by calculating the F1-score for each class individually, taking into account the number of instances in each class:

\begin{equation}
\text{Weighted Macro-F1} = \frac{\sum_{i=1}^{C} w_i \times \text{F1}_i}{\sum_{i=1}^{C} w_i}
\end{equation}

\noindent
where:
\begin{align*}
&C \text{ is the total number of classes,} \\
&w_i \text{ is the weight, based on class frequency, assigned to class } i, \\
&\text{F1}_i \text{ is the F1 score for class } i, \\
&\text{Precision}_i \text{ is the precision for class } i, \\
&\text{Recall}_i \text{ is the recall for class } i.
\end{align*}

\noindent
The F1 score for each class \(i\) is also calculated as the harmonic mean between precision and recall (\cref{eq:f1}).

\subsection{Fine-grained Skill Categorization}\label{subsec:fine}

In the context of skill categorization, this can be posed as an entity linking problem, where we attempt to match a span to an existing taxonomy code. Formally, we process the input document $\mathcal{D} = \{w_1, \ldots, w_r\}$, a collection of entity mentions denoted as $\mathcal{MD} = \{m_1, \ldots, m_n\}$, and a knowledge base (KB), ESCO in this case: $\mathcal{E} = \{e_1, \ldots, e_{13890}, \text{\texttt{UNK}}\}$, where the number 13,890 is the total amount of skills in ESCO and \texttt{UNK} is a label for a non-linkable skill. The objective of an entity linking model is to generate a list of mention-entity pairs $\{(m_i, e_i)\}_{i=1}^{n}$, where each entity $e$ corresponds to an entry in the KB.

\paragraph{Evaluation of Skill Categorization}

For the evaluation of linking performance, we assess the accuracy of generated mention-entity pairs in comparison to the ground truth. Here, we use the evaluation metric Accuracy@$k$, following prior research~\citep{logeswaran-etal-2019-zero,wu-etal-2020-scalable, zaporojets2022tempel}. We calculate the correctness between mentions and entities in the KB as the sum of correct hits or true positives if the ground truth for instance $i$ is in the top-$k$ predictions, formally:

\begin{equation}
\begin{aligned}
\text{Accuracy@$k$} = \frac{1}{n} \sum_{i=1}^{n} \text{TP in top-$k$ for instance $i$.}
\end{aligned}
\end{equation}

\noindent
In summary, in this section, we formalized the task of skill extraction and closely related tasks. In the next sections, we will be talking about several peculiarities in SE and how we analyze them.

\section{Annotating Data}

This section will provide a concise overview of practical considerations within data annotation. Specifically, we will delve into the considerations of handling personally identifiable information (PII) within annotated datasets, recognizing its significance in ethical and privacy dimensions. Additionally, we will explore the incorporation of active learning techniques to enhance efficiency in data labeling. 

\subsection{Personally Identifiable Information}
Job advertisement data contains information that pertains to individual privacy. This data typically comprises predominantly \emph{general} information about individuals, as opposed to \emph{sensitive} information, as defined by the European General Data Protection Regulation (GDPR;~\citealp{regulation2016regulation}). General information encompasses details such as names, addresses, email addresses, phone numbers, and similar identifiers. On the other hand, sensitive information includes data related to racial or ethnic backgrounds, religious beliefs, social security numbers, and so forth. Ensuring the privacy of individuals is a legal mandate, enforced by various legislations such as the US Health Insurance Portability and Accountability Act (HIPAA;~\citealp{act1996health}) and the GDPR. For this thesis, our concern is primarily with complying with ethical and legal requirements to safeguard the privacy of subjects, specifically about the general information contained in job advertisements. In~\cref{chap:chap1}, we investigate whether we can train an NLP system to remove parts of the text that can lead to the identification of individuals.

\subsection{Active Learning}

\begin{figure*}[ht]
    \centering
    \includegraphics[width=.8\linewidth]{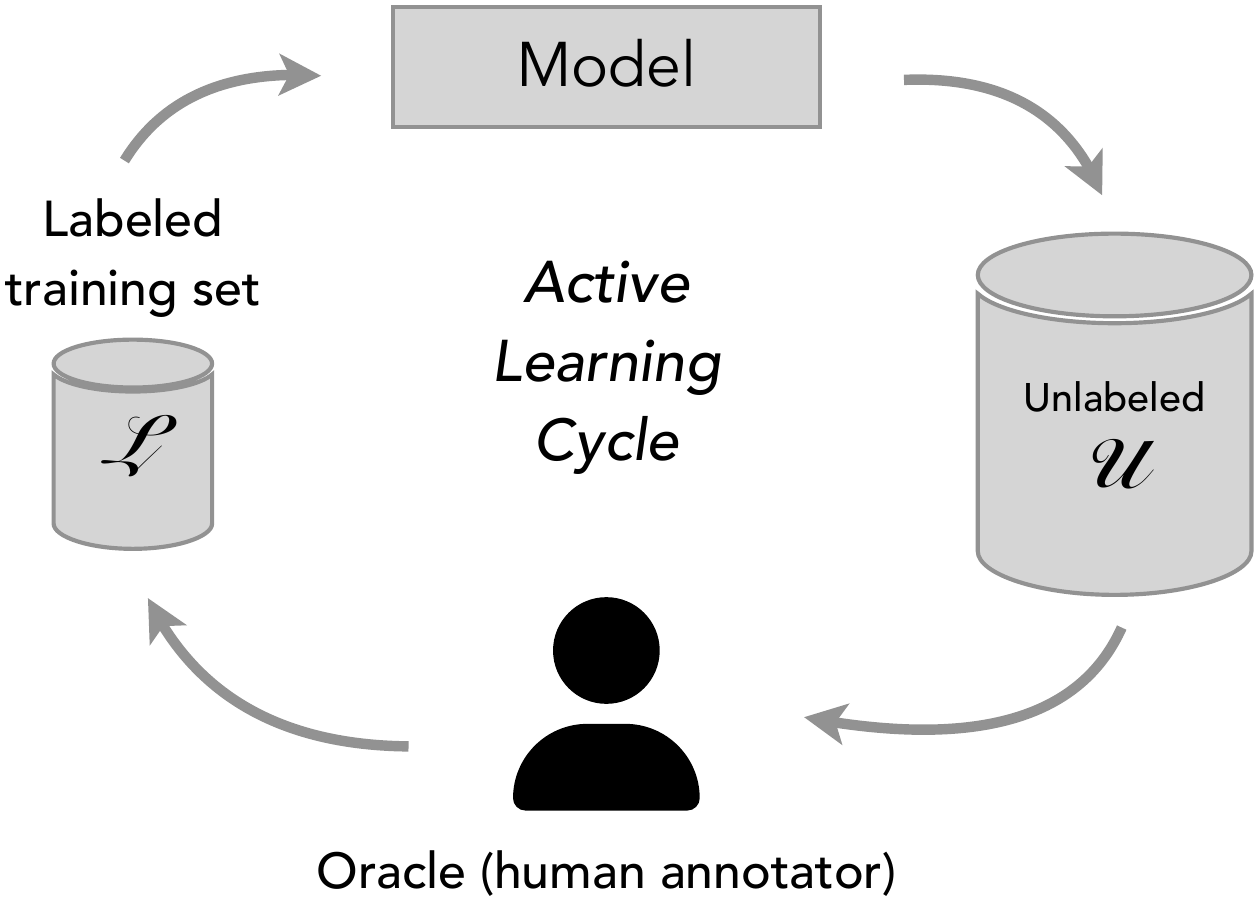}
    \caption{\textbf{Pool-based Active Learning Cycle.} The human annotator starts to annotate a small labeled set $\mathcal{L}$ to train an initial model. Then, the model is applied to an unlabeled pool of data $\mathcal{U}$. Given a certain acquisition function or score, the model selects the instances that are the most \emph{informative}. This set of instances is given to the oracle to annotate.}
    \label{fig:al-cycle}
\end{figure*}

The main idea of Active Learning (AL) is that a machine learning model can achieve greater accuracy with fewer labeled training instances if it can \emph{choose} the data from which it learns. Generally, it is a widely-used method to tackle the time-consuming and expensive collection and manual labeling of data~\citep{cohn1994active, lewis1994heterogeneous, settles2009active}. One can consider injecting such a process into an annotation cycle. An example of an AL cycle (pool-based in this case) is shown in~\cref{fig:al-cycle}. We start with a human annotator (also called an oracle) to annotate a couple of examples. The number of initial examples varies, but usually, the initial labeled set $\mathcal{L}$ should be enough to give a machine learning some signal of the task. Once the model is trained, this is applied to an unlabeled data pool $\mathcal{U}$. Given some acquisition function or score, the model selects the instances having the highest \emph{informativeness}. This set of unlabeled examples is given to the oracle for annotation. This cycle continues for a couple of rounds until a desired performance is achieved or other stopping criteria (e.g., convergence). In~\cref{chap:chap2}, we introduce a novel acquisition function, where we show a new form of estimating \emph{informativeness} during model training.

\subsection{Weak Supervision}\label{def:weaksupervision}
In machine learning, a significant challenge comes from the demand for extensive sets of manually annotated data. Weak supervision emerges as a strategic solution to address this issue by leveraging human expertise to generate noisy labels, often termed as \emph{weak} labels~\citep{stephan-roth-2022-weanf, zhu-etal-2023-weaker}. This approach commonly involves the incorporation of keywords, heuristics, or insights from external data sources, usually called \emph{distant supervision}~\citep{craven1999constructing, mintz2009distant}. 

In this thesis, we interchangeably use the terms weak supervision and distant supervision. However, these can be interpreted as one being a part of another, where distant supervision is a type of weak supervision~\citep{stephan-roth-2022-weanf}. In~\cref{chap:chap4} and~\cref{chap:chap5}, we leverage the ESCO Application Programming Interface (API) in two different approaches: First, in~\cref{chap:chap4}, we directly query the API with free form text to obtain noisy taxonomy codes for skill spans. Then, in~\cref{chap:chap5}, we extract the taxonomy entries from ESCO, and try to match them to n-grams in job posting text via heuristics (e.g., exact or fuzzy string matching) and embedding approaches.

\section{Analysis}
There are a couple peculiarities in skill extraction that can be linked back to the challenges mentioned in~\cref{sub:challenges}, such as the long length of skills and the long-tail distribution of skills. In this section, we discuss several approaches to analysis for skill extraction that have been used in several chapters of this thesis.

\subsection{Span Length}
Skills may be of considerable length, as will be demonstrated in \cref{chap:chap3}, where manually annotated skills were observed to span up to 20 tokens.\footnote{One example here could be: ``assist local authors in writing and delivering articles while ensuring that local articles are in high quality and that language is at eye level with our employees and comply with corporate standards''. Note that this skill is also part of the \emph{coordinated conjunction} challenge mentioned  in~\cref{sub:challenges}.} Our focus lies in assessing the predictive efficacy of language models, specifically in the context of handling long skills. To gauge the robustness of these language models on span length, we adopt a structured approach by categorizing the gold standard spans into different buckets based on their token length. Subsequently, we calculate the Span-F1 (\cref{eq:f1}) score for each distinct bucket. 

Formally, let $\{B_1, B_2, \ldots, B_k\}$ represent the token length buckets, where each $B_i$ corresponds to a range of token lengths. The gold standard spans are then assigned to the appropriate bucket $B_i$ based on their token length. The Span-F1 score is computed independently for each bucket, providing an evaluation of the language model's predictive performance across varying lengths of skills. This approach allows us to gain insights into model performance concerning the length of the predicted spans, thereby improving our understanding of its capabilities in handling diverse skill lengths. We use this type of analysis in \cref{chap:chap3} and \cref{chap:chap6}.

\subsection{Long-tail Skills}
The distribution of skills across different industries can vary, indicating a tendency for certain skills to be more popular than others. To ensure that our models account for the natural patterns of skill occurrence, we investigate how models perform on skills with lower frequency (\cref{chap:chap6}, \cref{chap:chap7}). Similar to our approach with span length, we categorize gold standard spans based on their frequency counts and proceed to calculate the Span-F1 (\cref{eq:f1}) score for each frequency bucket.

Formally, let \(S_{g}\) be the set of gold standard spans in the dataset, and \(f(s_i)\) represent the frequency count of each skill \(s_i\). We define non-overlapping frequency buckets \(B = \{B_1, B_2, \ldots, B_k\}\) based on the distribution of gold standard spans. A function \(\phi: S_{g} \rightarrow B\) assigns each span \(s_i\) to the corresponding frequency bucket \(B_j\) based on \(f(s_i)\). This approach allows us to gain insights into whether the model performs well even under limited data of skills. We use this analysis in \cref{chap:chap6} and \cref{chap:chap7}.

\subsection{False Positives}
The ESCO taxonomy, comprised of 13,890 labels, defines a finite label space. There is, however, a ``field unknown'' label for knowledge components. Continuous improvement of ESCO is crucial to sustain its value for the labor market, as well as education and training systems. 
Employers consistently demand new emerging occupations and skills, while changes in curricula and terminology are regularly integrated into education and training programs. 
To effectively address these, it is imperative to actively share feedback, suggestions, and proposals on enhancing the content and management of the classification. 
Organizations using ESCO, along with other stakeholders, play a key role in this collaborative effort to ensure the ongoing relevance and effectiveness of ESCO.
When confronted with the emergence of skills within the ESCO taxonomy, one way of finding them involves inspecting instances of false positives generated by the model. False positives, in this context, denote predictions that deviate from the gold annotations, typically indicative of overprediction tendencies. However, a qualitative inspection, as undertaken in various works in this thesis (\cref{chap:chap6}, \cref{chap:chap7}, and \cref{chap:chap8}), reveals instances where the model generates predictions resembling skills. This qualitative analysis indicates that a manual inspection of false positives can be useful for finding new skills or missed annotations and getting realistic estimations of performance.

\section{Transfer to the Job Market Domain}
\subsection{Motivation}
Most contemporary NLP systems are designed and trained with a focus on generic data (e.g., Wikipedia or large-scale web data). A fundamental problem in machine learning is that (self-)supervised learning systems strongly depend on the data they were trained on. Changes in the characteristics of this training data has strong effects on model transferability~\citep{kittredge1986analyzing}. This issue of ``domain divergence'' has prompted comprehensive surveys on how to best adapt language models trained on one domain to more specific targets~\citep{ramponi2020neural, kashyap2021domain, saunders2022domain}, and remains an open issue, even with language models of increasing size~\citep{ling2023domain,singhal2023large,wu2023bloomberggpt}. This is no different from text in the job market domain. In~\cref{chap:chap3}, \cref{chap:chap4}, \cref{chap:chap6}, and \cref{chap:chap7}, we show that models that are tuned on domain-specific data (i.e., job descriptions) outperform models that have been pre-trained on general data. 

\subsection{Multi-task Learning}
Machine learning generally involves training a model to perform a single task. However, a model can benefit by learning from auxiliary, related tasks. Multitask Learning (MTL;~\citealp{caruana1997multitask}) is an inductive transfer mechanism. The goal is to improve generalization performance. MTL improves generalization by leveraging the domain or task-specific information contained in the training signals of related tasks. It does this by training tasks in parallel while sharing (a portion of) its weights. In effect, the training signals for the extra tasks serve as an inductive bias. The seminal work by~\citet{caruana1997multitask} mostly discusses the concept of \emph{hard parameter sharing}: In neural network-based models, it is usually applied by sharing the hidden layers (i.e., the input representations and encoder in~\cref{fig:seq-label-nn}) between all tasks while keeping task-specific output layers~\citep{ruder2017overview}, which we use in this thesis. In~\cref{chap:chap1}, we show such an example of MTL for the task of de-identification in job postings. We combine several datasets from different domains (e.g., medical) and investigate whether multi-task learning is beneficial and improves on the target task of de-identification.

\subsection{Transfer Learning}
In the conventional supervised machine learning framework, the premise is to train a model for a specific task and domain. This entails having labeled data corresponding to the task and domain in question. By using this labeled dataset, we can train a model with the expectation that it will exhibit proficient performance when applied to previously unseen data associated with the same task and domain. Essentially, our expectation is rooted in the assumption that the data is independently and identically distributed. Conversely, when presented with data for an alternative task or domain, we once again need labeled data specific to that task or domain. This new set of labeled data is then used to train a distinct model, with the expectation that it will excel when applied to data of this particular type.

Once there is an insufficient amount of labeled data available for the intended task or domain, this will hinder the training of a reliable model. Transfer learning emerges as a solution to address this limitation, allowing us to tackle such scenarios by capitalizing on data from a related task or domain, termed the source task and source domain. Instead of training the model to perform multiple tasks simultaneously to improve the performance of another related task as in multi-task learning, transfer learning aims to transfer knowledge learned from one task (source task) to improve performance on a different, but related, task (target task). In practical terms, the objective is to transfer as much knowledge as possible from the source context to our target task or domain. The nature of this knowledge can manifest in diverse forms, contingent upon the specific task and data at hand. For example, we have the same task (e.g., skill extraction) in one language (e.g., English) and we investigate the performance of the English-based model in another language, for example, Danish. This is an example of \emph{transductive transfer learning} and specifically \emph{cross-lingual transfer learning}, where the source and target downstream tasks are the same~\citep{blitzer-etal-2007-biographies, arnold2007comparative}, but the source and target domains are different~\citep{ruder2017overview}, i.e., the text is written in a different language. We explore cross-lingual transfer learning in~\cref{chap:chap4} and~\cref{chap:chap6}.

\newpage
\part{Annotating Data}
\newpage

\chapter{{D}e-identification of {P}rivacy {R}elated {E}ntities in {J}ob {P}ostings}
\chaptermark{{JobStack}}
\label{chap:chap1}

\newcommand{\berto}{BERT\textsubscript{Overflow}}
\newcommand{\bertb}{BERT\textsubscript{base}}
\newcommand{\bertl}{BERT\textsubscript{large}}
\newcommand{\std}[2]{#1\textsubscript{$\pm$#2}}

The work presented in this chapter is based on a paper that has been published as: \bibentry{jensen-etal-2021-de}.

\newpage

\section*{Abstract}
De-identification is the task of detecting privacy-related entities in text, such as person names, emails and contact data. It has been well-studied within the medical domain. The need for de-identification technology is increasing, as privacy-preserving data handling is in high demand in many domains. In this paper, we focus on job postings.  We present \jobstack{}, a new corpus for de-identification of personal data in job vacancies on Stackoverflow. We introduce baselines, comparing Long-Short Term Memory (LSTM) and Transformer models. To improve upon these baselines, we experiment with contextualized embeddings and distantly related auxiliary data via multi-task learning. Our results show that auxiliary data improves de-identification performance. Surprisingly, vanilla BERT turned out to be more effective than a BERT model trained on other portions of Stackoverflow.

\section{Introduction}
It is becoming increasingly important to anonymize privacy-related information in text, such as person names and contact details. The task of de-identification is concerned with detecting and anonymizing such information. Traditionally, this problem has been studied in the medical domain by e.g.,~\citet{szarvas2007state,friedrich-etal-2019-adversarial, trienes2020comparing} to anonymize (or pseudo-anonymize) person-identifiable information in electronic health records (EHR).  With new privacy regulations (\cref{sec:relwork1}) de-identification is becoming more important for broader types of text. For example, a company or public institution might seek to de-identify documents before sharing them. On another line, de-identification can benefit society and technology at scale. Particularly auto-regressive models trained on massive text collections pose a potential risk for exposing private or sensitive information~\citep{carlini2019secret, carlini2020extracting}, and de-identification can be one way to address this.

In this paper, we analyze how effective sequence labeling models are in identifying privacy-related entities in job posts. To the best of our knowledge, we are the first study that investigates de-identification methods applied to job vacancies. In particular, we examine: 

\begin{itemize}
    \item How do Transformer-based models compare to LSTM-based models on this task?
    \item How does BERT compare to \berto{}~\citep{tabassum2020code}
    \item To what extent can we use existing medical de-identification data and Named Entity Recognition (NER) data to improve de-identification performance?
\end{itemize}

To answer these questions, we put forth a new corpus, \jobstack{}, annotated with around 22,000 sentences in English job postings from Stackoverflow for person names, contact details, locations, and information about the profession of the job post itself.

\paragraph{Contributions.} We present \jobstack{}, the first job postings dataset with professional and personal entity annotations from  Stackoverflow. Our experiments on entity de-identification with neural methods show that Transformers outperform bi-LSTMs, but surprisingly a BERT variant trained on another portion of Stackoverflow is less effective. We find auxiliary tasks from both news and the medical domain to help boost performance.

\section{Related Work}\label{sec:relwork1}

\subsection{De-identification in the Medical Domain}
De-identification has mostly been investigated in the medical domain (e.g.,~\citet{szarvas2007state, meystre2010automatic, liu2015automatic, jiang2017identification, friedrich-etal-2019-adversarial, trienes2020comparing}) to ensure the privacy of a patient in the analysis of their medical health records. 
Apart from an ethical standpoint, it is also a legal requirement imposed by multiple legislations such as the US Health Insurance Portability and Accountability Act (HIPAA;~\citealp{act1996health}) and the European General Data Protection Regulation (GDPR;~\citealp{regulation2016regulation}). 

Many prior works in the medical domain used the I2B2/UTHealth dataset~\citep{stubbs_annotating_2015} to evaluate de-identification. The dataset consists of clinical narratives, which are free-form medical texts written as a first person account by a clinician. Each of the documents describes a certain event, consultation or hospitalization. All of the texts have been annotated with a set of Protected Health Information (PHI) tags (e.g.\ name, profession, location, age, date, contact, IDs) and subsequently replaced by realistic surrogates.  The dataset was originally developed for use in a shared task for automated de-identification systems. Systems tend to perform very well on this set, in the shared task three out of ten systems achieved F1 scores above 90 \citep{stubbs_annotating_2015}. More recently, systems reach over 98 F1 with neural models~\citep{dernoncourt2017identification, liu2017identification, khin2018deep, trienes2020comparing, johnson2020deidentification}. We took I2B2 as inspiration for annotation of \jobstack{}.

Past methods for de-identification in the medical domain can be categorised in three categories. (1) Rule-based approaches, (2) traditional machine learning (ML)-based systems (e.g., feature-based Conditional Random Fields (CRFs;~\citealp{lafferty2001conditional}), ensemble combining CRF and rules, data augmentation, clustering), and (3) neural-based approaches.

\subsection{Rule-based} First, \citet{gupta2004evaluation} made use of a set of rules, dictionaries, and fuzzy string matching to identify protected health information (PHI). In a similar fashion,~\citet{neamatullah2008automated} used lexical look-up tables, regular expressions, and heuristics to find instances of PHI.

\subsection{Traditional ML} Second, classical ML approaches employ feature-based 
CRFs~\citep{aberdeen2010mitre, he2015crfs}. Moreover, earlier work showed the use of CRFs in an ensemble with rules~\citep{stubbs_annotating_2015}. Other ML approaches include data augmentation by~\citet{mcmurry2013improved}, where they added public medical texts to properly distinguish common medical words and phrases from PHI and trained decision trees on the augmented data.

\subsection{Neural methods} Third, regarding neural methods,~\citet{dernoncourt2017identification} were the first to use Bi-LSTMs, which they used in combination with character-level embeddings. Similarly,~\citet{khin2018deep} performed %
de-identification by using a Bi-LSTM-CRF architecture with ELMo embeddings~\citep{peters2018deep}. 
~\citet{liu2017identification} used four individual methods (CRF-based, Bi-LSTM, Bi-LSTM with features, and rule-based methods) for de-identification, and used an ensemble learning method to combine all PHI instances predicted by the three methods.
~\citet{trienes2020comparing} opted for a Bi-LSTM-CRF as well, but applied it with contextual string embeddings~\citep{akbik2018contextual}. 
Most recently,~\citet{johnson2020deidentification} fine-tuned \bertb{} and \bertl{}~\citep{devlin2019bert} for de-identification. Next to ``vanilla'' BERT, they experiment with fine-tuning different domain specific pre-trained language models, such as SciBERT~\citep{beltagy2019scibert} and BioBERT~\citep{lee2020biobert}. They achieve state-of-the art performance in de-identification on the I2B2 dataset with the fine-tuned \bertl{} model. 
From a different perspective, the approach of ~\citet{friedrich-etal-2019-adversarial} is based on adversarial learning, which automatically pseudo-anonymizes EHRs.

\begin{figure}
    \centering
\noindent\fbox{%
    \parbox{\textwidth}{%
    \small
...\\
13. Job description:\\ 
14. \textbf{\lbrack XXX\textsubscript{Organization}\rbrack} is a modern multi tenant, microservices based solution and Floor Planning is one major functional solution vertical of the \textbf{\lbrack XXX\textsubscript{Organization}\rbrack} platform.\\
15. What you’ll be doing:\\
16. As a \textbf{\lbrack XXX\textsubscript{Profession}\rbrack} for \textbf{\lbrack XXX\textsubscript{Organization}\rbrack}, you will be one of the founding members of our \textbf{\lbrack XXX\textsubscript{Location}\rbrack} based floor planning development team.\\
17. You will be in charge for development of future floor planning capabilities on the \textbf{\lbrack XXX\textsubscript{Organization}\rbrack} platform and be the software architect for the capability.\\
18. You will drive the team to improve the coding practices and boost performance.\\ 
19. You will also be a member of our \textbf{\lbrack XXX\textsubscript{Organization}\rbrack} and have a major influence on feature roadmap and technologies we use.\\
...
        }%
}
    \caption{\textbf{Snippet \jobstack{}.} Snippet of a job posting, full job posting can be found in~\cref{app:A1}.}
    \label{fig:snippet1}
\end{figure}

\subsection{De-identification in other Domains}
Data protection in general however is not only limited to the medical domain. Even though work outside the clinical domain is rare, personal and sensitive data is in abundance in all kinds of data. For example, \citet{eder2019identification} pseudonymized German emails.~\citet{bevendorff2020crawling} published a large preprocessed email corpus, where only the email addresses themselves were anonymized.
Apart from emails, several works went into the de-identification of SMS messages~\citep{treurniet2012collection, patel2013approaches, chen2013creating} in Dutch, French, English, and Mandarin respectively. Both ~\citet{treurniet2012collection, chen2013creating} conducted the same strategy and automatically anonymized all occurrences of dates, times, decimal amounts, and numbers with more than one digit (telephone numbers, bank accounts, et cetera), email addresses, URLs, and IP addresses. All sensitive information was replaced with a placeholder.~\citet{patel2013approaches} introduced a system to anonymize SMS messages by using dictionaries. It uses a dictionary of first names and  anti-dictionaries (of ordinary language and of some forms of SMS writing) to identify the words that require anonymization. 
In our work, we study de-identification for names, contact information, addresses, and professions, as further described in \cref{sec:dataset1}.

\section{\jobstack{}}\label{sec:dataset1}
In this section, we describe the \jobstack{} dataset.
There are two basic approaches to removing privacy-bearing data from job postings.
First, anonymization identifies instances of personal data (e.g. names, email addresses, phone numbers) and replaces these strings with some placeholder (e.g.\ \texttt{\{name\}}, \texttt{\{email\}}, \texttt{\{phone\}}). 
The second approach, pseudonymization preserves the information of personal data by replacing these privacy-bearing strings with randomly chosen alternative strings from the same privacy type (e.g. replacing a name with ``John Doe''). 
The term de-identification subsumes both anonymization and pseudonymization. In this work, we focus on anonymization.\footnote{\citet{meystre2015identification} notes that de-identification means removing or replacing personal identifiers to make it difficult to reestablish a link between the individual and his or her data, but it does not make this link impossible.} 

\citet{eder2019identification} argues that the anonymization approach might be appropriate to eliminate privacy-bearing data in the medical domain, but would be inappropriate for most Natural Language Processing (NLP) applications since crucial discriminative information and contextual clues will be erased by anonymization.

If we shift towards pseudonymization, we argue that there is still the possibility to resurface the original personal data. Henceforth, our goal is to anonymize job postings to the extent that one would not be able to easily identify a company from the job posting.
However, as job postings are public, we are aware that it would be simple to find the original company that posted it with a search engine.
Nevertheless, we abide to GDPR compliance which requires us to protect the personal data and privacy of EU citizens for transactions that occur within EU member states~\citep{regulation2016regulation}. In job postings, this would be the names of employees, and their corresponding contact information.\footnote{\url{https://t.ly/yVjEq}}

Over a period of time, we scraped 2,755 job postings from Stackoverflow and selected 395 documents to annotate, the subset ranges from June 2020 to September 2020.
We manually annotated the job postings with the following five entities: \texttt{Organization}, \texttt{Location}, \texttt{Contact}, \texttt{Name}, and \texttt{Profession}. 

To make the task as realistic as possible, we kept all sentences in the documents. The statistics provided in the following therefore reflect the natural distribution of entities in the data.
A snippet of an example job post can be seen in \cref{fig:snippet1}, the full job posting can be found in \cref{sec:jobpost1}.

\begin{table}[t]
\centering
\begin{tabular}{lrrr|r}
\toprule
 \multicolumn{1}{l}{Timespan Docs.$\rightarrow$}                               & 06--08.2020  & \multicolumn{2}{c|}{09.2020}   &            \\
Statistic$\downarrow$                            & Train             & Dev              & Test  & Total       \\\midrule

\# Documents                & 313                & 41                & 41               & 395                 \\
\# Sentences                & 18,055             & 2082              & 2092             & 22,219              \\
\# Tokens                   & 195,425            & 22,049             & 21,579          & 239,053             \\
\# Entities                 & 4,057              & 462               & 426              & 5,154               \\\midrule

average \# sentences        & 57.68             & 50.78             & 51.02            &  53.16     \\ 
average tokens / sent.      & 10.82             & 10.59             & 10.32            &  10.78     \\ 
average entities / sent.    & 0.22              & 0.22              & 0.20             &  0.21      \\ 
density                     & 14.73             & 14.31             & 14.58            &  14.54     \\\midrule

\texttt{Organization}       & 1803              & 215               & 208               &       2226        \\
\texttt{Location}           & 1511              & 157               & 142               &       1810        \\
\texttt{Profession}         & 558               & 63                & 64                &       685         \\
\texttt{Contact}            & 99               & 10                & 7                  &       116         \\
\texttt{Name}               & 86                & 17                & 5                 &       108         \\
\bottomrule
\end{tabular}
\caption{\textbf{Statistics of \jobstack{}.} We document the statistics of our \jobstack{} dataset. We show the surface-level statistics (e.g., the number of documents). We also show several sentence-level statistics (e.g., average tokens per sentence). Last, we note the number of the entities annotated in our dataset. }
\label{tab:datastats1}
\end{table}

\subsection{Statistics}

\cref{tab:datastats1} shows the statistics of our dataset.
We split our data in 80\% train, 10\% development, and 10\% test. 
Besides of a regular document-level random split, ours is further motivated based on time.
The training set covers the job posts posted between June to August 2020 and the development- and test set are posted in September 2020.
To split the text into sentences, we use the \texttt{sentence-splitter} library used for processing the Europarl corpus \citep{koehn2005europarl}.
In the training set, we see that the average number of sentences is higher than in the development- and test set (6-7 more). We therefore also calculate the density of the entities, meaning the percentage of sentences with at least one entity.
The table shows that 14.5\% of the sentences in \jobstack{} contain at least one entity. Note that albeit having %
document boundaries, we treat the task of de-identification as a standard word-level sequence labeling task.

\subsection{Annotation Schema}\label{sec:schema1}
The aforementioned entity tags are based on the English I2B2/UTHealth corpus \citep{stubbs_annotating_2015}. The tags are more coarse-grained than the I2B2 tags. For example, we do not distinguish between zip code and city, but tag them with \texttt{Location}. We give a brief explanation of the tags.\\
\texttt{Organization}: This includes all companies and their legal entity mentioned in the job postings. The tag is not limited to the company that authored the job posting but does also include mentions of stakeholders or any other company.\\
\texttt{Location}: This is the address of the company in the job posting. The location also refers to all other addresses, zip codes, cities, regions, and countries mentioned throughout the text. This is not limited to the company  address, but should be used for all location names in the job posting, including abbreviations.\\
\texttt{Contact}: The label includes, URLs, email addresses and phone numbers. This could be, but is not limited to, contact info of an employee from the authoring company.\\
\texttt{Name}: This label covers the names of people. This could be, but is not limited  to, a person from the company, such as the contact person, CEO, or manager. All names appearing in the job posting should be annotated no matter the relation to the job posting itself. Titles such as Dr.\ are not part of the annotation. Apart from people names in our domain, difficulties could arise with other types of names. An example would be project names, with which one could identify a company. In this work, we did not annotate such names.\\
\texttt{Profession}: This label covers the profession that is being searched for in the job posting or desired prior relevant jobs for the current profession. We do not annotate additional meta information such as gender (e.g. Software Engineer (f/m)). We also do not annotate mentions of colleague positions in either singular or plural form. For example: ``\textit{As a Software Engineer, you are going to work with Security Engineers}''. Here we annotate Software Engineer as profession, but we do not annotate Security Engineers. While this may sound straightforward, however, there are difficulties in regards to annotating professions. A job posting is free text, meaning that one can write anything they prefer to make the job posting as clear as possible (e.g., \textit{Software Engineer (at a unicorn start-up based in [..]}). The opposite is also possible when they are looking for one applicant to fill in one of multiple positions. For example, ``\textit{We are looking for an applicant to fill in the position of DevOps/Software Engineer}''. From our interpretation, they either want a ``DevOps Engineer'' or a ``Software Engineer''. We decided to annotate the full string of characters ``DevOps/Software Engineer'' as a profession.

\begin{table}[t]
\centering
\begin{tabular}{lrrr}
\toprule
                                        & Token    & Entity    & Unlabeled          \\\midrule
A1 -- A2                                & 0.889    & 0.767     & 0.892             \\
A1 -- A3                                & 0.898    & 0.782     & 0.904             \\
A2 -- A3                                & 0.917    & 0.823     & 0.920             \\\midrule
Fleiss' $\kappa$                        & 0.902    & 0.800     & 0.906             \\\bottomrule
\end{tabular}
\caption{\textbf{Inter-annotator Agreement.} We show agreement over pairs with Cohen's $\kappa$ and all annotators with Fleiss' $\kappa$.}
\label{tab:iaahuman1}
\end{table}

\subsection{Annotation Quality}
To evaluate our annotation guidelines, a sample of the data was annotated by three annotators, one with a background in Linguistics (A1) and two with a background in Computer Science (A2, A3). 
We used an open-source text annotation tool named \texttt{Doccano} \citep{doccano}. 
There are around 1,500 overlapping sentences that we calculated agreement on. 
The annotations were compared using Cohen's $\kappa$~\citep{fleiss1973equivalence} between pairs of annotators, and Fleiss' $\kappa$~\citep{fleiss1971measuring}, which generalizes Cohen's $\kappa$ to more than two concurrent annotations. 
\cref{tab:iaahuman1} shows three levels of $\kappa$ calculations, we follow~\citet{balasuriya2009named}'s approach of calculating agreement in NER. (1) \texttt{Token} is calculated on the token level, comparing the agreement of annotators on each token (including non-entities) in the annotated dataset. 
(2) \texttt{Entity} is calculated on the agreement between named entities alone, excluding agreement in cases where all annotators agreed that a token was not a named entity. 
(3) \texttt{Unlabeled} refers to the agreement between annotators on the exact span match over the surface string, regardless of the type of named entity (i.e., we only check the position of tag without regarding  the type of the named entity).
\citet{landis1977measurement} state that a $\kappa$ value greater than 0.81 indicates almost perfect agreement. 
Given this, all annotators are in strong agreement. 

After this annotation quality estimation, we finalized the guidelines. They formed the basis for the professional linguist annotator, who annotated and finalized the entire final \jobstack{} dataset.

\section{Methods}\label{sec:methods1}

For entity de-identification we use a classic Named Entity Recognition (NER) approach using a Bi-LSTM with a CRF layer. On top of this we evaluate the performance of Transformer-based models with two different pre-trained BERT variants. Furthermore, we evaluate the helpfulness of auxiliary tasks, both using data close to our domain, such as de-identification of medical notes, and more general NER, which covers only a subset of the entities. Further details on the data are given in~\cref{sec:auxtasks1}.

\subsection{Models}

Firstly, we test a Bi-LSTM sequence tagger (Bilty) \citep{plank-etal-2016}, both with and without a CRF layer. The architecture is similar to the widely used models in previous works. For example, preliminary results of Bilty versus~\citet{trienes2020comparing} show accuracy almost identical to each other: 99.62\% versus 99.76\%. Next, we test a Transformer based model, namely the MaChAmp \citep{vandergoot-etal-2020-machamp} toolkit. Current research shows good results for NER using a Transformer model without a CRF layer~\citep{martin-etal-2020-camembert}, hence we tested MaChAmp both with and without a CRF layer for predictions. For both models, we use their default parameters.

\subsection{Embeddings}

For embeddings, we tested with no pre-trained embeddings, pre-trained Glove \citep{pennington2014glove} embeddings, and Transformer-based pre-trained embeddings. For Transformer-based embeddings, we focused our attention on two BERT models, \bertb{}~\citep{devlin2019bert} and \berto{}~\citep{tabassum2020code}. When using the Transformer-based embeddings with the Bi-LSTM, the embeddings were fixed and did not get updated during training. 

Using the MaChAmp \citep{vandergoot-etal-2020-machamp} toolkit, we fine-tune the BERT variant with a Transformer encoder. For the Bi-LSTM sequence tagger, we first derive BERT representations as input to the tagger. The tagger further uses word and character embeddings which are updated during model training.

The \berto{} model is a transformer  with the same architecture as \bertb{}. It has been trained from scratch on a large corpus of text from the Q\&A section of Stackoverflow, making it closer to our text domain than the ``vanilla'' BERT model. However, \berto{} is not trained on the job postings portion of Stackoverflow.

\subsection{Auxiliary tasks}
\label{sec:auxtasks1}
Both the Bi-LSTM~\citep{plank-etal-2016} and the MaChAmp~\citep{vandergoot-etal-2020-machamp} toolkit are capable of multi-task Learning (MTL;~\citealp{caruana1997multitask}). We, therefore, set up a number of experiments testing the impact of three different auxiliary tasks. The auxiliary tasks and their datasets are as follows:

\begin{itemize}
    \item I2B2/UTHealth~\citep{stubbs_annotating_2015} - Medical de-identification;
    \item CoNLL 2003~\citep{sang2003introduction} - News Named Entity Recognition;
    \item The combination of the above.
\end{itemize}

The data of the two tasks are similar to our dataset in two different ways. The I2B2 lies in a different text domain, namely medical notes, however, the label set of the task is close to our label set, as mentioned in \cref{sec:schema1}.
For CoNLL, we have a general corpus of named entities but fewer types (location, organization, person, and miscellaneous), but the text domain is presumably closer to our data. 
We test the impact of using both auxiliary tasks along with our own dataset.

\begin{table*}[t]
\centering
\resizebox{\linewidth}{!}{
\begin{tabular}{lrrr}
\toprule
Model                     & Precision                   & Recall                     & F1-score                   \\ \midrule
Bilty                     & \std{79.00}{1.10}           & \std{65.80}{3.72}          & \std{71.76}{2.57}          \\
Bilty + CRF               & \std{84.09}{1.90}           & \std{67.96}{0.81}          & \std{75.15}{0.66}          \\
Bilty + Glove 50d         & \std{79.21}{2.19}           & \std{67.03}{2.76}          & \std{72.53}{0.83}          \\
Bilty + Glove 50d + CRF   & \std{82.93}{0.87}           & \std{64.93}{3.93}          & \std{72.74}{2.23}          \\
Bilty + \bertb            & \std{83.70}{0.58}           & \std{73.01}{1.34}          & \std{77.99}{0.91}          \\
Bilty + \bertb{} + CRF    & \textbf{\std{88.23}{0.87}}  & \std{73.30}{1.47}          & \std{80.09}{0.60}          \\
Bilty + \berto            & \std{70.86}{0.68}           & \std{41.27}{4.19}          & \std{52.01}{3.15}          \\
Bilty + \berto{} + CRF    & \std{77.79}{1.20}           & \std{40.33}{2.98}          & \std{53.08}{2.88}          \\ \midrule
MaChAmp + \bertb          & \std{86.66}{0.73}           & \std{84.78}{0.44}          & \std{85.70}{0.13}          \\
MaChAmp + \bertb{} + CRF  & \std{86.40}{0.62}           & \textbf{\std{86.15}{0.00}} & \textbf{\std{86.27}{0.31}} \\
MaChAmp + \berto          & \std{70.88}{0.17}           & \std{61.47}{0.81}          & \std{65.84}{0.48}          \\
MaChAmp + \berto{} + CRF  & \std{77.27}{3.68}           & \std{63.06}{2.11}          & \std{69.35}{0.96}          \\ 
\bottomrule
\end{tabular}}
\caption{\textbf{\jobstack{} Development Set Results.} Results on the development set across three runs using our \jobstack{} dataset.}
\label{tab:resultsdev1}
\end{table*}

\begin{table}[t]
\centering
\begin{adjustbox}{width=\textwidth}
\begin{tabular}{llrrr}
\toprule
Model                                                             & Auxiliary tasks              & Precision                     & Recall                       & F1 Score                     \\ \midrule
                                    Bilty + \bertb{} + CRF        & \jobstack{} + CoNLL           & \std{86.91}{1.94}             & \std{77.49}{1.87}            & \std{81.90}{0.32}            \\
                                                                  & \jobstack{} + I2B2            & \std{83.61}{2.61}             & \std{75.18}{2.59}            & \std{79.15}{2.19}            \\
                                                                  & \jobstack{} + CoNLL + I2B2    & \std{84.92}{1.67}             & \std{78.28}{4.34}            & \std{81.37}{2.01}            \\
                                    Bilty + \berto{} + CRF        & \jobstack{} + CoNLL           & \std{79.34}{2.34}             & \std{46.54}{1.99}            & \std{58.62}{1.46}            \\
                                                                  & \jobstack{} + I2B2            & \std{72.03}{6.48}             & \std{46.10}{2.55}            & \std{55.99}{1.93}            \\
                                                                  & \jobstack{} + CoNLL + I2B2    & \std{71.20}{4.80}             & \std{50.86}{3.31}            & \std{59.15}{2.15}            \\\midrule
                                    MaChAmp + \bertb{} + CRF      & \jobstack{} + CoNLL           & \std{87.24}{1.94}             & \textbf{\std{87.23}{1.24}}   & \textbf{\std{87.20}{0.34}}   \\
                                                                  & \jobstack{} + I2B2            & \textbf{\std{88.44}{0.84}}    & \std{84.92}{0.44}            & \std{86.64}{0.53}            \\
                                                                  & \jobstack{} + CoNLL + I2B2    & \std{86.13}{0.50}             & \std{86.00}{0.87}            & \std{86.06}{0.66}            \\
                                    MaChAmp + \berto{} + CRF      & \jobstack{} + CoNLL           & \std{75.65}{1.41}             & \std{66.24}{0.98}            & \std{70.62}{0.64}            \\
                                                                  & \jobstack{} + I2B2            & \std{80.26}{1.32}             & \std{68.47}{1.03}            & \std{73.88}{0.16}            \\
                                                                  & \jobstack{} + CoNLL + I2B2    & \std{77.66}{0.82}             & \std{69.41}{0.89}            & \std{73.29}{0.22}            \\\bottomrule
\end{tabular}
\end{adjustbox}
\caption{\textbf{Multi-task Learning Results.} Performance of multi-task learning on the development set across three runs.}
\label{tab:mtlresults1}
\end{table}

\section{Evaluation}
All results are mean scores across three different runs.\footnote{We sampled three random seeds: $3477689$, $4213916$, $8749520$ which are used for all experiments.} The metrics are all calculated using the \texttt{conlleval} script\footnote{\url{https://www.clips.uantwerpen.be/conll2000/chunking/output.html}} from the original CoNLL-2000 shared task. \cref{tab:resultsdev1} shows the results from training on \jobstack{} only, \cref{tab:mtlresults1} shows the results of the MTL experiments described in \cref{sec:auxtasks1}. Both report results on the development set. Lastly, \cref{tab:testscores1} shows the scores from evaluating selected best models as found on the development set, when tested on the final held-out test set.

\begin{table}[t]
\centering
\resizebox{\linewidth}{!}{
\begin{tabular}{llrrr}
\toprule
Model                                                               & Auxiliary tasks                & Precision                     & Recall                        & F1 Score                    \\ \midrule
Bilty + \bertb{} + CRF                                              & \jobstack{}                     & \textbf{\std{82.44}{0.95}}   & \std{75.90}{1.39}             & \std{78.99}{0.32}           \\ \midrule
MaChAmp + \bertb{} + CRF                                            & \jobstack{}                     & \std{75.92}{0.39}            & \std{84.35}{0.49}             & \std{79.91}{0.38}           \\
                                                                    & \jobstack{} + CoNLL             & \std{77.84}{1.19}            & \std{85.06}{0.91}             & \std{81.27}{0.28}           \\
                                                                    & \jobstack{} + I2B2              & \std{80.30}{0.99}            & \std{83.88}{0.67}             &\textbf{\std{82.05}{0.80}}   \\
                                                                    & \jobstack{} + CoNLL + I2B2      & \std{77.66}{0.58}            & \textbf{\std{85.68}{0.57}}    & \std{81.47}{0.43}           \\ \bottomrule
\end{tabular}%
}
\caption{\textbf{Test Set Results.} Evaluation of the best-performing models on the test set across three runs.}
\label{tab:testscores1}
\end{table}

\subsection{Is a CRF layer necessary?}

In \cref{tab:resultsdev1}, as expected, adding the CRF for the Bi-LSTM clearly helps, and consistently improves precision and thereby F1 score. For the stronger BERT model the overall improvement is smaller and does not necessarily stem from higher precision. We note that on average across the three seed runs, MaChAmp with \bertb{} and no CRF mistakenly adds an I-tag following an O-tag $8$ times out of $426$ gold entities. In contrast, the MaChAmp with \bertb{} and CRF, makes no such mistake in any of its three seed runs. 
Earlier research, such as~\citet{souza2019portuguese} show that BERT models with a CRF layer improve or perform similarly to its simpler variants when comparing the overall F1 scores. Similarly, they note that in most cases it shows higher precision scores but lower recall, as in our results for the development set. However, interestingly, the precision drops during test for the Transformer-based model. As the overall F1 score increases slightly, we use the CRF layer in all subsequent experiments. The main takeaway here is that both models benefit from an added CRF layer for the task, but the Transformer model to a smaller degree.

\subsection{LSTM versus Transformer}
Initially, LSTM networks dominated the de-identification field in the medical domain. Up until recently, large-scale pre-trained language models have been ubiquitous in NLP, although rarely used in this field. On both development and test results (\cref{tab:resultsdev1}, \cref{tab:testscores1}), we show that a Transformer-based model outperforms the LSTM-based approaches with non-contextualized and contextualized representations.

\subsection{Poor performance with \berto{}}
\bertb{} is the best embedding method among all experiments using Bilty, with \berto{} being the worst with a considerable margin. Being able to fine-tune \bertb{} does give a good increase in performance overall. The same trend is apparent with fine-tuning \berto{}, but it is not enough to catch up with \bertb{}. We see that overall MaChAmp with \bertb{} and CRF is the best model. However, Bilty with \bertb{} and CRF does have the best precision. 

We hypothesized the domain-specific \berto{} representations would be beneficial for this task. 
Intuitively, \berto would help with detecting profession entities.
Profession entities contain specific skills related to the IT domain, such as \textit{Python developer}, \textit{Rust developer}, \textit{Scrum master}.
Although the corpus it is trained on is not one-to-one to our vacancy domain, we expected to see at most a slight performance drop. 
This is not the case, as the drop in performance turned out to be high. It is not fully clear to us why this is the case. It could be the Q\&A data it is trained on consists of more informal dialogue than in job postings. In the future, we would like to compare these results to train a BERT model on job postings data.

\subsection{Auxiliary data increases performance}
Looking at the results from the auxiliary experiments in~\cref{tab:mtlresults1} we see that all auxiliary sources are beneficial, for both types of models. 
A closer look reveals that once again MaChAmp with \bertb{} is the best performer across all three auxiliary tasks. Also, we see that Bilty with \bertb{} has good precision, though not the best this time around. For a task like de-identification recall is preferable, thereby showing that fine-tuning BERT is better than the classic Bi-LSTM-CRF. Moreover, we see that \berto{} is under-performing compared to \bertb{}. However, \berto{} is able to get a 4 point increase in F1 with I2B2 as auxiliary task in MaChAmp. For Bilty with \berto{} we see a slightly greater gain with both CoNLL and I2B2 as auxiliary tasks. When comparing the auxiliary data sources to each other, we note that the closer text domain (CoNLL news) is more beneficial than the closer label set (I2B2) from a more distant medical text source. This is consistent for the strongest models.

In general, it can be challenging to train multi-task networks that outperform or even  match their single-task counterparts~\citep{alonso2017multitask, clark2019bam}.~\citet{ruder2017overview} mentions training on a large number of tasks is known to help regularize multi-task models. A related benefit of MTL is the transfer of learned ``knowledge'' between closely related tasks. In our case, it has been beneficial to add auxiliary tasks to improve our performance on both development and test compared to a single-task setting. In particular, it seemed to have helped with pertaining a high recall score.

\subsection{Performance on the test set}
Finally, we evaluate the best performing models on our held out test set. The best models are selected based on their performance on F1, precision, and recall. The results are seen in \cref{tab:testscores1}. Comparing the results to those seen in \cref{tab:resultsdev1} and \cref{tab:mtlresults1} it is clear to see that Bilty with \bertb{} sees a smaller drop in F1 compared to that of MaChAmp with \bertb. We do also see an increase in recall for Bilty compared to its performance on the development set. In general, we see that recall for each model is staying quite stable without any significant drops. It is also interesting to see that, the internal ranking between MTL MaChAmp with \bertb{} has changed, with \jobstack{} + I2B2 being the best-performing model in terms of F1.

\begin{table}[!ht]
\centering
\begin{tabular}{llrr}
\toprule
                                                                     &                           & \multicolumn{2}{c}{MaChAmp + \jobstack{}}               \\
                                     Entity Type$\downarrow$         &                           & + CoNLL                    & + I2B2                     \\ \midrule
                                    \textbf{Organization} (208)      &                      F1   & $77.51 \pm 0.81$           & $78.34 \pm 1.32$           \\
                                                                     &                      P    & $73.73 \pm 1.66$           & $77.86 \pm 1.60$           \\
                                                                     &                      R    & $81.73 \pm 0.96$           & $78.85 \pm 1.74$           \\ \midrule
                                    \textbf{Location} (142)          &                      F1   & $86.88 \pm 1.51$           & $86.67 \pm 1.80$           \\
                                                                     &                      P    & $83.86 \pm 1.82$           & $83.47 \pm 1.19$           \\
                                                                     &                      R    & $90.14 \pm 1.41$           & $90.14 \pm 2.54$           \\ \midrule
                                    \textbf{Profession} (64)         &                      F1   & $80.20 \pm 2.76$           & $83.88 \pm 0.90$           \\
                                                                     &                      P    & $77.44 \pm 3.82$           & $82.42 \pm 0.63$           \\
                                                                     &                      R    & $83.33 \pm 4.51$           & $85.42 \pm 1.80$           \\ \midrule
                                    \textbf{Contact} (7)             &                      F1   & $87.91 \pm 3.81$           & $75.48 \pm 4.30$           \\
                                                                     &                      P    & $90.47 \pm 8.25$           & $71.03 \pm 4.18$           \\
                                                                     &                      R    & $85.71 \pm 0.00$           & $80.95 \pm 8.24$           \\ \midrule
                                    \textbf{Name} (5)                &                      F1   & $86.25 \pm 8.08$           & $85.86 \pm 4.38$           \\
                                                                     &                      P    & $76.39 \pm 12.03$          & $75.40 \pm 6.87$           \\
                                                                     &                      R    & $100.00 \pm 0.00$          & $100.00 \pm 0.00$           \\ \bottomrule
\end{tabular}
\caption{Performance of the two different auxiliary tasks. Reported is the F1, Precision (P), and Recall (R) per entity. The number behind the entity name is the gold label instances in the test set.}
\label{tab:entityanalysis1}
\end{table}

\subsection{Per-entity Analysis} In \cref{tab:entityanalysis1}, we show a deeper analysis on the test set: the performance of the two different auxiliary tasks in a multi-task learning setting, namely CoNLL and I2B2. We hypothesized different performance gains with each auxiliary task. For I2B2, we expected \texttt{Contact} and \texttt{Profession} to do better than CoNLL, since I2B2 contains contact information entities (e.g., phone numbers, emails, et cetera) and professions of patients. Surprisingly, this is not the case for \texttt{Contact}, as CoNLL outperforms I2B2 on all three metrics. We do note however this result could be due to little instances of \texttt{Contact} and \texttt{Name} being present in the gold test set. Additionally, both named entities are predicted six to nine times by both models on all three runs on the test set. This could indicate a strong difference in performance.
For \texttt{Profession}, it shows that I2B2 is beneficial for this particular named entity as expected. For the other three named entities, the performance is similar. As \texttt{Location}, \texttt{Name}, and \texttt{Organization} are in both datasets, we did not expect any difference in performance. The results confirm this intuition.

\section{Conclusions}
In this work, we introduce \jobstack{}, a dataset for de-identification of English Stackoverflow job postings.
Our implementation is publicly available.\footnote{\url{https://github.com/kris927b/JobStack}} The dataset is freely available upon request.
We present neural baselines based on LSTM and Transformer models. 
Our experiments show the following: (1) Transformer-based models consistently outperform Bi-LSTM-CRF-based models that have been standard for de-identification in the medical domain. (2) Stackoverflow-related BERT representations are not more effective than regular BERT representations on Stackoverflow job postings for de-identification. (3) MTL experiments with BERT representations and related auxiliary data sources improve our de-identification results; the auxiliary task trained on the closer text type was the most beneficial, yet results improved with both auxiliary data sources. This shows the benefit of using multi-task learning for de-identification in job vacancy data.

\clearpage
\section{Appendix}\label{app:A1}
\subsection{Example Job Posting}\label{sec:jobpost1}
Below, we show an example snippet of a de-identified job posting, there is a cut-off for brevity.

\noindent\fbox{%
    \parbox{\textwidth}{%
    \small
1. \textbf{\lbrack XXX\textsubscript{Profession}\rbrack}  \\
2. \textbf{\lbrack XXX\textsubscript{Organization}\rbrack} \\
3. $<$ADDRESS$>$, $<$ADDRESS$>$, \textbf{\lbrack XXX\textsubscript{Location}\rbrack} , - , \textbf{\lbrack XXX\textsubscript{Location}\rbrack} \\
$<$cutoff$>$\\

13. Job description:\\ 
14. \textbf{\lbrack XXX\textsubscript{Organization}\rbrack} is a modern multi tenant, microservices based solution and Floor Planning is one major functional solution vertical of the \textbf{\lbrack XXX\textsubscript{Organization}\rbrack} platform. \\

15. What you’ll be doing:\\
16. As a \textbf{\lbrack XXX\textsubscript{Profession}\rbrack} for \textbf{\lbrack XXX\textsubscript{Organization}\rbrack}, you will be one of the founding members of our \textbf{\lbrack XXX\textsubscript{Location}\rbrack} based floor planning development team.\\
17. You will be in charge for development of future floor planning capabilities on the \textbf{\lbrack XXX\textsubscript{Organization}\rbrack} platform and be the software architect for the capability.\\
18. You will drive the team to improve the coding practices and boost performance.\\ 
19. You will also be a member of our \textbf{\lbrack XXX\textsubscript{Organization}\rbrack} and have a major influence on feature roadmap and technologies we use.\\

20. What you’ll bring to the table:\\
$<$cutoff$>$\\
28. Ability to work on-site in our \textbf{\lbrack XXX\textsubscript{Location}\rbrack} office, with flexible remote work possibilities \\
$<$cutoff$>$\\
    }%
}

\chapter{{C}artography {A}ctive {L}earning}
\label{chap:chap2}
The work presented in this chapter is based on a paper that has been published as: \bibentry{zhang-plank-2021-cartography-active}.

\newpage 

\section*{Abstract}
We propose Cartography Active Learning (CAL), a novel Active Learning (AL) algorithm that exploits  the behavior of the model on individual instances \textit{during training} as a proxy to find the most informative instances for labeling. CAL is inspired by data maps, which were recently proposed to derive insights into dataset quality~\citep{swayamdipta-etal-2020-dataset}.
We compare our method on popular text classification tasks to commonly used AL strategies, which instead rely on post-training behavior. We demonstrate that CAL is competitive to other common AL methods, showing that training dynamics derived from small seed data can be successfully used for AL. We provide insights into our new AL method by analyzing batch-level statistics utilizing the data maps. Our results further show that CAL results in a more data-efficient learning strategy, achieving comparable or better results with considerably less training data.

\section{Introduction}
Active Learning (AL) is a widely-used method to tackle the time-consuming and expensive collection and manual labeling of data. 
In recent years,  many AL strategies were proposed. The simplest and most widely used is \emph{uncertainty sampling}~\citep{lewis1994sequential, lewis1994heterogeneous}, where the learner queries instances that it is most uncertain about. Uncertainty sampling is myopic:\ it only measures the information content of a single data instance.
Alternative AL algorithms instead focus on selecting a \emph{diverse} batch~\citep{geifman2017deep, sener2017active, gissin2019discriminative, zhdanov2019diverse} or to estimate the \emph{uncertainty} distribution of the learner~\citep{houlsby2011bayesian, gal2016dropout}. However, these methods are usually limited in their notion of informativeness, which is tied to post-training model uncertainty and batch diversity.  

Recently, \citet{swayamdipta-etal-2020-dataset} introduced \emph{data maps}, 
to visualize the behaviour of the model on individual instances during training (\emph{training dynamics}). The plotted data maps (\cref{fig:trec-agnews}) reveal distinct regions in a dataset:\ groups of \emph{ambiguous} instances useful for high performance and linked to high informativeness, \emph{easy-to-learn} instances which aid optimization, and \emph{hard-to-learn} instances which frequently correspond to mislabeled or erroneous instances.

We propose \emph{Cartography Active Learning} (CAL), which automatically selects the the most informative instances that contribute optimally to model learning. To do so, we leverage a largely ignored source of information:\  insights derived \emph{during training}, i.e., training dynamics derived from \emph{limited} data maps (see~\cref{sec:method3}) to choose informative instances at the boundary of ambiguous and hard-to-learn instances. We hypothesize that this region is where the model will \textit{learn the most} from.
Data maps provide the additional benefit that we can use them to measure informativeness of a batch with straightforward metrics and visualize dataset properties. These distinct regions in the data maps have their own respective statistics. Therefore, as a second research question we investigate whether data map statistics help to assess why some AL algorithms work better than others.

\paragraph{Contributions.} 
In this paper, our contributions are twofold. (1) We present Cartography Active Learning, a novel AL algorithm that exploits data maps for AL. 
We compare our results against other competitive and widely used AL algorithms and outperform them in early AL iterations. (2) Additionally, we leverage the data maps 
to inspect what instances AL methods select. We show that our approach optimally selects informative instances avoiding only \emph{hard-to-learn} and \textit{easy-to-learn} cases, which leads to better AL and comparable or better results than full dataset training. 

\section{Related Work}\label{background}
 AL has seen many usage scenarios in the Natural Language Processing (NLP) field~\citep{shen2017deep,lowell-etal-2019-practical, ein-dor-etal-2020-active, margatina2021active}. The perspective of AL is that if a model is allowed to select the data from which it will \textit{learn the most}, it will achieve comparable (or better) performance with less training instances~\citep{siddhant2018deep}, and at the same time addressing the costly labeling process with a human annotator.

A popular scenario is pool-based active learning~\citep{lewis1994sequential,settles2009active, settles2012active}, which assumes a small set of labeled data $\,\mathcal{L}$ and a large pool of unlabeled data  $\,\mathcal{U}$.
Most AL algorithms start similarly:\ a model is fit to  $\,\mathcal{L}$ to get access to $P_\theta(y\mid \boldsymbol{x})$, then apply a query strategy to get the best scored instance from  $\,\mathcal{U}$, label this instance and add it to  $\,\mathcal{L}$ in an iterative process.

\subsection{Common Strategies} A commonly used query strategy is \textit{uncertainty sampling}~\citep{lewis1994sequential, lewis1994heterogeneous}. In this approach, the learner queries the instances which it is least certain about.
There are two popular approaches. (1) Uncertainty sampling based on entropy~\citep{shannon1948mathematical, dagan1995committee}, it uses the entropy of the label distribution as a measure for the uncertainty of the model on an instance. (2) Uncertainty sampling based on which best labeling is the least confident~\citep{culotta2005reducing}.

\subsection{Batch-mode Active Learning} It is inefficient and time-consuming to obtain sampled queries one by one for annotation in the context of Deep Neural Networks (DNNs).
In a real-world setting, consider having multiple annotators available. One can exploit this setting and label the instances in batches and parallel.
Batch-mode AL allows the learner to query instances in groups. To assemble the optimal batch, one can greedily pick the top-$k$ examples according to an instance-level acquisition function suitable for DNNs. 
There are many works on ways for making neural network posteriors accurately represent the confidence on a given example. One popular example is stochastic regularisation techniques such as dropout during inference time, known as the Monte Carlo Dropout technique~\citep{houlsby2011bayesian}.~\citet{gal2016dropout} refer to this as Bayesian Active Learning by Disagreement (BALD). This allows us to consider the model as a Bayesian neural network and calculate approximations of uncertainty estimates by analyzing its multiple predictions. However, if the information of these top-$k$ examples is similar, this will result in the model not generalizing well over the dataset. Therefore, alternative approaches take the \textit{diversity} of a batch into account.

\begin{figure*}[t]
    \centering
    \includegraphics[width=.45\linewidth]{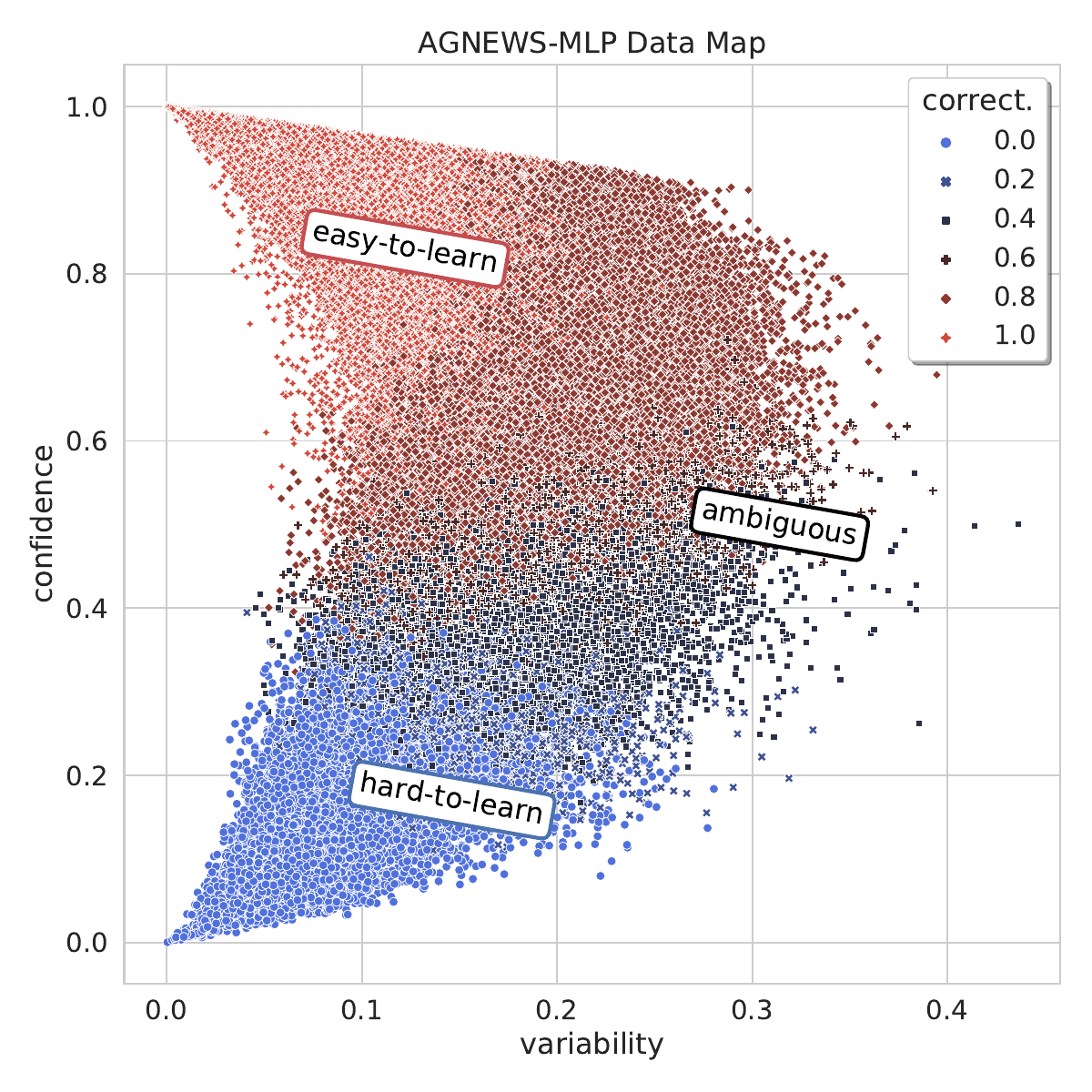}
    \hspace{1em}
    \includegraphics[width=.45\linewidth]{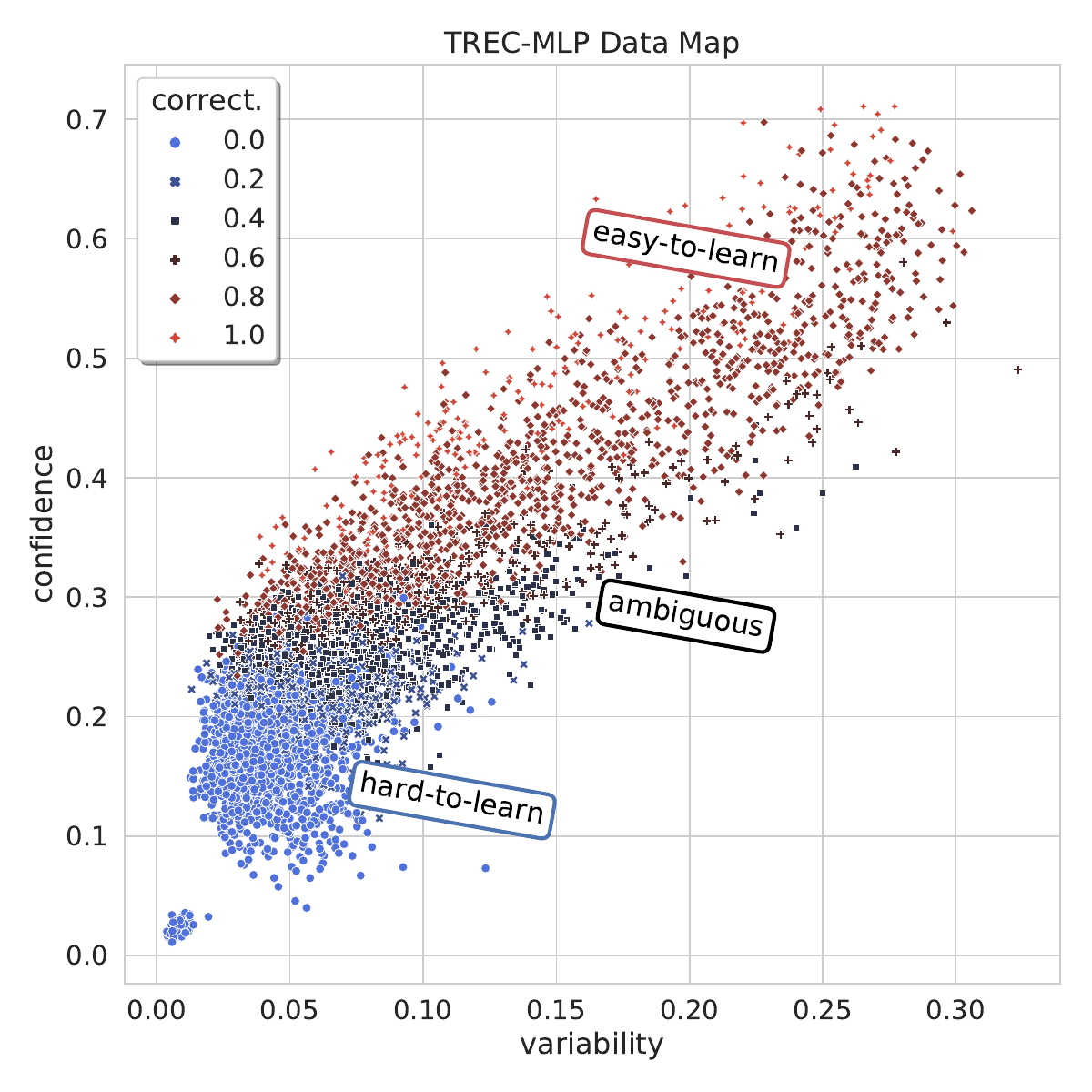}
     \caption{\textbf{Full Data Maps for AGNews \& TREC.} AGNews (120,000 instances) on the left, and  TREC  (5,452 instances) on the right, both w.r.t.\ an MLP training for ten epochs.  The x-axis shows \textbf{variability} and the y-axis the \textbf{confidence}. The colors and shapes indicate the \textbf{correctness}.}
    \label{fig:trec-agnews}
\end{figure*}

\subsection{Batch-aware Query Strategies}
Instead of greedily choosing the examples that maximize some score, one can instead try to find a batch that is as diverse as possible. 
One recently proposed effective strategy is Discriminative Active Learning (DAL;~\citealp{gissin2019discriminative}). This approach aims to select instances from $\,\mathcal{U}$ that make $\,\mathcal{L}$ representative of $\,\mathcal{U}$. In other words, the idea is to train a separate model to classify between $\,\mathcal{L}$ and $\,\mathcal{U}$. Then, to use that model to choose the instances which are most confidently classified as being from $\,\mathcal{U}$. 
If $\,\mathcal{U}$ and $\,\mathcal{L}$ become indistinguishable, the learner has successfully closed the data gap between $\,\mathcal{U}$ and $\,\mathcal{L}$. DAL was proposed for computer vision and was recently successfully used in NLP~\citep{ein-dor-etal-2020-active}. Alternative diversity AL strategies exist, such as core-set, which often rely on heuristics~\citep{sener2017active, geifman2017deep}.\footnote{Contemporary to our work and both appearing at EMNLP 2021,~\citet{margatina2021active} proposed CAL (Contrastive Active Learning), another AL method with the same acronym to our approach. The main difference to Cartography AL is using data points that are similar in the model feature space, while optimizing for maximally disagreeing predictive likelihoods.}

\section{Cartography Active Learning}\label{cal}
The key idea of CAL is to use model-independent measures, from fitting the model on the seed data $\,\mathcal{L}$, by using data maps~\citep{swayamdipta-etal-2020-dataset} for AL. Data maps help identify characteristics of instances within the broader trends of a dataset by leveraging their training dynamics (i.e., the behavior of a model \textit{during training}, such as mean and standard deviation of confidence and correctness with respect to the gold label). These model-dependent measures reveal distinct regions in a data map, by and large, reflecting instance properties (see~\cref{fig:trec-agnews} and details below on \textit{easy-to-learn}, \textit{ambiguous}, and \textit{hard-to-learn} instances). Training dynamics encapsulate information of data quality that has been largely ignored in AL:\ the sweet spot of instances at the boundary of \textit{hard-to-learn} and \textit{ambiguous} instances, which are quick to label while providing informative samples, as shown in full data training~\citep{swayamdipta-etal-2020-dataset}.

In the next part, we introduce training characteristics, first showing the resulting data maps on the full data. Then we introduce CAL, which proposes to learn a data map from the seed labeled data $\,\mathcal{L}$ and identifying regions of instances with a binary classifier, inspired by DAL~\citep{gissin2019discriminative}. To identify these regions, we require:\ (1) a data map can be learned from limited data, and (2) a classifier to identify informative instances. The full algorithm, illustrated in~\cref{algo:cal}, is described later. 

\begin{figure*}[t]
    \centering
    \includegraphics[width=.45\linewidth]{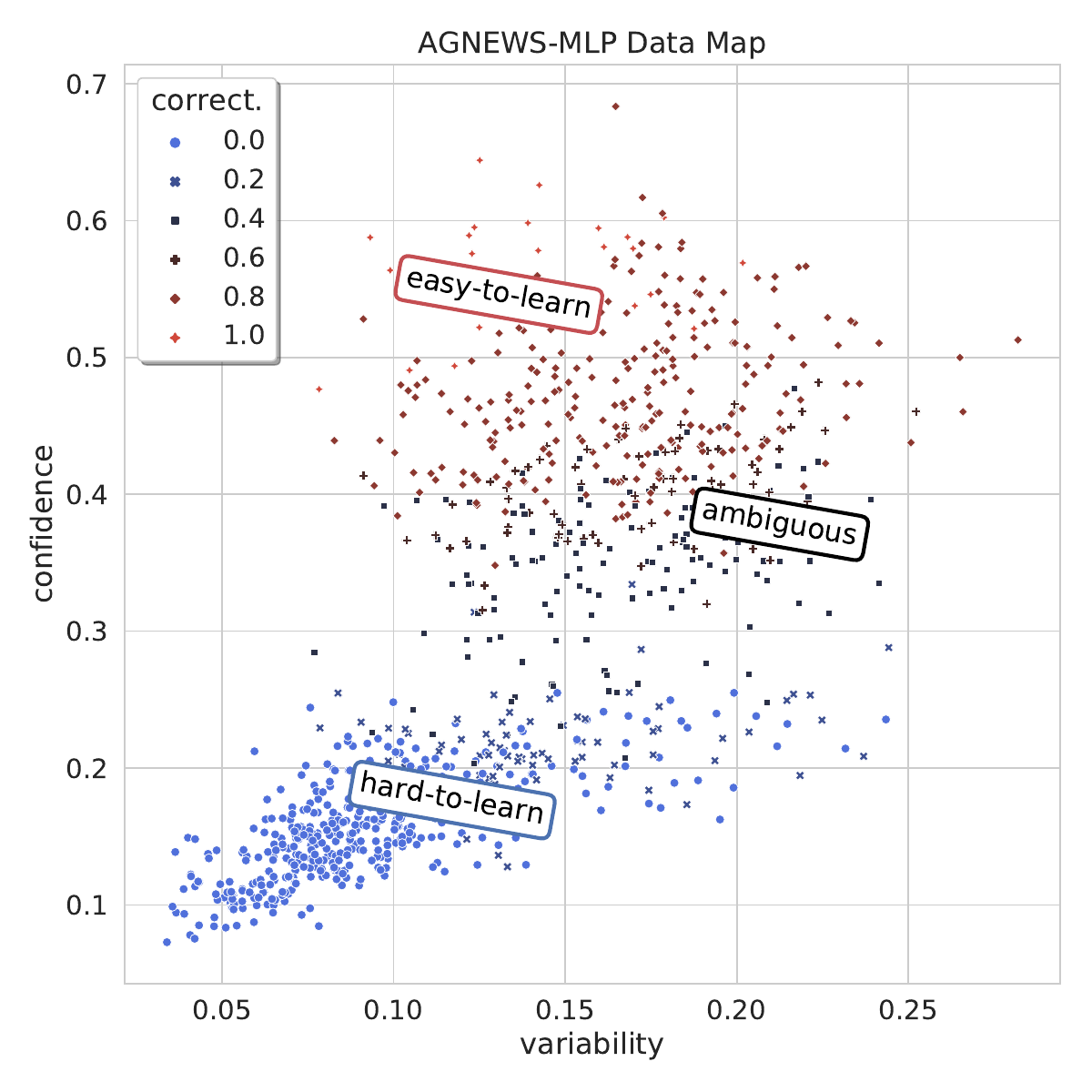}
    \hspace{1em}
    \includegraphics[width=.45\linewidth]{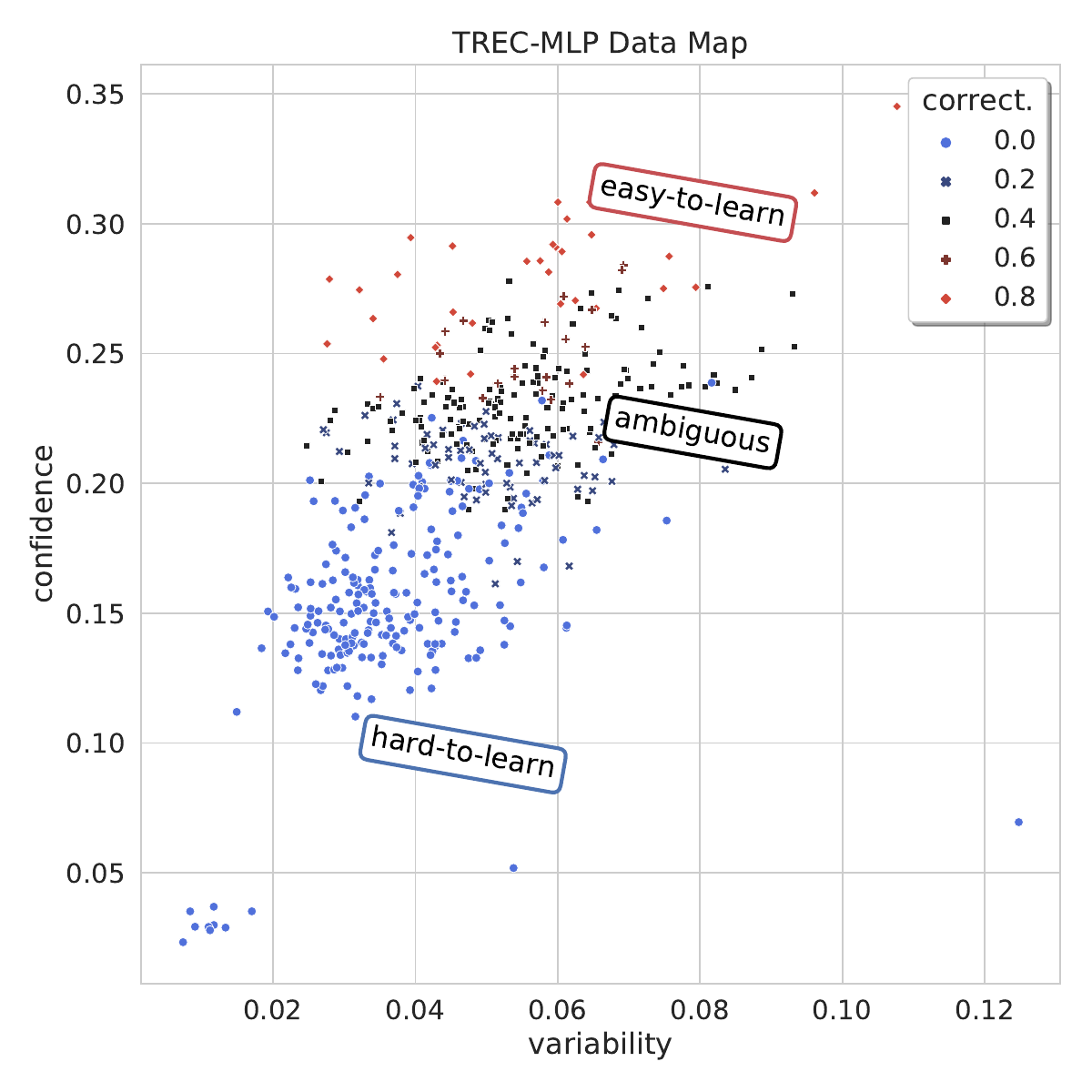}
    \caption{\textbf{Data Maps with Limited Seed Data.} Data map for the AGNews seed set (1,000 instances), and TREC seed set (500 instances). Both data maps are based on an MLP trained for ten epochs.}
    \label{fig:trec-agnews-seed}
\end{figure*}

\subsection{Mapping the Data}
Formally, the training dynamics of instance $i$ are defined as the statistics calculated over $E$ epochs. These statistics are then used as the coordinates in the plot. The following statistics are calculated, \textbf{confidence}, \textbf{variability}, and \textbf{correctness}, following the notation of~\citet{swayamdipta-etal-2020-dataset}:

\begin{equation}{\label{eq:conf}}
    \hat{\mu}_{i}=\frac{1}{E} \sum_{e=1}^{E} p_{\boldsymbol{\theta}^{(e)}}(y_{i}^{*} \mid \boldsymbol{x}_{i})
\end{equation}
\textbf{Confidence}\footnote{Similar to~\citet{swayamdipta-etal-2020-dataset}, we note that the term \textit{confidence} here is the output probability of the model over the gold label as opposed to the certainty of the predicted label as commonly used in AL literature.}~(\autoref{eq:conf}) is the mean model probability of the gold label $(y_{i}^{*})$ across epochs. Where $p_{\boldsymbol{\theta}^{(e)}}$ is the model's probability with parameters $\boldsymbol{\theta}^{(e)}$ at the end of the $e$\textsuperscript{th} epoch.

\begin{equation}{\label{eq:var}}
    \hat{\sigma}_{i}=\sqrt{\frac{\sum_{e=1}^{E}\big(p_{\boldsymbol{\theta}^{(e)}}(y_{i}^{*} \mid \boldsymbol{x}_{i})-\hat{\mu}_{i}\big)^2}{E}}
\end{equation}
Then, \textbf{variability}~(\autoref{eq:var}) is calculated as the standard deviation of $p_{\boldsymbol{\theta}^{(e)}}(y_{i}^{*} \mid \boldsymbol{x}_{i})$, the spread across epochs $E$.

\begin{equation}{\label{eq:corr}}
    \hat{\phi_i} = \frac{1}{E}\sum_{e=1}^{E}\boldsymbol{1}(\hat{y_i} = y_{i}^{*} \mid \boldsymbol{x}_{i})
\end{equation}

Last, \textbf{correctness}~(\autoref{eq:corr}) is denoted as the fraction of times the model correctly labels instance $\boldsymbol{x}_{i}$ across epochs $E$.

Given the aforementioned training dynamics and the obtained statistics per instance, we plot the data maps for both AGNews~\citep{Zhang2015CharacterlevelCN} and TREC~\citep{li-roth-2002-learning}, using all training data (\cref{fig:trec-agnews}). The data map is based on a Multi-layer Perceptron (MLP).
As shown by~\citet{swayamdipta-etal-2020-dataset}, data maps identify three distinct regions:\ \textit{easy-to-learn}, \textit{ambiguous}, and \textit{hard-to-learn}. The \textit{easy-to-learn} instances are consistently predicted correctly with high confidence, these instances can be found in the upper region of the plot. The \textit{ambiguous} samples have high variability and the model is inconsistent in predicting these correctly (middle region). The instances that are (almost) never predicted correctly, and have low confidence and variability, are referred to as \textit{hard-to-learn} cases. This confirms findings by
\citet{swayamdipta-etal-2020-dataset} where they show that training on the samples of these distinct regions, and in particular the \textit{ambiguous} instances, promote optimal performance. 
While uncertainty-based AL mostly focus on hard-cases, CAL instead focuses on ambiguous and possibly easier instances.

\subsection{Data Maps from Seed Data}
Given the distinct regions for data selection in the full data map in~\cref{fig:trec-agnews}, we first investigate whether regions are still identifiable if we have little amounts of training data, as this is a prerequisite for CAL.~\cref{fig:trec-agnews-seed} shows this for 1,000 training samples of AGNews and 500 of TREC. We can see that the data points are more scattered, where the \textit{easy-to-learn} and \textit{ambiguous} samples are mixed. However, it seems the \textit{hard-to-learn} region can still qualitatively be distinguished from the other regions.

\subsection{CAL Algorithm} The algorithm is detailed in~\cref{algo:cal} and described next. To select presumably informative instances, we train a binary classifier on the seed set $\,\mathcal{L}$ and apply it to $\,\mathcal{U}$ to select the instances that are the closest to the decision boundary between \textit{ambiguous} and \textit{hard-to-learn} instances. By visualizing a decision function in~\cref{fig:trec-agnews-seed} that separates the \textit{hard-to-learn} region from the \textit{ambiguous}/\textit{easy-to-learn} region, we select the instances that are the closest to this boundary. In other words, selecting instances with output probability 0.5 with respect to the binary classifier. This does two things, (1) it prevents the binary classifier from selecting only \textit{easy-to-learn} instances (low-variability, high-confidence), and (2) selecting some truly \textit{hard-to-learn} instances (low-variability, low-confidence) which are both not optimal for learning.

Similar to DAL, for the binary classification task, we map our original input space $\,\mathcal{X}$ to the learned representation~$\Psi$ of the last hidden layer of an MLP (\cref{configurations}). These are the features used in our binary classifier ($\theta'$). Formally, as we have three hidden layers, \[\Psi : \mathcal{X} \rightarrow \mathcal{\hat{X}},
\text{where } \Psi = \mathbf{h}_3 = f( W_3 \cdot \mathbf{h}_2 + \mathbf{b}_3).\]
For label space $\mathcal{Y}$, we consider the binary values $\{0,1\}$. This label depends on the \textit{correctness}.
The label $y_{\Psi(\hat{\boldsymbol{x}}_i)}$ for the learned representation of instance ${\hat{\boldsymbol{x}}_i}$ is labeled 1 when the \textit{correctness} using the limited data map at epoch $E$ is above the threshold $t_{cor} > 0.2$. We refer to these as \textbf{high-cor} cases. The samples that are rarely correct ($t_{cor}$ $\leq 0.2$) are labeled as 0 and we refer to these as \textbf{low-cor} cases. To give a better intuition, we refer to~\cref{fig:trec-agnews-seed}, where the regions of \textit{hard-to-learn} and \textit{ambiguous/easy-to-learn} are visually separable with this \textit{correctness} threshold $t_{cor} =$ 0.2. This threshold is empirically chosen by investigating the influence of different \textit{correctness} thresholds on the performance of CAL in~\cref{analysis:threshold} (\cref{tab:threshold}).

\begin{algorithm}[t]
\SetAlgoLined
\textbf{input:} Labeled seed set $\mathcal{L}$, Unlabeled set $\mathcal{U}$, Total budget $K$, Number of queries $n$, Correctness threshold $t_{cor}=0.2$\;
\For{\text{i = 1, ..., n}}{
    $\Psi(\mathcal{L})$, $\Psi(\mathcal{U}) \leftarrow$ train main classifier $\theta$ on $\mathcal{L}$, get representations of $\mathcal{L}$ and $\mathcal{U}$\;
    $\hat{\mu}$, $\hat{\sigma}$, $\hat{\phi} \leftarrow$ get data map statistics of $\mathcal{L}$ with $\theta$\;
    $P_{\theta^\prime} \leftarrow$ train binary classifier $\theta^\prime$ on $\Psi(\mathcal{L})$ with $y_{\Psi(\hat{\boldsymbol{x}}_i)} = \begin{cases}
        1, & \text{if } \hat{\phi_i} > t_{cor}\\
        0, & \text{else}
    \end{cases}$\;
    \For{\text{j = 1, ..., $\frac{K}{n}$}}{
        $\hat{\boldsymbol{x}} \leftarrow \underset{\boldsymbol{x} \in \Psi(\mathcal{U})}{\mathrm{argmin}} \, \lvert 0.5 - P_{\theta^\prime}(\hat{y} = 1 \mid \boldsymbol{x}) \rvert$\;
        $\mathcal{L} \leftarrow \mathcal{L} \cup \hat{\boldsymbol{x}}$\;
        $\mathcal{U} \leftarrow \mathcal{U} \backslash \hat{\boldsymbol{x}}$\;
    }
    reset parameters $\theta$ and $\theta^\prime$\;
}
\textbf{return} $\mathcal{L}$, $\mathcal{U}$\;
\caption{Cartography Active Learning}
\label{algo:cal}
\end{algorithm}

\section{Experimental Setup}\label{sec:method3}
We focus on pool-based active learning.
Once trained on a seed set $\,\mathcal{L}$, we begin the simulated AL loop by iteratively selecting instances based on the scoring of an acquisition function. We take the top-50 instances, following prior work~\citep{gissin2019discriminative,ein-dor-etal-2020-active}. The selected instances are shown the withheld label and added to the labeled set $\,\mathcal{L}$ and removed from $\,\mathcal{U}$. We evaluate the performance of the trained model on a predefined held-out test set.
We run 30 AL iterations, over five random seeds, and report averages over these runs.

\subsection{Datasets}\label{sec:data3}
\begin{table}[t]
    \centering
    \begin{tabular}{l|rrrr}
    \toprule
        Dataset & Train     & Test      & Classes   & Seed set size \\ \midrule
        AGNews  & 120,000   & 7,600     & 4         & 1,000\\
        TREC    & 5,452     & 500       & 6         & 500\\\bottomrule
    \end{tabular}
    \caption{\textbf{Datasets.} Statistics of the two datasets.}
    \label{tab:datastat}
\end{table}
\noindent
In our AL setup, we consider two popular text classification tasks, namely \textbf{AGNews}~\citep{Zhang2015CharacterlevelCN} and \textbf{TREC}~\citep{li-roth-2002-learning}.\ The AGNews task entails classifying news articles into four classes:\ world, sports, business, science/technology. For TREC, the task is to categorize questions into one of six categories based on the subject of the question, such as questions about locations, persons, concepts, et cetera.
Statistics of the data can be found in~\cref{tab:datastat}.
We start with a seed set size of 1,000 for AGNews and 500 for TREC, both are stratified. This means after the AL iterations we will have 2,500 labeled instances for AGNews and 2,000 for TREC. Our motivation here is to keep the AL simulation realistic. We assume enough annotation budget to initially annotate 500--1,000 samples. Then, in every AL iteration annotate an additional 50 samples, which seems manageable for an annotator. Finally, we run 30 AL iterations to give a good overview of the performance of the acquisition functions over the iterations towards convergence.

\subsection{Acquisition Functions}\label{subsec:al}
We consider five acquisition functions.
We opt for a random sampling baseline (\textbf{Rand.}), four existing acquisition functions, and our proposed CAL algorithm. We chose these as they are state-of-the-art and cover a spectrum of acquisition functions (uncertainty, batch-mode and diversity-based). 

\paragraph{Least Confidence (LC;~\citealp{culotta2005reducing}).} It takes 
\[\underset{\boldsymbol{x}\in \mathcal{U}}{\mathrm{argmax}}\, 1 - P_\theta(\hat{y}\mid \boldsymbol{x})\] 
of the predictive (e.g.\ softmax) distribution as the model's uncertainty, and chooses instances with lowest predicted probability.

\paragraph{Max-Entropy (Ent.;~\citealp{dagan1995committee}).} Another popular example is entropy based sampling. Instances are selected according to the function
\[\underset{\boldsymbol{x}\in \mathcal{U}}{\mathrm{argmax}}\,- \sum_{y \in\mathcal{Y}} P_\theta(y\mid \boldsymbol{x}) \log_2 \big(P_\theta(y\mid \boldsymbol{x})\big)\]
and again, based on the \textit{a posteriori} probability distribution.

\paragraph{Bayesian Active Learning by Disagreement (BALD;~\citealp{houlsby2011bayesian, gal2016dropout}).} This approach entails applying dropout at test time, then estimating uncertainty as the disagreement between outputs realized via multiple passes through the model. We use the Monte Carlo Dropout technique on ten inference cycles, with the max-entropy acquisition function.

\paragraph{Discriminative Active Learning (DAL;~\citealp{gissin2019discriminative}).} This approach poses AL as a binary classification task, it uses a separate binary classifier as a proxy to select instances that make $\mathcal{L}$ representative of the entire dataset (i.e., making the labeled set indistinguishable from the unlabeled pool set). The input space for the binary classifier is task-agnostic. One maps the original input space $\mathcal{X}$ to a learned representation $\hat{\mathcal{X}}$ as the input space, with label space $\,\mathcal{Y} = \{l,u\}$ referring to labeled and unlabeled. In the original paper, the learned representation is defined as the logits of the last hidden layer of the main classifier which solves the original task.
Formally, it selects the top-$k$ instances that satisfy
\[\underset{\boldsymbol{x}\in \mathcal{U}}{\mathrm{argmax}}\, \hat{P_\theta}(\hat{y} = u\mid \Psi(\boldsymbol{x}))\]
where $\hat{P_\theta}$ is the trained binary classifier given the learned representations $\Psi$ of instances $\boldsymbol{x}$. 

\subsection{Configurations}\label{configurations}
This work uses two models for the AL setup solving the classification tasks. All the code is open source and available to reproduce our results.\footnote{\url{https://github.com/jjzha/cal}}

\paragraph{Main Classifier}
We use a Multi-layer Perceptron (MLP), with three $d_{\text{emb}}= $ 300 ReLu layers, dropout probability $p = 0.3$, minimize cross-entropy,
AdamW optimizer~\citep{loshchilov2017decoupled}, with a learning rate of $1\mathrm{e}{-4} \text{, } \beta_1 = 0.9 \text{, } \beta_2 = 0.999 \text{, } \epsilon  = 1\mathrm{e}{-8}$.

\paragraph{Binary Classifier} This model is suited for the binary classification task of DAL and CAL. In this case it is a single $d_{\text{emb}}= $ 300 ReLu layer. We minimize the cross-entropy as well. 
We use the AdamW optimizer with a learning rate of $5\mathrm{e}{-5}$.
For more details regarding reproducibility, we refer to~\cref{reproducibility}. 

\paragraph{Training the Binary Classifier} 
For both the binary classification task of DAL and CAL,
we empirically determined that setting the number of epochs for the binary classifier to 30 yielded good results. In the context of DAL, for AGNews it reaches around 98\% accuracy during training, and for TREC it achieves around 89\% accuracy.
In contrast, with CAL, we start with little amounts of data for a binary classifier to train on. In the early stage of the AL iterations, the binary classifier does not achieve a high accuracy for both AGNews and TREC (around random). After it reaches the fifth or sixth AL iteration it starts to properly distinguish the \textbf{low-cor}/\textbf{hig-cor} samples, as it probably has enough samples to learn from. It achieves around 65--75\% accuracy for AGNews, and towards 85\% accuracy for TREC. The classification accuracy on AGNews seems low. However, further tuning of the binary classifier (e.g., increasing the number of epochs) slightly increases binary classification accuracy, but did not result in better performance for the overall AL setup.

\paragraph{Significance}
Recently, the Almost Stochastic Order test~\citep[ASO;\  ][]{dror2019deep}\footnote{Implementation of \citet{dror2019deep} can be found at~\url{https://github.com/Kaleidophon/deep-significance}~\citep{dennis_ulmer_2021_4638709}} has been proposed to test statistical significance for DNNs over multiple runs.
Generally, the ASO test determines whether a stochastic order~\citep{reimers2018comparing} exists between two models or algorithms based on their respective sets of evaluation scores. Given the single model scores over multiple random seeds of two algorithms $A$ and $B$, the method computes a test-specific value ($\epsilon_{min}$) that indicates how far algorithm $A$ is from being significantly better than algorithm $B$. When distance $\epsilon_{min} = 0.0$, one can claim that $A$ stochastically dominant over $B$ with a predefined significance level. When $\epsilon_{min} < 0.5$ one can say $A \succeq B$. On the contrary, when we have $\epsilon_{min} = 1.0$, this means $B \succeq A$. For $\epsilon_{min} = 0.5$, no order can be determined. We took 0.05 for the predefined significance level $\alpha$ and measure it across all 30 AL iterations.

\begin{figure*}[ht]
    \centering
    \includegraphics[width=.49\linewidth]{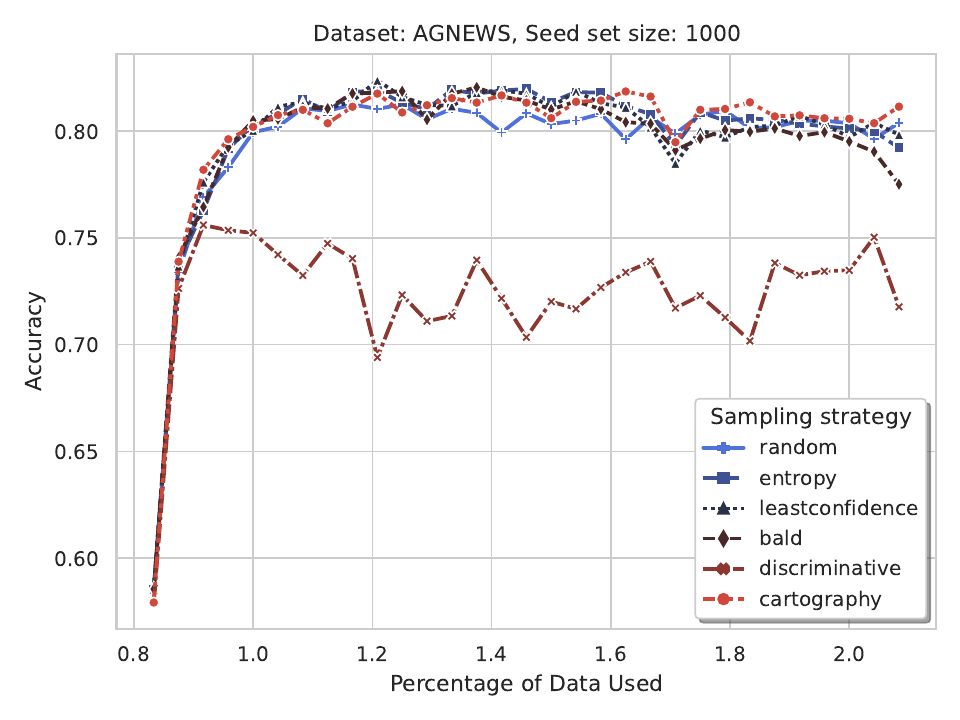}
    \includegraphics[width=.49\linewidth]{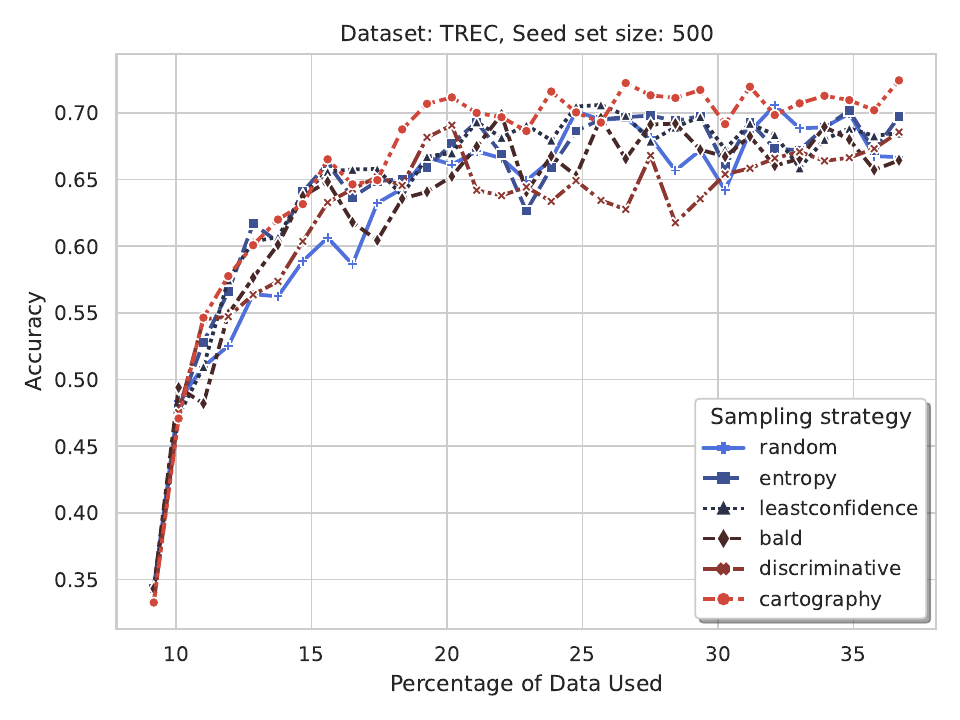}
    \caption{\textbf{Performance AL strategies.} Performance of the various AL strategies in terms of accuracy. The accuracy shown over the AL iterations is the average over five random seeds. Note that for both datasets we added the same number of instances to the seed set (+1,500 instances). The x-axis correspond to the fraction of the total size of the respective dataset. }
    \label{fig:trec-agnews-result}
\end{figure*}
\begin{table*}[ht]
    \centering
    \resizebox{.44\linewidth}{!}{
    \begin{tabular}{l|l|l|l|l|l|l}
    \toprule
    & \rotatebox[origin=c]{90}{Rand.} & \rotatebox[origin=c]{90}{LC} & \rotatebox[origin=c]{90}{Ent.} & \rotatebox[origin=c]{90}{BALD} & \rotatebox[origin=c]{90}{DAL} & \rotatebox[origin=c]{90}{CAL} \\
    \midrule
    Rand.     &  \cellcolor{gray}  &          1.00    & 1.00                & 1.00             & \textit{0.01}         & 1.00\\\hline
    LC        &  \textbf{0.00}     & \cellcolor{gray} & 1.00                & 1.00             & \textit{0.01}         & 1.00\\\hline
    Ent.      &  \textbf{0.00}     & \textbf{0.00}    & \cellcolor{gray}    & 1.00             & \textit{0.01}         & 1.00\\\hline
    BALD      &  \textbf{0.00}     & \textbf{0.00}    & 0.00                & \cellcolor{gray} & \textit{0.01}         & 1.00\\\hline
    DAL       &  0.99              & 0.99             & 0.99                & 0.99             & \cellcolor{gray}      & 1.00\\\hline
    CAL       &  \textbf{0.00}     & \textbf{0.00}    & \textbf{0.00}       & \textbf{0.00}    & \textbf{0.00}         & \cellcolor{gray}\\
    \bottomrule
    \end{tabular}}
    \hspace{2em}
    \resizebox{.44\linewidth}{!}{
    \begin{tabular}{l|l|l|l|l|l|l}
    \toprule
    & \rotatebox[origin=c]{90}{Rand.} & \rotatebox[origin=c]{90}{LC} & \rotatebox[origin=c]{90}{Ent.} & \rotatebox[origin=c]{90}{BALD} & \rotatebox[origin=c]{90}{DAL} & \rotatebox[origin=c]{90}{CAL} \\
    \midrule
    Rand.     &  \cellcolor{gray}& 1.00             & 1.00              & 1.00              & 1.00                  & 1.00\\\hline
    LC        &  \textbf{0.00}   & \cellcolor{gray} & 1.00              & 0.78              & 0.57                  & 1.00  \\\hline
    Ent.      &  \textbf{0.00}   & \textbf{0.00}    & \cellcolor{gray}  & 0.87              & 0.60                  & 1.00  \\\hline
    BALD      &  \textbf{0.00}   & \textit{0.22}    & \textit{0.13}     & \cellcolor{gray}  & 1.00                  & 1.00\\\hline
    DAL       &  \textbf{0.00}   & \textit{0.43}    & \textit{0.40}     & 1.00              & \cellcolor{gray}      & 1.00\\\hline
    CAL       &  \textbf{0.00}   & \textbf{0.00}    & \textbf{0.00}     & \textbf{0.00}     & \textbf{0.00}         & \cellcolor{gray}\\
    \bottomrule
    \end{tabular}}
    \caption{\textbf{Almost Stochastic Order Scores of AGNews (left) \& TREC (right).} ASO scores expressed in $\epsilon_{min}$ 
    The significance level $\alpha =$ 0.05 is adjusted accordingly by using the Bonferroni correction~\citep{bonferroni1936teoria}. \textbf{Bold} numbers indicate stochastic dominance and \textit{cursive} means that one algorithm is better than the other, e.g., for AGNews, LC (row) is stochastically dominant over the random baseline (column) with $\epsilon_{min}$ value of 0.00). Numbers are obtained across all 30 AL iterations}
    \label{tab:significance}
\end{table*}

\begin{figure*}[ht]
    \centering
    \includegraphics[width=\linewidth]{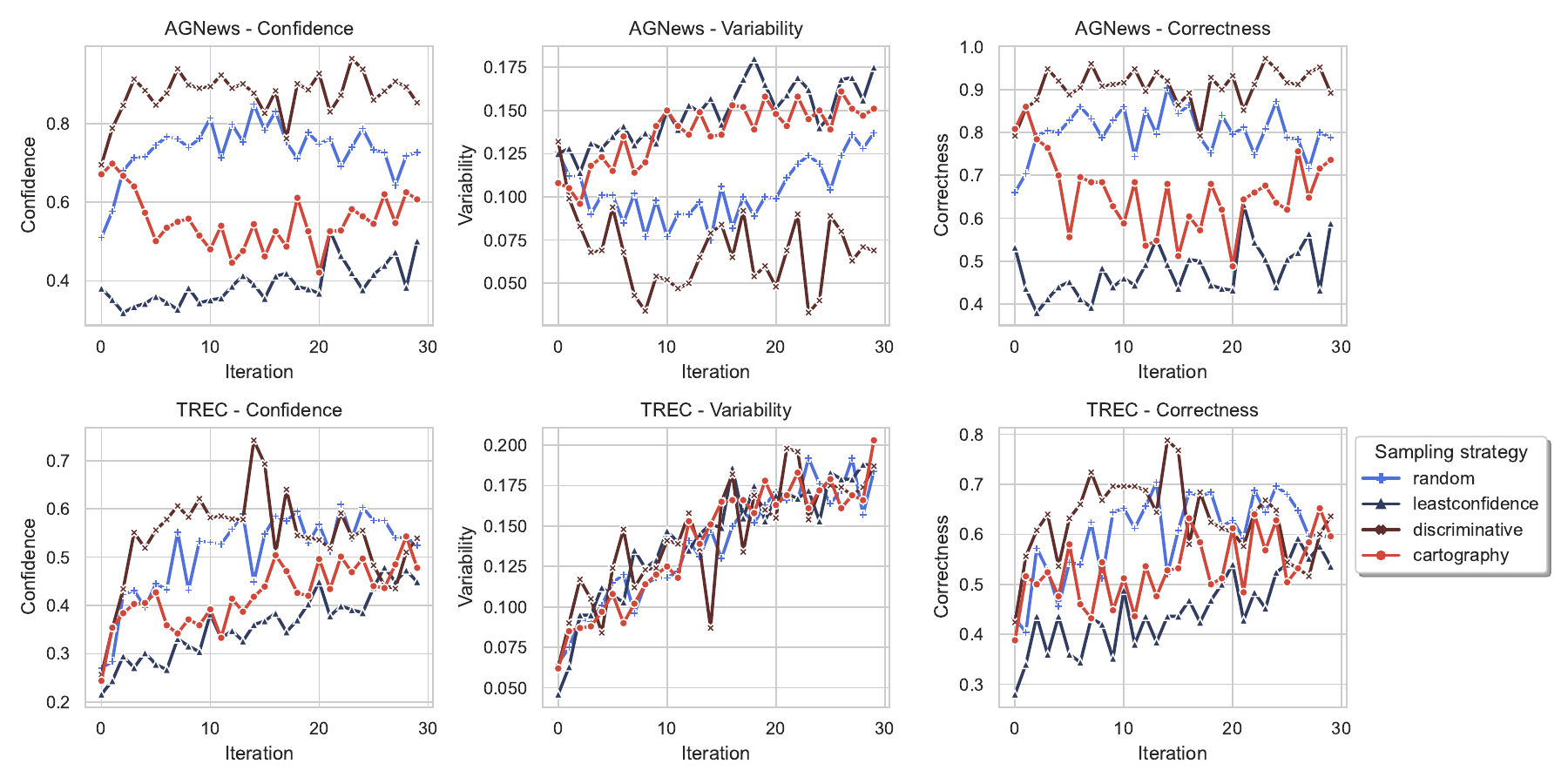}
    \caption{\textbf{Average Statistics AGNews \& TREC.}~Values correspond to the mean statistics~(\cref{cal}) of instances over the five random seeds. We calculate the statistics of each top-50 batch \textit{after} being added to the seed set. Therefore, we only have statistics of 29 runs, as in the 30th run we stop the AL cycle.}
    \label{fig:statistics-agnews-trec}
\end{figure*}

\section{Results \& Analysis}\label{sec:results}
We plot the accuracy of the AL algorithms~(\cref{subsec:al}) on each dataset in~\cref{fig:trec-agnews-result}. For AGNews and TREC, all AL strategies except DAL outperform the random baseline. CAL is statistically dominant over BALD (AGNews) and DAL (AGNews, TREC), and competitive with LC and Entropy (\cref{tab:significance}). This shows that CAL reaches strong results. CAL (illustrated as \texttt{cartography} in the figure with a red-dotted line).

\subsection{Why does CAL work?}
To gain insight on why CAL works better, we investigate the average statistics of each selected batch of samples using the data maps. 
In~\cref{fig:statistics-agnews-trec}, we check the mean confidence, variability and correctness over each selected batch of 50 for sampling strategies Random, LC, DAL, and CAL for both AGNews and TREC. 
The statistics of the instances are extracted after the selected top--50 batch is added to the seed set. Once the main model is trained again on the increased seed set, we obtain the statistics of the previously added batch of 50.~\cref{fig:statistics-agnews-trec} shows that \texttt{leastconfidence} and \texttt{cartography}  has the higher variability amongst the other methods (middle graph). There is a similar signal as the findings of~\citet{swayamdipta-etal-2020-dataset}, the \textit{ambiguous} samples that the model \textit{learns the most} from are usually the instances that have the highest variability and average confidence---as we can see in~\cref{fig:statistics-agnews-trec} (especially pronounced on AGNews), these are the instances CAL selects.
In contrast, LC selects instances that have relatively low confidence and low variability in the early stages, but catches up in later ones. This suggests that LC chooses only \textit{hard-to-learn} instances at the start. In general, CAL seems to select more \textit{informative} samples as opposed to the random strategy.

DAL seems to start by choosing \textit{ambiguous} samples with high variability. However, it quickly picks mostly \textbf{high-cor} samples in later iterations, in contrast to CAL. Consequently, the performance for DAL drops as it leads to picking the \textit{easy-to-learn} samples. Picking too many \textit{easy-to-learn} instances results in worse optimization~\citep{swayamdipta-etal-2020-dataset}. The drop for DAL is visible in Figure~\ref{fig:trec-agnews-result}, which shows that CAL and DAL are close at start, and the accuracy of DAL then drops. 

\subsection{What is the influence of the \textit{correctness} threshold?}

\begin{table}[htpb]
    \centering
    \begin{tabular}{l|rrrrr}
        \toprule
        $t_{cor}$      & $0.0$    & $\mathbf{\leq0.2}$ & $\leq0.4$ & $\leq0.6$ & $\leq0.8$\\
               \midrule
        AGNews & 0.789 & \textbf{0.811}    & 0.796    & 0.788    & 0.781\\
        TREC   & 0.676 & \textbf{0.724}    & 0.701    & 0.706    & --\\
        \bottomrule
    \end{tabular}
    \caption{\textbf{Influence of the Correctness Threshold.} 
    Influence of the correctness threshold $t_{cor}$ on the final average accuracy score per dataset over the five random seeds. $t_{cor}$ considers from what boundary we should consider \textbf{low-cor} cases. In \textbf{bold} is what threshold we are using.}
    \label{tab:threshold}
\end{table}

Here we investigate how changing the correctness threshold for the binary classification task impacts accuracy. As shown in~\cref{tab:threshold}, the selected threshold worked well for the two datasets studied. In particular, the accuracy does not drop substantially on AGNews if we move the \textit{correctness} threshold to a higher value.~\citet{swayamdipta-etal-2020-dataset} indicates that the \textit{ambiguous} region contains the instances with the highest variability. For AGNews, these instances have a \textit{correctness} range from 0.2--0.8 (as seen in~\cref{fig:trec-agnews}). Therefore, we assume that most of these instances are informative for the model.
In the case of TREC, the final accuracy also stays similar with different \textit{correctness} thresholds and does not drop substantially. We find in the full data map for TREC that the area with $t_{cor} = \{0.0, 0.4\}$ thresholds is more dense compared to $t_{cor} =$ 0.2 (details in~\cref{stats:full}). In other words, there are more samples, but also ones that are detrimental to performance. We note that the \textit{correctness} threshold could vary for different datasets and models.

\begin{table}
    \centering
\begin{tabular}{l|rr}
\toprule
                            & AGNews    & TREC\\
\midrule
CAL $\cap$ DAL              &  4                &  114 \\
CAL $\cap$ LC               &  2                &  101 \\
CAL $\cap$ Rand.            &  7                &  82 \\
\midrule
DAL $\cap$ LC               &  3                &  81 \\
DAL $\cap$ Rand.            &  2                &  87 \\
LC  $\cap$ Rand.            &  2                &  100\\
\bottomrule
\end{tabular}
    \caption{\textbf{Number of Overlapping Instances on AGNews \& TREC.}~The values corresponds to the total overlapping instances out of 1,500 $*$ 5 random seeds $=$ 7,500.}
    \label{tab:overlap_agnews_trec}
\end{table}

\subsection{Do AL strategies select the same instances for labeling?}\label{analysis:threshold}
We measured the overlap between the batches selected by each pair of strategies (Random, LC, DAL, CAL) on AGNews and TREC (\cref{tab:overlap_agnews_trec}). The batch overlap for CAL is low, with the highest overlap being 7 instances for AGNews with CAL $\cap$ Rand.\ followed by CAL $\cap$ DAL with 4 overlapping instances. The highest overlap for TREC is 114 instances for CAL $\cap$ DAL. Note this is the total overlap over five seeds. These results indicate that the AL algorithms choose different instances.

A popular approach for improving classification performance is combining (complementary) AL strategies. For example,~\citet{zhdanov2019diverse} proposed the idea to combine uncertainty sampling and diversity sampling for image classification.
As DAL and CAL have few overlapping instances, they can be complementary to each other. To test this, we combined DAL and CAL using a simple heuristic, by providing both of them half of the annotation budget (i.e., take top-25 batch of each AL strategy). This resulted in an accuracy score of 0.695 for TREC and 0.789 for AGNews. This suggests that it could have a positive effect if there is a more sophisticated approach. This is an open research topic that requires further investigation.

\subsection{How data-efficient is CAL in comparison to full data training?} If a model is able to choose the instances that it can learn the most from, it can reach comparable results or even outperform a model trained on all data by using fewer training instances. This is noted by~\citet{siddhant2018deep}, where they achieve 98--99\% of the full dataset performance while labeling only 20\% of the samples. The overall accuracy for AGNews trained on all data is 0.820 accuracy on test. For TREC, this results in 0.634. With CAL, we achieve around 99\% of the full dataset performance while using only 2\% training data for AGNews. For TREC, we outperform the full dataset performance by 0.090 accuracy (0.634 vs.\ 0.724), the full dataset performance is already reached by using around 14\% training data. This is appealing, as active learning can provide more data-effective learning solutions.

\section{Conclusion}
In this paper, we introduced a new AL algorithm, Cartography Active Learning. The AL objective is transformed into a binary classification task~\citep{gissin2019discriminative}, where we optimize for selecting the most informative data with respect to a model by leveraging insights from data maps~\citep{swayamdipta-etal-2020-dataset}.  Data maps help to identify distinct regions in a dataset based on training dynamics (\textit{hard-to-learn} and \textit{easy-to-learn/ambiguous} instances), which have shown to play an important role in model optimization and stability in full dataset training~\citep{swayamdipta-etal-2020-dataset}. We use these insights in low-data regimes and propose CAL. 
Specifically for the task of Visual Question Answering, contemporary work by~\citet{karamcheti-etal-2021-mind} use data maps for analysis into AL acquisition functions and the effect of outliers. In contrast, this work instead applies data maps directly for AL by training a discriminator on limited seed data maps to select the most informative instances.
We show empirically that our method is competitive or significantly outperforms various popular AL methods, and provide intuitions on why this is the case by using training dynamics.

\newpage

\section{Appendix}

\subsection{Reproducibility}\label{reproducibility}
We initialize the main model with English $d_{\text{emb}}= $ 300 FastText embeddings~\citep{bojanowski2017enriching} and keep it frozen during training and inference, we sum the embeddings over the sequence of tokens as motivated by~\citet{banea-etal-2014-simcompass}. For AGNews we impose a maximum sequence length of 200 and a batch size of 64. For TREC, a maximum sentence length of 42 and a batch size of 16. We run both models for 30 epochs, with no early stopping. Per AL iteration, we do a weight reset on all models. We average our results over five randomly generated seeds (398048, 127003, 259479, 869323, 570852). All experiments were ran on an NVIDIA\textsuperscript{\textregistered} A100 SXM4 40 GB GPU
and an AMD EPYC\textsuperscript{\texttrademark} 7662 64-Core CPU. Specifically for CAL, a single AL batch (50) selection iteration takes 11 seconds on average assuming TREC. For AGNews, one AL iteration takes 62 seconds on average. Both runtimes are with respect to the models depicted in~\cref{configurations} and hardware mentioned above.

\subsection{Full Data Map}\label{stats:full}

\begin{figure*}[t]
    \centering
    \includegraphics[width=0.9\linewidth]{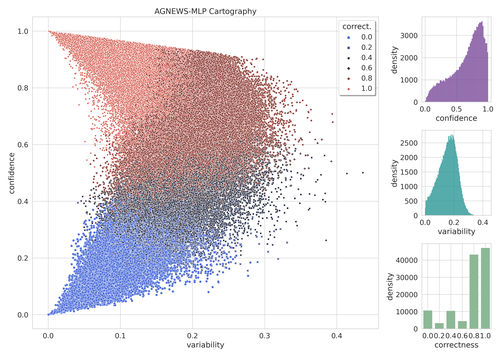}
    \caption{\textbf{Density of AGNews.} Density statistics of AGNews over ten epochs.}
    \label{fig:agnews-results}
\end{figure*}

\begin{figure*}[t]
    \centering
    \includegraphics[width=0.9\linewidth]{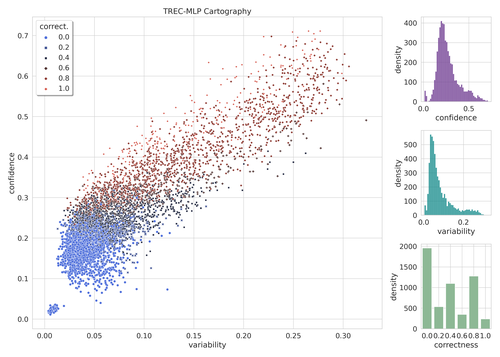}
    \caption{\textbf{Density of TREC.} Density statistics of TREC over ten epochs.}
    \label{fig:trec-results}
\end{figure*}

\cref{fig:agnews-results} and~\cref{fig:trec-results} show the full data maps for AGNews and TREC respectively. Identically to~\citet{swayamdipta-etal-2020-dataset}, we show the density of data points in the plots. We can see a clear difference in density between the datasets. For AGNews, we can see the majority of data points have a high confidence ($\sim$0.8) and high correctness. In contrast, TREC contains plenty of instances that have low confidence ($\sim$0.3) and low correctness.

\chapter{\textsc{{S}kill{S}pan}: {H}ard and {S}oft {S}kill {E}xtraction from {E}nglish {J}ob {P}ostings}
\chaptermark{{SkillSpan}}
\label{chap:chap3}
The work presented in this chapter is based on a paper that has been published as: \bibentry{zhang-etal-2022-skillspan}.

\definecolor{Gray}{gray}{0.9}
\definecolor{LightCyan}{rgb}{0.88,1,1}
\newcolumntype{?}{!{\vrule width 1pt}}
\newcolumntype{g}{>{\columncolor{LightCyan}}r}

\newcommand{\spanb}{SpanBERT}
\newcommand{\sk}[1]{\colorbox{pink}{$\textrm{[#1]}_\textrm{\tiny SKILL}$}}
\newcommand{\kn}[1]{\colorbox{yellow}{$\textrm{[#1]}_\textrm{\tiny KNOWLEDGE}$}}
\newcommand{\guide}[1]{\noindent\fbox{\parbox{\textwidth}{#1}}}
\newcommand{\bjo}{\textsc{Big}}
\newcommand{\hou}{\textsc{House}}
\newcommand{\tech}{\textsc{Tech}}
\newcommand*\circled[1]{\tikz[baseline=(char.base)]{
            \node[shape=circle,draw,inner sep=.6pt] (char) {#1};}}
\renewcommand{\floatpagefraction}{.99} %

\newpage

\section*{Abstract}
Skill Extraction (SE) is an important and widely-studied task useful to gain insights into labor market dynamics. However, there is a lacuna of datasets and annotation guidelines; available datasets are few and contain crowd-sourced labels on the span-level or labels from a predefined skill inventory.
To address this gap, we introduce \textsc{SkillSpan}, a novel SE dataset consisting of 14.5K sentences and over 12.5K annotated spans. We release its respective guidelines created over three different sources annotated for hard \emph{and} soft skills by domain experts. We introduce a BERT baseline~\citep{devlin2019bert}. To improve upon this baseline, we experiment with language models that are optimized for long spans~\citep{joshi-etal-2020-spanbert, beltagy2020longformer}, continuous pre-training on the job posting domain~\citep{han-eisenstein-2019-unsupervised, gururangan-etal-2020-dont}, and multi-task learning~\citep{caruana1997multitask}. Our results show that the domain-adapted models significantly outperform their non-adapted counterparts, and single-task outperforms multi-task learning. 

\section{Introduction}

Job markets are under constant development---often due to developments in technology, migration, and digitization---so are the skill sets required. Consequentially, job vacancy data is emerging on a variety of platforms in big quantities and can provide insights on labor market skill demands or aid job matching~\citep{balog2012expertise}. SE is to extract the competences necessary from unstructured text. 

Previous work in SE shows promising progress, but is halted by a lack of available datasets and annotation guidelines. Two out of 14 studies release their dataset, which limit themselves to crowd-sourced labels~\citep{sayfullina2018learning} or annotations from a predefined list of skills on the document-level~\citep{bhola-etal-2020-retrieving}. Additionally, none of the 14 previously mentioned studies release their annotation guidelines, which obscures the meaning of a competence. Job markets change, as do the skills in, e.g., the European Skills, Competences, Qualifications and Occupations (ESCO;~\citealp{le2014esco}) taxonomy (\cref{skillknowledge2}). Hence, it is important to cover for possible emerging skills.

\begin{figure}
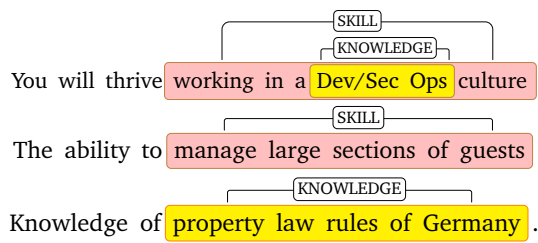

\centering
\resizebox{.6\linewidth}{!}{
\begin{dependency}[edge slant=0pt, edge vertical padding=3pt, edge style={-}]
    \begin{deptext}
    You \& will \& thrive \& working \& in \& a \& Dev/Sec \& Ops \& culture \& . \\
    \end{deptext}
    \wordgroup[group style={fill=pink, draw=brown, inner sep=.3ex}]{1}{4}{9}{sk}
    \wordgroup[group style={fill=yellow, draw=brown, inner sep=.1ex}]{1}{7}{8}{kn}
    \depedge[edge height=3.5ex]{4}{9}{SKILL}
    \depedge[edge height=1ex, edge horizontal padding=-18pt, edge end x offset=10pt]{7}{8}{KNOWLEDGE}
\end{dependency}}
\resizebox{.6\linewidth}{!}{
\begin{dependency}[edge slant=0pt, edge vertical padding=2pt, edge style={-}]
    \begin{deptext}
    The \& ability \& to \& manage \& large \& sections \& of \& guests  \& .\\
    \end{deptext}
    \wordgroup[group style={fill=pink, draw=brown, inner sep=.1ex}]{1}{4}{8}{sk}
    \depedge[edge height=1ex]{4}{8}{SKILL}
\end{dependency}}
\resizebox{.6\linewidth}{!}{
\begin{dependency}[edge slant=0pt, edge vertical padding=2pt, edge style={-}]
    \begin{deptext}
    Knowledge \& of \& property \& law \& rules \& of \& Germany \& .\\
    \end{deptext}
    \wordgroup[group style={fill=yellow, draw=orange, inner sep=.1ex}]{1}{3}{7}{kn}
    \depedge[edge height=1ex]{3}{7}{KNOWLEDGE}
\end{dependency}}
    \caption{\textbf{Examples of Skills \& Knowledge Components.} Annotated samples of passages in varying job postings. More details are given in \cref{sec:ann}.}
    \label{fig:doccano2}
\end{figure}

We propose \textsc{SkillSpan}, a novel SE dataset annotated at the span-level for \emph{skill} and \emph{knowledge} components (SKCs) in job postings (JPs). As illustrated in~\cref{fig:doccano2}, SKCs can be nested inside skills. \textsc{SkillSpan} allows for extracting possibly undiscovered competences and to diminish the lack of coverage of predefined skill inventories.

Our analysis~(\cref{fig:violin2}) shows that SKCs contain on average longer sequences than typical Named Entity Recognition (NER) tasks. Albeit we additionally study models optimized for long spans~\citep{joshi2020spanbert, beltagy2020longformer}, some underperform. Overall, we find specialized domain BERT models~\citep{alsentzer-etal-2019-publicly, lee2020biobert, gururangan2020don, nguyen-etal-2020-bertweet} perform better than their non-adapted counterparts. We explore the benefits of domain-adaptive pre-training on the JP domain~\citep{han-eisenstein-2019-unsupervised, gururangan2020don}. Last, given the examples from~\cref{fig:doccano2}, we formulate the task as both as a sequence labeling and a multi-task learning (MTL) problem, i.e., training on both skill and knowledge components jointly~\citep{caruana1997multitask}. 

\paragraph{Contributions.} In this paper: 
\begin{enumerate}
    \itemsep0em
    \item We release \textsc{SkillSpan}, a novel skill extraction dataset, 
with annotation guidelines, and our open-source code.\footnote{\url{https://github.com/kris927b/SkillSpan}}
    \item We present strong baselines for the task including a new SpanBERT~\citep{joshi2020spanbert} trained from scratch, and domain-adapted variants~\citep{gururangan2020don}, which we will release on the \texttt{HuggingFace} platform~\citep{wolf-etal-2020-transformers}. To the best of our knowledge, we are the first to investigate the extraction of skills and knowledge from job postings with state-of-the-art language models.
    \item We give an analysis of single-task versus multi-task learning in the context of skill extraction, and show that for this particular task single-task learning outperforms multi-task learning. 
\end{enumerate}

\begin{landscape}
\begin{table*}[t]
\centering
    \begin{adjustbox}{width=1.5\textwidth}
    \begin{tabular}{l|lllllll}
    \toprule
Paper & \textbf{Annotations} & \textbf{Approach} & \textbf{Size} & \textbf{Skill Type} & \textbf{(Baseline) Model(s)} & \faPencil & \faBook\\
\midrule
 \citet{kivimaki-etal-2013-graph}   & Document-level        &  Automatic        & N/A               & Hard &  LogEnt., TF-IDF, LSA, LDA         & \xmark & \xmark \\
 \citet{zhao2015skill}              & Sentence-level        &  Automatic        & N/A               & Hard & Word2Vec                           & \xmark & \xmark\\
 \citet{javed2017large}             & Span-level           &  Skill Inventory  & N/A               & Both & Word2Vec                            & \xmark & \xmark\\
 \citet{jia2018representation}      & Span-level       &  Automatic        & 21,158 JPs*       & Hard & LSTM                                    & \xmark & \xmark\\
 \citet{sayfullina2018learning}     & Span-level        &  Crowdsourcing    & 4,863 Sent.       & Soft & CNN, LSTM                            & \cmark & \xmark\\
 \citet{smith2019syntax}            & Span-level           &  Manual           & 100 JPs           & Hard & Pattern Matching                     & \xmark & \xmark\\
 \citet{gugnani2020implicit}        & Span-level           &  Domain Experts   & $\sim$200 JPs     & Hard & Word2Vec, Doc2Vec                   & \xmark & \xmark\\
 \citet{li2020deep}                 & Document-level        &  Proprietary      & N/A               & Hard & FastText                           & \xmark & \xmark\\
 \citet{shi2020salience}            & Span-level           &  Proprietary      & N/A               & Hard & FastText, USE, BERT                 & \xmark & \xmark\\
 \citet{tamburri2020dataops}        & Sentence-level        &  Domain Experts   & $\sim$3,000 Sent. & Both & BERT                               & \xmark & \xmark\\
 \citet{chernova2020occupational}   & Span-level            &  Manual           & 100 JPs           & Both & FinBERT                            & \xmark & \xmark\\
 \citet{bhola-etal-2020-retrieving} & Document-level        &  Skill Inventory  & 20,298 JPs*       & Hard & BERT                               & \cmark & \xmark\\
 \citet{smith2021skill}             & Span-level           &  Manual           & 100 JPs           & Hard & Pattern Match., Word2Vec            & \xmark & \xmark\\
 \citet{liu2021learning}            & Document-level        &  Crowdsourcing    & N/A               & Hard & GNN                                & \xmark & \xmark\\
 \midrule
 \textbf{This work}                 & Span-level    &  Domain Experts   & 391 JPs           & Both & (Domain-adapted) BERT           & \cmark  & \cmark \\
    \bottomrule
    \end{tabular}
    \end{adjustbox}
    \caption{\textbf{Contributions of Related Work}. We list the recent works of Skill Extraction. Note that (*) indicates labels that are automatically inferred from some source (e.g., a predefined skill inventory) and not \emph{manually} annotated. With respect to the annotation approach, ``Manual'' indicates uncertainty whether they used domain experts or not. Also note that many works do not release their dataset with annotations (\faPencil)  nor guidelines (\faBook). The list is inspired by \citet{khaouja2021survey}.}
    \label{tab:relwork25}
\end{table*}
\end{landscape}

\section{Related Work}

There is a pool of prior work relating to SE. We summarize it in~\cref{tab:relwork25}, depicting state-of-the-art approaches, level of annotations, what kind of competences are annotated, the modeling approaches, the size of the dataset (if available), type of skills annotated for, baseline models, and whether they release their annotations and guidelines.

As can be seen in~\cref{tab:relwork25}, many works do \textit{not release their data} (apart from~\citealp{sayfullina2018learning} and~\citealp{bhola-etal-2020-retrieving}) and \textbf{none release their annotation guidelines}. In addition, none of the previous studies approach SE as a span-level extraction task with state-of-the-art language models, nor did they release a dataset of this magnitude with manually annotated (long) spans of competences by domain experts.

Although \citet{sayfullina2018learning} annotated on the span-level (thus being useful for SE) and release their data, they instead explored several approaches to \emph{Skill Classification}. To create the data, they extracted all text snippets containing \emph{one} soft skill from a predetermined list. Crowdworkers then annotated the highlighted skill whether it was a soft skill referring to the candidate or not. They show that an LSTM~\citep{hochreiter1997long} performs best on classifying the skill in the sentence. In our work, we annotated a dataset three times their size (\cref{tab:num_post2}) for both hard \emph{and} soft skills. In addition, we also extract the specific skills from the sentence.

\citet{tamburri2020dataops} classifies sentences that contain skills in the JP. The authors manually labeled their dataset with domain experts. They annotated whether a sentence contains a skill or not. Once the sentence is identified as containing a skill, the skill cited within is extracted. In contrast, we directly annotate for the span within the sentence.

\citet{bhola-etal-2020-retrieving} cast the task of skill extraction as a multi-label skill classification at the document-level. There is a predefined set of unique skills given the job descriptions and they predict multiple skills that are connected to a given job description using BERT~\citep{devlin2019bert}. In addition, they experiment with several additional layers for better prediction performance. We instead explore domain-adaptive pre-training for SE.

The work closest to ours is by \citet{chernova2020occupational}, who approach the task similarly with span-level annotations (including longer spans) but approach this for the Finnish language. It is unclear whether they annotated by domain experts. Also, neither the data nor the annotation guidelines are released. For a comprehensive overview with respect to SE, we refer to~\citet{khaouja2021survey}.

\section{Skill \& Knowledge Definition}\label{skillknowledge2}
There is an abundance of competences and there have been large efforts to categorize them. For example, the The International Standard Classification of Occupations~\citep[ISCO;\ ][]{elias1997occupational} is one of the main international classifications of occupations and skills. It belongs to the international family of economic and social classifications. Another example, the European Skills, Competences, Qualifications and Occupations~\citep[ESCO;\ ][]{le2014esco} taxonomy is the European standard terminology linking skills and competences and qualifications to occupations and derived from ISCO. The ESCO taxonomy mentions three categories of competences: \emph{Skill}, \emph{knowledge}, and \emph{attitudes}. ESCO defines knowledge as follows:
\begin{quote}
    ``Knowledge means the outcome of the assimilation of information through learning. Knowledge is the body of facts, principles, theories and practices that is related to a field of work or study.''~\footnote{\url{https://ec.europa.eu/esco/portal/escopedia/Knowledge}}
\end{quote}
For example, a person can acquire the Python programming language through learning. This is denoted as a \emph{knowledge} component and can be considered a \emph{hard skill}. However, one also needs to be able to apply the knowledge component to a certain task. This is known as a \emph{skill} component. ESCO formulates it as:
\begin{quote}
    ``Skill means the ability to apply knowledge and use know-how to complete tasks and solve problems.''~\footnote{\url{https://ec.europa.eu/esco/portal/escopedia/Skill}}
\end{quote}
 
\noindent
In ESCO, the \emph{soft skills} are referred to as \emph{attitudes}. ESCO considers attitudes as skill components:

\begin{quote}
    ``The ability to use knowledge, skills and personal, social and/or methodological abilities, in work or study situations and professional and personal development.''~\footnote{\url{http://data.europa.eu/esco/skill/A}}
\end{quote}

\noindent
To sum up, hard skills are usually referred to as \emph{knowledge} components, and applying these hard skills to something is considered a \emph{skill} component. Then, soft skills are referred to as \emph{attitudes}, these are part of skill components. There has been no work, to the best of our knowledge, in annotating skill and knowledge components in JPs.

\begin{table}[ht!]
\centering
    \resizebox{.9\linewidth}{!}{
    \begin{tabular}{l|lrrr|g}
    \toprule
& $\downarrow$ \textbf{Statistics}, \textbf{Src.} $\rightarrow$ & \textbf{\bjo{}} & \textbf{\hou{}} & \textbf{\tech{}} & \textbf{Total}\\
    \midrule
\multirow{6}{*}{\rotatebox[origin=c]{90}{\textbf{Train}}}
   & \textbf{\# Posts}                  & 60        & 60        & 80        & 200       \\
   & \textbf{\# Sentences}              & 1,036     & 1,674     & 3,156     & 5,866     \\
   & \textbf{\# Tokens}                 & 29,064    & 36,995    & 56,549    & 122,608   \\
   & \textbf{\# Skill Spans}            & 1,086     & 984       & 1,237     & 3,307     \\
   & \textbf{\# Knowledge Spans}        & 439       & 781       & 2,188     & 3,408     \\
   & \textbf{\# Overlapping Spans}      & 45        & 29        & 135       & 209       \\
   \midrule
\multirow{6}{*}{\rotatebox[origin=c]{90}{\textbf{Development}}}
   & \textbf{\# Posts}                  & 30        & 30        & 30        & 90     \\
   & \textbf{\# Sentences}              & 783       & 1,022     & 2,187     & 3,992  \\
   & \textbf{\# Tokens}                 & 11,762    & 19,173    & 21,149    & 52,084 \\
   & \textbf{\# Skill Spans}            & 469       & 525       & 545       & 1,539  \\
   & \textbf{\# Knowledge Spans}        & 126       & 287       & 806       & 1,219  \\
   & \textbf{\# Overlapping Spans}      & 12        & 17        & 32        & 61     \\
 \midrule
\multirow{6}{*}{\rotatebox[origin=c]{90}{\textbf{Test}}}
   & \textbf{\# Posts}                  & 36        & 33        & 32        & 101    \\
   & \textbf{\# Sentences}              & 1,112     & 1,216     & 2,352     & 4,680  \\
   & \textbf{\# Tokens}                 & 14,720    & 21,923    & 20,885    & 57,528 \\
   & \textbf{\# Skill Spans}            & 634       & 637       & 459       & 1,730  \\
   & \textbf{\# Knowledge Spans}        & 242       & 350       & 834       & 1,426  \\
   & \textbf{\# Overlapping Spans}      & 12        & 8         & 9         & 29     \\
   \midrule
  \rowcolor{LightCyan}
   & \textbf{\# Posts}                  & 126       & 123       & 142       & 391       \\
  \rowcolor{LightCyan}
   & \textbf{\# Sentences}              & 2,931     & 3,912     & 7,695     & 14,538    \\
    \rowcolor{LightCyan}
   & \textbf{\# Tokens}                 & 55,546    & 78,091    & 98,583    & 232,220   \\
    \rowcolor{LightCyan}
   & \textbf{\# Skill Spans}            & 2,189     & 2,146     &  2,241    & 6,576     \\
   \rowcolor{LightCyan}
   & \textbf{\# Knowledge Spans}        & 807       & 1,418     &  3,828    & 6,053     \\
   \rowcolor{LightCyan}
   \multirow{-6}{*}{\rotatebox[origin=c]{90}{\textbf{Total}}}
   & \textbf{\# Overlapping Spans}      & 69        & 54        & 178       & 301     \\
    \midrule
    \rowcolor{Gray}
   & \textbf{\# Posts}                  & \multicolumn{4}{r}{126,769} \\
     \rowcolor{Gray}
   & \textbf{\# Sentences}              & \multicolumn{4}{r}{3,195,585} \\
      \rowcolor{Gray}
   \multirow{-3}{*}{\rotatebox[origin=c]{90}{\textbf{\textsc{$\mathcal{U}$}}}}
   & \textbf{\# Tokens}                 & \multicolumn{4}{r}{460,484,670} \\
\bottomrule
    \end{tabular}}
    \caption{\textbf{Statistics of Dataset.} Indicated is the number of JPs across splits \& source and their respective number of sentences, tokens, and spans. The total is reported in the cyan column and rows. We report the overall statistics of the unlabeled JPs ($\mathcal{U}$) in the gray rows.}
    \label{tab:num_post2}
\end{table}
\clearpage

\section{\textsc{SkillSpan} Dataset}\label{sec:ann}

\subsection{Data} We continuously collected JPs via web data extraction between June 2020--September 2021.\footnote{Our data statement~\citep{bender-friedman-2018-data} can be found in \cref{app:datastatement2}} Our JPs come from three sources:

\begin{enumerate}
\itemsep0em
    \item \textbf{\bjo{}}: A large job platform with various types of JPs, with several type of positions;
    \item \textbf{\hou{}}: A \emph{static} in-house dataset consisting of similar types of jobs as \bjo{}. Dates range from 2012--2020;
    \item \textbf{\tech{}}: The StackOverflow JP platform that consisted mostly of technical jobs (e.g., developer positions).
\end{enumerate}

We release the anonymized raw data and annotations of the parts with permissible licenses, i.e., \hou{} (from a govermental agency which is our collaborator) and \tech{}.\footnote{Links to our data can be found at \url{https://github.com/kris927b/SkillSpan}.} For anonymization, we perform it via manual annotation of job-related sensitive and personal data regarding \texttt{Organization}, \texttt{Location}, \texttt{Contact}, and \texttt{Name} following the work by~\citet{jensen-etal-2021-de}.

\cref{tab:num_post2} shows the statistics of \textsc{SkillSpan}, with 391 annotated JPs from the three sources containing 14.5K sentences and 232.2K tokens. The unlabeled JPs (only to be released as pre-trained model) consist of 126.8K posts, 3.2M sentences, and 460.5M tokens. What stands out is that there are 2--5 times as many annotated knowledge components in \tech{} in contrast to the other sources, despite a similar amount of JPs. We expect this to be due the numerous KCs depicted in this domain (e.g., programming languages), while we observe considerably fewer soft skills (e.g., ``work flexibly''). The amount of skills is more balanced across the three sources. Furthermore, overlapping spans follow a consistent trend among splits, with the train split containing the most.

\begin{figure}[t!]
    \centering
    \includegraphics[width=.9\linewidth]{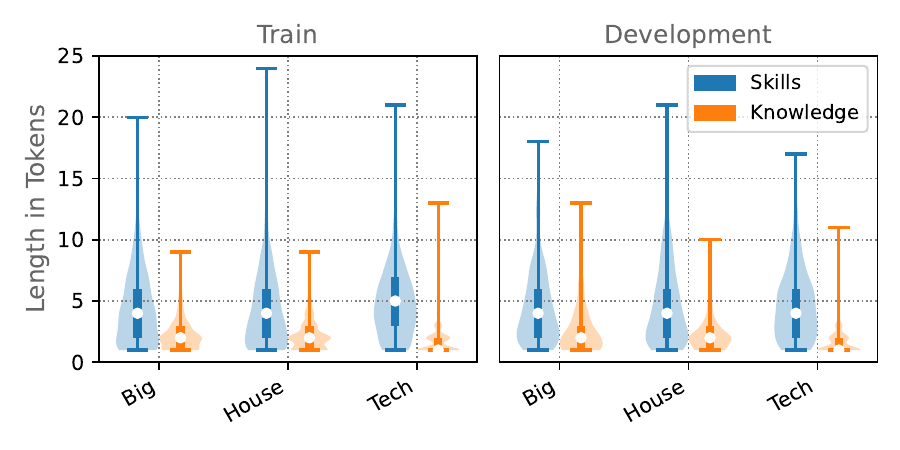}
    \caption{\textbf{Violin Plots of Annotated Components.} Indicated are the distributions regarding the length of spans in each type of annotated component (i.e., length of skills and knowledge components). The white dot is the median length, the bars range from the first quartile to the third quartile, and the colored line ranges from the lower adjacent value to the higher adjacent value.}
    \label{fig:violin2}
\end{figure}

\subsection{Data Annotation}

We annotate competences related to SKCs in two levels as illustrated in~\cref{fig:doccano2}. We started the process in March 2021, with initial annotation rounds to construct and refine the annotation guidelines (as outlined further below). The annotation process spanned eight months in total.
Our final annotation guidelines can be found in~\cref{sec:ann_guide2}. The guidelines were developed by largely following example spans given in the ESCO taxonomy. 
However, at this stage, we focus on span identification, and  we do not take the fine-grained taxonomy codes from ESCO for labeling the spans, 
leaving the mapping to ESCO and taxonomy enrichment as future work.

\subsection{Further Details on the Annotation Process} The development of the annotation guidelines and our annotation process is depicted as follows: \circled{\textbf{1}} We wrote base guidelines derived from a small number of JPs. \circled{\textbf{2}} We had three pre-rounds consisting of three JPs each. After each round, we modified, improved and finalized the guidelines. \circled{\textbf{3}} Then, we had three longer-lasting annotation rounds consisting of 30 JPs each. We re-annotated the previous 11 JPs in \circled{1} and \circled{2}.
\circled{\textbf{4}} After these rounds, one of the annotators (the hired linguist) annotated JPs in batches of 50. The data in \circled{1}, \circled{2}, and \circled{3} was annotated by three annotators (101 JPs).

We used an open source text annotation tool named \textsc{Doccano} \citep{doccano}. 
There are around 57.5K tokens (approximately 4.6K sentences, in 101 job posts) that we calculated agreement on. 
The annotations were compared using Cohen's $\kappa$~\citep{fleiss1973equivalence} between pairs of annotators, and Fleiss' $\kappa$~\citep{fleiss1971measuring}, which generalises Cohen's $\kappa$ to more than two concurrent annotations. We consider two levels of $\kappa$ calculations: \textbf{\textsc{Token}} is calculated on the token level, comparing the agreement of annotators on each token (including non-entities) in the annotated dataset.~\textbf{\textsc{Span}} refers to the agreement between annotators on the exact span match over the surface string, regardless of the type of SKC, i.e., we only check the position of tag without regarding the type of the entity. 
The observed agreements scores over the three annotators from step \circled{3} are between 0.70--0.75 Fleiss' $\kappa$ for both levels of calculation which is considered a \emph{substantial agreement}~\citep{landis1977measurement} and a $\kappa$ value greater than 0.81 indicates \emph{almost perfect agreement}. Given the difficulty of this task, we consider the aforementioned $\kappa$ score to be strong. Particularly, we observed a large improvement in annotation agreement from the earlier rounds (step \circled{1} and \circled{2}), where our Fleiss' $\kappa$ was 0.59 on token-level and 0.62 for the span-level. 

Overall, we observe higher annotator agreement for knowledge components (3--5\% higher) compared to skills 
which tend to be longer. The \textsc{Tech} domain is the most consistent for agreement while \textsc{Big} shows more variation over rounds, likely due to the broader nature of the domains of JPs. 

\subsection{Annotation Span Statistics}
A challenge of annotating spans is the length (i.e., boundary), SKCs being in different domains (e.g., business versus technical components), and frequently written differently, e.g., ``being able to work together'' v.s.\ ``teamwork''). \cref{fig:violin2} shows the statistics of our annotations in violin plots. For the training set, the median length (white dot) of skills is around 4 for \bjo{} and \hou{}, for \tech{} this is a median of 5. In the development set, the median stays at length 4 across all sources. Another notable statistic is the upper and lower percentile of the length of skills and knowledge, indicated with the thick bars. Here, we highlight the fact that skill components could consist of many tokens, for example, up to length 7 in the \hou{} source split (see blue-colored violins). For knowledge components, the spans are usually shorter, where it is consistently below 5 tokens (see orange-colored violins). All statistics follow a similar distribution across train, development, and sources in terms of length and distribution. This gives a further strong indication that consistent annotation length has been conducted across splits and sources.

\begin{table}[t]
\centering
    \resizebox{\linewidth}{!}{
    \begin{tabular}{l|lll}
    \toprule
& \textbf{\bjo{}} & \textbf{\hou{}} & \textbf{\tech{}}\\
\midrule
\multirow{5}{*}{\rotatebox[origin=l]{90}{\textbf{\textsc{Skill}}}}
    & ambitious              & structured           & hands-on                      \\
    & proactive              & teaching             & communication skills             \\
    & work independently     & communication skills & leadership                    \\
    & attention to detail    & project management   & passionate                    \\
    & motivated              & drive                & open-minded                   \\
    \midrule
\multirow{5}{*}{\rotatebox[origin=l]{90}{\textbf{\textsc{knowledge}}}}
    & full uk driving licence   & english              & java        \\
    & sap energy assessments    & supply chain         & javascript  \\
    & right to work in the uk   & project management   & aws         \\
    & sen                       & powders              & docker      \\
    & acca/aca                  & machine learning     & node.js     \\
    \bottomrule
    \end{tabular}}
    \caption{\textbf{Most Frequent Skills in the Development Data.} Top-5 skill components in our data in terms of frequency on different sources. A larger example can be found in~\cref{tab:freq-skill2} and~\cref{tab:freq-knowledge2} (\cref{app:quali2}).}
    \label{tab:freq-top52}
\end{table}

\subsection{Qualitative Analysis of Annotations} 
Qualitative differences in SKCs over the three sources are shown (lowercased) in \cref{tab:freq-top52}. With respect to skill components, all sources follow a similar usage of skills. The annotated skills mostly relate to the \emph{attitude} of a person 
and hence mostly consist of soft skills.
With respect to knowledge components, we observe differences between the three sources. First, on the source-level, the knowledge components vastly differ between \bjo{} and \tech{}. \bjo{} postings seem to cover more \emph{business} related components, whereas \tech{} has more \emph{engineering} components. \hou{} seems to be a mix of the other two sources. Lastly, note that both the skill and knowledge components between the splits diverge in terms of the type of annotated spans, which indicates a variation in the annotated components. We show the top--10 skills annotated in the train, development, and test splits for SKCs in~\cref{app:quali2}.
From a syntactic perspective, skills frequently consist of noun phrases, verb phrases, or adjectives (for soft skills). Knowledge components usually consists of nouns or proper nouns, such as ``python'', ``java'', and so forth. 

\section{Experimental Setup}

The task of SE is formulated as a sequence labeling problem. Formally, we consider a set of JPs $\mathcal{D}$, where $d \in \mathcal{D}$ is a set of sequences (i.e., entire JPs) with the $i^\text{th}$ input sequence $\mathcal{X}^i_{d} = \{x_1, x_2, ..., x_T\}$ and a target sequence of \texttt{BIO}-labels $\mathcal{Y}^i_{d} = \{y_1, y_2, ..., y_T\}$ (e.g., ``\texttt{B-SKILL}'', ``\texttt{I-KNOWLEDGE}'', ``\texttt{O}''). The goal is to use $\mathcal{D}$ to train a sequence labeling algorithm $h: \mathcal{X} \mapsto \mathcal{Y}$ to accurately predict entity spans by assigning an output label $y_t$ to each token $x_t$. 

As baseline we consider BERT and we investigate more recent variants, and we also train models from scratch. Models are chosen due to their state-of-the-art performance, or in particular, for their strong performance on  longer spans.

\subsection{\bertb{}~\citep{devlin2019bert}} An out-of-the-box \bertb{} model (\texttt{bert-base-cased}) from the \texttt{HuggingFace} library~\citep{wolf-etal-2020-transformers} functioning as a baseline.

\subsection{SpanBERT~\citep{joshi2020spanbert}} A BERT-style model that focuses on span representations as opposed to single token representations. SpanBERT is trained by masking contiguous spans of tokens and optimizing two objectives: (1) masked language modeling, which predicts each masked token from its own vector representation. (2) The span boundary objective, which predicts each masked token from the representations of the unmasked tokens at the start and end of the masked span.

We train a SpanBERT\textsubscript{base} model from scratch on the BooksCorpus~\citep{Zhu_2015_ICCV} and English Wikipedia using cased Wordpiece tokens~\citep{wu2016google}. We use AdamW~\citep{kingma2014adam} for 2.4M training steps with batches of 256 sequences of length 512. The learning rate is warmed up for 10K steps to a maximum value of 1$e$-4, after which it has a decoupled weight decay~\citep{loshchilov2017decoupled} of 0.1. We add a dropout rate of 0.1 across all layers. Pretraining was done on a v3-8 TPU on the GCP and took 14 days to complete. We take the official TensorFlow implementation of SpanBERT by~\citet{ram2021few}.

\subsection{JobBERT} We apply domain-adaptive pre-training~\citep{gururangan2020don} to a \bertb~model using the 3.2M unlabeled JP sentences~(\cref{tab:num_post2}).\footnote{\url{https://huggingface.co/jjzha/jobbert-base-cased}} Domain-adaptive pre-training relates to the continued self-supervised pre-training of a large language model on domain-specific text. This approach improves the modeling of text for downstream tasks within the domain. We continue training the BERT model for three epochs (default in \texttt{HuggingFace}) with a batch size of 16.

\subsection{JobSpanBERT} We apply domain-adaptive pre-training to our SpanBERT on 3.2M unlabeled JP sentences.\footnote{\url{https://huggingface.co/jjzha/jobspanbert-base-cased}} We keep parameters identical to the vanilla SpanBERT, but change the number of steps to 40K to have three passes over the unlabeled data.

\subsection{Experiments}
We have 391 annotated JPs (\cref{tab:num_post2}) that we divide across three splits: Train, dev.\, and test set. We use 101 JPs that all three annotators annotated as the gold standard test set with aggregated annotations via majority voting. The 101 postings are divided between the sources as: 36 \bjo{}, 33 \hou{}, and 32 \tech{}. The remaining 290 JPs were annotated by one annotator. We use 90 JPs (30 from each source, namely \bjo{}, \hou{}, and \tech{}) as the dev.\ set. The remaining 200 JPs are used as the train set. The sources in the train set are divided into 60 \bjo{}, 60 \hou{}, and 80 \tech{}.

\subsection{Setup} The data is structured as \texttt{CoNLL} format~\citep{tjong2003introduction}. For the nested annotations, the skill tags are appearing only in the first column and the knowledge tags are only appearing in the second column of the file and they are allowed to overlap with each other. 
We perform experiments with single-task learning (STL) on either the skill or knowledge components, MTL for predicting both skill and knowledge tags at the same time, while evaluating the MTL models also on either skills or knowledge components. We used a single joint MTL model with hard-parameter sharing~\citep{caruana1997multitask}.
All models are with a final Conditional Random Field~\citep[CRF;\ ][]{lafferty2001conditional} layer. Earlier research, such as~\citet{souza2019portuguese, jensen-etal-2021-de} show that BERT models with a CRF-layer improve or perform similarly to its simpler variants when comparing the overall F1 and make no tagging errors (e.g., \texttt{B}-tag follows \texttt{I}-tag). In the case of MTL we use one for each tag type (skill and knowledge). In the STL experiments we use one CRF for the given tag type.

We use the \textsc{MaChAmp} toolkit~\citep{van-der-goot-etal-2021-massive} for our experiments. For each setup we do five runs (i.e., five random seeds).\footnote{For reproducibility, we refer to~\cref{app:hyper2}.} 
For evaluation we use span-level precision, recall, and F1, where the F1 for the MTL setting is calculated as described in~\citet{benikova2014nosta}.

\begin{figure*}[t!]
    \centering
    \includegraphics[width=\linewidth]{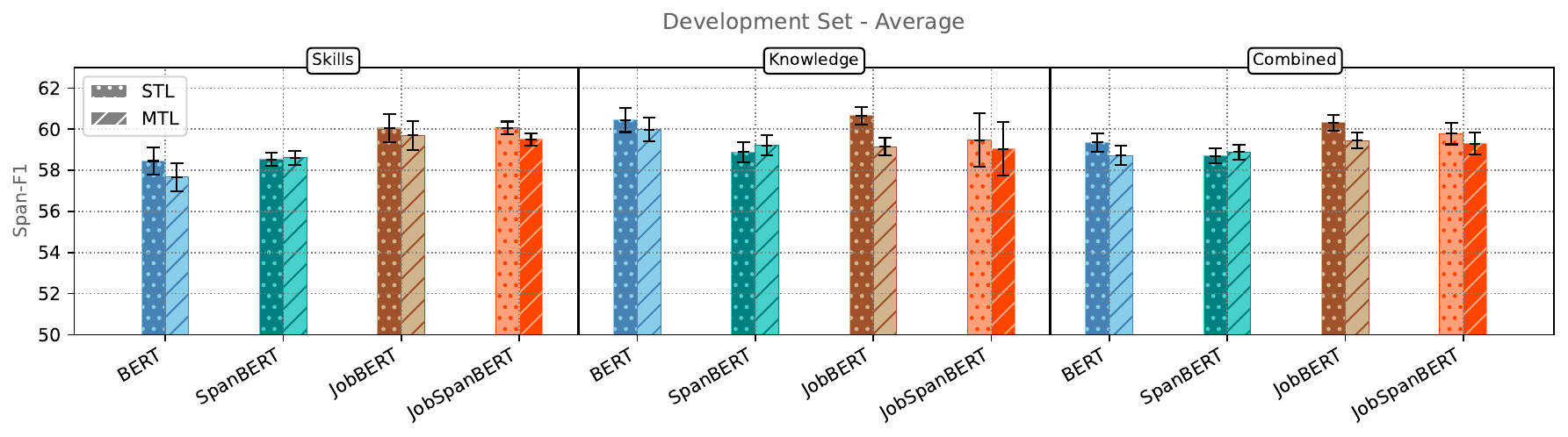}
    \includegraphics[width=\linewidth]{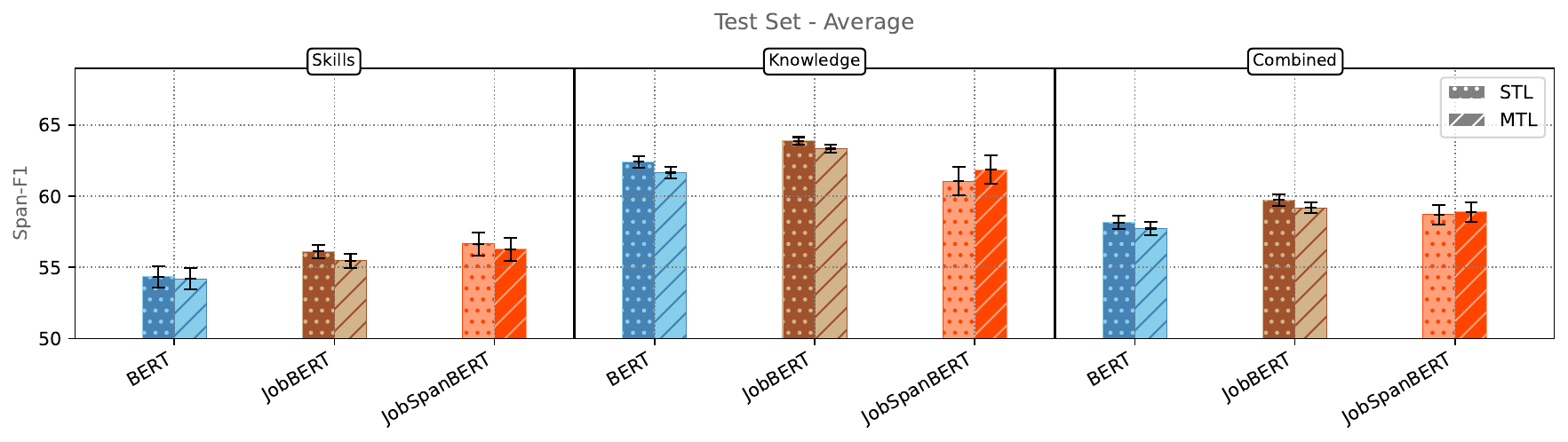}
 
    \caption{\textbf{Performance of Models.} We test the models on \textbf{\textsc{Skills}}, \textbf{\textsc{Knowledge}}, and \textbf{\textsc{Combined}}. We report the span-F1 and standard deviation (error bars) of runs on five random seeds. \textbf{Note that the y-axis starts from 50 span-F1}. \textbf{STL} indicates single-task learning and \textbf{MTL} indicates the multi-task model. Differences can be seen on the test set: JobSpanBERT performs best on \textsc{Skills}, JobBERT is best on \textsc{Knowledge}, and JobBERT achieves best in \textsc{Combined}. Exact numbers of the plots are in~\cref{tab:results2} (\cref{app:source-perf}).}
    \label{fig:results2}
\end{figure*}
\clearpage

\section{Results}

The results of the experiments are given in \cref{fig:results2}. We show the average performance of each model in F1 and respective standard deviation over the development and test split.
Exact scores on each source split and other metric details are provided in~\cref{app:source-perf}. As mentioned before, we experiment with the following settings: \textbf{\textsc{Skill}}, we train and predict only on skills. \textbf{\textsc{Knowledge}}, train and only predict for knowledge. \textbf{\textsc{Combined}}, we merge the STL predictions of both skills and knowledge. We also train the models in an MTL setting, predicting both skills and knowledge simultaneously. We evaluate the MTL model on both \textbf{\textsc{Skill}} and \textbf{\textsc{Knowledge}} separately, and also compare it against the aggregated STL predictions.

\subsection{Performance on Development Set} In \cref{fig:results2}, we show the results on the development set in the upper plot. We observe similar performance between the domain-adapted STL models---JobBERT and JobSpanBERT---have similar span-F1 for \textsc{Skill}: \std{60.05}{0.70} vs.\ \std{60.07}{0.70}. In contrast, for \textsc{Knowledge}, \bertb{} and JobBERT are closest in predictive performance: \std{60.44}{0.58} vs.\ \std{60.66}{0.43}. In the \textsc{Combined} setting, JobBERT performs highest with a span-F1 of \std{60.32}{0.39}. On average, JobBERT performs best over all three settings.
Surprisingly, the models for both \textsc{Skill} and \textsc{Knowledge} perform similarly (around 60 span-F1), despite the sources' differences in properties and length~\cref{fig:violin2}.
In addition, we find that MTL is not performing better than STL across sources. 
For exact numbers and source-level (i.e., \bjo{}, \hou{}, \tech{}), we refer to~\cref{app:source-perf}.

\subsection{Performance on Test Set} We select the best performing models in the development set evaluation and apply it to the test set. Results are in~\cref{fig:results2} in the bottom plot. Since JobBERT and JobSpanBERT are performing similarly, we apply both to the test set and \bertb{}. 
We observe a deviation from the development set to the test set: JobSpanBERT \std{60.07}{0.30}$\rightarrow$\std{56.64}{0.83} on \textsc{Skill}, JobBERT \std{60.66}{0.43}$\rightarrow$\std{63.88}{0.28}
on \textsc{Knowledge}. For \textsc{Combined}, JobBERT performs slightly worse: \std{60.32}{0.39}$\rightarrow$\std{59.73}{0.38}. Similar to the development set, we find that on all three methods of evaluation (i.e., \textsc{Skill}, \textsc{Knowledge}, and \textsc{Combined}), STL still outperforms MTL. For \textsc{Skill} and \textsc{Knowledge}, STL is almost stochastically dominant over MTL (i.e., significant), and for \textsc{Combined} there is stochastic dominance of STL over MTL, indicated in the next paragraph. 

\subsection{Significance}
\begin{figure*}[t]
    \centering
    \includegraphics[width=.32\linewidth]{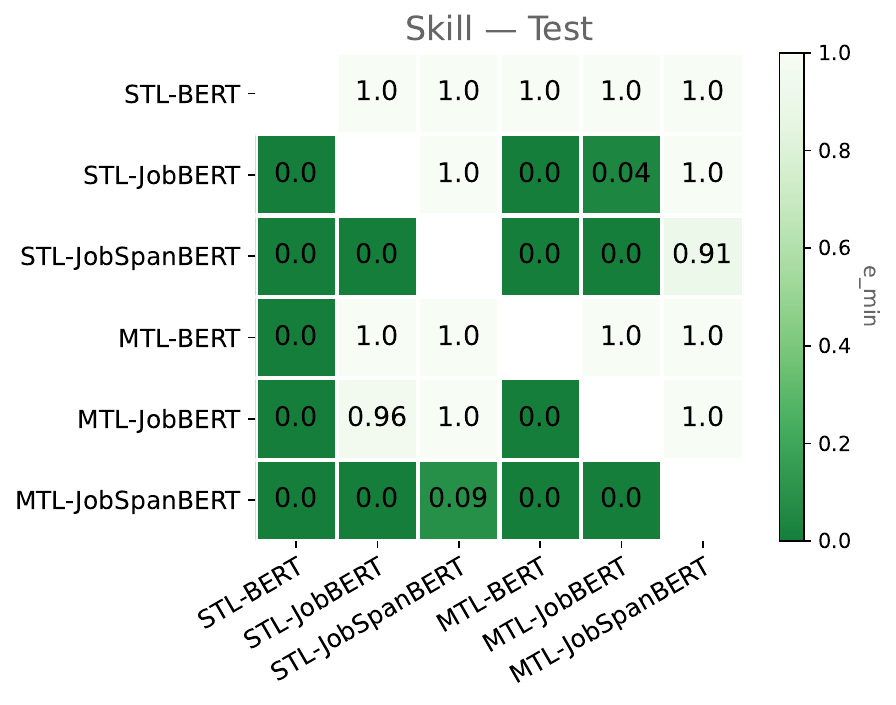}
    \includegraphics[width=.32\linewidth]{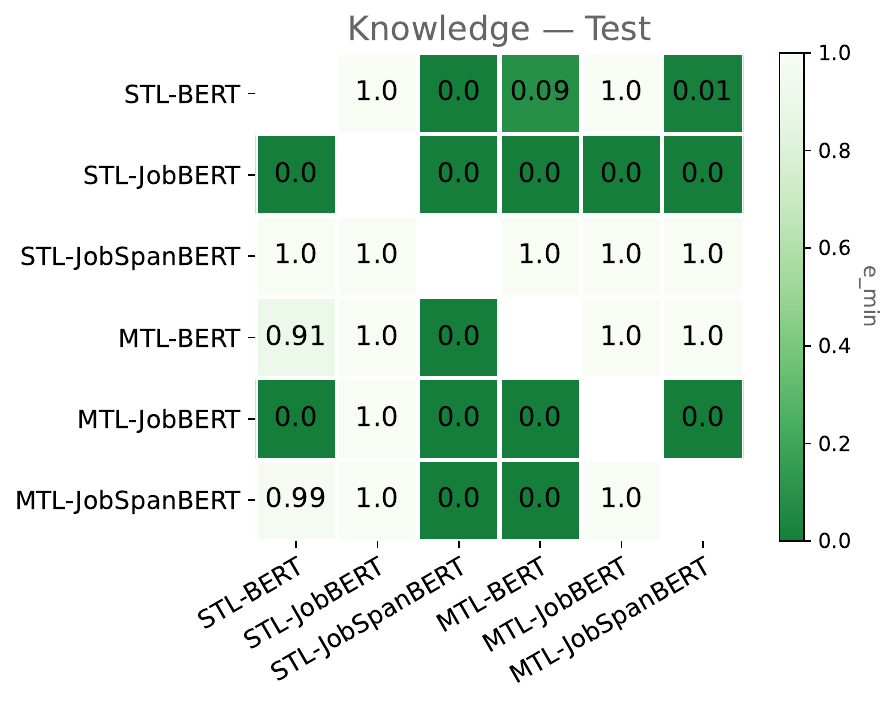}
    \includegraphics[width=.32\linewidth]{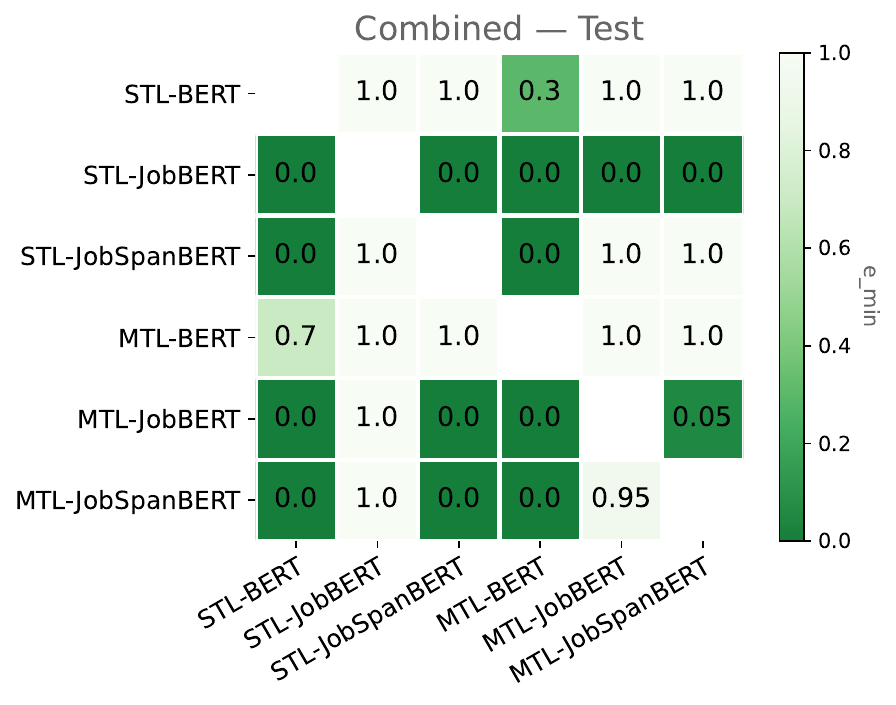}
    \caption{\textbf{Almost Stochastic Order Scores of the Test Set.}
    ASO scores expressed in $\epsilon_\text{min}$.
    The significance level $\alpha =$ 0.05 is adjusted accordingly by using the Bonferroni correction~\citep{bonferroni1936teoria}. Read from row to column: E.g., in \textsc{Combined} STL-JobBERT (row) is stochastically dominant over STL-\bertb{} (column) with $\epsilon_\text{min}$ of 0.00.}
    \label{fig:significance-test2}
\end{figure*}

We compare all pairs of models based on five random seeds each using Almost Stochastic Order~\citep[ASO;\  ][]{dror2019deep} tests with a confidence level of $\alpha =$ 0.05.
The ASO scores of the test set are indicated in~\cref{fig:significance-test2}. 
We show that MTL-JobSpanBERT for \textsc{Skill} shows almost stochastic dominance ($\epsilon_\text{min} < $ 0.5) over all other models. For \textsc{Knowledge} and \textsc{Combined}, We show that STL-JobBERT is stochastically dominant ($\epsilon_\text{min} = $ 0.0) over \emph{all} the other models.
For more details, we refer to~\cref{app:sign-per-source} for ASO scores on the development set.

\section{Discussion}

\subsection{What Did Not Work} Additionally, we experiment whether representing the entire JP for extracting tokens yields better results than the experiments so far, which were sentence-by-sentence processing setups. To handle entire JPs and hence much longer sequences we use a pre-trained Longformer$_\text{base}$~\citep{beltagy2020longformer} model. The document length we use in the experiments is 4096 tokens. Results of the Longformer on the test set are lower: For skills, JobSpanBERT against Longformer results in \std{56.64}{0.83} vs.\ \std{52.55}{2.39}. For \textsc{Knowledge}, JobBERT against Longformer shows \std{63.88}{0.28} vs.\ \std{57.26}{1.05}. Last, for \textsc{Combined}, JobBERT against Longformer results in \std{59.73}{0.38} vs.\ \std{55.05}{0.71}. This drop in performance is difficult to attribute to a concrete reason: e.g., the Longformer is trained on more varied sources than BERT, but not specifically for JPs, which may have contributed to this gap.
Since the vanilla Longformer already performs worse than \bertb~overall, we did not opt to apply domain-adaptive pre-training. Overall, we show that representing the full JP is not beneficial for SE, at least not in the Longformer setup tested here.

\subsection{Continuous Pretraining helps SE}

As previously mentioned, due to the domain specialization of the domain-adapted pre-trained BERT models, they predict more skills and frequently perform better in terms of precision, recall, and F1 as compared to their non-adaptive counterparts. This is especially encouraging as we confirm findings that continuous pre-training helps to adapt models to a specific domain~\citep{alsentzer-etal-2019-publicly, lee2020biobert, gururangan2020don, nguyen-etal-2020-bertweet}. However, there are exceptions. Particularly in~\cref{tab:results2} on \textsc{Test} for \textsc{Knowledge}, \bertb{} comes closer in predictive performance to JobBERT (difference of 1.5 F1) than on \textsc{Skills}. Our intuition is that knowledge components are often already in the pre-training data (e.g., Wikipedia pages of certain competences like Python, Java etc.) and therefore adaptive pre-training does not substantially boost performance.

\begin{figure}[t]
    \centering
    \includegraphics[width=.9\linewidth]{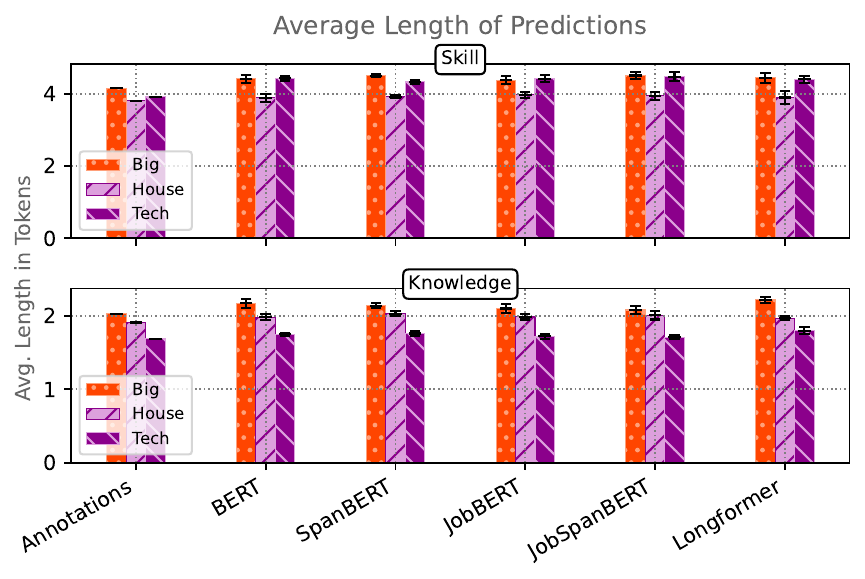}
    \caption{\textbf{Average Length of Predictions of Single Models.} We show the average length of the predictions versus the length of our annotated skills and knowledge components on the \emph{test set} and the total number of predicted skills and knowledge tags in each respective split (\#). There is a consistent trend over the three sources.}
    \label{fig:avglength2}
\end{figure}

\subsection{Difference in Length of Predictions} 
The main motivation of selecting models optimized for long spans was the length of the annotations (\cref{fig:violin2}). We investigate the average length of predictions of each model~(\cref{fig:avglength2}) to find out whether the models that are adapted to handle longer sequences truly predict longer spans. Interestingly, the average length of predicted skills are longer than the annotations over all three sources. There is a consistent trend among \textsc{Skill}: \bjo{} and \tech{} have similar length over predictions ($>$4), while \hou{} is usually lower than length 3.
For both \bjo{} and \tech{}, JobSpanBERT predicts the longest skill spans (4.51 and 4.48 respectively). We suspect due to the domain-adaptive pre-training on JPs, it improved the span prediction performance. In contrast, the Longformer predicts shorter spans. Note that the Longformer is not domain-adapted to JPs. 

Regarding \textsc{Knowledge}, there is also a consistent trend: \bjo{} has the overall longest prediction length while \tech{} has the lowest. The Longformer predicts the longest spans on average for \bjo{} and \tech{}. Knowledge components are representative of a normal-length NER task and might not need a specialized model for long sequences. We show the exact numbers in~\cref{tab:avglength} (\cref{app:source-perf}) and the number of predicted \textsc{Skill} and \textsc{Knowledge}: JobBERT and JobSpanBERT have higher recall than the other models.

\begin{figure*}[t!]
    \centering
    \includegraphics[width=\linewidth]{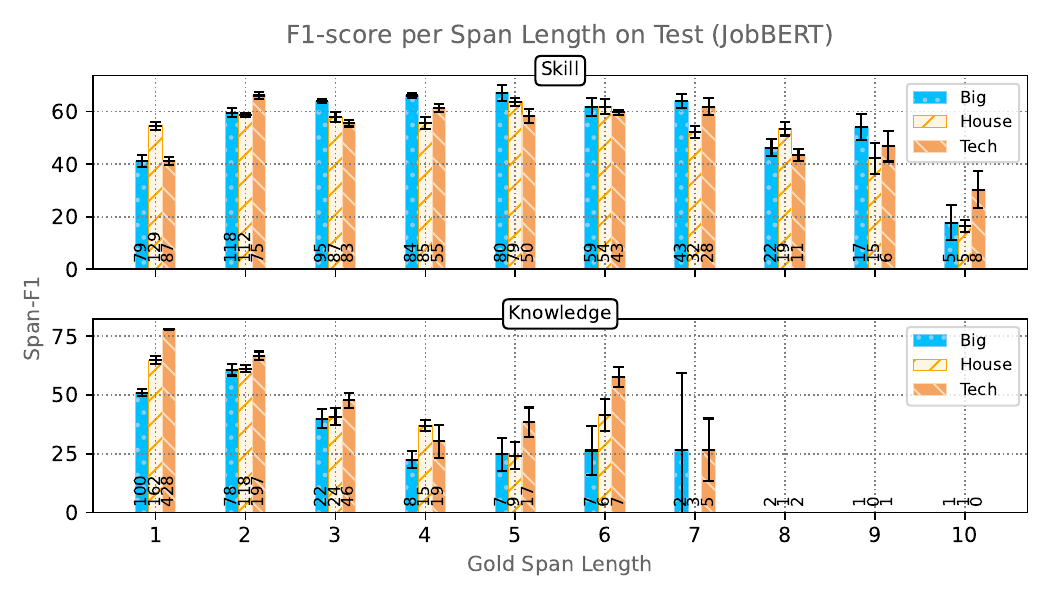}
    \caption{\textbf{Average Span-F1 per Span Length.} We bucket the performance of JobBERT according to the length of the spans until 10 tokens and show the performance on each length, averaged over five random seeds. Indicated per bar is the support. The model performs best on medium-length skill spans (i.e., spans with token length of 4-5). For knowledge spans, on average, it performs best on short-length spans (i.e., spans with token length of 1-2).}
    \label{fig:predlength2}
\end{figure*}

\subsection{Performance per Span Length}
\textsc{Skills} are generally longer than \textsc{Knowledge} components in our dataset (\cref{fig:violin2}). The previous overall results on the test set (\cref{fig:results2}) show that performance on \textsc{Skill} is substantially lower than \textsc{Knowledge}. We therefore investigate whether this performance difference is attributed to the longer spans in \textsc{Skill}. In~\cref{fig:predlength2}, we show the average performance of the best performing model (JobBERT) on the three sources (test set) based on the gold span length, until a length of 10.

In \textsc{Skill} components (upper plot), we see much support for spans with length 1 and 2, which then lowers once the spans become longer. Spans with length of 1 shows low performance on \bjo{} and \tech{} (around 40 span-F1), which influences the total span-F1. Short skills are usually soft skills, such as ``passionate'', which can be used as a skill or not. This might confuse the model. In contrast, performance effectively stays similar (around 60 span-F1) for span length of 2 till 7 for all sources. Afterwards, it drops in performance. Thus, the weak performance on \textsc{Skill} seem to be due to lower performance on the short spans.

For the \textsc{Knowledge} components (lower plot), they are generally shorter.  We see that there is a gap in support between the sources, \tech{} has a larger number of gold labels compared to \bjo{} and \hou{}. Unlike soft skills, KCs usually consist of proper nouns such as ``Python'', ``Java'', and so forth, which connects to the high performance on \tech{} (around 76 span-F1). Furthermore, support for spans longer than 2 drops considerably. In this case, if the model predicts a couple of instances correctly, it would substantially increase span-F1. Contrary to \textsc{Skill}, high performance of \textsc{Knowledge} can be attributed to its strong performance on short spans.

\section{Conclusion}
We present a novel dataset for skill extraction on English job postings--- \textsc{SkillSpan}---and domain-adapted BERT models---JobBERT and JobSpanBERT. We outline the dataset and annotation guidelines, created for hard \emph{and} soft skills annotation on the \emph{span-level}.
Our analysis shows that domain-adaptive pre-training helps to improve performance on the task for both skills and knowledge components. Our domain-adapted JobSpanBERT performs best on skills and JobBERT on knowledge. Both models achieve almost stochastic dominance over all other models for skills and knowledge extraction, whereas JobBERT in the STL setting achieves stochastic dominance over other models. 

With the rapid emergence of new competences, our new approach of skill extraction has future potential, e.g., to enrich knowledge bases such as ESCO with unseen skills or knowledge components, and in general, contribute to providing insights into labor market dynamics. We hope our dataset encourages research into this emerging area of computational job market analysis.

\clearpage

\section{Appendix}
\subsection{Data Statement \textsc{SkillSpan}}\label{app:datastatement2}
Following~\citet{bender-friedman-2018-data}, the following outlines the data statement for \textsc{SkillSpan}:
\begin{enumerate}[A.]
\itemsep0em
    \item \textsc{Curation Rationale}: Collection of job postings in the English language for span-level sequence labeling, to study the impact of sequence labeling on the extraction of skill and knowledge components from job postings.
    \item \textsc{Language Variety}: The non-canonical data was collected from the StackOverflow job posting platform, an in-house job posting collection from our national labor agency collaboration partner (The Danish Agency for Labor Market and Recruitment\footnote{\url{https://www.star.dk/en/}}), and web extracted job postings from a large job posting platform. US (en-US) and British (en-GB) English are involved.
    \item \textsc{Speaker Demographic}: Gender, age, race-ethnicity, socioeconomic status are unknown.
    \item \textsc{Annotator Demographic}: Three hired project participants (age range: 25--30), gender: one female and two males, white European and Asian (non-Hispanic). Native language: Danish, Dutch. Socioeconomic status: higher-education students. Female annotator is a professional annotator with a background in Linguistics and the two males with a background in Computer Science.
    \item \textsc{Speech Situation}: Standard American or British English used in job postings. Time frame of the data is between 2012--2021.
    \item \textsc{Text Characteristics}: Sentences are from job postings posted on official job vacancy platforms.
    \item \textsc{Recording Quality}: N/A.
    \item \textsc{Other}: N/A.
    \item \textsc{Provenance Appendix}: The job posting data in \tech{} is from Stackoverflow jobs, and is licensed under the CC BY-SA license. The job posting data from \hou{} is from our collaborators: The Danish Agency for Labour Market and Recruitment (STAR).
\end{enumerate}

\subsection{Annotation Guidelines}\label{sec:ann_guide2}

\subsubsection{Span Specifications}
\guide{
\textbf{Legend:} \colorbox{pink}{Skill}, \colorbox{yellow}{Knowledge}, ``•'' indicates an example sentence.
}

\guide{
    1. A skill starts with a \textbf{VERB}, otherwise (\textbf{ADJECTIVE}) + \textbf{NOUN}\\\\
    1.1 Modal verbs are not tagged:
    \begin{itemize}\setlength\itemsep{0em}
        \item \st{Can} \sk{put personal touch on the menu}.
        \item \st{Will} \sk{train new staff}.
    \end{itemize}
}

\guide{
2. Split up phrases with prepositions and/or conjunctions\\\\
2.1 \textbf{Unless} the conjunction coordinates two nouns functioning as one argument:
\begin{itemize}
    \item \sk{Coordinate parties and conferences}.
\end{itemize}
2.2 Do not tag skills with anaphoric pronouns, only tag preceding skill:
\begin{itemize}
    \item \sk{Prioritizing tasks} and identifying those that are most important.
\end{itemize}
2.3 Split nouns and adjectives that are coordinated if they do not have a verb attached:
\begin{itemize}
    \item Be \sk{inquisitive} and \sk{proactive}.
    \item Prior in-house experience with \kn{media}, \kn{publishing} or \kn{internet companies}.
\end{itemize}
2.4 If there is a listing of skill tags and they lead up to different subtasks, we split them:
\begin{itemize}
    \item \sk{keep up the high level of quality in our team} through \sk{reviews}, \sk{pairing} and \sk{mentoring}.
\end{itemize}
}

\guide{
    3. If there is relevant information appended after irrelevant information (e.g., info specific to a company) we try to make the skill as short as possible:
    \begin{itemize}
        \item \sk{providing the best solution} \st{for Siemens Gamesa in a very} \sk{structured} and \sk{analytic} manner.
    \end{itemize}
}

\guide{
    4. Note also the words skills and knowledge can be included in the span of the component if leaving it out makes it nonsensical:
    \begin{itemize}
        \item \sk{personal skills} → just [personal] would make it nonsensical.
    \end{itemize}
}

\guide{
    5. Parentheses after a skill tag are included if they elaborate the component before them or if they are an abbreviation of the component.
}
\guide{
6. \textbf{Inclusion of adverbials in components}. Adverbials are included if it concerns the manner of doing something. All others are excluded:
\begin{itemize}
    \item   \st{like to} \sk{solve technical challenges \underline{independently}}.
    \item   \sk{communicates \underline{openly}}.
    \item   \sk{striving for the best} \st{in all that they do}.
    \item   \sk{Deliver first class customer service} \st{to our guests}.
    \item   \sk{Making the right decisions} \st{early in the process}.
\end{itemize}
}   
\guide{
7. \textbf{Attitudes as skills.} We annotate attitudes as a skill:
\begin{itemize}
    \item \st{a} \sk{can-do-approach} → we leave out articles from the attitude.
\end{itemize}
8. Attitudes are not tagged if they contain skill/knowledge components---then only the span of the skill is tagged.

\begin{itemize}
    \item \st{like to} \sk{solve technical challenges independently}.
    \item \st{Passion for} \kn{automation}.
    \item \st{enjoy} \sk{working in a team}.
\end{itemize}
}

\guide{
9. \textbf{Miscellaneous:}\\\\
9.1 Do \underline{not} tag ironic skills (e.g., lazy).\\\\
9.2 Avoid nesting of skills, annotate it as one span.\\\\
9.3  We annotate all skills that are part of sections such as ``requirements'', ``good-to-haves'', ``great-to-knows'', ``optionals'', ``after this $x$ months of training you'll be able to...'', ``At the job you're going to...''.\\\\
9.4 When there is a general standard that can be added to the skill, we add these: 
\begin{itemize}
    \item \sk{Process payments according to the [...] standards}.
\end{itemize}
}

\subsection{Knowledge Specifications}\label{sub:knowledge2}
\guide{
1. \textbf{Rule-of-thumb}: knowledge is something that one possesses, and cannot (usually) physically execute:
\begin{itemize}
    \item \kn{Python} (programming language).
    \item \kn{Business}.
    \item \kn{Relational Databases}.
\end{itemize}
}

\guide{
2. If there is a component between parentheses that belongs to the knowledge component, we add it:
\begin{itemize}
    \item \kn{(non-) relational databases}.
    \item \kn{Driver License (UK/EU)}.
\end{itemize}
}

\guide{
3. \textbf{Licenses and certifications}: We add the additional words ``certificate'', ``card'', ``license'', et cetera. to the knowledge component.
}

\guide{
4. If the knowledge component looks like a skill, but the preceding verb is vague and empty (e.g., \textit{follow}, \textit{use}, \textit{comply with}, \textit{work with}) → only tag the knowledge component:
\begin{itemize}
    \item \st{Comply with} \kn{Food Code of Practice}.
    \item \st{Work with} \kn{AWS infrastructure}.
\end{itemize}
}

\guide{
5. We annotate only specified knowledge components:
\begin{itemize}
    \item \kn{MongoDB} or other \kn{NoSQL database}.
    \item \kn{JEST} \st{or other test libraries}. → ``other test libraries'' is under-specified.
\end{itemize}
}

\guide{
6. Knowledge components can be nested in skill components.
\begin{itemize}
    \item \sk{Design, execution and analysis...} 
    \item[] \sk{of \kn{phosphoproteomics} experiments}.
\end{itemize}
}

\guide{
7. If all components coordinate/share one knowledge tag, we annotate it as one:
\begin{itemize}
    \item \kn{application, data and infrastructure architecture}. → The knowledge tags coordinate to ``architecture''. 
    \item \kn{chemical/biochemical engineering}.
\end{itemize}
}

\guide{
8. If there is a listing of knowledge tags, we annotate all knowledge tags separately:
\begin{itemize}
    \item \kn{Bachelor Degree} in \kn{Mathematics}, \kn{Computer Science}, or \kn{Engineering}.
\end{itemize}
}

\subsection{Other Specifications}
\guide{
1. \textbf{Rule-of-thumb}: If in doubt, annotate it as a skill.
}

\guide{
2. We are preferring skills over knowledge components.
}

\guide{
3. We prioritize skills over attitudes; if there is a skill within the attitude, only tag the skill:
\begin{itemize}
    \item \st{Passionate around} \sk{solving business problems} through \\ \kn{innovation \& engineering practices}.
\end{itemize}

}

\guide{
4. Skill or knowledge components in the top headlines of the JP are not tagged (e.g., title of a JP). If it is a sub-headline or in the rest of the posting, tag it.
}

\guide{
5. We try to keep the skill/knowledge components as short as possible (i.e., exclude information at the end if it makes it too specific for the job).
}

\guide{
6. We do not include ``fluff'' and ``triggers'' (i.e., words that indicate a skill or knowledge component will follow:
``\st{advanced knowledge of} \kn{...}'') around the components, including degree. This goes for both before and after:
\begin{itemize}
    \item \st{Working proficiency in} \kn{developmental toolsets}.
    \item \st{Knowledge of} \kn{application data and architecture} \st{disciplines}.
    \item \sk{Manual handling} \st{tasks}.
    \item \kn{CI/CD} \st{experience}.
    \item \st{You master} \kn{English} \st{on level C1}.
    \item \st{Proficient in} \kn{Python} \st{and} \kn{English}.
    \item \st{Fluent in spoken and written} \kn{English}.
\end{itemize}
}

\guide{
7. Pay attention to expressions such as ``participation in...'', ``contributing'', and ``transfer (knowledge)''. These are usually not considered skills.
\begin{itemize}
    \item \st{Contribute to the enjoyable and collaborative work environment}.
    \item \st{Participation in the Department's regular research activities}.
    \item \st{Desire to be part of something meaningful and innovative}.
\end{itemize}
}

\guide{
8. Skills and Knowledge components that are found in not-so-straightforward places (e.g., project descriptions) are annotated as well, if they relate to the position.
}

\guide{
9. In the pattern of ``skill'' followed by some elaboration, see if it can be annotated with a skill and a knowledge tag:
\begin{itemize}
    \item \sk{Ensure food storage and prep.\ areas are maintained} according to \kn{Health \& Safety and Audit standards}.
\end{itemize}
}
\guide{
10. Occupations and positions in companies/academia should be excluded.
}
\guide{
11. If there's a knowledge/skill component in the position, we exclude it as well.
\begin{itemize}
\item Experienced Java Engineer. → completely untagged.
\end{itemize}
}
\guide{
12. Only annotate the skills that are \underline{related} to the position.\\\\
12.1. This includes skills that are specific for the position as well (e.g., skills of a ruminants professor versus math professor).\\\\
12.2 Also skills that the person for the position is expected to do in the future.\\\\
12.3 This does \underline{not} include skills, knowledge or attitudes describing only the company, the group you will join in the department, and so on. \underline{Only annotate} if it is specified or implied that the employee should possess the skill as well.
}
\guide{
13. We annotate industries and fields (that the employee will be working in) as knowledge components.}

\subsection{Type of Skills Annotated}\label{app:quali2}
In both~\cref{tab:freq-skill2} and~\cref{tab:freq-knowledge2}, we show the top-10 skill and knowledge components that have been annotated. We split the top-10 among the data splits (i.e., train, development, and test set), and also between source splits (i.e., \bjo{}, \hou{}, \tech{}).

\subsection{Reproducibility}\label{app:hyper2}

\begin{table}[t]
    \centering
    \begin{tabular}{l|r|r}
    \toprule
    \textbf{Parameter} & \textbf{Value} & \textbf{Range} \\
    \midrule
    Optimizer                           & AdamW                & \\
    $\beta_\text{1}$, $\beta_\text{2}$  & 0.9, 0.99            & \\
    Dropout                             & 0.2                  & 0.1, 0.2, 0.3\\
    Epochs                              & 20                   & \\
    Batch Size                          & 32                   & \\
    Learning Rate (LR)                  & 1e-4                 & 1e-3, 1e-4, 1e-5\\
    LR scheduler                        & Slanted triangular   & \\
    Weight decay                        & 0.01                 & \\
    Decay factor                        & 0.38                 & 0.35, 0.38, 0.5\\
    Cut fraction                        & 0.2                  & 0.1, 0.2, 0.3\\
    \bottomrule
    \end{tabular}
    \caption{Hyperparameters of \textsc{MaChAmp}.}
    \label{tab:hyperparameters2}
\end{table}

\noindent
We use the default hyperparameters in \textsc{MaChAmp}~\citep{van-der-goot-etal-2021-massive} as shown in~\cref{tab:hyperparameters2}. For more details we refer to their paper. For the five random seeds we use 3477689, 4213916, 6828303, 8749520, and 9364029. All experiments with \textsc{MaChAmp} were ran on an NVIDIA\textsuperscript{\textregistered} TITAN X (Pascal) 12 GB GPU
and an Intel\textsuperscript{\textregistered} Xeon\textsuperscript{\textregistered} Silver 4214 CPU.

\begin{table*}[ht]
\centering
\resizebox{\linewidth}{!}{
    \begin{tabular}{ll|rr|rr|rr}
    \toprule
    & \textbf{Evaluation} $\rightarrow$ & \multicolumn{2}{c|}{\textbf{\textsc{Skill}}} & \multicolumn{2}{c|}{\textbf{\textsc{Knowledge}}} & \multicolumn{2}{c}{\textbf{\textsc{Combined}}}\\
    \midrule
    \textbf{Src.} & $\downarrow$ \textbf{Model}, \textbf{Task} $\rightarrow$ & \textbf{STL} & \textbf{MTL} & \textbf{STL} & \textbf{MTL} & \textbf{STL (*2)} & \textbf{MTL}\\
    \midrule
    \multirow{4}{*}{\rotatebox[origin=c]{90}{\textbf{\bjo{}}}}
    & \bertb        & \std{59.55}{0.97} & \std{58.88}{1.14} & \std{50.68}{3.25} & \std{51.10}{1.67} & \std{57.46}{1.19} & \std{57.00}{0.91}\\
    & SpanBERT      & \std{59.78}{0.44} & \std{60.02}{2.15} & \std{50.65}{2.32} & \std{51.79}{2.12} & \std{57.71}{0.53} & \std{58.00}{2.07}\\
    & JobBERT       & \std{60.60}{0.81} & \std{59.76}{0.60} & \std{50.29}{1.86} & \std{47.59}{1.11} & \std{58.19}{0.49} & \std{56.75}{0.50}\\
    & JobSpanBERT   & \std{60.16}{0.61} & \std{59.44}{1.11} & \std{45.20}{2.76} & \std{47.69}{3.38} & \std{56.56}{0.49} & \std{56.58}{0.63}\\
    \midrule
    \multirow{4}{*}{\rotatebox[origin=c]{90}{\textbf{\hou{}}}}
    & \bertb        & \std{56.83}{1.29} & \std{55.89}{1.90} & \std{55.00}{1.11} & \std{54.05}{1.00} & \std{56.17}{0.92} & \std{55.20}{1.35}\\
    & SpanBERT      & \std{57.54}{1.08} & \std{57.30}{0.84} & \std{52.01}{1.72} & \std{51.48}{1.01} & \std{55.55}{1.10} & \std{55.09}{0.74}\\
    & JobBERT       & \std{59.81}{1.17} & \std{59.97}{0.85} & \std{54.94}{1.15} & \std{54.23}{2.60} & \std{58.02}{0.93} & \std{57.80}{1.50}\\
    & JobSpanBERT   & \std{59.97}{1.03} & \std{59.62}{0.74} & \std{55.66}{1.51} & \std{53.10}{1.27} & \std{58.37}{1.07} & \std{57.14}{0.56}\\
    \midrule
    \multirow{4}{*}{\rotatebox[origin=c]{90}{\textbf{\tech{}}}}
    & \bertb        & \std{59.05}{0.71} & \std{58.34}{0.75} & \std{64.08}{1.04} & \std{63.77}{1.18} & \std{62.10}{0.67} & \std{61.65}{0.62}\\
    & SpanBERT      & \std{58.39}{0.46} & \std{58.61}{1.14} & \std{62.68}{0.60} & \std{63.40}{0.93} & \std{61.02}{0.35} & \std{61.56}{0.81}\\
    & JobBERT       & \std{59.81}{0.75} & \std{59.36}{0.90} & \std{64.57}{0.42} & \std{63.15}{0.94} & \std{62.69}{0.40} & \std{61.67}{0.90}\\
    & JobSpanBERT   & \std{60.09}{1.43} & \std{59.48}{0.61} & \std{63.40}{1.51} & \std{63.23}{0.64} & \std{62.09}{0.85} & \std{61.80}{0.54}\\
    \midrule
    \rowcolor{Gray}
    & \bertb        & \std{58.45}{0.68} & \std{57.67}{1.01} & \std{60.44}{0.58} & \std{59.98}{0.75} & \std{59.35}{0.46} & \std{58.72}{0.48}\\
    \rowcolor{Gray}
    & SpanBERT      & \std{58.53}{0.33} & \std{58.60}{0.83} & \std{58.89}{0.49} & \std{59.21}{0.78} & \std{58.69}{0.36} & \std{58.88}{0.64}\\
    \rowcolor{Gray}
    & JobBERT       & \std{60.05}{0.70} & \std{59.69}{0.62} & \textbf{\std{60.66}{0.43}}\textsuperscript{*} & \std{59.15}{1.07} & \textbf{\std{60.32}{0.39}}\textsuperscript{*} & \std{59.44}{0.81}\\
    \rowcolor{Gray}
    \multirow{-4}{*}{\rotatebox[origin=c]{90}{\textbf{\textsc{Average}}}}
    & JobSpanBERT   & \textbf{\std{60.07}{0.30}}\textsuperscript{$\dagger$} & \std{59.51}{0.68} & \std{59.47}{1.31} & \std{59.04}{0.65} & \std{59.79}{0.53} & \std{59.29}{0.43}\\
    \midrule
    \rowcolor{LightCyan}
    & \bertb        & \std{54.34}{0.74} & \std{54.20}{0.68} & \std{62.43}{0.41} & \std{61.66}{0.83} & \std{58.16}{0.47} & \std{57.73}{0.66} \\
    \rowcolor{LightCyan}
    & JobBERT       & \std{56.11}{0.49} & \std{55.46}{0.75} & \textbf{\std{63.88}{0.28}}\textsuperscript{*} & \std{63.35}{0.30} & \textbf{\std{59.73}{0.38}}\textsuperscript{$\dagger$} & \std{59.18}{0.37}\\
    \rowcolor{LightCyan}
    \multirow{-3}{*}{\rotatebox[origin=c]{90}{\textbf{\textsc{Test}}}}
    & JobSpanBERT   & \textbf{\std{56.64}{0.83}}\textsuperscript{*} & \std{56.27}{0.55} & \std{61.06}{0.99} & \std{61.87}{0.55} & \std{58.72}{0.69} & \std{58.90}{0.48}\\
    \bottomrule
    \end{tabular}
}
\caption{\textbf{Performance of Models.} We test the models on \textbf{skills}, \textbf{\textsc{Knowledge}}, and \textbf{\textsc{Combined}} (MTL). We report the span-F1 and standard deviation of runs on five random seeds on the \emph{development set} (\textbf{\textsc{Average}}, in gray). Results on the \emph{test set} are below in the \textbf{\textsc{Test}} rows (in cyan). \textbf{STL} indicates single-task learning and \textbf{MTL} indicates the multi-task model. 
\textbf{Bold} numbers indicate best performing model in that experiment. A ($\dagger$) means that it is stochastically dominant over \emph{all} the other models. (*) denotes \emph{almost stochastic dominance }($\epsilon_\text{min} <$ 0.5) over---at minimum---one other model.}
    \label{tab:results2}
\end{table*}

\begin{table*}[t]
\centering
\resizebox{\linewidth}{!}{
    \begin{tabular}{ll|rr|rr|rr}
    \toprule
& \textbf{Evaluation} $\rightarrow$ & \multicolumn{2}{c|}{\textbf{\textsc{Skill}}} & \multicolumn{2}{c|}{\textbf{\textsc{Knowledge}}} & \multicolumn{2}{c}{\textbf{\textsc{Multi}}}\\
\midrule
\textbf{Src.} & $\downarrow$ \textbf{Model} & \textbf{Precision} & \textbf{Recall} & \textbf{Precision} & \textbf{Recall} & \textbf{Precision} & \textbf{Recall}\\
\midrule
\multirow{4}{*}{\rotatebox[origin=c]{90}{\textbf{\bjo{}}}}
    & \bertb        & \std{57.09}{1.70} & \std{62.27}{1.28}       & \std{43.95}{4.17} & \std{60.00}{1.65}               & \std{52.63}{1.32} & \std{62.19}{0.87} \\
    & SpanBERT      & \std{58.28}{0.59} & \std{61.36}{0.68}       & \std{45.80}{2.89} & \std{56.82}{3.39}               & \std{54.02}{1.81} & \std{62.63}{2.60} \\
    & JobBERT       & \std{57.90}{1.25} & \std{63.59}{0.99}       & \std{43.45}{1.98} & \std{59.84}{3.44}               & \std{51.13}{0.48} & \std{63.74}{0.79} \\
    & JobSpanBERT   & \std{58.39}{1.03} & \std{62.09}{1.85}       & \std{38.55}{3.12} & \std{54.76}{3.18}               & \std{52.22}{0.35} & \std{61.75}{1.22} \\
    \midrule
\multirow{4}{*}{\rotatebox[origin=c]{90}{\textbf{\hou{}}}}
    & \bertb        & \std{55.95}{2.46} & \std{57.79}{0.67}      & \std{52.84}{0.65} & \std{57.42}{2.76}                & \std{51.65}{1.11} & \std{59.28}{2.07} \\
    & SpanBERT      & \std{56.70}{1.59} & \std{58.44}{1.16}      & \std{49.87}{2.57} & \std{54.49}{3.09}                & \std{52.27}{0.64} & \std{58.25}{1.50} \\
    & JobBERT       & \std{58.16}{1.30} & \std{61.56}{1.53}      & \std{51.18}{2.18} & \std{59.37}{1.34}                & \std{53.72}{1.57} & \std{62.56}{1.47} \\
    & JobSpanBERT   & \std{59.04}{0.85} & \std{60.99}{2.58}      & \std{51.36}{2.70} & \std{60.84}{1.19}                & \std{53.91}{0.77} & \std{60.79}{0.54} \\
     \midrule
\multirow{4}{*}{\rotatebox[origin=c]{90}{\textbf{\tech{}}}}
    & \bertb        & \std{58.28}{1.30} & \std{59.89}{1.39}      & \std{60.79}{1.89} & \std{67.79}{1.20}                & \std{58.19}{1.12} & \std{65.55}{0.75} \\
    & SpanBERT      & \std{58.62}{0.32} & \std{58.16}{0.76}      & \std{59.43}{1.21} & \std{66.35}{1.18}                & \std{58.34}{0.97} & \std{65.17}{1.41} \\
    & JobBERT       & \std{58.81}{1.38} & \std{60.88}{1.51}      & \std{61.38}{1.11} & \std{68.14}{1.36}                & \std{57.69}{0.93} & \std{66.25}{0.90} \\
    & JobSpanBERT   & \std{59.86}{3.07} & \std{60.40}{0.68}      & \std{59.78}{2.43} & \std{67.57}{1.97}                & \std{58.26}{0.82} & \std{65.82}{0.91} \\
    \midrule
\multirow{4}{*}{\rotatebox[origin=c]{90}{\textbf{\textsc{Average}}}}
    & \bertb        & \std{57.11}{1.65} & \std{59.90}{0.95}      & \std{56.86}{1.33} & \std{64.54}{1.31}                & \std{55.02}{0.85} & \std{62.98}{0.93} \\
    & SpanBERT      & \std{57.85}{0.65} & \std{59.23}{0.52}      & \std{55.65}{1.09} & \std{62.58}{1.56}                & \std{55.61}{0.61} & \std{62.58}{1.25} \\
    & JobBERT       & \std{58.29}{1.08} & \std{61.94}{1.16}      & \std{56.73}{1.41} & \std{65.22}{1.03}                & \std{55.03}{0.84} & \std{64.62}{0.77} \\
    & JobSpanBERT   & \std{59.11}{1.59} & \std{61.12}{1.49}      & \std{55.11}{2.41} & \std{64.66}{1.38}                & \std{55.64}{0.56} & \std{63.46}{0.69} \\
     \midrule
     \multirow{3}{*}{\rotatebox[origin=c]{90}{\textbf{\textsc{Test}}}}
    & \bertb        & \std{56.02}{1.50}  & \std{52.79}{1.18}      & \std{59.09}{0.85} & \std{66.20}{1.69}                & \std{55.82}{1.03} & \std{59.79}{0.87} \\
    & JobBERT       & \std{55.94}{1.19} & \std{56.29}{0.49}      & \std{60.03}{1.13} & \std{68.30}{1.46}                & \std{55.87}{0.29} & \std{62.89}{0.56}\\
    & JobSpanBERT   & \std{57.57}{1.24} & \std{55.77}{1.65}      & \std{57.83}{1.03} & \std{64.71}{2.10}                & \std{57.06}{0.74} & \std{60.89}{0.42} \\
    \bottomrule
    \end{tabular}}

\caption{\textbf{Precision and Recall of Models}. We test the models on \emph{skills}, \emph{knowledge}, and \emph{multi-task} setting. We report the average precision, recall and standard deviation of runs on five random seeds on the \emph{development set} (\textbf{\textsc{Average}}). Results on the \emph{test set} are below in the \textbf{\textsc{Test}} rows.}
    \label{tab:results2-precision-recall}
\end{table*}

\begin{table*}[ht]
    \centering
\begin{adjustbox}{width=\textwidth}
    \begin{tabular}{l|rr|rr|rr}
    \toprule
    \textbf{Source} $\rightarrow$& \multicolumn{2}{c|}{\textbf{\bjo{}}} & \multicolumn{2}{c|}{\textbf{\hou{}}} & \multicolumn{2}{c}{\textbf{\tech{}}}\\
    \midrule
    $\downarrow$ \textbf{Model} & \textsc{\textbf{Skills}} (\#) & \textbf{\textsc{Knowledge}} (\#) & \textsc{\textbf{Skills}} (\#) & \textbf{\textsc{Knowledge}} (\#) & \textsc{\textbf{Skills}} (\#) & \textbf{\textsc{Knowledge}} (\#)\\
\midrule
    \textsc{Annotations}      & 4.16 (634)             & 2.03 (242)             & 3.81 (637)             & 1.91 (350)             & 3.92 (459) & 1.69 (834) \\
     \bertb                    & \std{4.42}{0.11} (628) & \std{2.17}{0.06} (307) & \std{3.89}{0.11} (615) & \std{1.98}{0.04} (461) & \std{4.43}{0.06} (449) & \std{1.75}{0.02} (885)\\
     SpanBERT                  & \std{4.50}{0.04} (621) & \std{2.14}{0.03} (298) & \std{3.92}{0.04} (597) & \std{2.03}{0.03} (441) & \std{4.33}{0.06} (444) & \std{1.76}{0.03} (869)\\
     JobBERT                   & \std{4.38}{0.11} (670) & \std{2.10}{0.06} (313) & \std{3.97}{0.08} (650) & \std{1.99}{0.04} (470) & \std{4.42}{0.10} (479) & \std{1.72}{0.03} (932)\\
     JobSpanBERT               & \std{4.51}{0.09} (629) & \std{2.08}{0.05} (313) & \std{3.95}{0.11} (623) & \std{2.01}{0.06} (452) & \std{4.48}{0.12} (439) & \std{1.71}{0.03} (875)\\
     Longformer                & \std{4.45}{0.14} (653) & \std{2.22}{0.04} (298) & \std{3.90}{0.17} (639) & \std{1.97}{0.03} (483) & \std{4.40}{0.10} (472) & \std{1.80}{0.05} (864)\\
\bottomrule
    \end{tabular}
    \end{adjustbox}
    \caption{\textbf{Average Length of Predictions of Single Models.} We show the average length of the predictions versus the length of our annotated skills and knowledge components on the \emph{test set} and the total number of predicted skills and knowledge tags in each respective split (\#).}
    \label{tab:avglength}
\end{table*}

\subsection{Exact Number of Performance}\label{app:source-perf}
In~\cref{tab:results2}, we show the exact numbers of the plot indicated in~\cref{fig:results2}. In addition, we also show the results of each respective split.

For the STL models, we observe differences in performances over the sources which is particularly pronounced for knowledge components: The \textsc{Tech} source is the easiest to process (and has most SKCs), while SKCs identification performance is the lowest for \textsc{Big}. This might be due to the broad nature of this source. 

In the exact results table (\cref{tab:results2}) we add a ($\dagger$) next to the highest span-F1 if the model is truly stochastically dominant ($\epsilon_\text{min} = $ 0.0) over \emph{all} the other models. (*) denotes that the best model achieved \emph{almost stochastic dominance} ($\epsilon_\text{min} <$ 0.5) over---at minimum---one other model (e.g., in \textsc{Test} rows w.r.t \textsc{Combined}: MTL-JobBERT $\succeq$ MTL-JobSpanBERT with $\epsilon_\text{min} =$ 0.06) and stochastically dominant over the rest.

In~\cref{tab:results2-precision-recall}, we report the precision and recall of the models, \textsc{Skill} and \textsc{Knowledge} show the precision and recall of the STL models. \textsc{Multi} shows the precision and recall of the MTL models.

Last, in \cref{tab:avglength}, we show the exact numbers of the length of predictions~\cref{fig:avglength2}. We also add the number of predicted \textsc{Skill} and \textsc{Knowledge} Overall, JobBERT and JobSpanBERT predict more skills in general than the other models. This is also the case for knowledge components. We hypothesize that this might be due to the BERT models now being more specialized towards the JP domain and recognizing more SKCs.

\begin{figure*}
    \centering
    \includegraphics[width=.5\linewidth]{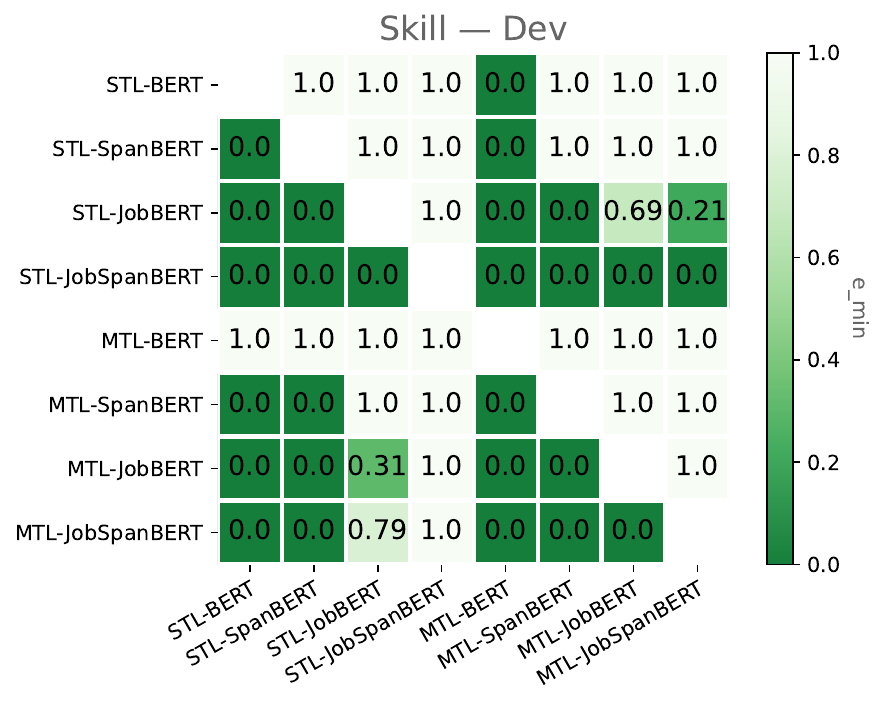}
    \includegraphics[width=.5\linewidth]{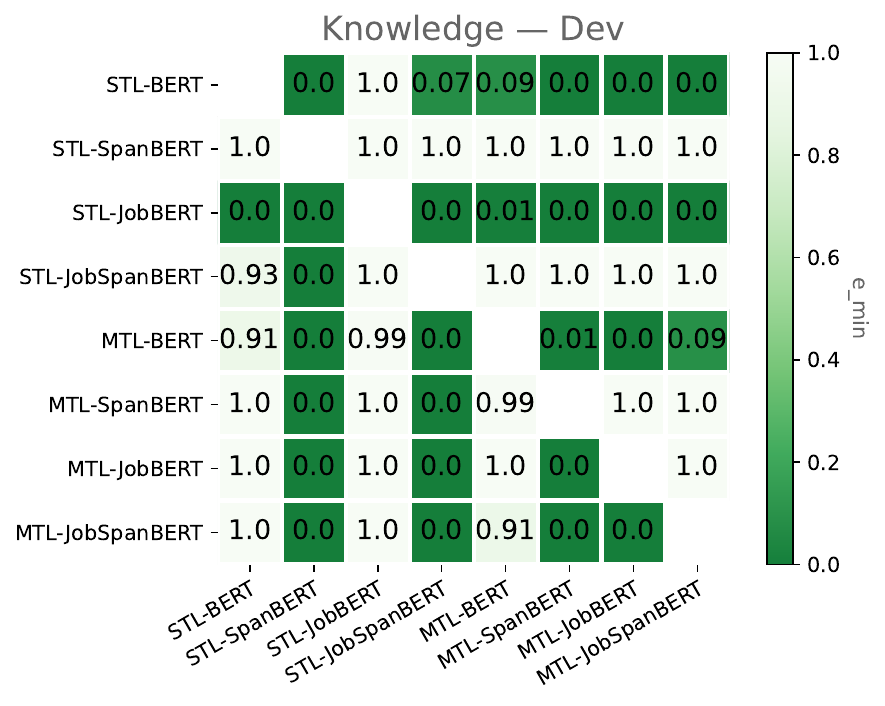}
    \includegraphics[width=.5\linewidth]{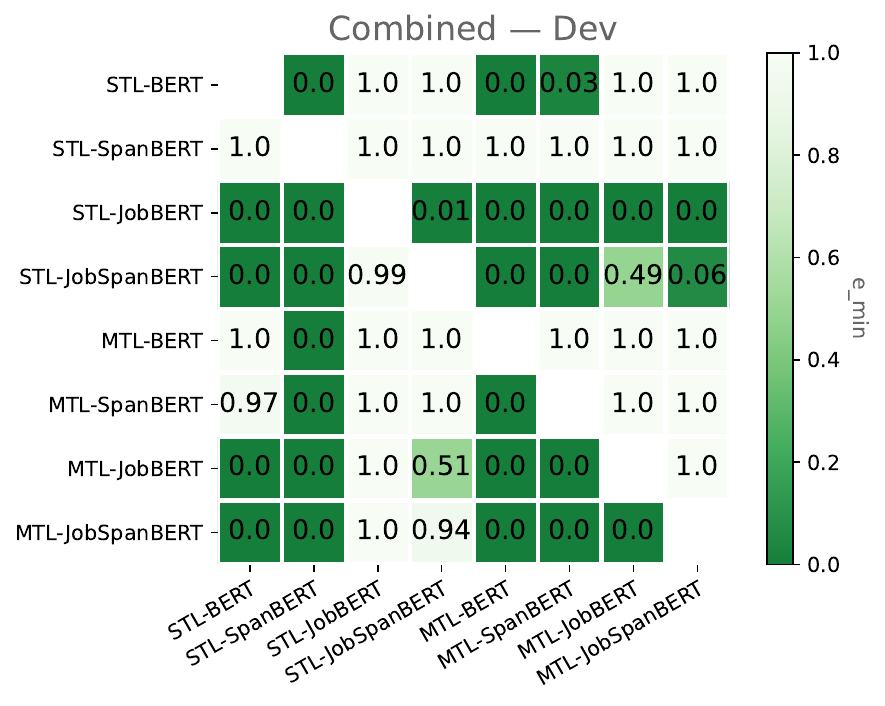}
    \caption{\textbf{Almost Stochastic Order Scores of the Development Set.}
    ASO scores expressed in $\epsilon_\text{min}$.
    The significance level $\alpha =$ 0.05 is adjusted accordingly by using the Bonferroni correction~\citep{bonferroni1936teoria}. Read from row to column: E.g., STL-JobBERT (row) is stochastically dominant over STL-\bertb{} (column) with $\epsilon_\text{min}$ of 0.00.}
    \label{fig:significance-dev2}
\end{figure*}

\subsection{Significance Testing}\label{app:sign-per-source}
Recently, the ASO test~\citep{dror2019deep}\footnote{Implementation of \citet{dror2019deep} can be found at~\url{https://github.com/Kaleidophon/deep-significance}~\citep{dennis_ulmer_2021_4638709}} has been proposed to test statistical significance for deep neural networks over multiple runs.
Generally, the ASO test determines whether a stochastic order~\citep{reimers2018comparing} exists between two models or algorithms based on their respective sets of evaluation scores. Given the single model scores over multiple random seeds of two algorithms $\mathcal{A}$ and $\mathcal{B}$, the method computes a test-specific value ($\epsilon_\text{min}$) that indicates how far algorithm $\mathcal{A}$ is from being significantly better than algorithm $\mathcal{B}$. When distance $\epsilon_\text{min} = 0.0$, one can claim that $\mathcal{A}$ stochastically dominant over $\mathcal{B}$ with a predefined significance level. When $\epsilon_\text{min} < 0.5$ one can say $\mathcal{A} \succeq \mathcal{B}$. On the contrary, when we have $\epsilon_\text{min} = 1.0$, this means $\mathcal{B} \succeq \mathcal{A}$. For $\epsilon_\text{min} = 0.5$, no order can be determined. We took 0.05 for the predefined significance level. In \cref{fig:significance-dev2}, we show the ASO scores on the development set.

\begin{landscape}
\begin{table}
\centering
\resizebox{\linewidth}{!}{
    \begin{tabular}{l|lll}
    \toprule
& \multicolumn{3}{c}{\textbf{Skills}}\\
\midrule
\textbf{Src.} & \textbf{Train} & \textbf{Development} & \textbf{Test}\\
\midrule
\multirow{10}{*}{\rotatebox[origin=l]{90}{\textbf{\bjo{}}}}
    & enthusiastic          & ambitious                              & customer service                                              \\
    & flexible              & proactive                              & communicator                                                  \\
    & team player           & work independently                     & flexible                                                      \\
    & friendly              & attention to detail                    & attention to detail                                           \\
    & attention to detail   & motivated                              & ambitious                                                     \\
    & communicator          & reliable                               & design and refine every touchpoint of the customer journey    \\
    & passionate            & flexible                               & enable inclusion                                              \\
    & communication         & willingness to learn                   & communicate effectively                                       \\
    & confident             & self-motivated                         & interpersonal skills                                          \\
    & flexible approach     & work as part of an established team    & proactive                                                     \\
    \midrule
\multirow{10}{*}{\rotatebox[origin=l]{90}{\textbf{\hou{}}}}
    & communication skills  & structured              & teaching              \\
    & motivated             & teaching                & research              \\
    & structured            & communication skills    & communication skills  \\
    & proactive             & project management      & outgoing              \\
    & analytical            & drive                   & flexible              \\
    & communication         & problem solving         & energetic             \\
    & self-driven           & communication           & responsible           \\
    & team player           & visit customers         & enthusiastic          \\
    & teaching              & curious                 & team player           \\
    & curious               & work independently      & communication         \\
    \midrule
\multirow{10}{*}{\rotatebox[origin=l]{90}{\textbf{\tech{}}}}
    & communication skills                                                      & hands-on               & solving business problems                             \\
    & passionate                                                                & communication skills   & apply your depth of knowledge and expertise           \\
    & apply your depth of knowledge and expertise                               & leadership             & partner continuously with your many stakeholders      \\
    & partner continuously with your many stakeholders                          & passionate             & achieve organizational goals                          \\
    & solving business problems through innovation and engineering practices    & open-minded            & building an innovative culture                        \\
    & work in large collaborative teams                                         & code reviews           & stay focused on common goals                          \\
    & hands-on                                                                  & independent            & work in large collaborative teams                     \\
    & building an innovative culture                                            & software development   & design                                                \\
    & team player                                                               & pioneer new approaches & development                                           \\
    & develop                                                                   & analytical skills      & communicate                                           \\
    \bottomrule
    \end{tabular}}
    \caption{\textbf{Most Frequent Skills in the Data.} Top--10 skill components in our data in terms of frequency.}
    \label{tab:freq-skill2}
\end{table}
\end{landscape}

\begin{table*}
\centering
\resizebox{\linewidth}{!}{
    \begin{tabular}{l|lll}
    \toprule
& \multicolumn{3}{c}{\textbf{Knowledge}}\\
\midrule
\textbf{Src.} & \textbf{Train} & \textbf{Development} & \textbf{Test}\\
\midrule
\multirow{10}{*}{\rotatebox[origin=l]{90}{\textbf{\bjo{}}}}
 & english              & full uk driving licence                                & strategic planning    \\
 & driving license      & sap energy assessments                                 & english               \\
 & excel                & right to work in the uk                                & cscs card             \\
 & cscs card            & sen                                                    & pms                   \\
 & maths                & acca/aca                                               & reservation systems   \\
 & ppc                  & professional kitchen                                   & keynote               \\
 & service design       & cra calculations                                       & illustrator           \\
 & uk/emea policies     & email marketing                                        & aba                   \\
 & bachelor’s degree    & qualitative and quantitative social research methods   & sen                   \\
 & computer science     & care setting                                           & full driving license  \\
    \midrule
\multirow{10}{*}{\rotatebox[origin=l]{90}{\textbf{\hou{}}}}
 & english              & english                                    & english         \\
 & engineering          & supply chain                               & danish          \\
 & computer science     & project management                         & business        \\
 & product management   & powders                                    & java            \\
 & python               & machine learning                           & marketing       \\
 & finance              & phd degree                                 & plm             \\
 & project management   & muscle models with learning and adaptation & production      \\
 & agile                & walking robots                             & supply chain    \\
 & danish               & model rules                                & economics       \\
 & javascript           & capacity development                       & excel           \\
    \midrule
\multirow{10}{*}{\rotatebox[origin=l]{90}{\textbf{\tech{}}}}
 & javascript           & java                 & java          \\
 & python               & javascript           & python        \\
 & java                 & aws                  & .net          \\
 & agile                & docker               & financial services \\
 & financial services   & node.js              & c\#           \\
 & node.js              & typescript           & javascript        \\
 & english              & react                & cloud             \\
 & kubernetes           & linux                & english           \\
 & cloud                & amazon-web-services  & reactjs           \\
 & docker               & devops               & automation        \\
    \bottomrule
    \end{tabular}}
    \caption{\textbf{Most Frequent Knowledge in the Data.} Top--10 knowledge components in our data in terms of frequency.}
    \label{tab:freq-knowledge2}
\end{table*}

\chapter{{K}ompetencer: {F}ine-grained {S}kill {C}lassification in {D}anish {J}ob {P}ostings via {D}istant {S}upervision and {T}ransfer {L}earning}
\chaptermark{{Kompetencer}}
\label{chap:chap4}
The work presented in this chapter is based on a paper that has been published as: \bibentry{zhang-etal-2022-kompetencer}.

\newpage 

\section*{Abstract}
Skill Classification (SC) is the task of classifying job competences from job postings. This work is the first in SC applied to Danish job vacancy data. We release the first Danish job posting dataset: \textsc{Kompetencer} (\emph{en}: competences), annotated for nested spans of competences. To improve upon coarse-grained annotations, we make use of The European Skills, Competences, Qualifications and Occupations (ESCO;~\citealp{le2014esco}) taxonomy API to obtain fine-grained labels via distant supervision. We study two setups: The zero-shot and few-shot classification setting. We fine-tune English-based models and RemBERT~\citep{chung2020rethinking} and compare them to in-language Danish models. Our results show RemBERT significantly outperforms all other models in both the zero-shot and the few-shot setting.

\section{Introduction}

Job Posting data (JPs) is emerging on a variety of platforms in big quantities, and can provide insights on labor market skill set demands and aid job matching~\citep{balog2012expertise}. \emph{Skill Classification} (SC) is to classify competences (i.e., hard and soft skills) necessary for any occupation from unstructured text or JPs. 

Several works focus on Skill Identification~\citep{jia2018representation,sayfullina2018learning,tamburri2020dataops}. This is to classify whether a skill occurs in a sentence or job description. However, continuing the pipeline, there is little work in further categorizing the identified skills by leveraging taxonomies such as ESCO. Another limitation is the scope of language, where all previous work focus on English job postings. This hinders in particular local job seekers from finding an occupation suitable to their specific skills within their community via online job platforms. 

\begin{figure}[t!]
    \centering
    \includegraphics[width=.9\linewidth]{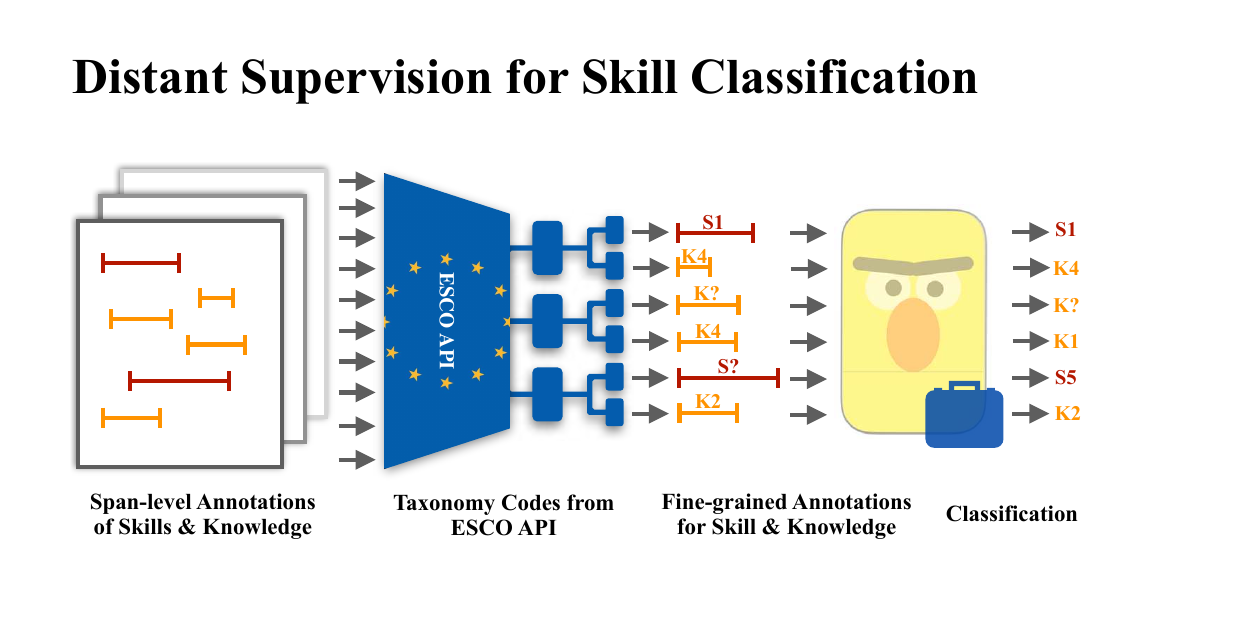}
    \looseness=-1
    \caption{\textbf{Pipeline for Fine-grained Danish Skill Classification.} We propose a distant supervision pipeline, where we have identified spans of skills and knowledge. We query the ESCO API and fine-tune a model on the distantly supervised labels.}
    \label{fig:pipeline}
\end{figure}

In this work, we look into the Danish labor market. We introduce \textsc{Kompetencer}, a novel Danish job posting dataset annotated on the \emph{span-level} for nested \emph{Skill} and \emph{Knowledge} Components (SKCs) in job postings. We do not directly annotate for the fine-grained taxonomy codes from e.g., ESCO, but rather annotate more generic spans of SKCs (\cref{fig:doccano}), and then exploit the ESCO API to bootstrap fine-grained SKCs via distant supervision~\citep{mintz2009distant} and create ``silver'' data for skill classification. Our proposed distant supervision pipeline is denoted in~\cref{fig:pipeline}.

Recently, Natural Language Processing has seen a surge of several transfer learning methods and architecture which help improve state-of-the-art significantly on several tasks~\citep{peters2018deep,howard2018universal,radford2018improving,devlin2019bert}.
In this work, we explore the benefits of zero-shot cross-lingual transfer learning with English \bertb{}~\citep{devlin2019bert} and a \bertb{} that we continuously pretrain~\citep{han-eisenstein-2019-unsupervised,gururangan2020don} on 3.2M English JP sentences and test on Danish and compare it to in-language models: Danish BERT and our domain-adapted Danish BERT model on 24.5M Danish JP sentences. We analyze the zero-shot transfer of English to Danish SC. Last, we experiment with few-shot training: We fine-tune a multilingual model~\citep{chung2020rethinking} on English JPs with a few Danish JPs and show how zero-shot transfer compares to training on a small amount of in-language data.

\paragraph{Contributions.}\circled{1} We release \textsc{Kompetencer},\footnote{\url{https://github.com/jjzha/kompetencer}} the first Danish Skill Classification dataset with distantly supervised fine-grained labels using the ESCO taxonomy. \circled{2} We furthermore present experiments and analysis with in-language Danish models vs.\ a zero-shot cross-lingual transfer from English to Danish with domain-adapted BERT models. \circled{3} We target a few-shot learning setting with a multilingual model trained on both English and a few Danish JPs.

\section{\textsc{Kompetencer} Dataset}\label{sec:data}

\begin{table}[t]
    \centering
    \begin{tabular}{lrr}
    \toprule
    $\downarrow$ \textbf{Statistics}, \textbf{Language} $\rightarrow$ & \textsc{\textbf{English (EN)}} & \textsc{\textbf{Danish (DA)}} \\
    \midrule
    \textbf{\# Posts}                   & 391      &  60    \\
    \textbf{\# Sentences}               & 14,538   &  1,479 \\
    \textbf{\# Tokens}                  & 232,220  &  20,369\\
    \textbf{\# Skill Spans}             & 6,576    &  665   \\
    \textbf{\# Knowledge Spans}         & 6,053    &  255   \\
    $\mathbf{\bar{x}}$ \textbf{Skill Span}        & 3.97  & 3.71\\
    $\mathbf{\bar{x}}$ \textbf{Knowledge Span}    & 1.80  & 1.73\\
    $\mathbf{\tilde{x}}$ \textbf{Skill Span}      & 4     & 3   \\
    $\mathbf{\tilde{x}}$ \textbf{Knowledge Span}  & 1     & 1   \\
    \textbf{Skill [90\%]}                         & [1, 9]& [1, 9]    \\
    \textbf{Knowledge [90\%]}                     & [1, 5]& [1, 4]    \\
    \textbf{Silver fine-grained labels}                      & \cmark & \xmark \\
    \textbf{Gold fine-grained labels}                        & \xmark & \cmark \\
    \bottomrule
    \end{tabular}
    \caption{\textbf{Statistics of Annotated Dataset.} We report the total number of JPs across languages and their respective number of sentences, tokens, and SKCs. Below, we show the mean length of SKCs ($\mathbf{\bar{x}}$), median length of SKCs ($\mathbf{\tilde{x}}$), and the 90th percentile of length [90\%] starting from length 1. We also indicate the type of labels in both sets (silver or gold labels). The EN set is larger than the DA split.}
    \label{tab:num-posts}
\end{table}

\subsection{Skill \& Knowledge Definition}
There is an abundance of competences and there have been large efforts to categorize them. 
The European Skills, Competences, Qualifications and Occupations (ESCO;~\citealp{le2014esco}) taxonomy is the standard terminology linking skills, competences and qualifications to occupations. The ESCO taxonomy mentions three categories of competences: \emph{Knowledge}, \emph{skill}, and \emph{attitudes}. ESCO defines knowledge as follows:
\begin{quote}
    ``Knowledge means the outcome of the assimilation of information through learning. Knowledge is the body of facts, principles, theories and practices that is related to a field of work or study.''~\footnote{\scriptsize{\href{https://ec.europa.eu/esco/portal/escopedia/Knowledge}{\url{ec.europa.eu/esco/portal/escopedia/Knowledge}}}}
\end{quote}
For example, a person can acquire the Python programming language through learning. This is denoted as a \emph{knowledge} component and can be considered generally a \emph{hard skill}. However, one also needs to be able to apply the knowledge component to a certain task. This is known as a \emph{skill} component. ESCO formulates it as:
\begin{quote}
    ``Skill means the ability to apply knowledge and use know-how to complete tasks and solve problems.''~\footnote{\scriptsize{\href{https://ec.europa.eu/esco/portal/escopedia/Skill}{\url{ec.europa.eu/esco/portal/escopedia/Skill}}}}
\end{quote}
 
\noindent
In ESCO, the \emph{soft skills} are referred to as \emph{attitudes}. ESCO considers attitudes as skill components:

\begin{quote}
    ``The ability to use knowledge, skills and personal, social and/or methodological abilities, in work or study situations and professional and personal development.''~\footnote{\scriptsize{\href{http://data.europa.eu/esco/skill/A}{\url{data.europa.eu/esco/skill/A}}}}
\end{quote}

\noindent
To sum up, hard skills are usually referred to as \emph{knowledge} components, and applying these hard skills to something is considered a \emph{skill}. Then, soft skills are referred to as \emph{attitudes}, these are part of skill components. There has been no work, to the best of our knowledge, in annotating skill and knowledge components in JPs.

\begin{figure}[t]
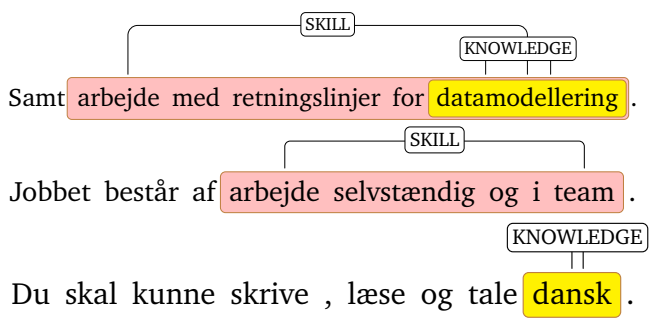

\centering
\begin{adjustbox}{width=.7\textwidth}
\begin{dependency}[edge slant=0pt, edge vertical padding=3pt, edge style={-}]
    \begin{deptext}
    Samt \& arbejde \&  med \&  retningslinjer \&  for \&  datamodellering \& . \\
    \end{deptext}
    \wordgroup[group style={fill=pink, draw=brown, inner sep=.3ex}]{1}{2}{6}{sk}
    \wordgroup[group style={fill=yellow, draw=brown, inner sep=.1ex}]{1}{6}{6}{kn}
    \depedge[edge height=4ex]{2}{6}{SKILL}
    \depedge[edge height=2ex, edge horizontal padding=-18pt, edge end x offset=10pt]{6}{6}{KNOWLEDGE}
\end{dependency}
\end{adjustbox}

\begin{adjustbox}{width=.7\textwidth}
\begin{dependency}[edge slant=0pt, edge vertical padding=2pt, edge style={-}]
    \begin{deptext}
    Jobbet \& består \& af \& arbejde \& selvstændig \& og \& i \& team \&.\\
    \end{deptext}
    \wordgroup[group style={fill=pink, draw=brown, inner sep=.1ex}]{1}{4}{8}{sk}
    \depedge[edge height=2ex]{4}{8}{SKILL}
\end{dependency}
\end{adjustbox}

\begin{adjustbox}{width=.7\textwidth}
\begin{dependency}[edge slant=0pt, edge vertical padding=2pt, edge style={-}]
    \begin{deptext}
    Du \& skal \& kunne \& skrive \& , \& læse \& og \& tale \& dansk \& .\\
    \end{deptext}
    \wordgroup[group style={fill=yellow, draw=brown, inner sep=.1ex}]{1}{9}{9}{kn}
    \depedge[edge height=2ex]{9}{9}{KNOWLEDGE}
\end{dependency}
\end{adjustbox}

    \caption{\textbf{Examples of Skills and Knowledge Components.} Annotated samples of passages in varying Danish job postings. SKCs can be nested as shown in the first example.}
    \label{fig:doccano}
\end{figure}

\subsection{Dataset Statistics}
Both the English and Danish data comes from a large job platform with various types of JPs.\footnote{We release the annotated spans in~\url{https://github.com/jjzha/kompetencer/tree/master/data}} The English JPs are from~\citet{zhang-etal-2022-skillspan}.  In~\cref{tab:num-posts}, we show the statistics of both the annotated English and Danish data split. We note that the number of English JPs is larger than the Danish split. For Danish, there are fewer knowledge spans proportional to English. Apart from this, both the English and Danish JPs follow a similar trend in terms of statistics. The mean length of skills and knowledge ($\mathbf{\bar{x}}$) is slightly shorter for Danish, 3.97 vs. 3.71 and 1.80 vs. 1.73 respectively. The median length of skills ($\mathbf{\tilde{x}}$) is one token shorter for Danish. However, we note again that the length of skills can vary substantially, ranging from 1--9 for both languages. Then, for knowledge components this ranges from 1--5 and 1--4 for English and Danish respectively. The similarity in statistics shows the consistency of annotations, which we elaborate on in the next section.

\cref{fig:doccano} shows some examples of the annotated SKCs. ``Samt arbejde med retningslinjer for datamodellering'' (\emph{en}: ``As well as working with guidelines for data modeling''), shows a nesting example: ``datamodellering'' shows a knowledge component (i.e., something that one can learn), and the skill is to apply it. ``Jobbet består af arbejde selvstændig og i team'' (\emph{en}: The job consists of working independently and in a team) indicates an \emph{attitude} as ``working independently or in a team'' is a social ability. We furthermore consider languages a knowledge component, as one can acquire the language through schooling. Overall, the classification of the spans could be a short sentence (i.e., $\leq$9 tokens) or single token classification.

\begin{figure}[t!]
    \centering
    \includegraphics[width=\linewidth]{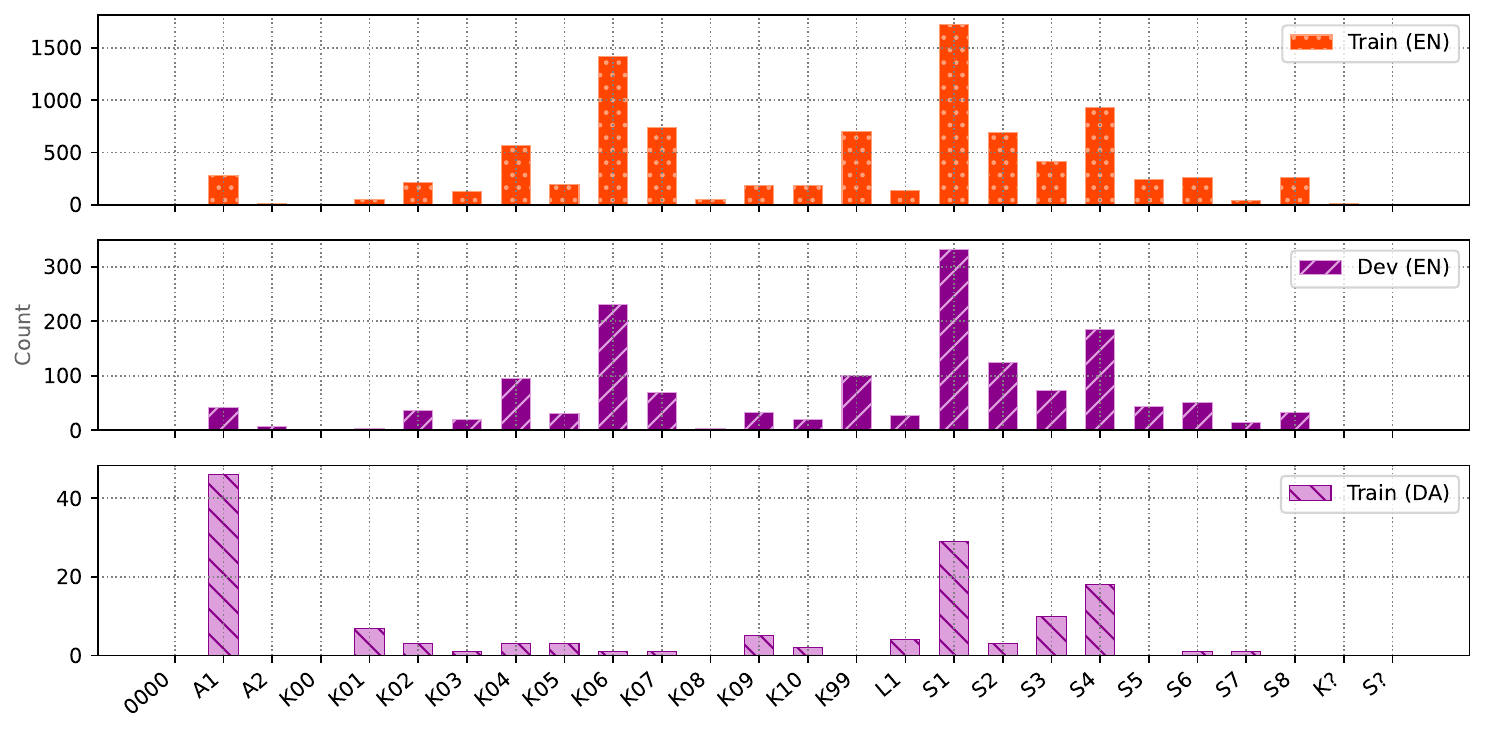}
    \caption{\textbf{Label Distribution of Distantly Supervised Labels.} In the top and middle barplot we show the fine-grained label distribution of the English training and development split respectively. The splits follow a similar distribution. For the Danish training split on the bottom, there is a large increase of \texttt{A1} labels, which indicate more \emph{attitude}-like skills. All splits have a larger fraction of the label \texttt{S1}, which encapsulates communicative skills. Explanations of labels are given in~\cref{tab:lab1},~\cref{tab:lab2}, and~\cref{tab:lab3} (\cref{labelmeaning}, Appendix).}
    \label{fig:distribution}
\end{figure}
\subsection{Annotations}

\paragraph{Skill Identification Annotations} We annotate with the annotation guidelines denoted in~\citet{zhang-etal-2022-skillspan} used on the English data split to identify the SKCs in a JP. There are around 57.5K tokens (approximately 4.6K sentences, in 101 job posts) that was used to calculated agreement on. 
The annotations were compared using Cohen's $\kappa$~\citep{fleiss1973equivalence} between pairs of annotators, and Fleiss' $\kappa$~\citep{fleiss1971measuring}, which generalizes Cohen's $\kappa$ to more than two concurrent annotations. We consider two levels of $\kappa$ calculations: \textbf{\textsc{Token}} is calculated on the token level, comparing the agreement of annotators on each token (including non-entities) in the annotated dataset.~\textbf{\textsc{Span}} refers to the agreement between annotators on the exact span match over the surface string, regardless of the type of named entity, i.e., we only check the position of tag without regarding the type of the named entity. The observed agreement scores over the three annotators is between 0.70--0.75 Fleiss' $\kappa$ for both levels of calculation which is considered a \emph{substantial agreement}~\citep{landis1977measurement}. Then, for the Danish data split, we use the same guidelines as for English. Here, we consider one annotator that annotates for the SKCs.

\begin{algorithm}[t]
\SetAlgoLined
    $X \gets \text{Top-100 query results from ESCO}$\;
    $X \gets \{x \mid \texttt{typeof}(x) = \text{Type}\}$\;
    $d \gets \infty$\;
    $r \gets \text{None}$\;
    \For{$x \in X$}{ \hfill\tcp{Find Skill in the ESCO API}
        $D \gets \text{levenshtein}(x, \text{Skill})$\;
        \If{$D = 0$}{
            \textbf{return} $x$ \tcp*{Perfect match}
        }
        \If{$D < d$}{
            $r \gets x$\;
            $d \gets D$\;
        }
    }
    \textbf{return} $r$ \tcp*{Best match based on Levenshtein distance}
\caption{Getting the best match for a skill in the ESCO API using Levenshtein distance}
\label{alg:cap}
\end{algorithm}

\begin{figure}[t!]
    \centering
    \includegraphics[width=\linewidth]{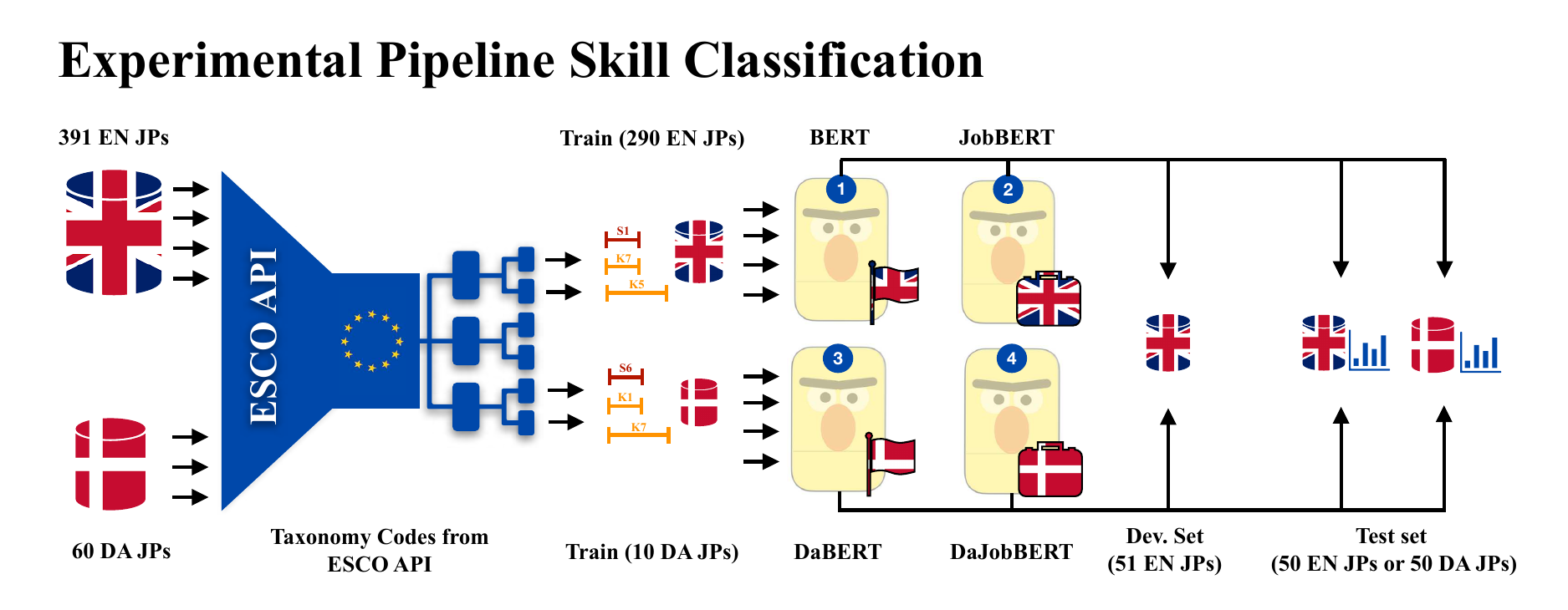}
    \caption{\textbf{Experimental Pipeline for Fine-grained Danish Skill Classification.} Read from left to right, we start with each respective dataset for English (EN) and Danish (DA). We obtain the labels from the ESCO API and train for each language split two models: For EN, these are (1) \bertb{} and (2) JobBERT. For DA, these are (3) DaBERT and (4) DaJobBERT. The Danish data is split into 10/50 train/test and the English data in to 290/51/50 train/dev/test JPs. The Danish models are fine-tuned on the Danish train set, and use \emph{no} in-language development set (i.e., English dev.). In the end all models are applied to the Danish and English test set separately.}
    \label{fig:experiment-pipeline}
\end{figure}

\paragraph{Fine-grained Annotations}\label{sec:fine-anno} Currently, our proposed dataset consists of identified SKCs. To obtain fine-grained labels of each span, we explore distant supervision using the ESCO API, where the setup is broadly depicted in~\cref{fig:pipeline}. The annotated spans are queried to the API, then via~\cref{alg:cap}, we determine whether the obtained SKC is ``relevant'' or not via Levenshtein distance matching~\citep{levenshtein1966binary}. In addition, we determine the quality of the distant supervised labels by human evaluation. We manually check each of the annotated spans to its obtained label from the ESCO API. After checking a subset 2,622 English labels --- without correcting --- and its distantly supervised labels, we obtain 41.3\% accuracy on the correctness of the distantly supervised labels. We note that across all 9,473 labels in the original English training and development data (details of train/dev/test splits in~\cref{sec:expsetup}), a total of 7.4\% is unidentified by the ESCO database, and is thus labeled by \texttt{K99} from ESCO in the resulting train and development data here. For the Danish data, we obtain 70.4\% accuracy on the training set and 20.2\% is missing, albeit the Danish training set only contains 138 SKCs. For the Danish test set, we correct the distantly supervised labels to create a gold test set. Here, 14.1\% was initially correct and 23.5\% missed a label. In~\cref{fig:distribution}, we show the distantly supervised fine-grained label distribution of the English training and development set split, and the Danish training split. The following labels: \texttt{0000}, \texttt{K?}, and \texttt{S?} are artifacts of querying the ESCO API (i.e., unidentified skills). We did not employ any post-processing and left them as is. We presumed they would not influence the model significantly as their numbers are low.

\section{Methodology}\label{sec:method}
In the current setting, we have annotated spans of SKCs. We extract the spans from the JPs and query the ESCO API to obtain silver labels. We formulate this task as a text classification problem. We consider a set of JPs $\mathcal{D}$, where $d \in \mathcal{D}$ is a set of extracted spans (\emph{and not full sentences}) with the $i^\text{th}$ span $\mathcal{X}^i_{d} = \{x_1, x_2, ..., x_T\}$ and a target class $c \in \mathcal{C}$, where $\mathcal{C} = \{\text{\texttt{S*}}, \text{\texttt{K*}}\}$. The labels \texttt{S*} and \texttt{K*} depend on the distantly supervised ESCO taxonomy code (e.g., S4: Management Skills,\footnote{\url{http://data.europa.eu/esco/skill/S4}} K2: Arts and Humanities\footnote{\url{http://data.europa.eu/esco/isced-f/02}}). The goal of this task is then to use $\mathcal{D}$ to train an algorithm $h: \mathcal{X} \mapsto \mathcal{C}$ to accurately predict skill tags by assigning an output label $c$ for input $\mathcal{X}^i_{d}$.

\subsection{Encoders}
As baseline for Danish SC, we consider a Danish BERT (\textbf{DaBERT}) encoder.\footnote{\url{https://huggingface.co/Maltehb/danish-bert-botxo}} Following~\citet{gururangan2020don}, we continuously pretrain DaBERT on 24.5M Danish JP sentences for \emph{one} epoch, we name this \textbf{DaJobBERT}.\footnote{\url{https://huggingface.co/jjzha/dajobbert-base-cased}}
To test zero-shot performance from English to Danish for SC, we use \textbf{\bertb{}}~\citep{devlin2019bert} and a domain-adapted \bertb{} model on 3.2M JP sentences, namely \textbf{JobBERT}~\citep{zhang-etal-2022-skillspan}. We assume that domain-adapted models like JobBERT and DaJobBERT would improve SC as the ``domain'' is the same.

\subsection{Multilingual Encoder}
We explore whether using a multilingual encoder would benefit the classification of skills for Danish in a low-resource setting. For the experiments we use \textbf{RemBERT}~\citep{chung2020rethinking}, it has recently shown to outperform mBERT~\citep{devlin2019bert} on several tasks. All models are using a final Softmax layer for the classification of spans.

\subsection{Experimental Setup}\label{sec:expsetup}
Our detailed experimental setup is shown in~\cref{fig:experiment-pipeline}. We start with 391 English and 60 Danish job postings (\cref{tab:num-posts}) annotated with spans of SKCs. The spans are then queried to the ESCO API (\cref{fig:pipeline}). We split the English data into 290 train (9,472 SKCs), 51 dev (1,577 SKCs), and 50 JPs for test (1,578 SKCs), and for the Danish data we split this into 10 JPs (138 SKCs) for training and 50 JPs for test (782 SKCs). For the label distribution we refer back to~\cref{fig:distribution} (excl. test).

We fine-tune \bertb{} and JobBERT on the spans in 290 English JPs. Next, we fine-tune DaBERT and DaJobBERT on the 10 Danish JPs. For RemBERT, we fine-tune in three ways: Only on English, only on Danish, and on both English and Danish together.
For all setups, we choose the model with the best score on the English dev.\ set. As pointed out by~\citet{artetxe2020call}: Pure unsupervised cross-lingual transfer should not use any cross-lingual signal by definition. As our attention is on Danish, we do not use any Danish labeled training data \emph{nor} dev.\ data in the zero-shot setting. All models in the end will be tested on the held-out 50 English and Danish JPs separately.\footnote{The English test set contains silver labels (distantly supervised), while the Danish test set is human corrected (gold).} In summary, we have three setups: (1) Fine-tuned on English JPs only (BERT, JobBERT, RemBERT), (2) fine-tuned on Danish JPs only (DaBERT, DaJobBERT, RemBERT), and (3) fine-tuned on both English and Danish JPs (RemBERT). We consider (1) a zero-shot setting, while (2) and (3) do have access to some Danish training data, hence this is a few-shot setting. Throughout the experiments, we use the \textsc{MaChAmp} (v0.3) toolkit~\citep{van-der-goot-etal-2021-massive} for classification. All reported results are the average over five runs with different random seeds on weighted macro-F1.

\begin{figure}[t!]
    \centering
    \includegraphics[width=\linewidth]{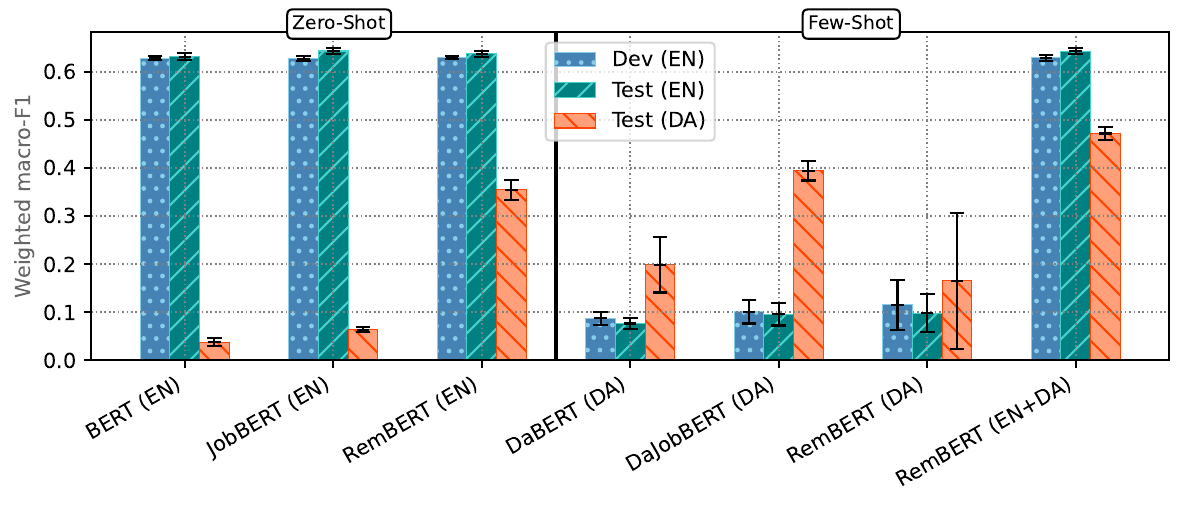}
    \looseness=-1
    \caption{\textbf{Performance of Models on English and Danish.} We test seven setups on several splits of data: English development (\textsc{\textbf{Dev (EN)}}), English test set (\textsc{\textbf{Test (EN)}}), and Danish test set (\textsc{\textbf{Test (DA)}}). Reported is the weighted macro-F1. The whiskers indicate each respective standard deviation of runs on five random seeds. Left side of the black vertical line indicates a full zero-shot setting on \textsc{Test (DA)}, on the right shows the few-shot setting on the same test set. With respect to the models, language abbreviation in brackets (e.g., \textbf{BERT (EN)}) indicates what it has been fine-tuned on. Exact numbers including significance testing are noted in~\cref{tab:exactresults} (\cref{app:results}, Appendix).}
    \label{fig:results4}
\end{figure}

\section{Analysis of Results}

We show the experimental results in~\cref{fig:results4}. Plotted is the weighted macro-F1 of all three setups with seven models and their corresponding standard deviation on the English development set, English test set, and the Danish test set. All models left of the black vertical line are the zero-shot setup, applied to Danish. On the right, these models are in the few-shot setting, this is due to the model having access to some target language training data (DA). 

\subsection{Performance Zero-shot Setting}
For the models trained on English only (BERT, JobBERT, and RemBERT (EN)) when applied to the English development set, all three models perform similarly. They achieve around 0.63--0.64 weighted macro-F1 with little standard deviation: \bertb{} \std{0.628}{0.004}, JobBERT \std{0.628}{0.006}, and RemBERT (EN) \std{0.629}{0.003} weighted macro-F1. Similarly for the English test set: \bertb{} \std{0.632}{0.007}, JobBERT \std{0.644}{0.006}, and RemBERT (EN) \std{0.637}{0.007} weighted macro-F1, where JobBERT significantly outperforms all other models (details in \cref{app:results}).

It is a tacit that the English-based models perform better than the baseline (DaBERT) on English, both dev.\ and test. 
Conversely, the English-based models perform poorly on the Danish test set: \bertb{} \std{0.038}{0.008} and JobBERT \std{0.063}{0.005} weighted macro-F1. However, given a multilingual encoder (RemBERT) only trained on English, gives a significant gain in zero-shot performance (\std{0.354}{0.021}) with little standard deviation and significantly outperforms the other zero-shot setting models including the target-language baseline (DaBERT). We strongly suspect this is due to Danish being included in the pretraining data of RemBERT.

\subsection{Performance Few-shot Setting} Apart from RemBERT (EN+DA) having access to English data, all other models fine-tuned on Danish perform poorly on English dev.\ and test. The performance of RemBERT (DA) is slightly better than the best performing Danish-only model DaJobBERT (\std{0.098}{0.040} vs.\std{0.096}{0.024} weighted macro-F1 on English test), where our intuition again goes to the pretraining data.

For DA test, DaBERT is a strong baseline, achieving \std{0.199}{0.058} weighted macro-F1 with little Danish training data. RemBERT (DA) did not result in significant gains having pretrained on multiple languages and another intuition could be that this is a result of negative transfer~\citep{rosenstein2005transfer}. Then, DaJobBERT performs better than DaBERT on the Danish test set: \std{0.395}{0.021} weighted macro-F1. Note that we conducted domain adaptive pretraining from the DaBERT checkpoint on 24.5M Danish JP sentences for one epoch with the Masked Language Modeling objective. This shows that in-language \emph{and} in-domain pretraining is beneficial for this specific task of SC.

\subsection{Combining Training Data} Last, giving RemBERT all training data (English and Danish) results in substantial improvement over all other models in the zero-shot and few-shot setting alike: \std{0.472}{0.014}, which significantly outperforms all other models on Danish test. Henceforth, it is helpful to have a bit of target-language training data for higher resulting performance.

\subsection{Is Domain Adaptive Pretraining Worth It?} In light of the results, domain adaptive pretraining shows its benefit for both English and Danish fine-tuning. Specifically for Danish, from the baseline (DaBERT), the improvement is close to 0.2 weighted macro-F1 with DaJobBERT. The domain adaptive pretraining took $\sim$35 hours, using 4 GPUs, to pass once over the unlabeled data (24.5M Danish JP sentences). The largest gain is obtained with combining both English and Danish training data: The improvement is around 0.27 weighted macro-F1. However, the 391 EN and 60 DA JPs took around two months of non-stop annotating. In short, there is a trade-off between continuous pretraining on unlabeled text and annotating: (1) Domain adaptive pretraining gives short-term gains with little costs, but there needs to be enough unlabeled data in the right domain. (2) Annotating extra data results in larger gains long-term, but there is more costs involved.

\begin{figure}[t]
    \centering
        \includegraphics[width=.75\linewidth]{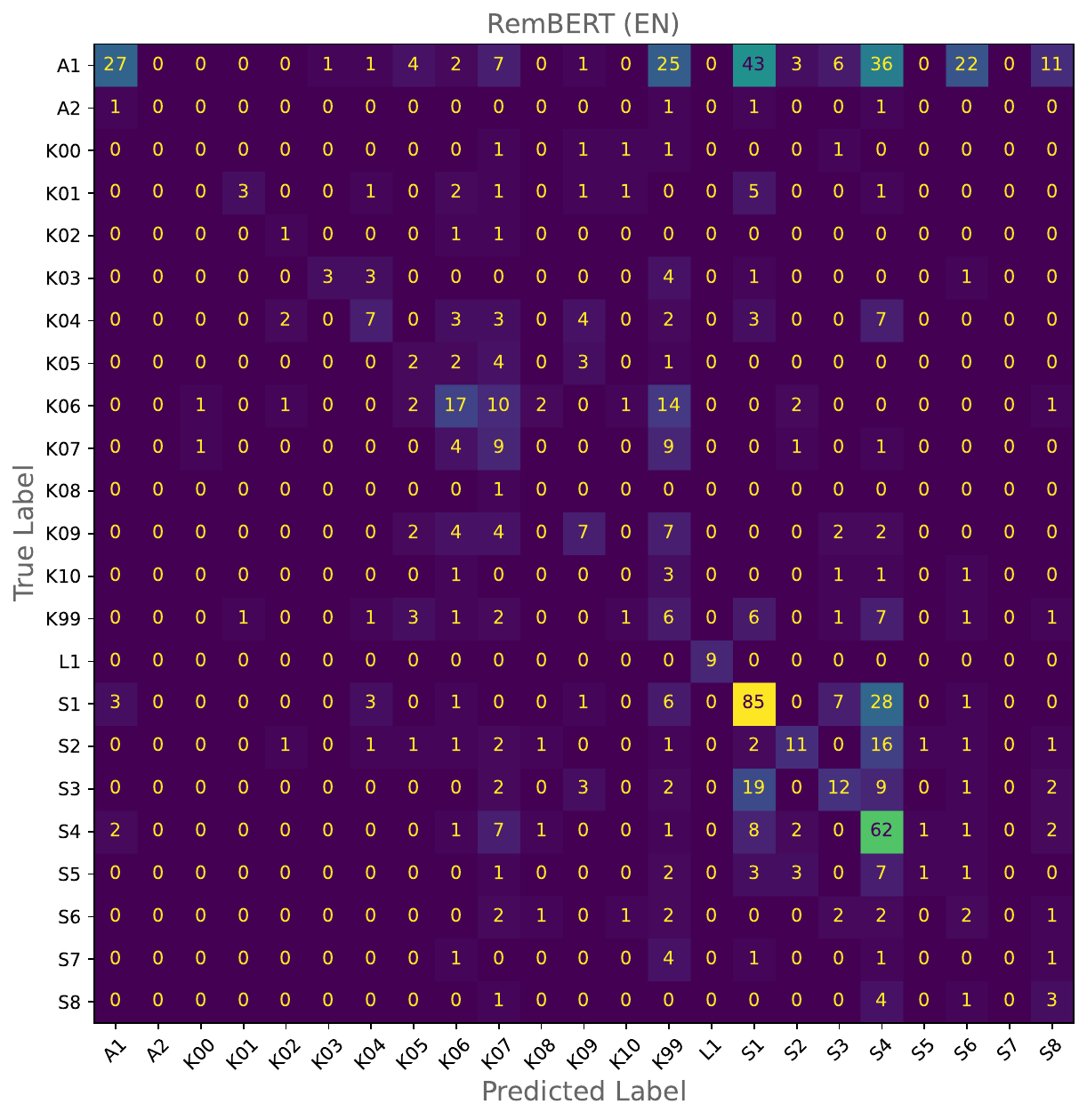}
    \caption{\textbf{Confusion Matrix of RemBERT (EN).} We show the confusion matrix of the zero-shot setting with RemBERT (EN). On the diagonal are the correctly predicted labels. Most of the ``confusion'' is with respect to the labels that encompass the larger fraction of the test set: \texttt{A1}: Attitudes and \texttt{S1}: Communication, collaboration and creativity.}
    \label{fig:cm-en}
\end{figure}

\subsection{Analysis of Predictions} In~\cref{fig:cm-en}, we show the confusion matrix of the best performing zero-shot model on the test set of the best run and investigate what the model does not predict correctly. In the matrix, the model mostly confuses the label \texttt{A1}, which relates to \emph{attitudes} and gets predicted as \texttt{S1}: Communication, collaboration and creativity. There could be some overlap between these labels as for example the skill ``effektiv'' (\emph{en}: efficient/effective). This is officially labeled as an attitude by ESCO, but a grey area is that ``effective'' could relate to ``creativity''. 

There is also a small cluster of confusion from \texttt{S1-4}. These are rather distinct classes of skills. For example, \texttt{S4} means management skills. A specific example is ``fagligt velfunderet'' (\emph{en}: professionally sound), this could be an attitude. This is all hard to determine since there is no context around the skill. Overall, there is some confusion between the skills when taken out of context. We leave the exploration of fine-grained skill classification \emph{with context} for future work.

\subsection{Qualitative Analysis Distant Supervision}\label{sec:quality}
We analyze the label selection method and the missing labels in the English dataset as mentioned in \cref{sec:fine-anno}. 
We find that the missing labels in the English data is predominantly coming from technical skills. We found that the missing spans are mostly knowledge components in the form of technologies used today by developers, such as ReactJS, Django, AWS, etc.\ This lack of coverage could either be due to specificity or the ever-growing list of technologies. In ESCO, there are several technologies that are listed (e.g., NoSQL, Drupal, WordPress to name a few), but there are also a lot missing (e.g., TensorFlow, Data Science, etc.).

\section{Related Work}

Many works focus on the identification of skills in job descriptions, i.e., whether a sentence contains a skill or not~\citep{sayfullina2018learning,tamburri2020dataops} or what the necessary skills are inferred from an entire job posting~\citep{bhola-etal-2020-retrieving}. We instead identified the SKCs manually in the job descriptions on the sentence-level, as this gives us the highest quality of identified SKCs. Furthermore, there are several works in fine-grained SC (i.e., categorize the skills), but mostly focus on English job descriptions. A straightforward approach is to do exact matching with a predefined list of skills~\citep{malherbe2016bridge,papoutsoglou2017mining,sibarani2017ontology} or do a frequency analysis of skills, cluster them by hand and attach a more general category to them e.g.,~\citet{gardiner2018skill}. 

Some works have used the ESCO taxonomy directly~\citep{boselli2018classifying,giabelli2020graphlmi}. For example,~\citet{boselli2018classifying} classified both titles and descriptions for its most suitable ISCO~\citep{elias1997occupational} code (what ESCO is partially based on). However, they only gave one label to each data point (i.e., full job posting), which is unrealistic as most occupations require multiple competences.

Overall, to the best of our knowledge, there seems to be little to no work in directly classifying the identified SKC to a specific ESCO label. In addition, this work is the first of its kind for Danish JPs.

\section{Conclusion}
We present a novel skill classification dataset for competences in Danish: \textsc{Kompetencer}.\footnote{We release the Danish anonymized raw data and annotations of the parts with permissible licenses from a govermental agency which is our collaborator. Links to our English data can be found at \url{https://github.com/kris927b/SkillSpan}. For anonymization, we perform it via manual annotation of job-related sensitive and personal data regarding \texttt{Organization}, \texttt{Location}, \texttt{Contact}, and \texttt{Name} following the work by~\citet{jensen-etal-2021-de}.} In addition, we transform the coarse-grained human annotated spans to more fine-grained labels via distant supervision with the ESCO API. Our human evaluation shows that the distantly supervised labels give a signal of correctly annotated spans, where we achieve 41.3\% accuracy on a large English label subset, and 70.4\% accuracy on the Danish dev set, and 14.1\% accuracy on the Danish test set. We manually correct the Danish test set with the correct labels from ESCO to create a gold annotated set and keep the English labels as is, and thus silver labels. 

Furthermore, domain adaptive pretraining helps to improve performance on the task specifically for English. The best performance is achieved with RemBERT on both the zero-shot setting (\std{0.354}{0.021} weighted macro-F1) and few-shot setting (\std{0.472}{0.014} weighted macro-F1), where they significantly outperform the other models. The strong performance is likely due to the pretraining data that contains both Danish and English.

Last, since the annotations are on the token level, this work can be extended to, for example, sequence labeling. We hope this dataset initiates further research in the area of skill classification.

\clearpage
\section{Appendix}
\subsection{Data Statement \textsc{Kompetencer}}\label{app:datastatement}
Following~\citet{bender-friedman-2018-data}, the following outlines the data statement for \textsc{Kompetencer}:
\begin{enumerate}[A.]
    \itemsep0em
    \item \textsc{Curation Rationale}: Collection of job postings in the English and Danish language for skill classification, to study the impact of skill changes from job postings.
    \item \textsc{Language Variety}: The non-canonical data was collected from the StackOverflow job posting platform, an in-house job posting collection from our national labor agency collaboration partner (the Danish Agency for Labor Market and Recruitment\footnote{https://www.star.dk/en/}), and web extracted job postings from a large job posting platform. US (en-US), British (en-GB) English, and Danish (da-DK) are involved.
    \item \textsc{Speaker Demographic}: Gender, age, race-ethnicity, socioeconomic status are unknown.
    \item \textsc{Annotator Demographic}: Three hired project participants (age range: 25--30), gender: one female and two males, white European and Asian (non-Hispanic). Native language: Danish, Dutch. Socioeconomic status: higher-education students. Female annotator is a professional annotator with a background in Linguistics and the two males with a background in Computer Science.
    \item \textsc{Speech Situation}: Standard American, British English or Danish is used in job postings. Time frame of the data is between 2012--2021.
    \item \textsc{Text Characteristics}: Sentences are from job postings posted on official job vacancy platforms.
    \item \textsc{Recording Quality}: N/A.
    \item \textsc{Other}: N/A.
    \item \textsc{Provenance Appendix}: The Danish job posting data is from our collaborators: The Danish Agency for Labour Market and Recruitment (STAR).
\end{enumerate}

\begin{table}[t]
    \centering
    \begin{tabular}{lrr}
    \toprule
    \textsc{\textbf{Parameter}} & \textsc{\textbf{Value}} & \textsc{\textbf{Range}} \\
    \midrule
    Optimizer                           & AdamW                & \\
    $\beta_\text{1}$, $\beta_\text{2}$  & 0.9, 0.99            & \\
    Dropout                             & 0.2                  & 0.1, 0.2, 0.3\\
    Epochs                              & 20                   & \\
    Batch Size                          & 32                   & \\
    Learning Rate (LR)                  & 1e-4                 & 1e-3, 1e-4, 1e-5\\
    LR scheduler                        & Slanted triangular   & \\
    Weight decay                        & 0.01                 & \\
    Decay factor                        & 0.38                 & 0.35, 0.38, 0.5\\
    Cut fraction                        & 0.2                  & 0.1, 0.2, 0.3\\
    \bottomrule
    \end{tabular}
    \caption{\textbf{Hyperparameters of \textsc{MaChAmp}.}}
    \label{tab:hyperparameters4}
\end{table}

\subsection{Reproducibility}\label{app:hyper}

We use the default hyperparameters in \textsc{MaChAmp}~\citep{van-der-goot-etal-2021-massive} as shown in~\cref{tab:hyperparameters4}. For more details we refer to their paper. For the five random seeds we use 3477689, 4213916, 6828303, 8749520, and 9364029. All experiments with \textsc{MaChAmp} were ran on an NVIDIA\textsuperscript{\textregistered} NVIDIA A100-SXM4 40GB GPU
and an AMD\textsuperscript{\textregistered} EPYC 7662 64-Core Processor.

\subsection{Label Meaning}\label{labelmeaning}
In~\cref{tab:lab1}, \cref{tab:lab2}, and \cref{tab:lab3}, we show the definitions of the labels.
\begin{table*}[t]
    \footnotesize
    \centering
    \begin{tabularx}{\textwidth}{lXX}
    \toprule
    \textsc{\textbf{Label}} & \textsc{\textbf{Subject}} & \textsc{\textbf{Definition}} \\
    \midrule
    0000    & \texttt{ARTIFACT} & \texttt{ARTIFACT}  \\\hline
    A1      & Attitudes & Individual work styles that can affect how well someone performs a job.  \\\hline
    A2      & Values & Principles or standards of behavior, revealing one's judgment of what is important in life.   \\\hline
    K00     & Generic programmes and qualifications & Generic programmes and qualifications are those providing fundamental and personal skills education which cover a broad range of subjects and do not emphasise or specialise in a particular broad or narrow field. \\\hline
    K01     &  Education & \texttt{NO-DEFINITION} \\\hline
    K02     &  Arts and humanities & \texttt{NO-DEFINITION} \\\hline
    K03     &  Social sciences, journalism and information & \texttt{NO-DEFINITION}\\\hline
    K04     &  Business, administration and law & \texttt{NO-DEFINITION} \\\hline
    K05     &  Natural sciences, mathematics and statistics & \texttt{NO-DEFINITION}\\\hline
    K06     &  Information and communication technologies (icts) & \texttt{NO-DEFINITION}\\\hline
    K07     &  Engineering, manufacturing and construction not elsewhere classified & \texttt{NO-DEFINITION}\\\hline
    K08     & Agriculture, forestry, fisheries and veterinary & \texttt{NO-DEFINITION}\\\hline
    K09     & Health and welfare & \texttt{NO-DEFINITION}\\\hline
    K10     & Services & \texttt{NO-DEFINITION}\\\hline
    K99     & Field unknown & \texttt{NO-DEFINITION}\\\hline
    L1      & Languages & Ability to communicate through reading, writing, speaking and listening in the mother tongue and/or in a foreign language. \\
    \bottomrule
    \end{tabularx}%
    \caption{\textbf{Definition of ESCO Labels (Part 1).} Indicated are the definitions of the ESCO labels used in this work taken from the ESCO taxonomy. Artifacts of the ESCO API are \texttt{K?} and \texttt{S?}, and \texttt{0000}, this means that no component was found.}
    \label{tab:lab1}
\end{table*}
\begin{table*}[t]
    \footnotesize
    \centering
    \begin{tabularx}{\textwidth}{lXX}
    \toprule
    \textsc{\textbf{Label}} & \textsc{\textbf{Subject}} & \textsc{\textbf{Definition}} \\
    \midrule
    S1      & Communication, collaboration and creativity &  Communicating, collaborating, liaising, and negotiating with other people, developing solutions to problems, creating plans or specifications for the design of objects and systems, composing text or music, performing to entertain an audience, and imparting knowledge to others.\\\hline
    S2      & Information skills & Collecting, storing, monitoring, and using information; Conducting studies, investigations and tests; maintaining records; managing, evaluating, processing, analysing and monitoring information and projecting outcomes.\\\hline
    S3      & Assisting and caring & Providing assistance, nurturing, care, service and support to people, and ensuring compliance to rules, standards, guidelines or laws.\\\hline
    S4      & Management skills & Managing people, activities, resources, and organisation; developing objectives and strategies, organising work activities, allocating and controlling resources and leading, motivating, recruiting and supervising people and teams.\\\hline
    S5      & Working with computers & Using computers and other digital tools to develop, install and maintain ICT software and infrastructure and to browse, search, filter, organise, store, retrieve, and analyse data, to collaborate and communicate with others, to create and edit new content.\\
    \bottomrule
    \end{tabularx}%
    \caption{\textbf{Definition of ESCO Labels (Part 2).} Indicated are the definitions of the ESCO labels used in this work taken from the ESCO taxonomy. Artifacts of the ESCO API are \texttt{K?} and \texttt{S?}, and \texttt{0000}, this means that no component was found.}
    \label{tab:lab2}
\end{table*}
\begin{table*}[t]
    \footnotesize
    \centering
    \begin{tabularx}{\textwidth}{lXX}
    \toprule
    \textsc{\textbf{Label}} & \textsc{\textbf{Subject}} & \textsc{\textbf{Definition}} \\
    \midrule
    S6      & Handling and moving & Sorting, arranging, moving, transforming, fabricating and cleaning goods and materials by hand or using handheld tools and equipment. Tending plants, crops and animals.\\\hline
    S7      & Constructing & Building, repairing, installing and finishing interior and exterior structures.\\\hline
    S8      & Working with machinery and specialised equipment & Controlling, operating and monitoring vehicles, stationary and mobile machinery and precision instrumentation and equipment. \\\hline
    K?      & \texttt{ARTIFACT} & \texttt{ARTIFACT}  \\\hline
    S?      & \texttt{ARTIFACT} & \texttt{ARTIFACT}  \\
    \bottomrule
    \end{tabularx}%
    \caption{\textbf{Definition of ESCO Labels (Part 3).} Indicated are the definitions of the ESCO labels used in this work taken from the ESCO taxonomy. Artifacts of the ESCO API are \texttt{K?} and \texttt{S?}, and \texttt{0000}, this means that no component was found.}
    \label{tab:lab3}
\end{table*}

\begin{figure}[t]
    \centering
    \includegraphics[width=.5\linewidth]{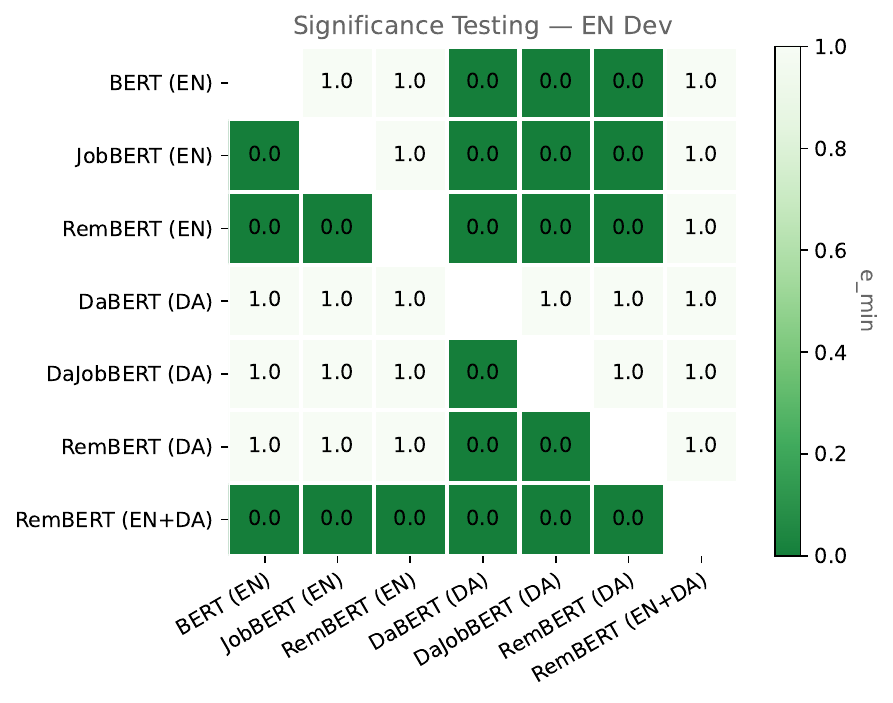}
    \includegraphics[width=.5\linewidth]{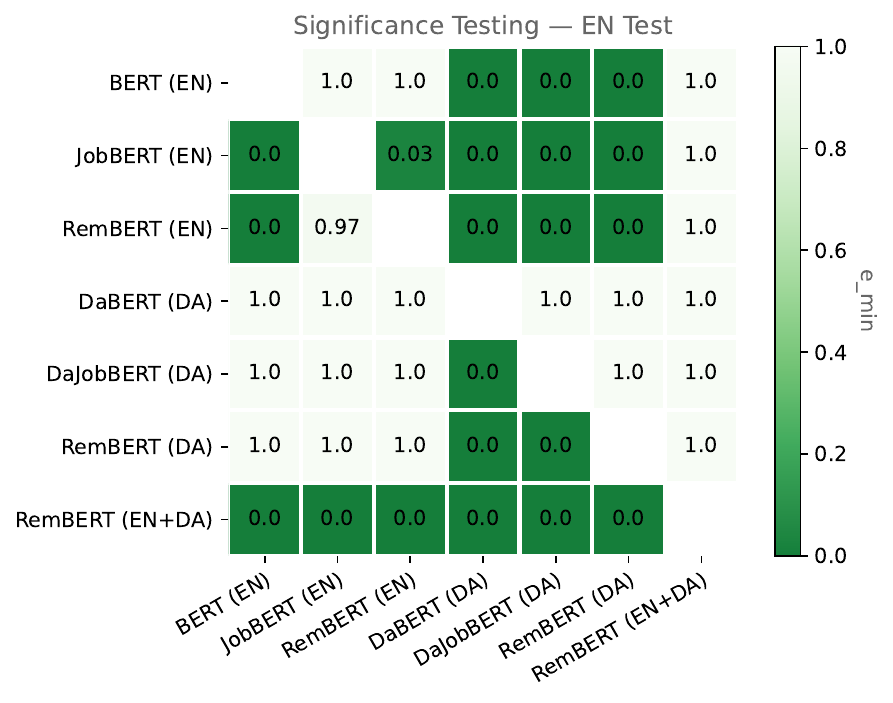}
    \includegraphics[width=.5\linewidth]{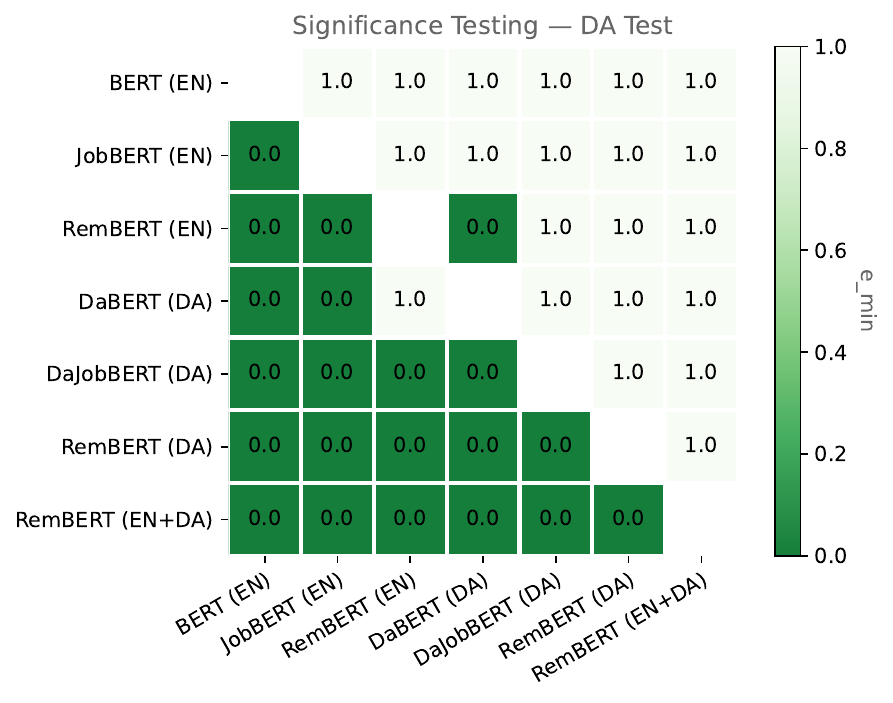}
    \caption{\textbf{Results Almost Stochastic Order.}
    ASO scores are expressed in $\epsilon_\text{min}$.
    The significance level $\alpha =$ 0.05 is adjusted accordingly by using the Bonferroni correction~\protect\citep{bonferroni1936teoria}. Almost stochastic dominance ($\epsilon_\text{min} < 0.5$) is indicated in the colored boxes: On \textsc{\textbf{EN Test}}, JobBERT is almost stochastically dominant over RemBERT (EN), with $\epsilon_\text{min} = 0.03$.}
    \label{fig:aso_test}
\end{figure}

\begin{table}[t]
    \centering
    \begin{tabular}{l|r|r|r}
    \toprule
    \textsc{\textbf{Model}} & \textsc{\textbf{EN Dev}} & \textsc{\textbf{EN Test}} & \textsc{\textbf{DA Test}} \\
    \midrule
    \bertb{} (EN)        & \std{0.628}{0.004}  & \std{0.632}{0.007}     & \std{0.038}{0.008}\\
    JobBERT (EN)         & \std{0.628}{0.006}  & \textbf{\std{0.644}{0.006}*}     & \std{0.063}{0.005}\\
    RemBERT (EN)         & \textbf{\std{0.629}{0.003}}  & \std{0.637}{0.007}     & \std{0.354}{0.021}\\
    \midrule
    DaBERT (DA)         & \std{0.088}{0.013}  & \std{0.076}{0.012}      & \std{0.199}{0.058}\\
    DaJobBERT (DA)      & \std{0.101}{0.024}  & \std{0.096}{0.024}      & \std{0.395}{0.021}\\
    RemBERT (DA)        & \std{0.116}{0.052}  & \std{0.098}{0.040}      & \std{0.166}{0.141}\\
    RemBERT (EN+DA)     & \textbf{\std{0.629}{0.006}*}  & \std{0.643}{0.006}      & \textbf{\std{0.472}{0.014}*}\\
    \bottomrule
    \end{tabular}
    \caption{\textbf{Exact Results on Splits.} Indicated are the exact results of the bar plots in~\cref{fig:results4}. Significance tested with Almost Stochastic Order~\protect\citep{dror2019deep} test with Bonferroni correction~\protect\citep{bonferroni1936teoria}. Bold indicates highest average weighted macro-F1 and asterisk indicates significance.}
    \label{tab:exactresults}
\end{table}

\subsection{Exact Results from Plots}\label{app:results}

In~\cref{tab:exactresults}, we show the exact results of the plots from~\cref{fig:results4} on English dev, English test, and Danish test respectively. In addition, we do significance testing. Recently, the Almost Stochastic Order (ASO) test~\citep{dror2019deep}\footnote{Implementation of~\citet{dror2019deep} can be found at~\url{https://github.com/Kaleidophon/deep-significance}~\citep{dennis_ulmer_2021_4638709}} has been proposed to test statistical significance for deep neural networks over multiple runs.
Generally, the ASO test determines whether a stochastic order~\citep{reimers2018comparing} exists between two models or algorithms based on their respective sets of evaluation scores. Given the single model scores over multiple random seeds of two algorithms $\mathcal{A}$ and $\mathcal{B}$, the method computes a test-specific value ($\epsilon_\text{min}$) that indicates how far algorithm $\mathcal{A}$ is from being significantly better than algorithm $\mathcal{B}$. When distance $\epsilon_\text{min} = 0.0$, one can claim that $\mathcal{A}$ stochastically dominant over $\mathcal{B}$ with a predefined significance level. When $\epsilon_\text{min} < 0.5$ one can say $\mathcal{A} \succeq \mathcal{B}$. On the contrary, when we have $\epsilon_\text{min} = 1.0$, this means $\mathcal{B} \succeq \mathcal{A}$. For $\epsilon_\text{min} = 0.5$, no order can be determined. We compared all pairs of models based on five random seeds each using ASO with a confidence level of $\alpha =$ 0.05 (before adjusting for all pair-wise comparisons using the Bonferroni correction~\citep{bonferroni1936teoria}). Almost stochastic dominance ($\epsilon_\text{min} < 0.5$) is indicated in~\cref{fig:aso_test} over all the splits.

\begin{figure}[ht]
    \centering
        \includegraphics[width=\linewidth]{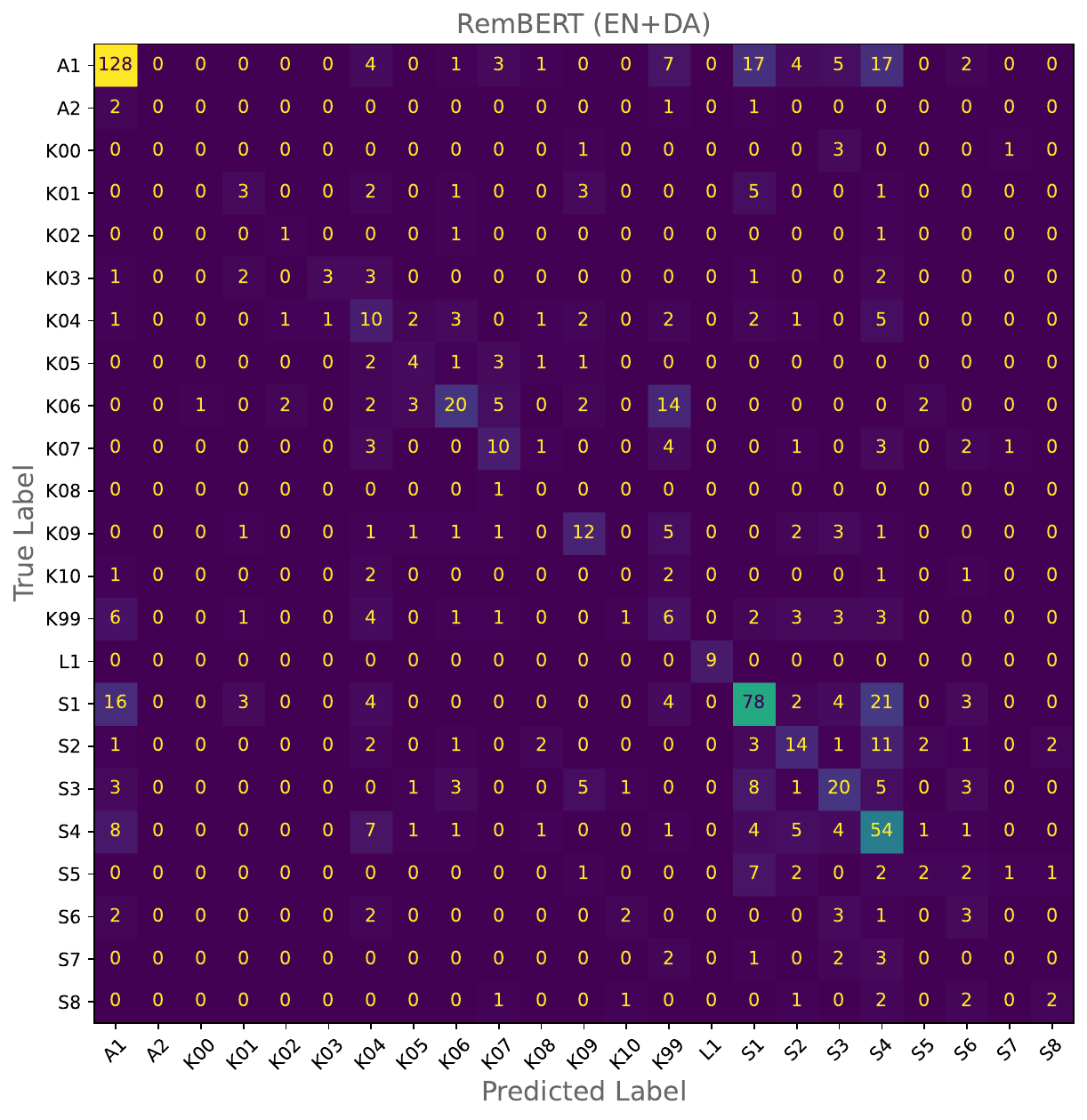}
    \caption{\textbf{Confusion Matrix of RemBERT (EN+DA).} We show the confusion matrix of the few-shot setting with RemBERT (EN+DA). On the diagonal are the correctly predicted labels. There is less confusion in this model as compared to RemBERT (EN). We suspect the additional Danish data benefits the prediction of \texttt{A1}.}
    \label{fig:cm}
\end{figure}

\subsection{Confusion Matrix Few-Shot}

In~\cref{fig:cm}, we show the confusion matrix of the best performing few-shot model on the test set of the best run and investigate what the model does not predict correctly. Dissimilar from \cref{fig:cm-en}, we only seem some confusion in the small cluster of \texttt{S1-4}. Giving the model a few Danish JPs substantially improved the prediction of \texttt{A1}, which relates to \emph{attitudes} and gets predicted as \texttt{S1}: Communication, collaboration, and creativity.

\newpage
\part{Modeling Occupational Skills}
\newpage

\chapter{{S}kill {E}xtraction from {J}ob {P}ostings using {W}eak {S}upervision}
\chaptermark{{Skill Extraction with Weak Supervision}}
\label{chap:chap5}
The work presented in this chapter is based on a paper that has been published as: \bibentry{zhang2022skill}.

\newcommand{\say}{\textit{Sayfullina}}
\newcommand{\sks}{\textit{SkillSpan}}

\newpage 

\section*{Abstract}
Aggregated data obtained from job postings provide powerful insights into labor market demands, and emerging skills, and aid job matching. However, most extraction approaches are supervised and thus need costly and time-consuming annotation. To overcome this, we propose Skill Extraction with Weak Supervision. We leverage the European Skills, Competences, Qualifications and Occupations taxonomy to find similar skills in job ads via latent representations. The method shows a strong positive signal, outperforming baselines based on token-level and syntactic patterns.

\section{Introduction}\label{intro}

The labor market is under constant development---often due to changes in technology, migration, and digitization---and so are the skill sets required~\citep{brynjolfsson2011race,brynjolfsson2014second}. Consequentially, large quantities of job vacancy data is emerging on a variety of platforms. Insights from this data on labor market skill set demands could aid, for instance, job matching~\citep{balog2012expertise}. The task of automatic \emph{skill extraction} (SE) is to extract the competences necessary for any occupation from unstructured text. 

Previous work on supervised SE frame it as a sequence labeling task (e.g.,~\citep{sayfullina2018learning,tamburri2020dataops,chernova2020occupational,zhang-jensen-plank:2022:LREC,zhang-etal-2022-skillspan,green-maynard-lin:2022:LREC,gnehm-bhlmann-clematide:2022:LREC}) or multi-label classification \citep{bhola-etal-2020-retrieving}. Annotation is a costly and time-consuming process with little annotation guidelines to work with. This could be alleviated by using predefined skill inventories. 

\begin{figure}
    \centering
    \includegraphics[width=.6\columnwidth]{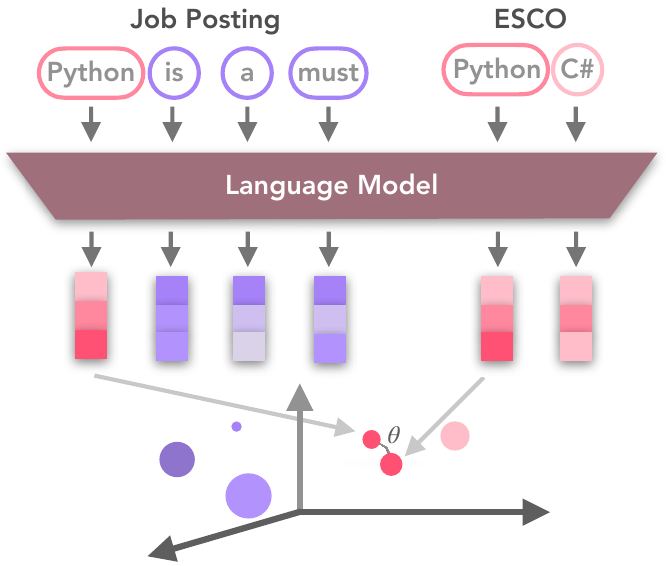}
    \caption{\textbf{Weakly Supervised Skill Extraction.} All ESCO skills and n-grams are extracted and embedded through a language model, e.g., RoBERTa~\citep{liu2019roberta}, to get representations. We label \emph{spans} from job postings close in vector space to the ESCO skill.}
    \label{fig:fig1}
\end{figure}

In this work, we approach span-level SE with weak supervision: 
We leverage the European Skills, Competences, Qualifications and Occupations (ESCO;~\citep{le2014esco}) taxonomy and find similar spans that relate to ESCO skills in embedding space (\cref{fig:fig1}). The advantages are twofold: First, labeling skills becomes obsolete, which mitigates the cumbersome process of annotation. Second, by extracting skill phrases, this could possibly enrich skill inventories (e.g., ESCO) by finding paraphrases of existing skills. We seek to answer: \emph{How viable is Weak Supervision in the context of SE?} We contribute: \circled{1} A novel weakly supervised method for SE; \circled{2} A linguistic analysis of ESCO skills and their presence in job postings;
\circled{3} An empirical analysis of different embedding pooling methods for SE for two skill-based datasets.\footnote{\url{https://github.com/jjzha/skill-extraction-weak-supervision}}

\begin{figure}[t]
    \centering
    \includegraphics[width=\linewidth]{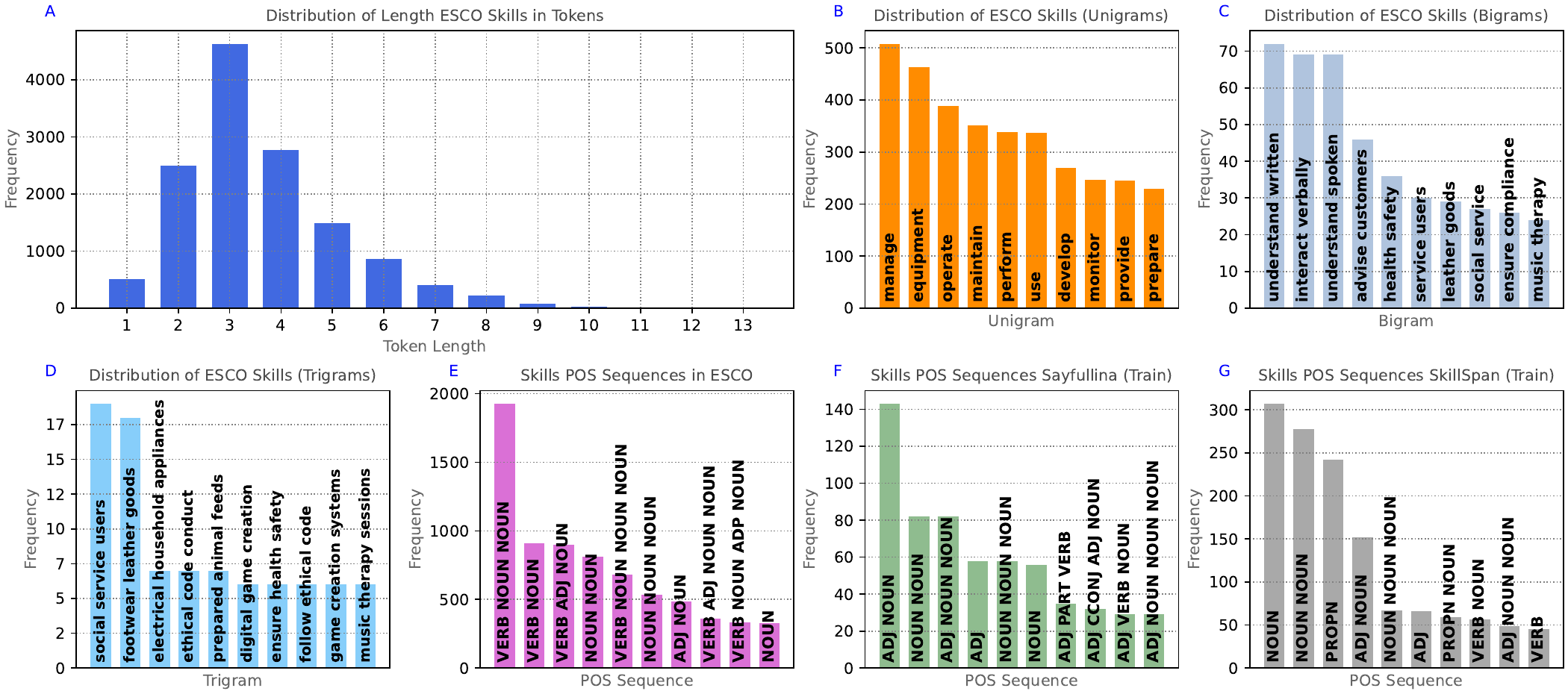} 
    \caption{\textbf{Surface-level Statistics of ESCO.} We show various statistics of ESCO. (\textbf{A}) ESCO skills token length, the mode is three tokens. (\textbf{B}) Most frequent unigrams of ESCO skills. (\textbf{C}) Most frequent bigrams of ESCO skills. (\textbf{D}) Most frequent trigrams of ESCO skills. (\textbf{E}) Most frequent POS sequences of ESCO skills. Last, we show the POS sequences of unique skills in both train sets of \say{} and \sks{} (\textbf{F}--\textbf{G}).}
    \label{fig:esco-stats}
\end{figure}

\section{Methodology}
Formally, we consider a set of job postings $\mathcal{D}$, where $d \in \mathcal{D}$ is a set of sequences (e.g., job posting sentences) with the $i^\text{th}$ input sequence $\mathcal{T}^i_{d} = [t_1, t_2, ..., t_n]$ and a target sequence of \texttt{BIO}-labels $\mathcal{Y}^i_{d} = [y_1, y_2, ..., y_n]$ (e.g., ``\texttt{B-SKILL}'', ``\texttt{I-SKILL}'', ``\texttt{O}'').\footnote{Definition of labels can be found in~\citep{zhang-etal-2022-skillspan}.} The goal is to use an algorithm, which predicts skill spans by assigning an output label sequence $\mathcal{Y}^i_{d}$ for each token sequence $\mathcal{T}^i_{d}$ from a job posting based on representational similarity of a span to any skill in ESCO.

\begin{table}[t]
\centering
\begin{tabular}{llrr}
\toprule
& \textbf{Statistics} & \textbf{\say{}} & \textbf{\sks{}} \\
\midrule
\multirow{3}{*}{\rotatebox[origin=c]{90}{\textsc{\textbf{Train}}}}
  & \textbf{\# Sentences}              & 3,703     & 5,866     \\
  & \textbf{\# Tokens}                 & 53,095    & 122,608   \\
  & \textbf{\# Skill Spans}            & 3,703     & 3,325     \\
\midrule
\multirow{3}{*}{\rotatebox[origin=c]{90}{\textsc{\textbf{Dev.}}}}
  & \textbf{\# Sentences}              & 1,856     & 3,992     \\
  & \textbf{\# Tokens}                 & 26,519    & 52,084     \\
  & \textbf{\# Skill Spans}            & 1,856     & 2,697     \\
\midrule
\multirow{3}{*}{\rotatebox[origin=c]{90}{\textsc{\textbf{Test}}}}
  & \textbf{\# Sentences}              & 1,848     & 4,680     \\
  & \textbf{\# Tokens}                 & 26,569    & 57,528    \\
  & \textbf{\# Skill Spans}            & 1,848     & 3,093     \\
\midrule
 & \textbf{Avg.\ Len.\ Skills }             & 1.77     & 2.92     \\
  \bottomrule
    \end{tabular}
    \caption{\textbf{Statistics of Datasets.} Indicated is each dataset and their respective number of sentences, tokens, skill spans, and the average length of skills in tokens.}
    \label{tab:num_post}
\end{table}

\subsection{Data}
We use the datasets from~\citep{zhang-etal-2022-skillspan} (\sks{}) and a modification of~\citep{sayfullina2018learning} (\say{}).\footnote{In contrast to \sks{}, \say{} has a skill in every sentence, where they focus on categorizing sentences for soft skills.}
In~\cref{tab:num_post}, we show the statistics of both. \sks{} contain nested labels for skill and knowledge components~\citep{le2014esco}. To make it fit for our weak supervision approach, we simplify their dataset by considering both skills and knowledge labels as one label (i.e., \texttt{B-KNOWLEDGE} becomes \texttt{B-SKILL}).%

\subsection{ESCO Statistics} 
We use ESCO as a weak supervision signal for discovering skills in job postings. There are 13,890 ESCO skills.\footnote{Per 25-03-2022, taking ESCO \texttt{v1.0.9}.} In~\cref{fig:esco-stats}, we show statistics of the taxonomy: (A) On average most skills are 3 tokens long. In (C-D), we show 
n-grams frequencies with range [$1;3$]. We can see that the most frequent uni- and bigrams are verbs, while the most frequent trigrams consist of nouns. 

Additionally, we show an analysis of ESCO skills from a linguistic perspective. We tag the training data using the publicly available MaChAmp v0.2 model~\citep{van-der-goot-etal-2021-massive} trained on all Universal Dependencies 2.7 treebanks~\citep{11234/1-3687}.\footnote{A Udify-based~\citep{kondratyuk201975} multi-task model for POS, lemmatization, dependency parsing, built on top of the \texttt{transformers} library~\citep{wolf-etal-2020-transformers}, and specifically using mBERT~\citep{devlin2019bert}.} Then, we count the most frequent Part-of-Speech (POS) tags in all sources of data (E-G). ESCO's most frequent tag sequences are \texttt{VERB-NOUN}, these are not as frequent in \say{} nor \sks{}. \say{} mostly consists of adjectives, which is attributed to the categorization of soft skills. \sks{} mostly consists of \texttt{NOUN} sequences. Overall, we observe most skills consist of verb and noun phrases.

\begin{figure*}[t]
    \centering
    \includegraphics[width=\linewidth]{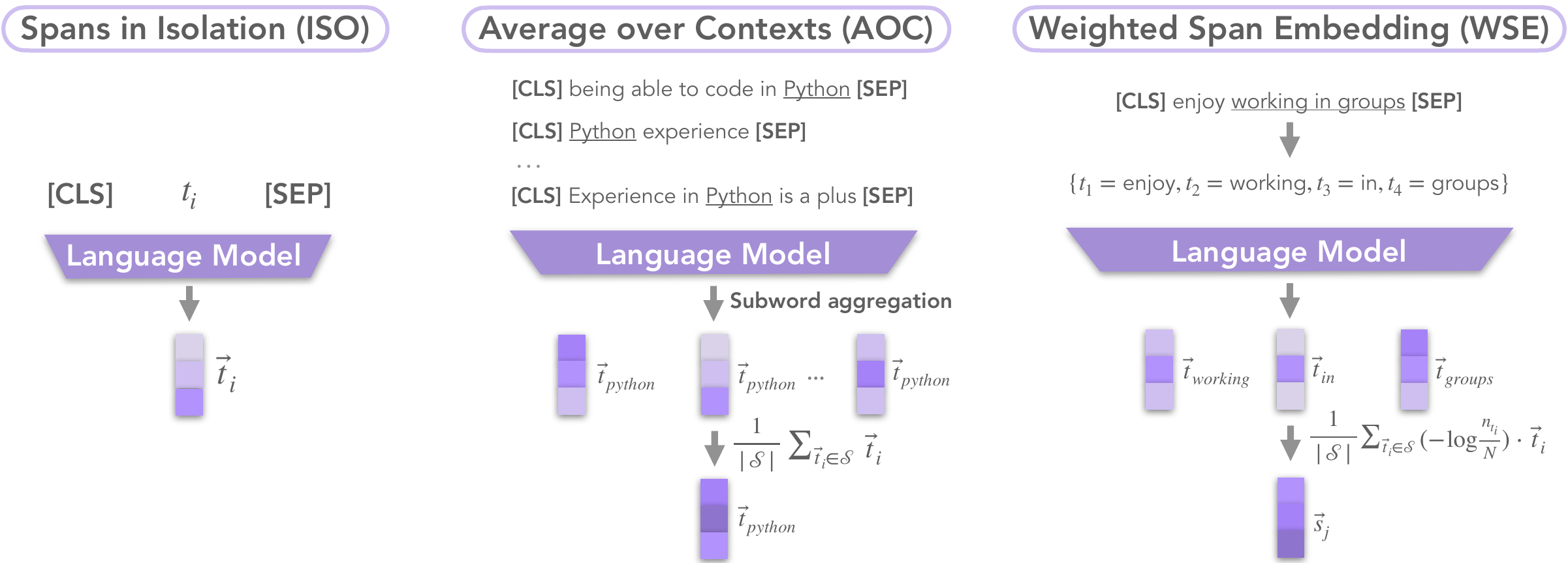}
    \caption{\textbf{Skill Representations.} We show different methods to embed ESCO skill phrases. The approaches are inspired by \protect\citet{litschko2022cross}. We embed a skill by encoding it directly without surrounding context (\textbf{left}). We aggregate different contextual representations of the same skill term (\textbf{middle}). Last, we encode the skill phrase via a weighted sum of embeddings with each token's inverse document frequency as weight (\textbf{right}). For the middle and right methods, $\mathcal{S}$ is the number of sentences where the ESCO skill appears.}
    \label{fig:vectors}
\end{figure*}

\subsection{Baselines}
As our approach is to find similar n-grams based on ESCO skills, we choose an n-gram range of $[1;4]$ (where $4$ is the median) derived from~\cref{fig:esco-stats} (A). For higher matching probability, we apply an additional pre-processing step to the ESCO skills by removing non-tokens (e.g., brackets) and words between brackets (e.g., ``Java (programming)'' becomes ``Java''). We have three baselines: 

\textbf{Exact Match}: We do exact substring matching with ESCO and the sentences in both datasets.

\textbf{Lemmatized Match}: ESCO skills are written in the infinitive form. We take the same approach as exact match on the training sets, now with the lemmatized data of both. The data is lemmatized with MaChAmp v0.2~\citep{van-der-goot-etal-2021-massive}.

\textbf{POS Sequence Match}: Motivated by the observation that certain POS sequences often overlap between sources (\cref{fig:esco-stats}, E-G), we attempt to match POS sequences within ESCO with the POS sequences in the datasets. For example \texttt{NOUN-NOUN}, \texttt{NOUN}, \texttt{VERB-NOUN} and \texttt{ADJ-NOUN} sequences are commonly occurring in all three sources.

\begin{algorithm}[t]
\SetAlgoLined
$M \gets \{\text{RoBERTa}, \text{JobBERT}\}$\;
$E \gets \{\text{ISO},\text{AOC},\text{WSE}\}$\;
$\tau \gets [0,1]$\;
$D$ \hfill\tcp*[f]{A set of sentences from job postings}\;
$L \gets \emptyset$\;
$P \gets D$ \hfill\tcp*[f]{A set of sentences from job postings}\;
$S \gets S_E$ \hfill\tcp*[f]{ESCO Skill embeddings of type $E$}\;
\For{$p \in P$}{
    $\theta \gets 0$\;
    \For{$n \in \text{split}(p)$}{ \hfill\tcp*[f]{Each ngram $n$ of size $1-4$}
        $E \gets M(n)$\;
        $\Theta \gets \text{CosSim}(S, E)$\;
        \If{$\max(\Theta) > \tau \land \max(\Theta) > \theta$}{
            $\theta \gets \max(\Theta)$\;
        }
    }
    $L \gets L \cup [\theta]$\;
}
\textbf{return} $L$\;
\caption{Weakly Supervised Skill Extraction}
\label{alg:cap5}
\end{algorithm}

\subsection{Skill Representations}\label{subsec:rep}

We investigate several encoding strategies to match n-gram representations to embedded ESCO skills, the approaches are inspired by~\citet{litschko2022cross}, where they applied them to Information Retrieval. The language models (LMs) used to encode the data are RoBERTa~\citep{liu2019roberta} and the domain-specific JobBERT~\citep{zhang-etal-2022-skillspan}. All obtained vector representations of skill phrases with the three previous encoding methods are compared pairwise with each n-gram created from \say{} and \sks{}. An explanation of the methods (see~\cref{fig:vectors}):

\textbf{Span in Isolation (ISO)}: We encode skill phrases $t$ from ESCO in isolation using the aforementioned LMs, without surrounding contexts.

\textbf{Average over Contexts (AOC)}: We leverage the surrounding context of a skill phrase $t$ by collecting all the sentences containing $t$. We use all available sentences in the job postings dataset (excluding \textsc{Test}). For a given job posting sentence, we encode $t$ by using one of the previous mentioned LMs. We average the embeddings of its constituent subwords to obtain the final embedding $t$.

\textbf{Weighted Span Embedding (WSE)}: We obtain all inverse document frequency (idf) values of each token $t_i$ via 

\begin{displaymath}
\text{idf} = -\text{log}\frac{n_{t_i}}{N},
\end{displaymath} 
where $n_{t_i}$ is the number of occurrences of $t_i$ and $N$ the total number of tokens in our dataset. We encode the input sentence and compute the weighted sum of the embeddings ($\vec{s}_j$) of the specific skill phrase in the sentence, where each $t_i$'s IDF scores are used as weights. Again, we only use the first subword token for each tokenized word. Formally, this is 

\begin{displaymath}
\vec{s}_j = \sum_{\vec{t}_i} (-\text{log}\frac{n_{t_i}}{N}) \cdot \vec{t}_i.
\end{displaymath}

\begin{figure*}[t]
    \centering
    \includegraphics[width=\linewidth]{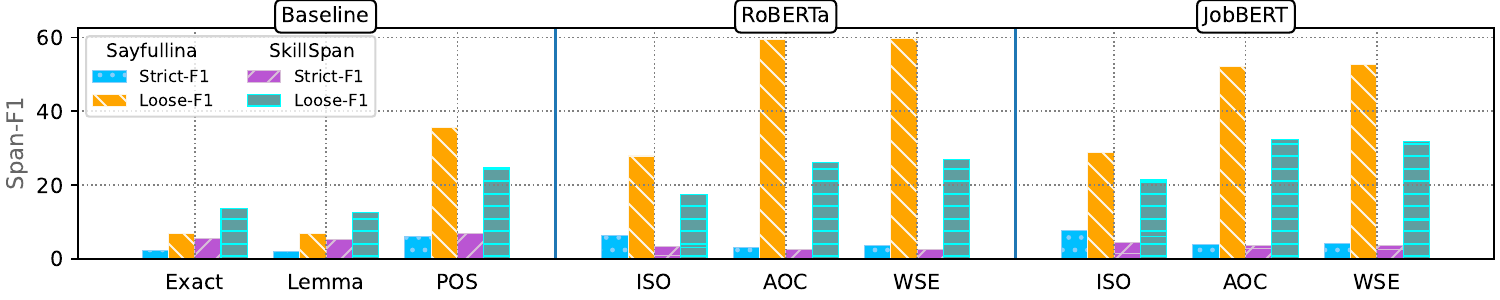}
    \looseness=-1
    \caption{\textbf{Results of Methods.} Results on \say{} and \sks{} are indicated by ``Baseline'' showing performance of Exact, Lemmatized (Lemma), and Part-of-Speech (POS). The performance of ISO, AOC, and WSE are separated by model, indicated by ``RoBERTa'' and ``JobBERT''. The performance of RoBERTa and JobBERT on \sks{} is determined by the best performing CosSim threshold (0.8).} %
    \label{fig:results-f1}
\end{figure*}

\begin{table*}[ht]
    \centering
    \begin{adjustbox}{width=\textwidth}
    \begin{tabular}{c|l|l}
    \toprule
    & \textit{\textbf{Gold}} & \textit{\textbf{Predicted}} \\
    \midrule
    \multirow{5}{*}{\rotatebox[origin=c]{90}{\textbf{\say{}}}}
    &...a \colorbox{yellow}{dynamic customer focused person} to join... & ...a dynamic \colorbox{pink}{customer focused person to} join... \\
    &...strong leadership and \colorbox{yellow}{team management skills}...                 & ...strong leadership \colorbox{pink}{and team management skills}...\\
    &...speak and \colorbox{yellow}{written english skills}...                          & ...speak and \colorbox{pink}{written english} skills...\\
    &...a \colorbox{yellow}{team environment and working independently skills}...                       & ...a team \colorbox{pink}{environment and working independently} skills...\\
    &...tangible business benefit extremely \colorbox{yellow}{articulate} and...        & ...tangible \colorbox{pink}{business benefit} extremely articulate and... \\
    \midrule
    \multirow{5}{*}{\rotatebox[origin=c]{90}{\textbf{\sks{}}}} 
    &...researcher within \colorbox{yellow}{machine learning} and \colorbox{yellow}{sensory system design}...     & ...researcher within machine \colorbox{pink}{learning and sensory system} design...\\
    &...standards and procedures \colorbox{yellow}{accessing and updating records}...   & ...standards and \colorbox{pink}{procedures accessing and updating} records...\\
    &...with a \colorbox{yellow}{passion for education} to... & ...with a passion for \colorbox{pink}{education} to...\\
    &...understands \colorbox{yellow}{Agile} as a mindset... & ...\colorbox{pink}{understands Agile as a} mindset...\\
    &...experience with \colorbox{yellow}{AWS} \colorbox{yellow}{GCP} \colorbox{yellow}{Microsoft Azure}... &  ...experience \colorbox{pink}{with AWS} GCP Microsoft...\\
    \bottomrule
    \end{tabular}
    \end{adjustbox}
    \caption{\textbf{Qualitative Examples of Predicted Spans.} We show the gold versus predicted spans of the best performing model on both datasets. The first 5 qualitative examples are from \say{} (RoBERTa with WSE), the last 5 are from \sks{}. Yellow the gold span and pink indicates the predicted span. The examples show many partial overlaps with the gold spans (but also incorrect ones), hence the high loose-F1.}
    \label{tab:quali5}
\end{table*}

\subsection{Matching} 
We rank pairs of ESCO embeddings $\vec{t}$ and encoded candidate n-grams $\vec{g}$ in decreasing order of cosine similarity (CosSim), calculated as 

\begin{displaymath}
\text{CosSim(}\vec{t},\vec{g}\text{)} = \frac{\vec{t}^T \vec{g}}{\|\vec{t}\| \|\vec{g}\|}.
\end{displaymath}

We show our pseudocode of the matching algorithm in \cref{alg:cap5}. Note that in \sks{} we have to set a threshold for CosSim, as there are sentences with no skills. A threshold allows us to have a ``no skill'' option. As seen in \cref{fig:results5}, \cref{sec:appendix-exact-num} the threshold sensitivity on \sks{} differs for JobBERT: Performance fluctuates, compared to RoBERTa. Precision goes up with a higher threshold, while recall goes down. For RoBERTa, it stays similar until CosSim$=0.9$. We use CosSim$=0.8$  as over 2 LMs and 3 methods it provides the best cutoff.

\section{Analysis of Results}\label{sec:results_chap5}

\subsection{Results}

Our main results (\cref{fig:results-f1}) show the baselines against ISO, AOC, and WSE of both datasets. We evaluate with two types of F1, following~\citet{van-der-goot-etal-2020-cross}: \texttt{strict} and \texttt{loose-F1}. For full model fine-tuning, RoBERTa achieves 91.31 and 98.55 strict and loose F1 on \say{} respectively. For \sks{}, this is 23.21 and 44.72 strict and loose F1 (on the available subsets of \sks{}). JobBERT achieves 90.18 and 98.19 strict and loose F1 on \say{}, 49.44 and 74.41 strict and loose F1 on \sks{}. The large difference between results is most likely due to lack of negatives in \say{}, i.e., all sentences contain a skill, which makes the task easier. These results highlight the difficulty of SE on \sks{}, where there are negatives as well (sentences with no skills).

The exact match baseline on \sks{} is higher than \say{}. We attribute this to \sks{} also containing ``hard skills'' (e.g., ``Python''), which is easier to match substrings with than ``soft skills''.\footnote{The exact numbers (+precision and recall) are in~\cref{tab:exact}, \cref{sec:appendix-exact-num}, including the definition of strict and loose-F1.}

For the performance of the skill representations on \say{}, RoBERTa and JobBERT outperform the Exact and Lemmatized baseline on strict-F1. For the POS baseline, only the ISO method of both models is slightly better. JobBERT performs better than RoBERTa in strict-F1 on both datasets. 

There is a substantial difference between strict and loose-F1 on both datasets. This indicates that there is partial overlap among the predicted and gold spans. RoBERTa performs best for \say{}, achieving $59.61$ loose-F1 with WSE. In addition, the best performing method for JobBERT is also WSE ($52.69$ loose-F1). For \sks{} we see a drop, JobBERT outperforms RoBERTa with AOC ($32.30$ vs.\ $26.10$ loose-F1) given a threshold of CosSim = 0.8. We hypothesize this drop in performance compared to \say{} could be attributed again to \sks{} containing negative examples as well (i.e., sentences with no skill). 

\subsection{Qualitative Analysis}
A qualitative analysis (\cref{tab:quali5}) reveals there is strong partial overlap with gold vs.\ predicted spans on both datasets, e.g., ``...strong leadership and \emph{team management skills}...'' vs.\ ``...strong leadership \emph{and team management skills}...'', indicating the viability of this method.

\section{Conclusion}
We investigate whether the ESCO skill taxonomy suits as weak supervision signal for Skill Extraction. We apply several skill representation methods based on previous work. We show that using representations of ESCO skills can aid us in this task. We achieve high loose-F1, indicating there is partial overlap between the predicted and gold spans, but need refined off-set methods to get the correct span out (e.g., human post-editing or automatic methods such as candidate filtering). Nevertheless, we see this approach as a strong alternative for supervised Skill Extraction from job postings. 

Future work could include going towards multilingual Skill Extraction, as ESCO consists of 27 languages, exact matching should be trivial. For the other methods several considerations need to be taken into account, e.g., a POS-tagger and/or lemmatizer for another language and a language-specific model.

\clearpage
\section{Appendix}

\begin{table}[t]
    \centering
    \begin{adjustbox}{width=\textwidth}
    \begin{tabular}{c|l|rr|rr}
    \toprule
    & \textbf{Dataset} $\rightarrow$ & \multicolumn{2}{c|}{\textsc{Sayfullina}} & \multicolumn{2}{c}{\textsc{SkillSpan}} \\
     & $\downarrow$ \textbf{Method}, \textbf{Metric} $\rightarrow$ & \textbf{Strict} (P $|$ R $|$ F1) & \textbf{Loose} (P $|$ R $|$ F1) & \textbf{Strict} (P $|$ R $|$ F1) & \textbf{Loose} (P $|$ R $|$ F1) \\
     \midrule
\multirow{3}{*}{\rotatebox[origin=c]{90}{\tiny\textbf{Baseline}}}
    & Exact         & 9.27 $|$ 1.30 $|$ 2.28    & 25.48 $|$ 3.95 $|$ 6.84     & 23.82 $|$ 3.21 $|$ 5.62         & 43.68 $|$ 8.27 $|$ 13.79  \\
    & Lemmatized    & 8.49 $|$ 1.19 $|$ 2.09    & 25.87 $|$ 4.00 $|$ 6.93     & 23.90 $|$ 2.97 $|$ 5.21         & 41.09 $|$ 7.49 $|$ 12.52 \\
    & POS           & 5.99 $|$ 5.95 $|$ 5.97    & 36.55 $|$ 34.51 $|$ 35.50   & 5.97 $|$ 7.88 $|$ 6.79          & 19.34 $|$ 34.71 $|$ 24.80    \\
    \midrule
\multirow{3}{*}{\rotatebox[origin=c]{90}{\tiny\textbf{RoBERTa}}}
    & ISO   & 6.26 $|$ 6.25 $|$ 6.26        & 26.90 $|$ 28.98 $|$ 27.90       & 2.90 $|$ 4.24 $|$ 3.43      & 12.69 $|$ 28.61 $|$ 17.56\\ 
    & AOC   & 3.24 $|$ 3.24 $|$ 3.24        & 64.04 $|$ 55.53 $|$ 59.48       & 2.23 $|$ 2.93 $|$ 2.53      & 20.08 $|$ 37.56 $|$ 26.10\\
    & WSE   & 3.67 $|$ 3.67 $|$ 3.67        & 64.64 $|$ 55.32 $|$ 59.61       & 2.29 $|$ 2.93 $|$ 2.57      & 20.90 $|$ 37.79 $|$ 26.85\\
    \midrule
\multirow{3}{*}{\rotatebox[origin=c]{90}{\tiny\textbf{JobBERT}}}
    & ISO   & 7.71 $|$ 7.72 $|$ 7.71        & 27.76 $|$ 29.95 $|$ 28.82                         & 4.17 $|$ 4.65 $|$ 4.39    & 17.07 $|$ 29.48 $|$ 21.61 \\ %
    & AOC   & 4.04 $|$ 4.05 $|$ 4.05        & 56.50 $|$ 48.41 $|$ 52.14                         & 4.44 $|$ 2.96 $|$ 3.54    & 33.64 $|$ 31.28 $|$ 32.30 \\
    & WSE   & 4.15 $|$ 4.16 $|$ 4.15        & 56.98 $|$ 49.00 $|$ 52.69                         & 4.78 $|$ 3.08 $|$ 3.74    & 34.01 $|$ 30.33 $|$ 31.95 \\
    \bottomrule
    \end{tabular}
    \end{adjustbox}
    \caption{We show the exact numbers of the performance of the methods.}
    \label{tab:exact}
\end{table}

\begin{landscape}
\begin{figure*}[t]
    \centering
    \includegraphics[width=.9\linewidth]{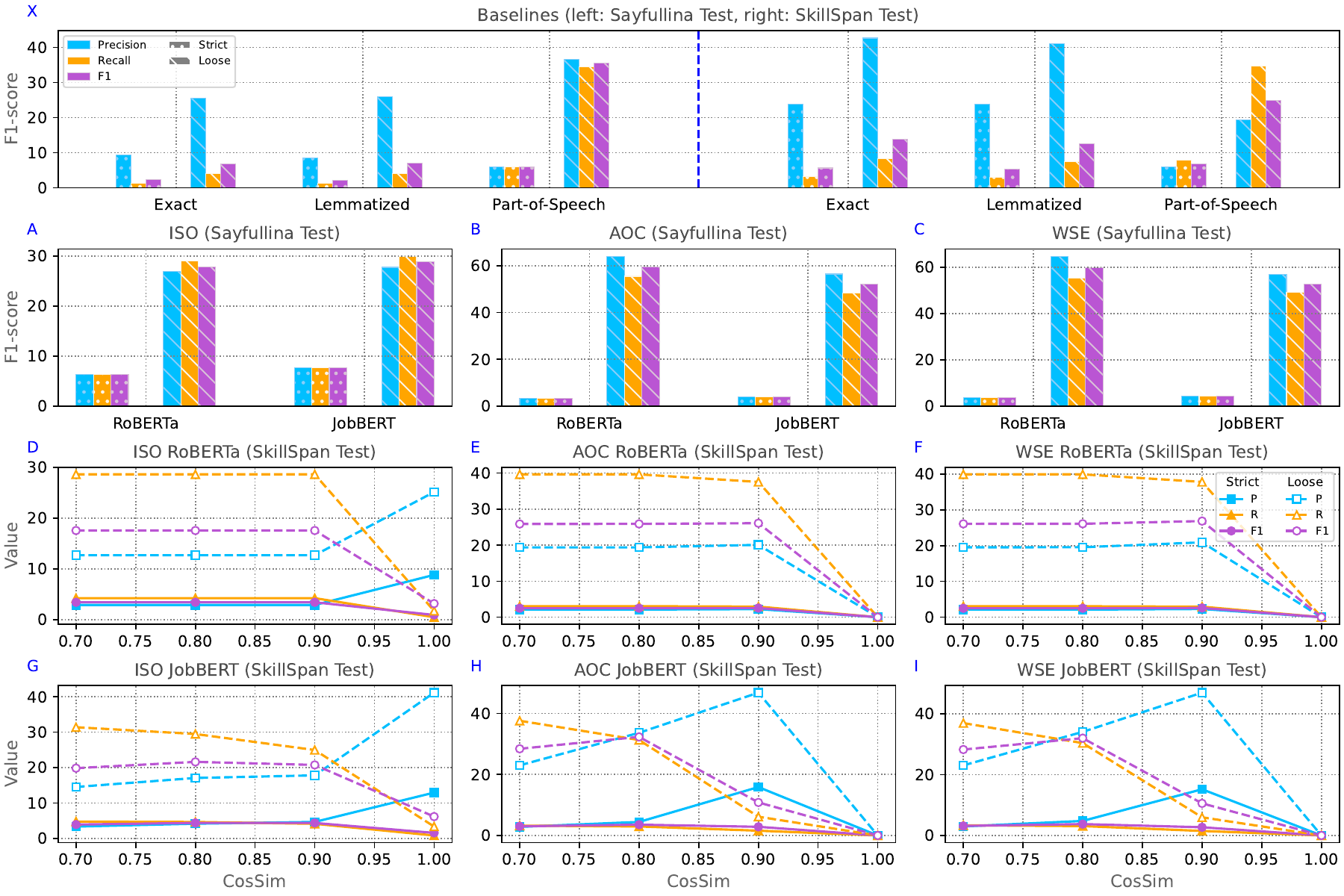}
    \looseness=-1
    \caption{\textbf{Results of Methods.} Results of the baselines are in (X), the performance of ISO, AOC, and WSE on \say{} in (A-C), and the same performance on \sks{} in (D-I) based on the model (RoBERTa or JobBERT). In D--F, we show the precision (P), recall (R), and F1 differences when taking an increasing CosSim.} %
    \label{fig:results5}
\end{figure*}
\end{landscape}
\subsection{Exact Results}\label{sec:appendix-exact-num}

\subsection{Definition F1} As mentioned, we evaluate with two types of F1-scores, following~\citet{van-der-goot-etal-2020-cross}. The first type is the commonly used span-F1, where only the correct span and label are counted towards true positives. This is called \texttt{strict-F1}. In the second variant, we seek for partial matches, i.e., overlap between the predicted and gold span including the correct label, which counts towards true positives for precision and recall. This is called \texttt{loose-F1}. We consider the loose variant as well, because we want to analyze whether the span is ``almost correct''.

\subsection{Exact Numbers Results} We show the exact numbers of \cref{fig:results-f1} in \cref{tab:exact} and more detailed results in~\cref{fig:results5}. Results show that there is high precision among the baseline approaches compared to recall. This is balanced using the representation methods for~\say{}. However, we observe that there is much higher recall for \sks{} than precision.

\chapter{{ESCOXLM-R}: {M}ultilingual {T}axonomy-driven {P}re-training for the {J}ob {M}arket {D}omain}
\chaptermark{{ESCOXLM-R}}
\label{chap:chap6}
The work presented in this chapter is based on a paper that has been published as: \bibentry{zhang2023escoxlm}.

\newpage

\section*{Abstract}
The increasing number of benchmarks for Natural Language Processing (NLP) tasks in the computational job market domain highlights the demand for methods that can handle job-related tasks such as skill extraction, skill classification, job title classification, and de-identification. While some approaches have been developed that are specific to the job market domain, there is a lack of generalized, multilingual models and benchmarks for these tasks. In this study, we introduce a language model called \escolmr{}, based on \xlmr{}, which uses domain-adaptive pre-training on the European Skills, Competences, Qualifications and Occupations (ESCO) taxonomy, covering 27 languages. The pre-training objectives for \escolmr{} include dynamic masked language modeling and a novel additional objective for inducing multilingual taxonomical ESCO relations.
We comprehensively evaluate the performance of \escolmr{} on 6 sequence labeling and 3 classification tasks in 4 languages and find that it achieves state-of-the-art results on 6 out of 9 datasets. Our analysis reveals that \escolmr{} performs better on short spans and outperforms \xlmr{} on entity-level and surface-level span-F1, likely due to ESCO containing short skill and occupation titles, and encoding information on the entity-level.

\section{Introduction}

The dynamic nature of labor markets, driven by technological changes, migration, and digitization, has resulted in a significant amount of job advertisement data (JAD) being made available on various platforms to attract qualified candidates~\citep{brynjolfsson2011race,brynjolfsson2014second,balog2012expertise}. This has led to an increase in tasks related to JAD, including skill extraction~\citep{kivimaki-etal-2013-graph,zhao2015skill,sayfullina2018learning,smith2019syntax,tamburri2020dataops,shi2020salience, chernova2020occupational,bhola-etal-2020-retrieving,zhang-etal-2022-skillspan,zhang-jensen-plank:2022:LREC, zhang2022skill,green-maynard-lin:2022:LREC,gnehm-bhlmann-clematide:2022:LREC,beauchemin2022fijo,decorte2022design, goyal-etal-2023-jobxmlc}, skill classification~\citep{decorte2022design,zhang-jensen-plank:2022:LREC}, job title classification~\citep{javed2015carotene,javed2016towards,decorte2021jobbert,green-maynard-lin:2022:LREC}, de-identification of entities in job postings~\citep{jensen2021identification}, and multilingual skill entity linking~\citep{esco-2022}.

While some previous studies have focused on JAD in non-English languages~\citep{zhang-jensen-plank:2022:LREC,gnehm-bhlmann-clematide:2022:LREC,beauchemin2022fijo}, their baselines have typically relied on language-specific models, either using domain-adaptive pre-training (DAPT; \citealp{gururangan2020don}) or off-the-shelf models. The lack of comprehensive, open-source JAD data in various languages makes it difficult to fully pre-train a language model (LM) using such data. In this work, we seek external resources that can help improve the multilingual performance on the JAD domain. We use the ESCO taxonomy~\citep{le2014esco}, which is a standardized system for describing and categorizing the skills, competences, qualifications, and occupations of workers in the European Union. The ESCO taxonomy, which has been curated by humans, covers over 13,000 skills and 3,000 occupations in 27 languages. Therefore, we seek to answer: \emph{To what extent can we leverage the ESCO taxonomy to pre-train a domain-specific and language-agnostic model for the computational job market domain?}

\begin{figure*}[t]
    \centering
    \includegraphics[width=\linewidth]{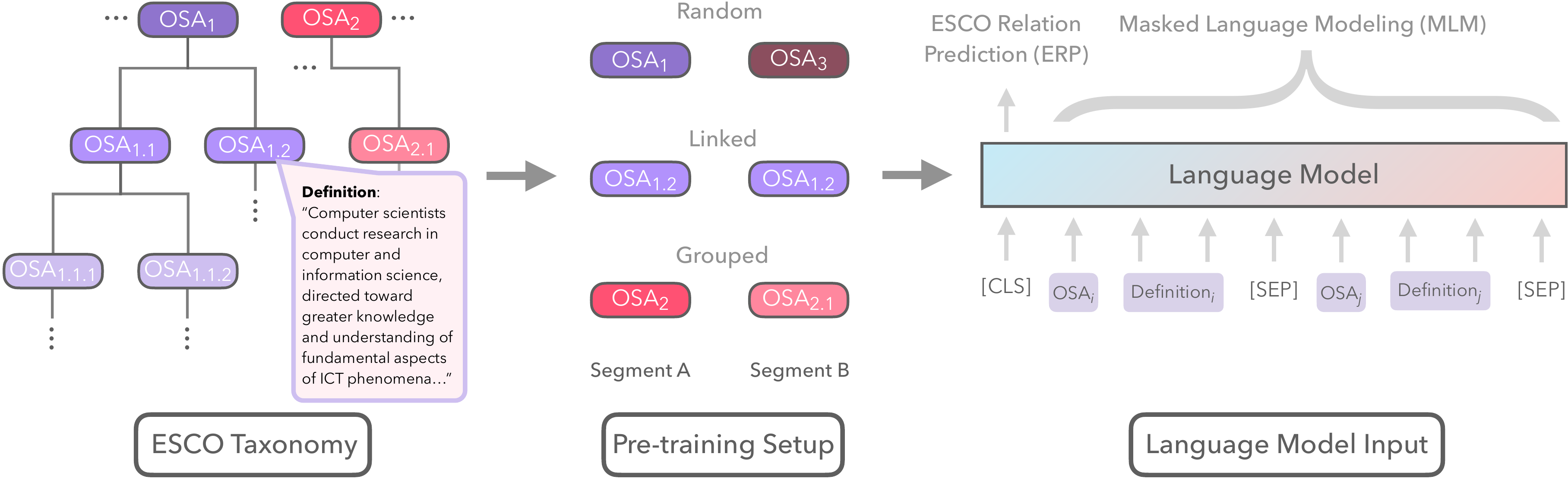}
    \caption{\textbf{ESCO Pre-training Objective}: From left to right, the figure illustrates the hierarchical structure of the ESCO taxonomy, which consists of occupations, skills, and aliases (OSA). Each OSA includes a definition. For the purposes of this study, we consider aliases of occupations to have the same definition as the occupation itself. In the middle of the figure, we show our pre-training setup. Pre-training instances are uniformly sampled in three ways: randomly, linked, or grouped (this is defined in~\cref{subsec:pretraining}). The selected instances (can be in different languages) are then fed to the language model, along with its description. We have two pre-training objectives: the regular MLM objective, and a new ESCO relation prediction objective, in which the goal is to predict which group the sampled instances belong to (Random, Linked, or Grouped).}
    \label{fig:1}
\end{figure*}

In this work, we release the first multilingual JAD-related model named \escolmr{}, a language model based on \xlmr{} that incorporates data from the ESCO taxonomy through the use of two pre-training objectives (\cref{fig:1}): Masked Language Modeling (MLM) and a novel ESCO relation prediction task (\cref{sec:escolmr}). We evaluate \escolmr{} on 9 JAD-related datasets in 4 different languages covering 2 NLP tasks (\cref{sec:exp6}). Our results show that \escolmr{} outperforms previous state-of-the-art (SOTA) on 6 out of 9 datasets (\cref{sec:results6}). In addition, our fine-grained analysis reveals that \escolmr{} performs better on short spans compared to \xlmr{}, and consistently outperforms \xlmr{} on entity-level and surface-level span-F1 (\cref{sec:disc}).

\paragraph{Contributions.} 

In this work, we present and release the following:
\begin{itemize}
   \itemsep0em
    \item \escolmr{}, an \xlmr-based model, which utilizes domain-adaptive pre-training on the 27 languages from ESCO.\footnote{The code for \escolmr{} is available as open-source: \url{https://github.com/mainlp/escoxlmr}. We further release \escolmr{} under an Apache License 2.0 on HuggingFace: \url{https://huggingface.co/jjzha/esco-xlm-roberta-large}.}
    \item The largest JAD evaluation study to date on 3 job-related tasks, comprising 9 datasets in 4 languages and 4 models.
    \item A fine-grained analysis of \escolmr's performance on different span lengths, and emerging entities (i.e., recognition of entities in the long tail).
\end{itemize}

\section{ESCOXLM-R}\label{sec:escolmr}

\subsection{Preliminaries}

In the context of pre-training, an LM is trained using a large number of unlabeled documents, $\mathcal{X} = {X^{(i)}}$, and consists of two main functions: $f_\text{encoder}(.)$, which maps a sequence of tokens $X = (x_1, x_2, ..., x_t)$ to a contextualized vector representation for each token, represented as $(h_1, h_2, ..., h_t)$, and $f_\text{head}(.)$, the output layer that takes these representations and performs a specific task, such as pre-training in a self-supervised manner or fine-tuning on a downstream application. For example, BERT~\citep{devlin2019bert} is pre-trained using two objectives: MLM and Next Sentence Prediction (NSP). In MLM, a portion of tokens in a sequence $X$ is masked and the model must predict the original tokens from the masked input. In the NSP objective, the model takes in two segments $(X_\text{A}, X_\text{B})$ and predicts whether segment $X_\text{B}$ follows $X_\text{A}$. RoBERTa~\citep{liu2019roberta} is a variation of BERT that uses dynamic MLM, in which the masking pattern is generated each time a sequence is fed to the LM, and does not use the NSP task.

\begin{figure*}[t]
    \centering
    \includegraphics[width=\linewidth]{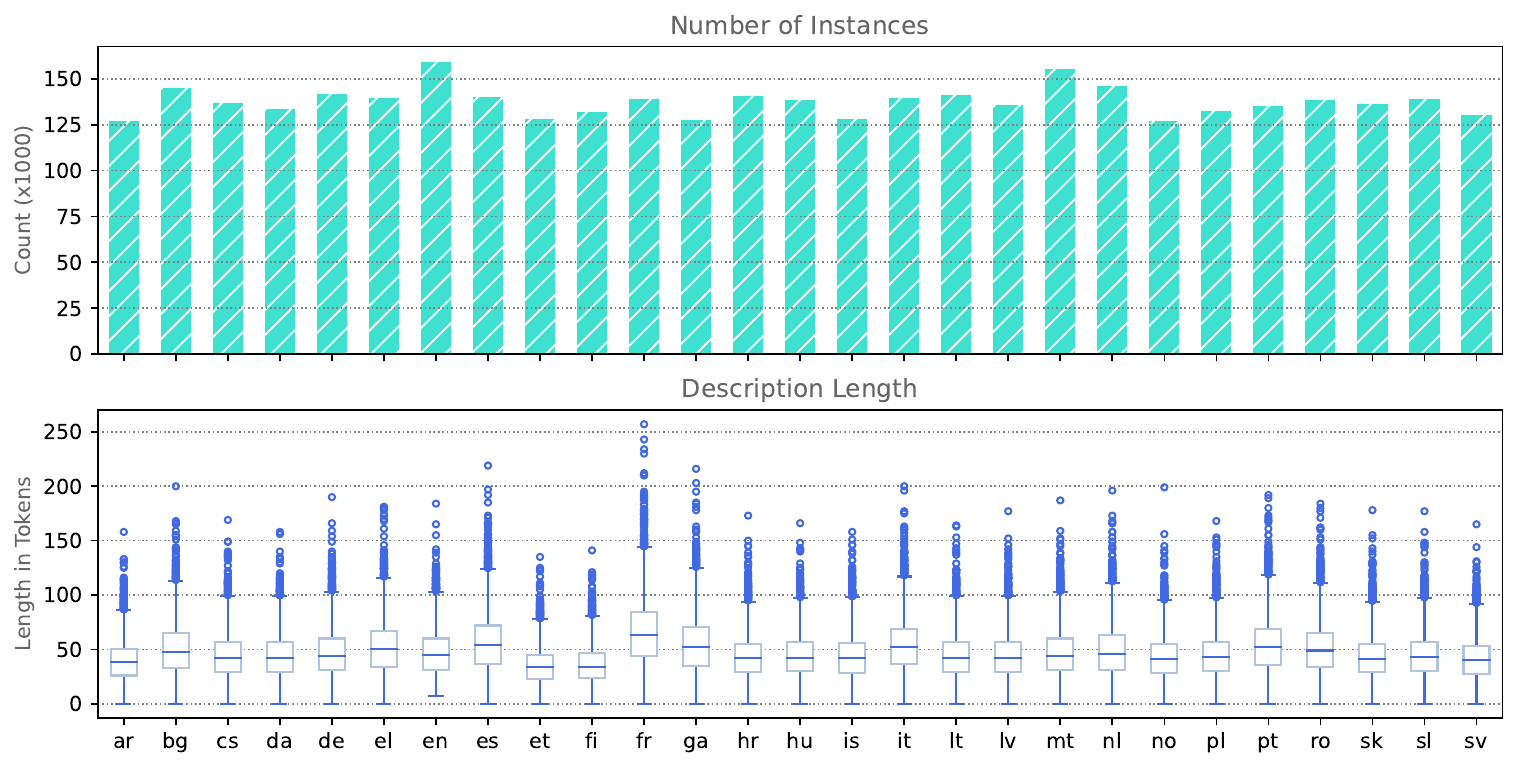}
    \caption{\textbf{Statistics of Pre-training Data.} The ESCO dataset contains descriptions in 27 languages, with a combined total of approximately 3.72 million descriptions (i.e., instances). On average, there are around 130,000 descriptions per language. The average length of each description is 26.3 tokens, with some descriptions reaching a maximum length of 150 or more tokens, as shown by the outliers in the boxplot.}
    \label{fig:esco_stats}
\end{figure*}

\subsection{Multilinguality} Both BERT and RoBERTa have been extended to support multiple languages, resulting in multilingual BERT (mBERT; \citealp{devlin2019bert}) and XLM-RoBERTa (\texttt{XLM-R}; \citealp{conneau2020unsupervised}). XLM-R was found to outperform mBERT on many tasks (e.g.,~\citealp{conneau2020unsupervised,hu2020xtreme, lauscher2020zero}) due to careful tuning, sampling, and scaling to larger amounts of textual data. Because of this, our \escolmr{} model is based on \xlmr{}.

\subsection{European Skills, Competences, Qualifications and Occupations Taxonomy}\label{subsec:esco}

The European Skills, Competences, Qualifications, and Occupations (ESCO;~\citealp{le2014esco}) taxonomy is a standardized system for describing and categorizing the skills, competences, qualifications, and occupations of workers in the European Union (EU). It is designed to serve as a common language for the description of skills and qualifications across the EU, facilitating the mobility of workers by providing a common reference point for the recognition of qualifications and occupations. The taxonomy is developed and maintained by the European Commission and is based on the International Classification of Occupations and the International Standard Classification of Education. It includes 27 European languages: Bulgarian (ar), Czech (cs), Danish (da), German (de), Greek (el), English (en), Spanish (es), Estonian (et), Finnish (fi), French (fr), Gaelic (ga), Croatian (hr), Hungarian (hu), Icelandic (is), Italian (it), Lithuanian (lt), Latvian (lv), Maltese (mt), Dutch (nl), Norwegian (no), Polish (pl), Portuguese (pt), Romanian (ro), Slovak (sk), Slovenian (sl), Swedish (sv), and Arabic (ar). Currently, it describes 3,008 occupations and 13,890 skills/competences (SKC) in all 27 languages.\footnote{Note that ESCO now also includes Ukrainian, but this model was trained before that inclusion. We use the ESCO V1.0.9 API to extract the data. ESCO contains an Apache 2.0 and a European Union Public License 1.2.}

The ESCO taxonomy includes a hierarchical structure with links between occupations, skills, and aliases (OSA). In this work, we focus on the occupation pages and extract the following information from the taxonomy:\footnote{An example of the extracted information can be found in~\cref{json-example} (\cref{sec:appendixa}), and the original page can be accessed at \url{https://bit.ly/3DY1zsX}.}
\begin{itemize}
\itemsep0em
\item \texttt{ESCO Code}: The taxonomy code for the specific occupation or SKC.
\item \texttt{Occupation Label}: The preferred occupation name (i.e., title of the occupation).
\item \texttt{Occupation Description/Definition}: A description of the responsibilities of the specific occupation.
\item \texttt{Major Group Name}: The name of the overarching group to which the occupation belongs, e.g., ``Veterinarians'' for the occupation ``animal therapist''.
\item \texttt{Alternative Labels}: Aliases for the specific occupation, e.g., ``animal rehab therapist'' for the occupation ``animal therapist''.
\item \texttt{Essential Skills}: All necessary SKCs for the occupation, including descriptions of these.
\item \texttt{Optional Skills}: All optional SKCs for the occupation, including descriptions of these.
\end{itemize}

In~\cref{fig:esco_stats}, we present the distribution of pre-training instances and the mean description lengths for each language in the ESCO taxonomy. Note that the number of descriptions is not the same for all languages, and we do not count empty descriptions (i.e., missing translations) for certain occupations or SKCs.

\begin{figure}[t]
    \centering
    \includegraphics[width=.75\linewidth]{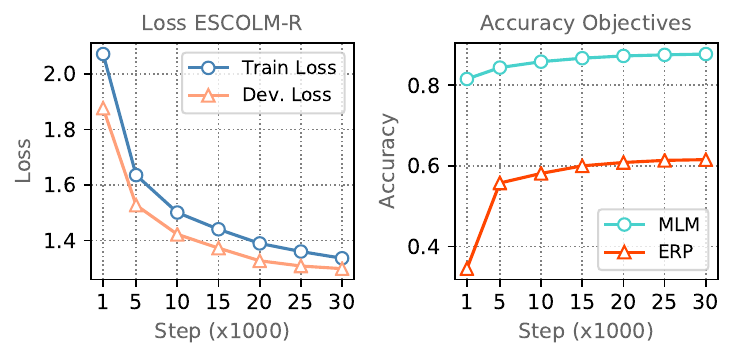}
    \caption{\textbf{Pre-Training Statistics.} The final log loss for the training set is 1.34, while the log loss for the development set is 1.30. The MLM accuracy is 84.3\%, while the Entity Relationship Prediction (ERP) accuracy is 60.0\%. These results were obtained after approximately 1.04 epochs of training on the total data. }
    \label{fig:pretraining}
\end{figure}

\subsection{Pre-training Setup}\label{subsec:pretraining}

To improve our \xlmr-based model, we employ domain-adaptive pre-training techniques as described in previous work such as~\citet{alsentzer2019publicly,han-eisenstein-2019-unsupervised,leebiobert,gururangan2020don,nguyen2020bertweet}. Given the limited amount of training data (3.72M sentences), we utilize the \xlmr{} checkpoint provided by the HuggingFace library~\citep{wolf-etal-2020-transformers} as a starting point.\footnote{\url{https://huggingface.co/xlm-roberta-large}} Our aim is to fine-tune the model to internalize domain-specific knowledge related to occupation and SKCs, while maintaining its general knowledge acquired during the original pre-training phase.

We introduce a novel self-supervised pre-training objective for \escolmr{}, inspired by LinkBERT from~\citet{yasunaga-etal-2022-linkbert}. We view the ESCO taxonomy as a graph of occupations and SKCs (\cref{fig:1}), with links between occupations or occupations and SKCs in various languages. By placing similar occupations or SKCs in the same context window and in different languages, we can learn from the links between (occupation $\leftrightarrow$ occupation) and (occupation $\leftrightarrow$ SKCs) in different languages for true cross-lingual pre-training. In addition to the MLM pre-training objective, which is used to learn concepts within contexts, we introduce another objective called ESCO Relation Prediction (ERP) to internalize knowledge of connections within the taxonomy in the LM. We take an anchor concept ($C_\text{A}$) by concatenating it with its description ($X_\text{A}$) from the ESCO taxonomy and sample an additional concept ($C_\text{B}$) concatenated with its description ($X_\text{B}$) to create LM input \texttt{[CLS]} $C_\text{A} X_\text{A}$ \texttt{[SEP]} $C_\text{B} X_\text{B}$ \texttt{[SEP]}.\footnote{The special tokens used in this example follow the naming convention of BERT for readability, \texttt{[CLS]} and \texttt{[SEP]}. However, since we use \xlmr{} there are different special tokens: \texttt{<s>} as the beginning of the sequence, \texttt{</s>} as the \texttt{SEP} token, and \texttt{</s></s>} as segment separators. Formally, given the example in the text: \texttt{<s> $C_\text{A}X_\text{A}$ </s></s> $C_\text{B}X_\text{B}$ </s>}.} We sample $C_\text{B} X_\text{B}$ in three ways with uniform probability:

\begin{enumerate}
    \itemsep0em
    \item \emph{Random}: We randomly sample $C_\text{B}X_\text{B}$ from the ESCO taxonomy, in any language;
    \item \emph{Linked}: We sample $C_\text{B}X_\text{B}$ in any language from the same occupation page, for example, an ``animal therapist'' (or an alias of the ``animal therapist'', e.g., ``animal rehab therapist'') should have knowledge of ``animal behavior'';
    \item \emph{Grouped}: We sample $C_\text{B}X_\text{B}$ from the same major group in any language. For the same example ``animal therapist'', it comes from major group 2: Professionals $\rightarrow$ group 22: Health professionals. Several other concepts, e.g., ``Nursing professionals'' fall under this major group.
\end{enumerate}

\begin{table*}[t]
\centering
\setlength{\arrayrulewidth}{0.25mm}
\resizebox{\linewidth}{!}{
\begin{tabular}[t]{lllllllrrr}
\toprule

Dataset Name           & Lang.\          & Loc.\   & License   & Task           & Metric        & Input Type    &   Train                   &      Dev.             &   Test \\
\midrule
\textsc{SkillSpan}     & en                & *       & CC-BY-4.0    & SL             & Span-F1       & Sentences       &   5,866               &      3,992                    &   4,680          \\
\textsc{Sayfullina}    & en                & UK      & Unknown    & SL             & Span-F1       & Sentences       &   3,706               &      1,854                    &   1,853          \\
\textsc{Green}         & en                & UK      & CC-BY-4.0    & SL             & Span-F1       & Sentences       &   8,670               &      963                      &   336            \\
\textsc{JobStack}      & en                & *       & RLT          & SL             & Span-F1       & Sentences       &   18,055              &      2,082                    &   2,092          \\
\textsc{Bhola}         & en                & SG      & CC-BY-4.0          & MLC            & MRR           & Documents       &   16,238              &      2,030                       &   2,030          \\
\textsc{Kompetencer}   & en                & DK      & CC-BY-4.0    & MCC            & W.\ Macro-F1  & Skills          &   9,472                &      1,577                    &   1,578         \\
\textsc{Kompetencer}   & \textbf{da}       & DK      & CC-BY-4.0    & MCC            & W.\ Macro-F1  & Skills          &   138                 &      -                    &   784    \\
\textsc{Gnehm}         & \textbf{de}       & CH      & CC-BY-NC-SA-4.0     & SL             & Span-F1       & Sentences        &   22,134              &      2,679                     &   2,943          \\
\textsc{Fijo}          & \textbf{fr}       & FR      & Unknown    & SL             & Span-F1       & Sentences       &   399                 &      50                       &   50             \\

\bottomrule
\end{tabular}}
\caption{\textbf{Dataset Statistics.} We show statistics for all 9 JAD datasets. There are 6 datasets in English and 3 in other languages (Danish, German, and French). We indicate the location the JAD originates from (whenever applicable, * indicates it comes from a variety of countries). We indicate the license of the dataset. Most of the task types consist of sequence labeling (e.g., span extraction, Named Entity Recognition, soft skill tagging). To maintain consistency, we use a single metric for each task type: Sequence Labeling (SL), Multilabel Classification (MLC), and Multiclass Classification (MCC). For \textsc{Kompetencer}, the statistics are provided in brackets for the Danish language.
}
\label{tab:num_post6}
\end{table*}

\subsection{Pre-training\ Objectives} 
The LM is trained using two objectives. First is the MLM objective, and the second is the ERP objective, where the task is to classify the relation $r$ of the \texttt{[CLS]} token in \texttt{[CLS]} $C_\text{A} X_\text{A}$ \texttt{[SEP]} $C_\text{B} X_\text{B}$ \texttt{[SEP]} ($r \in {\text{Random}, \text{Linked}, \text{Grouped}}$). The rationale behind this is to encourage the model to learn the relevance between concepts in the ESCO taxonomy. We formalize the objectives in~\cref{eq:pretrain}:

\begin{equation}\label{eq:pretrain}
\begin{aligned}
\mathcal{L} &= \mathcal{L}_{\mathrm{MLM}} + \mathcal{L}_{\mathrm{ERP}} \\
&=-\sum_{i} \log p\left(x_{i} \mid \mathbf{h}_{i}\right) - \log  p\left(r \mid \mathbf{h}_{\text{\texttt{[CLS]}}}\right),
\end{aligned}
\end{equation}

we define the overall loss $\mathcal{L}$ as the sum of the MLM loss $\mathcal{L}_{\mathrm{MLM}}$ and the ERP loss $\mathcal{L}_{\mathrm{ERP}}$. The MLM loss is calculated as the negative log probability of the input token $x_i$ given the representation $\mathbf{h}_i$. Similarly, the ERP loss is the negative log probability of the relationship $r$ given the representation of the start-token $\mathbf{h}_{\text{\texttt{[CLS]}}}$. In our implementation, we use \xlmr{} and classify the start-token \texttt{[CLS]} for ERP to improve the model's ability to capture the relationships between ESCO occupations and skills.

\subsection{Implementation}

For optimization we follow~\citep{yasunaga-etal-2022-linkbert}, we use the AdamW~\citep{loshchilov2017decoupled} optimizer with ($\beta_1$, $\beta_2$) = (0.9, 0.98). We warm up the learning rate \lr{1}{5} for a ratio of 6\% and then linearly decay it. The model is trained for 30K steps, which is equivalent to one epoch over the data, and the training process takes 33 hours on one A100 GPU with tf32. We use a development set comprising 1\% of the data for evaluation. In~\cref{fig:pretraining}, the pre-training loss and performance on the dev.\ set are plotted, it can be seen that the accuracy plateaus at 30K steps. Though the train and development loss hint that further gains could be obtained on the pretraining objective, we found through empirical analysis on downstream tasks that 30K steps performs best.

\section{Experimental Setup}\label{sec:exp6}
\cref{tab:num_post6} provides the details of the downstream datasets used in this study. Most of the datasets are in EN, with a smaller number in DA, DE, and FR. For each dataset, a brief description and the corresponding best-performing models are given. We put examples of each dataset (apart from JobStack due to the license) in~\cref{data:examples}.

\subsection{\textsc{SkillSpan}~\citep{zhang-etal-2022-skillspan}}

The job posting dataset includes annotations for skills and knowledge, derived from the ESCO taxonomy. The best model in the relevant paper, JobBERT, was retrained using a DAPT approach on a dataset of 3.2 million EN job posting sentences. This is the best-performing model which we will compare against.

\subsection{\textsc{Kompetencer}~\citep{zhang-jensen-plank:2022:LREC}}

This dataset is used to evaluate models on the task of classifying skills according to their ESCO taxonomy code. It includes EN and DA splits, with the EN set derived from \textsc{SkillSpan}. There are three experimental setups for evaluation: fully supervised with EN data, zero-shot classification (EN$\rightarrow$DA), and few-shot classification (a few DA instances). The best-performing model in this work is RemBERT~\citep{chung2020rethinking}, which obtains the highest weighted macro-F1 for both EN and DA. In this work, we use setup 1 and 3, where all available data is used.

\subsection{\textsc{Bhola}~\citep{bhola-etal-2020-retrieving}}
The task of this EN job posting dataset is multilabel classification: Predicting a list of necessary skills in for a given job description. It was collected from a Singaporean government website. It includes job requirements and responsibilities as data fields. Pre-processing steps included lowercasing, stopword removal, and rare word removal. Their model is BERT with a bottleneck layer~\citep{liu2017deep}. In our work, the bottleneck layer is not used and no additional training data is generated through bootstrapping. To keep comparison fair, we re-train their model without the additional layer and bootstrapping. We use Mean Reciprocal Rank as the main results metric.

\subsection{\textsc{Sayfullina}~\citep{sayfullina2018learning}}

This dataset is used for soft skill prediction, a sequence labeling problem. Soft skills are personal qualities that contribute to success, such as ``team working'', ``being dynamic'', and ``independent''. The models for this dataset include a CNN~\citep{kim-2014-convolutional}, an LSTM~\citep{hochreiter1997long}, and a Hierarchical Attention Network~\citep{yang2016hierarchical}. We compare to their best-performing LSTM model.

\subsection{\textsc{Green}~\citep{green-maynard-lin:2022:LREC}}

A sentence-level sequence labeling task involving labeling skills, qualifications, job domain, experience, and occupation labels. The job positions in the dataset are from the United Kingdom. The industries represented in the data vary and include IT, finance, healthcare, and sales. Their model for this task is a Conditional Random Field~\citep{lafferty2001conditional} model.

\subsection{\textsc{JobStack}~\citep{jensen2021identification}}

This corpus is used for de-identifying personal data in job vacancies on Stack Overflow. The task involves sequence labeling and predicting Organization, Location, Name, Profession, and Contact details labels. The best-performing model for this task is a transformer-based~\citep{vaswani2017attention} model trained in a multi-task learning setting. \citet{jensen2021identification} propose to use the I2B2/UTHealth corpus, which is a medical de-identification task~\citep{stubbs2015annotating}, as auxiliary data, which showed improvement over their baselines.

\subsection{\textsc{Gnehm}~\citep{gnehm-bhlmann-clematide:2022:LREC}}

A Swiss-German job ad dataset where the task is Information and Communications Technology (ICT)-related entity recognition, these could be ICT tasks, technology stack, responsibilities, and so forth. The used dataset is a combination of two other Swiss datasets namely the Swiss Job Market Monitor and an online job ad dataset~\citep{gnehm2020text, buchmann2022swiss}. Their model is dubbed JobGBERT and is based on DAPT with German \bertb{}~\citep{chan2020german}.

\subsection{\textsc{Fijo}~\citep{beauchemin2022fijo}}

A French job ad dataset with the task of labeling skill types using a sequence labeling approach. The skill groups are based on the AQESSS public skills repositories and proprietary skill sets provided by their collaborators. These skill types are divided into four categories: ``Thoughts'', ``Results'', ``Relational'', and ``Personal''. The best-performing model for this task is CamemBERT~\citep{martin2020camembert}.

\begin{table*}[t]
\centering
\setlength{\arrayrulewidth}{0.25mm}
\resizebox{\linewidth}{!}{
\begin{tabular}[t]{lllllrll}
\toprule
 Dataset                      &     Lang.       & Metric         &           Prev. SOTA                                 &       \xlmr                 &          \xlmr{} (+ DAPT)              & \escolmr{}                & \multicolumn{1}{c}{$\Delta$} \\
 \midrule
\textsc{SkillSpan}            &     EN             & Span-F1        &           \std{58.9}{4.5}                               &       \std{59.7}{4.6}     &         \std{62.0}{4.0}                 & \textbf{\std{62.6}{3.7}}     & $+$3.7          \\
\textsc{Sayfullina}           &     EN             & Span-F1        &           \std{73.1}{2.1}                               &       \std{89.9}{0.5}       &          \std{90.6}{0.4}              & \textbf{\std{92.2}{0.2}}     & $+$19.1         \\
\textsc{Green}                &     EN             & Span-F1        &           \std{31.8}{*}                                 &       \std{49.0}{2.4}       &          \std{47.5}{0.7}              & \textbf{\std{51.2}{2.1}}     & $+$19.4         \\
\textsc{Jobstack}             &     EN             & Span-F1        &           \textbf{\std{82.1}{0.8}}                      &       \std{81.2}{0.6}       &          \std{80.4}{0.7}              & \std{82.0}{0.7}              & $-$0.1          \\
\textsc{Kompetencer}          &     EN             & W. Macro-F1    &           \std{62.8}{2.8}                               &       \std{59.0}{9.5}       &         \textbf{\std{64.3}{0.5}}      & \std{63.5}{1.3}              & $-$0.7          \\
\textsc{Bhola}                &     EN             & MRR            &           \std{90.2}{0.2}                               &       \std{90.5}{0.3}       &          \std{90.0}{0.3}              & \textbf{\std{90.7}{0.2}}     & $+$0.5          \\
 \midrule
\textsc{Gnehm}                &     DE            & Span-F1        &           \std{86.7}{0.4}                               &       \std{87.1}{0.4}       &          \std{86.8}{0.2}            & \textbf{\std{88.4}{0.5}}     & $+$1.7          \\
\textsc{Fijo}                 &     FR             & Span-F1        &           \std{31.7}{2.3}                               &       \std{41.8}{2.0}       &         \std{41.7}{0.7}             & \textbf{\std{42.0}{2.3}}     & $+$10.3         \\
\textsc{Kompetencer}          &     DA             & W. Macro-F1    &           \std{45.3}{1.5}                               &       \std{41.2}{9.8}       &         \textbf{\std{45.6}{0.8}}    & \std{45.0}{1.4}              & $-$0.3          \\
\bottomrule
\end{tabular}}
\caption{\textbf{Results of Experiments.} The datasets and models are described in~\cref{sec:exp6}. We re-train the best-performing models of all papers to give us the standard deviation. The best-performing model is in bold. The difference in performance between \escolmr{} and the previous SOTA is shown as $\Delta$.
Note (*) that the results for \textsc{Green} are based on a CRF model where the data has been pre-split, and therefore, there is no standard deviation.}
\label{tab:res6}
\end{table*}

\section{Results}\label{sec:results6}

The results of the models are presented in~\cref{tab:res6}.  To evaluate the performance, four different models are used in total: \escolmr{}, the best-performing model originally reported in the relevant paper for the downstream task, vanilla \xlmr{}, and an \xlmr{} model that we continuously pre-trained using only MLM (DAPT; excluding the ERP objective) using the same pre-training hyperparameters as \escolmr{}. For more information regarding the hyperparameters of fine-tuning, we refer to~\cref{app:finetuning} (\cref{tab:hyperparameters}).

\subsection{English} \escolmr{} is the best-performing model in 4 out of 6 EN datasets. The largest improvement compared to the previous SOTA is observed in \textsc{Sayfullina} and \textsc{Green}, with over 19 F1 points. In 3 out of 4 datasets, \escolmr{} has the overall lower standard deviation. For \textsc{Jobstack}, the previous SOTA performs best, and for \textsc{Kompetencer}, \xlmr{} (+ DAPT) has the highest performance.

\subsection{Non-English} In 2 out of 3 datasets, \escolmr{} improves over the previous SOTA, with the largest absolute difference on French \textsc{Fijo} with 10.3 F1 points. In the Danish subset of \textsc{Kompetencer}, \xlmr{} (+ DAPT) has higher performance than \escolmr{}. Next, we will discuss potential reasons for these differences.

\subsection{Analysis} We highlight that the performance gains of \escolmr{} are generally much larger than any of the losses, indicating a largely positive effect of training on ESCO. The improved performance of \escolmr{} on JAD datasets in~\cref{tab:res6} is likely due to the focus on tasks with token-level annotation (i.e., sequence labeling). This suggests that pre-training on the ESCO taxonomy is particularly useful for these types of tasks. The under-performance of \escolmr{} on the \textsc{Kompetencer} dataset in both EN and DA may be because the task involves predicting the ESCO taxonomy code for a given skill \emph{without context}, where we expect ESCO to particularly help with tasks where having context is relevant. We suspect applying DAPT and ERP on ESCO specifically improves recognizing entities that are uncommon.  On the other hand, the poor performance on the \textsc{Jobstack} dataset may be due to the task of predicting various named entities, such as organizations and locations. By manual inspection, we found that ESCO does not contain entities related to organizations, locations, or persons, thus this reveals that there is a lack of relevant pre-training information to \textsc{Jobstack}.

\section{Discussion}\label{sec:disc}

\begin{figure}[t!]
    \centering
    \includegraphics[width=\linewidth]{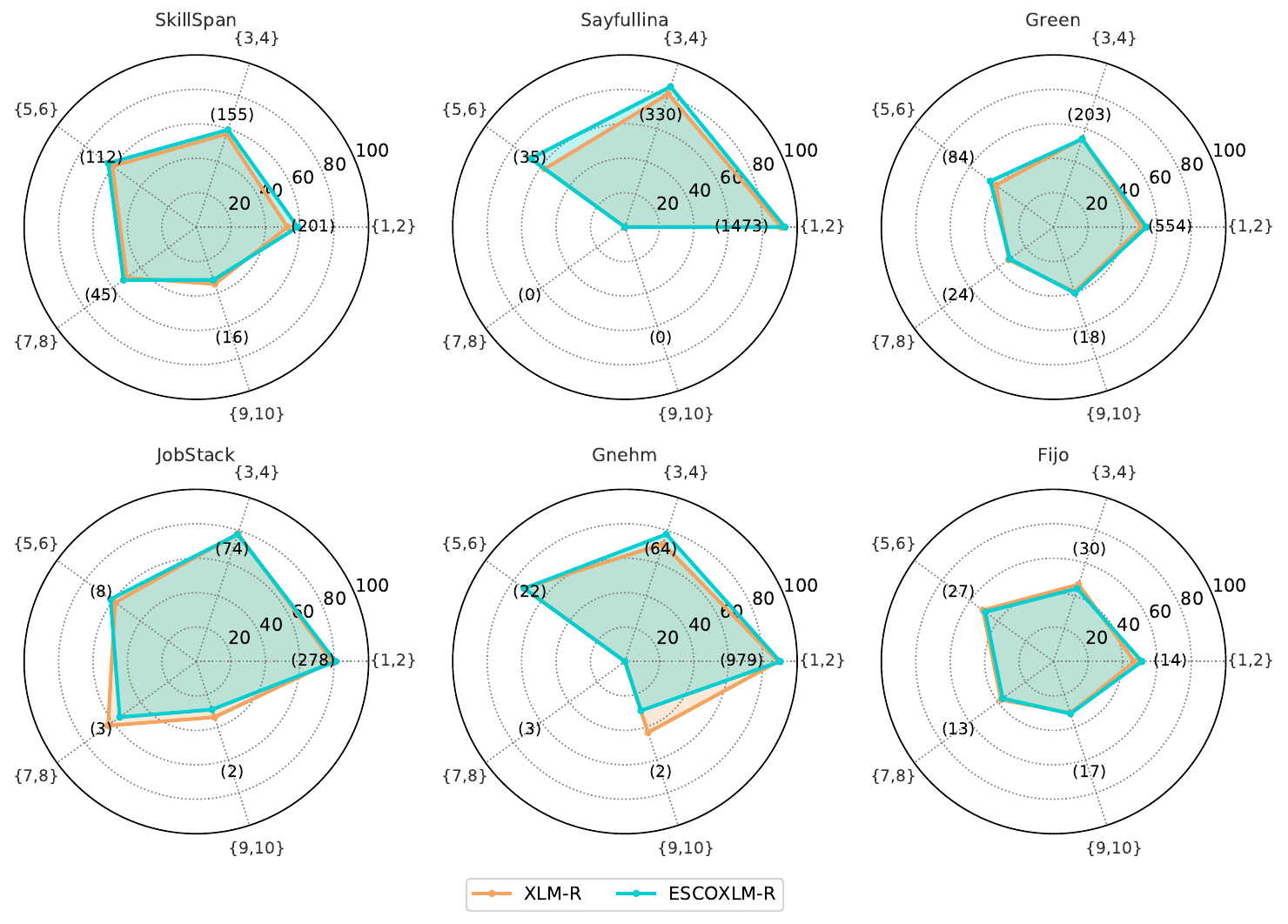}
    \caption{\textbf{Radar Charts of Span-F1 performance by Span Token Length.} We show the performance of \xlmr{} and \escolmr{} on different span lengths, we bucketed the performances of both models according to the length of the spans, up to 10 tokens, and presented the average performance over five random seeds. We did not include error bars in these plots. Note that in some plots, there are no instances in certain buckets (e.g., \textsc{Sayfullina} with 7-8, 9-10). Also, some outer rings only go up to 60 span F1, rather than 100.}
    \label{fig:radar}
\end{figure}

\subsection{Performance on Span Length}
We seek to determine whether the difference in performance between the \escolmr{} and \xlmr{} models is due to shorter spans, and to what extent. One application of predicting short spans well is the rise of technologies, for which the names are usually short in length.
~\citet{zhang2022skill} observes that skills described in the ESCO dataset are typically short, with a median length of approximately 3 tokens. We compare the average performance of both models on the test sets of each dataset, where span-F1 is used as measurement. We group gold spans into buckets of lengths 1-2, 3-4, 5-6, 7-8, and 9-10, and present the span-F1 for each model (\xlmr{} vs. \escolmr{}) in each bucket.

Shown in~\cref{fig:radar}, \escolmr{} outperforms \xlmr{} on shorter spans (i.e., 1-2 or 3-4) in 6 out of the 6 datasets, suggesting that pre-training on ESCO is beneficial for predicting short spans. However, there is a slight decline in performance on some datasets (e.g., \textsc{Skillspan}, \textsc{Jobstack}, and \textsc{Gnehm}) when the spans are longer (i.e., 7-8 or 9-10). It is worth noting that the number of instances in these longer span buckets is lower, and therefore errors may be less apparent in terms of their impact on overall performance.

\begin{table*}
    \centering
    \begin{tabular}{llllll}
    \toprule
    Dataset              &   Ratio         & \multicolumn{2}{c}{Span-F1 (Entity)}                  &     \multicolumn{2}{c}{Span-F1 (Surface)}        \\
                         &                       &    \texttt{XLM-R}             & \escolmr{}                    &  \texttt{XLM-R}                  & \escolmr{}            \\\midrule
    \textsc{SkillSpan}   &   0.90               &  \std{59.9}{7.9}     &  \textbf{\std{61.6}{6.6}}            &  \std{56.4}{5.7}        & \textbf{\std{57.9}{4.3}}     \\
    \textsc{Sayfullina}  &   0.22                &  \std{94.0}{0.2}     &  \textbf{\std{95.7}{0.3}}            &  \std{82.8}{0.6}        & \textbf{\std{87.2}{0.7}}     \\
    \textsc{Green}       &   0.79                &  \std{50.3}{2.4}     &  \textbf{\std{53.1}{2.1}}            &  \std{49.2}{2.4}        &  \textbf{\std{52.0}{2.1}}    \\
    \textsc{Jobstack}    &   0.41                &  \std{85.6}{0.7}     &  \textbf{\std{86.4}{0.5}}            &  \std{78.4}{1.2}        & \textbf{\std{79.8}{0.7}}     \\
    \textsc{Gnehm}       &   0.53                &  \std{89.3}{0.3}     &  \textbf{\std{89.6}{0.4}}            &  \std{87.3}{0.3}        & \textbf{\std{87.8}{0.6}}     \\
    \textsc{Fijo}        &   0.77                &  \std{34.4}{2.9}     &  \textbf{\std{35.7}{1.1}}            &  \std{34.4}{1.1}        &  \textbf{\std{35.7}{1.1}}     \\\bottomrule
    \end{tabular}
    \caption{\textbf{Entity vs.\ Surface-level span-F1 on Test.} In this table, the performance of two systems, \xlmr{} and \escolmr{}, was measured using entity-level and surface-level span-F1 scores. Entity-level span-F1 measures precision, recall, and harmonic mean at the entity level, while surface-level span-F1 measures a system's ability to recognize a range of entities. We include the ratio of surface entities to total entities in each \emph{training} set, with a higher ratio indicating more variety (a ratio of 1.00 indicates all entities are unique).
    }
    \label{tab:entsurf}
\end{table*}

\subsection{Entity-F1 vs. Surface-F1}

In this analysis, we adopt the evaluation method used in the W-NUT shared task on Novel and Emerging Entity Recognition~\citep{derczynski-etal-2017-results}. In this shared task, systems are evaluated using two measures: entity span-F1 and surface span-F1. Entity span-F1 assesses the precision, recall, and harmonic mean (F1) of the systems at the entity level, while surface span-F1 assesses their ability to correctly recognize a diverse range of entities, rather than just the most frequent surface forms.
This means surface span-F1 counts entity types, in contrast to entity tokens in the standard entity span-F1 metric. 

As shown in~\cref{tab:entsurf}, we first calculate the ratio of unique entities and total entities in each relevant train set (i.e., datasets where we do span labeling). A higher ratio number indicates a wider variety of spans. Both \xlmr{} and \escolmr{} tend to have lower performance when variety gets high (above 0.75). In addition, there are 2 datasets (\textsc{Sayfullina}, \textsc{Jobstack}) where we see a low variety of spans and large discrepancy between performance of entity span-F1 and surface span-F1. This difference is lower for \escolmr{} (especially in \textsc{Sayfullina}) suggesting that pre-training on ESCO helps predicting uncommon entities.

It is also noteworthy that the standard deviations for the scores at the entity span-F1 are generally lower than those for the surface span-F1. This suggests that the results for the entity span-F1 scores are more consistent across different runs, likely due to recognizing common entities more.

Overall, \escolmr{} consistently outperforms \xlmr{} in both the entity-level and surface-level F1 scores, indicating the benefits of using the ESCO dataset for pre-training on JAD tasks.

\section{Related Work}

To the best of our knowledge, we are the first to internalize an LM with ESCO for job-related NLP tasks. There are, however, several works that integrate factual knowledge (i.e., knowledge graphs/bases) into an LM. 
\citet{peters-etal-2019-knowledge} integrates multiple knowledge bases into LMs to enhance their representations with structured, human-curated knowledge and improve perplexity, fact recall and downstream performance on various tasks. \citet{zhang-etal-2019-ernie,he-etal-2020-bert,wang-etal-2021-kepler} combine LM training with knowledge graph embeddings. \citet{wang2021k} introduces K-Adapter for injecting knowledge into pre-trained models that adds neural adapters for each kind of knowledge domain.
~\citet{yu-etal-2022-dict} introduces Dict-BERT, which incorporates definitions of rare or infrequent words into the input sequence and further pre-trains a BERT model.

\citet{calixto2021wikipedia} introduced a multilingual Wikipedia hyperlink prediction intermediate task to improve language model pre-training. Similarly, \citet{yasunaga-etal-2022-linkbert} introduced LinkBERT which leverages links between documents, such as hyperlinks, to capture dependencies and knowledge that span across documents by placing linked documents in the same context and pre-training the LM with MLM and document relation prediction. 

\section{Conclusion}

In this study, we introduce \escolmr{} as a multilingual, domain-adapted LM that has been further pre-trained on the ESCO taxonomy. We evaluated \escolmr{}, to the best of our knowledge, on the broadest evaluation set in this domain on 4 different languages. The results showed that \escolmr{} outperformed \xlmr{} on job-related downstream tasks in 6 out of 9 datasets, particularly when the task was relevant to the ESCO taxonomy and context was important. It was found that the improvement of \escolmr{} was mainly due to its performance on shorter span lengths, demonstrating the value of pre-training on the ESCO dataset. \escolmr{} also demonstrated improved performance on both frequent surface spans and a wider range of spans. Overall, this work showed the potential of \escolmr{} as an LM for multilingual job-related tasks. We hope that it will encourage further research in this area.

\section*{Limitations}

There are several limitations to this study that should be considered. First, a key limitation is the lack of a variety of language-specific JAD. Here, we have four different languages namely EN, DA, FR, and DE. This means that our analysis is based on a limited subset of languages and may not be representative of JAD data outside of these four languages. 

In turn, the second limitation is that the ESCO taxonomy used as pre-training data only covers Europe and the datasets used in this work also covers mostly Europe. The results may not be generalizable to other regions. However, we see a slight improvement in the \textsc{Bhola} dataset, the data of which comes from Singapore, which hints that it could generalize to other cultures.

The ESCO relation prediction task aims for learning the relations between elements of the ESCO taxonomy. We acknowledge that we do not evaluate the effectiveness of the pre-training objective in relation-centered tasks. Unfortunately, to the best of our knowledge, there is no job-related dataset containing relations between skill/occupation concepts to benchmark our model on. We consider this interesting future work.

Finally, we did not conduct an ablation study on the ERP pre-training objective, i.e., which errors it makes. As the accuracy of the objective is 60\%, we are unable to determine which sampling method is detrimental to this accuracy. However, we suspect that the Linked sampling approach might be the hardest to predict correctly. For example, many occupations have a lot of necessary and optional skills, thus it is harder to determine if some skill truly belongs to a specific occupation. Nevertheless, we see that adding the ERP objective improves over regular MLM domain-adaptive pre-training.

Despite these limitations, we believe that this study provides valuable resources and insights into the use of \escolmr{} for analyzing JAD and suggests directions for future research. Future studies could address the limitations of this study by using a larger, more diverse datasets and by conducting ablation studies on the language model to better understand which parts contribute to the results.

\section*{Ethics Statement}
We also see a potential lack of language inclusiveness within our work, as we addressed in the Limitation section that ESCO mostly covers Europe (and the Arabic language). Nevertheless, we see \escolmr{} as a step towards inclusiveness, due to JAD frequently being English-only. In addition, to the best of our knowledge, ESCO itself is devoid of any gendered language, specifically, pronouns and other gender-specific terms in, e.g., occupations.  However, we acknowledge that LMs such as \escolmr{} could potentially be exploited in the process of hiring candidates for a specific job with unintended consequences (unconscious bias and dual use). There exists active research on fairer recommender systems (e.g., bias mitigation) for human resources (e.g., \citealp{mujtaba2019ethical, raghavan2020mitigating, deshpande2020mitigating, kochling2020discriminated, sanchez2020does, wilson2021building, vanimproving, arafan2022end}).

\clearpage

\section{Appendix}

\subsection{Example Extraction from ESCO}
\label{sec:appendixa}

\begin{listing*}
\begin{minted}[frame=single,
               framesep=3mm,
               linenos=true,
               xleftmargin=15pt,
               tabsize=2]{js}
{     
    "id": int, "esco_code": "2250.4", 
    "preferred_label": "animal therapist",
    "major_group": {
	    "title": "Veterinarians",
	    "description": "Veterinarians diagnose, [...]"
    },
    "alternative_label": [ "animal convalescence 
    therapist",
        [...]
    ],
    "description": "Animal therapists provide [...]",
    "essential_skills": [
	    {
	      "title": "anatomy of animals",
	      "description": "The study of animal body 
       parts, 
       [...]"
	    },
	    [...]
    ],
    "optional_skills": [...] 
}
\end{minted}
\caption{\textbf{Example Extraction.} An example of the information that is given for ESCO code 2250.4: animal therapist. The original page can be found here: \url{http://data.europa.eu/esco/occupation/0b2d3242-22a3-4de5-bd29-efd39cdf2c31}.} 
\label{json-example}
\end{listing*}

\clearpage

\subsection{Data Examples}\label{data:examples}

\begin{table}[ht]
    \centering
    \begin{tabular}{|l|l|}
    \hline
    \textsc{SkillSpan}      & \cref{json-skillspan6} \\
    \textsc{Sayfullina}     & \cref{json-sayfullina6} \\
    \textsc{Green}          & \cref{json-green6} \\
    \textsc{Bhola}          & \cref{json-bhola6} \\
    \textsc{Kompetencer}    & \cref{json-kompetencer6} \\
    \textsc{Fijo}           & \cref{json-fijo6} \\
    \textsc{Gnehm}          & \cref{json-gnehm6} \\
    \hline
    \end{tabular}
    \caption{\textbf{Data Examples.} Data example references for each dataset for fine-tuning.}
\end{table}

\begin{listing}[t]
\begin{minted}[frame=single,
               framesep=3mm,
               linenos=true,
               xleftmargin=15pt,
               tabsize=2]{xml}
Experience	O           O
in	        O           O
working     B-Skill     O
on          I-Skill     O
a           I-Skill     O
cloud-based I-Skill     O
application I-Skill     O
running     O           O
on          O           O
Docker      O           B-Knowledge
.           O           O

A           O           O
degree      O           B-Knowledge
in          O           I-Knowledge
Computer    O           I-Knowledge
Science     O           I-Knowledge
or          O           O
related     O           O
fields      O           O
.           O           O
\end{minted}
\caption{\textbf{Data Example SkillSpan.}} 
\label{json-skillspan6}
\end{listing}

\begin{listing}[t]
\begin{minted}[frame=single,
               framesep=3mm,
               linenos=true,
               xleftmargin=15pt,
               tabsize=2]{xml}
ability     O
to          O
work        B-Skill
under       I-Skill
stress      I-Skill
condition   O

due         O
to          O
the         O
dynamic     B-Skill
nature      O
of          O
the         O
group       O
environment O
,           O
the         O
ideal       O
candidate   O
will        O
\end{minted}
\caption{\textbf{Data Example Sayfullina.}} 
\label{json-sayfullina6}
\end{listing}

\begin{listing}[t]
\begin{minted}[frame=single,
               framesep=3mm,
               linenos=true,
               xleftmargin=15pt,
               tabsize=2]{xml}
A               O
sound           O
understanding   O
of              O
the             O
Care            B-Skill
Standards       I-Skill
together        O
with            O
a               O
Nursing         B-Qualification
qualification   I-Qualification
and             O
current         O
NMC             B-Qualification
registration    I-Qualification
are             O
essential       O
for             O
this            O
role            O
.               O
\end{minted}
\caption{\textbf{Data Example Green.}} 
\label{json-green6}
\end{listing}

\begin{listing*}[t]
\begin{minted}[frame=single,
               framesep=3mm,
               linenos=true,
               xleftmargin=15pt,
               tabsize=2]{xml}
department economics national university singapore 
invites applications teaching oriented positions 
level lecturer senior lecturer [...] <labels>
\end{minted}
\caption{\textbf{Data Example Bhola.}} 
\label{json-bhola6}
\end{listing*}

\begin{listing*}[t]
\begin{minted}[frame=single,
               framesep=3mm,
               linenos=true,
               xleftmargin=15pt,
               tabsize=2]{xml}
<English>
team worker                         S4
passion for developing your career  S1
liaise with internal teams          S1
identify system requirements        S2
plan out our new features           S4

<Danish>
arbejde med børn i alderen ½-3 år   S3
samarbejde                          S1
fokusere på god kommunikation       S1
bidrage til at styrke fællesskabet  S1
ansvarsbevidst                      A1
lyst til et aktivt udeliv           A1
\end{minted}
\caption{\textbf{Data Example Kompetencer.}} 
\label{json-kompetencer6}
\end{listing*}

\begin{listing}[t]
\begin{minted}[frame=single,
               framesep=3mm,
               linenos=true,
               xleftmargin=15pt,
               tabsize=2]{xml}
Participer      B-relationnel
au              I-relationnel
réseau          I-relationnel
téléphonique    I-relationnel
mis             O
sur             O
pied            O
lors            O
des             O
campagnes       O
d'inscription   O
pour            O
fournir         B-pensee
les             I-pensee
renseignements  I-pensee
nécessaires     I-pensee
aux             I-pensee
assurés         I-pensee
\end{minted}
\caption{\textbf{Data Example Fijo.}} 
\label{json-fijo6}
\end{listing}

\begin{listing}[t]
\begin{minted}[frame=single,
               framesep=3mm,
               linenos=true,
               xleftmargin=15pt,
               tabsize=2]{xml}
in                  O
mit                 O
guten               O
EDV-Kenntnissen     B-ICT

.                   O
Es                  O
erwartet            O
Sie                 O
eine                O
interessante        O
Aufgabe             O
in                  O
einer               O
Adressverwaltung    O
(                   O
Rechenzenter        B-ICT
)                   O
\end{minted}
\caption{\textbf{Data Example Gnehm.}} 
\label{json-gnehm6}
\end{listing}

\clearpage

\begin{table*}[t]
    \centering
    \resizebox{\linewidth}{!}{
    \begin{tabular}{lllll}
    \toprule
                            &   Learning rate                                               &  Batch size               &\texttt{max\_seq\_length}  &  Epochs               \\
                            \midrule
    \textsc{SkillSpan}      &   \{\lr{1}{4}, \lr{5}{5}, \lr{1}{5} \lr{5}{6}\}               &  \{16, 32, 64\}           & 128                       &  20                   \\
    \textsc{Kompetencer}    &   \{\lr{1}{4}, \lr{7}{5}, \lr{5}{5}, \lr{1}{5}, \lr{5}{6}\}   &  \{8, 16, 32\}            & 128                       &  20                   \\
    \textsc{Bhola}          &   \{\lr{1}{4}, \lr{7}{5}, \lr{5}{5}, \lr{1}{5}, \lr{5}{6}\}   &  \{4, 16, 32, 64, 128\}   & \{128, 256\}              &  10                   \\
    \textsc{Sayfullina}     &   \{\lr{1}{4}, \lr{5}{5}, \lr{1}{5}\}                         &  \{16, 32, 64\}           & 128                       &  10                   \\
    \textsc{Green}          &   \{\lr{1}{4}, \lr{5}{5}, \lr{1}{5}\}                         &  \{16, 32, 64\}           & 128                       &  10                   \\
    \textsc{JobStack}       &   \{\lr{1}{4}, \lr{7}{5}, \lr{5}{5}, \lr{1}{5}, \lr{5}{6}\}   &  \{16, 32, 64, 128\}      & 128                       &  20                   \\
    \textsc{Gnehm}          &   \{\lr{1}{4}, \lr{5}{5}, \lr{1}{5}\}                         &  \{16, 32, 64\}           & 128                       &  5                   \\
    \textsc{Fijo}           &   \{\lr{1}{4}, \lr{5}{5}, \lr{1}{5}\}                         &  \{8, 16, 32, 64\}        & 128                       &  10                   \\
    \bottomrule
    \end{tabular}}  
    \caption{\textbf{Hyperparameter Sweep for Fine-tuning.} We show a hyperparameter sweep for fine-tuning all models. Learning rate differs for both \xlmr{} and \escolmr{}, where \xlmr{} performs best on lower learning rate (e.g., \lr{1}{5}) and \escolmr{} on a bit of a higher learning rate (e.g., \lr{5}{5}). A batch size of 32 works best for all models. The max sequence length is usually the same, except for \textsc{Bhola} due to it containing long texts. Epochs are determined based on previous work (i.e., the relevant datasets).}      
    \label{tab:hyperparameters}
\end{table*}

\subsection{Fine-tuning Details}\label{app:finetuning}

For fine-tuning \xlmr{} (+ DAPT) and \escolmr{} on the downstream tasks, we use MaChAmp~\citep{van-der-goot-etal-2021-massive}. For more details we refer to their paper. We always include the original learning rate, batch size, maximum sequence length, and epochs from the respective downstream tasks in our search space (whenever applicable). Each model is trained on an NVIDIA A100 GPU with 40GBs of VRAM and an AMD Epyc 7662 CPU. The seed numbers the models are initialized with are 276800, 381552, 497646, 624189, 884832. We run all models with the maximun number of epochs indicated in~\cref{tab:hyperparameters} and select the best-performing one based on validation set performance in the downstream metric.

\chapter{{NNOSE}: {N}earest {N}eighbor {O}ccupational {S}kill {E}xtraction}
\chaptermark{{NNOSE}}
\label{chap:chap7}
The work presented in this chapter is based on a paper accepted as: \bibentry{zhang2024nnose}.

\newcommand{\vanilla}{\multirow{3}{*}{\rotatebox[origin=c]{90}{Vanilla}}}
\newcommand{\withknn}{\multirow{3}{*}{\rotatebox[origin=c]{90}{+\knn{}}}}
\newcommand{\dd}{\texttt{\{D\}}}
\newcommand{\dwt}{\texttt{{\{D\}}+WT}}
\newcommand{\ad}{$\forall$\texttt{D}}
\newcommand{\adwt}{$\forall$\texttt{D+WT}}
\newcommand{\knn}{$k$NN}
\newcommand{\nnose}{NNOSE}

\newcommand{\roberta}{RoBERTa\textsubscript{base}}
\newcommand{\up}[1]{{\color{blue} $\uparrow$\textsubscript{#1}}}
\newcommand{\down}[1]{{\color{red} $\downarrow$\textsubscript{#1}}}
\newcommand{\none}[1]{{--\textsubscript{#1}}}
\renewcommand{\floatpagefraction}{.99} %

\newpage

\section*{Abstract}
The labor market is changing rapidly, prompting increased interest in the automatic extraction of occupational skills from text. 
With the advent of English benchmark job description datasets, there is a need for systems that handle their diversity well. 
We tackle the complexity in occupational skill datasets tasks---combining and leveraging multiple datasets for skill extraction, to identify rarely observed skills within a dataset, and overcoming the scarcity of skills across datasets.
In particular, we investigate the retrieval-augmentation of language models, employing an external datastore for retrieving similar skills in a dataset-unifying manner. 
Our proposed method, \textbf{N}earest \textbf{N}eighbor \textbf{O}ccupational \textbf{S}kill \textbf{E}xtraction (NNOSE) effectively leverages multiple datasets by retrieving neighboring skills from other datasets in the datastore.  
This improves skill extraction \emph{without} additional fine-tuning. 
Crucially, we observe a performance gain in predicting infrequent patterns, with substantial gains of up to 30\% span-F1 in cross-dataset settings. 

\section{Introduction}
Labor market dynamics, influenced by technological changes, migration, and digitization, have led to the availability of job descriptions (JD) on platforms to attract qualified candidates~\citep{brynjolfsson2011race,brynjolfsson2014second,balog2012expertise}. 
JDs consist of a collection of skills that exhibit a characteristic \emph{long-tail pattern}, where popular skills are more common while niche expertise appears less frequently across industries~\citep{autor2003skill,autor2013growth}, such as ``teamwork'' vs.\ ``system design''.\footnote{Examples are from the \href{https://rb.gy/3zgld}{\texttt{CEDEFOP Skill Platform}}.} 
This pattern poses challenges for skill extraction (SE) and analysis, as certain skills may be underrepresented, overlooked, or emerging in JDs.
This complexity makes the extraction and analysis of skills more difficult, resulting in a \emph{sparsity of skills} in SE datasets. We tackle this by combining three different skill datasets.

To address the challenges in SE, we explore the potential of Nearest Neighbors Language Models (NNLMs;~\citealp{Khandelwal2020Generalization}). NNLMs calculate the probability of the next token by combining a parametric language model (LM) with a distribution derived from the k-nearest context--token pairs in the datastore. This enables the storage of large amounts of training instances without the need to retrain the LM weights, improving language modeling. However, the extent to which NNLMs enhance application-specific end-task performance beyond language modeling remains relatively unexplored. Notably, NNLMs offer several advantages, as highlighted by~\citet{Khandelwal2020Generalization}: 
First, explicit memorization of the training data aids generalization.
Second, a single LM can adapt to multiple domains without domain-specific training, by incorporating domain-specific data into the datastore (e.g., multiple datasets).
Third, the NNLM architecture excels at predicting rare patterns, particularly the long-tail.

\begin{figure}[!t]
    \centering
    \includegraphics[width=\linewidth]{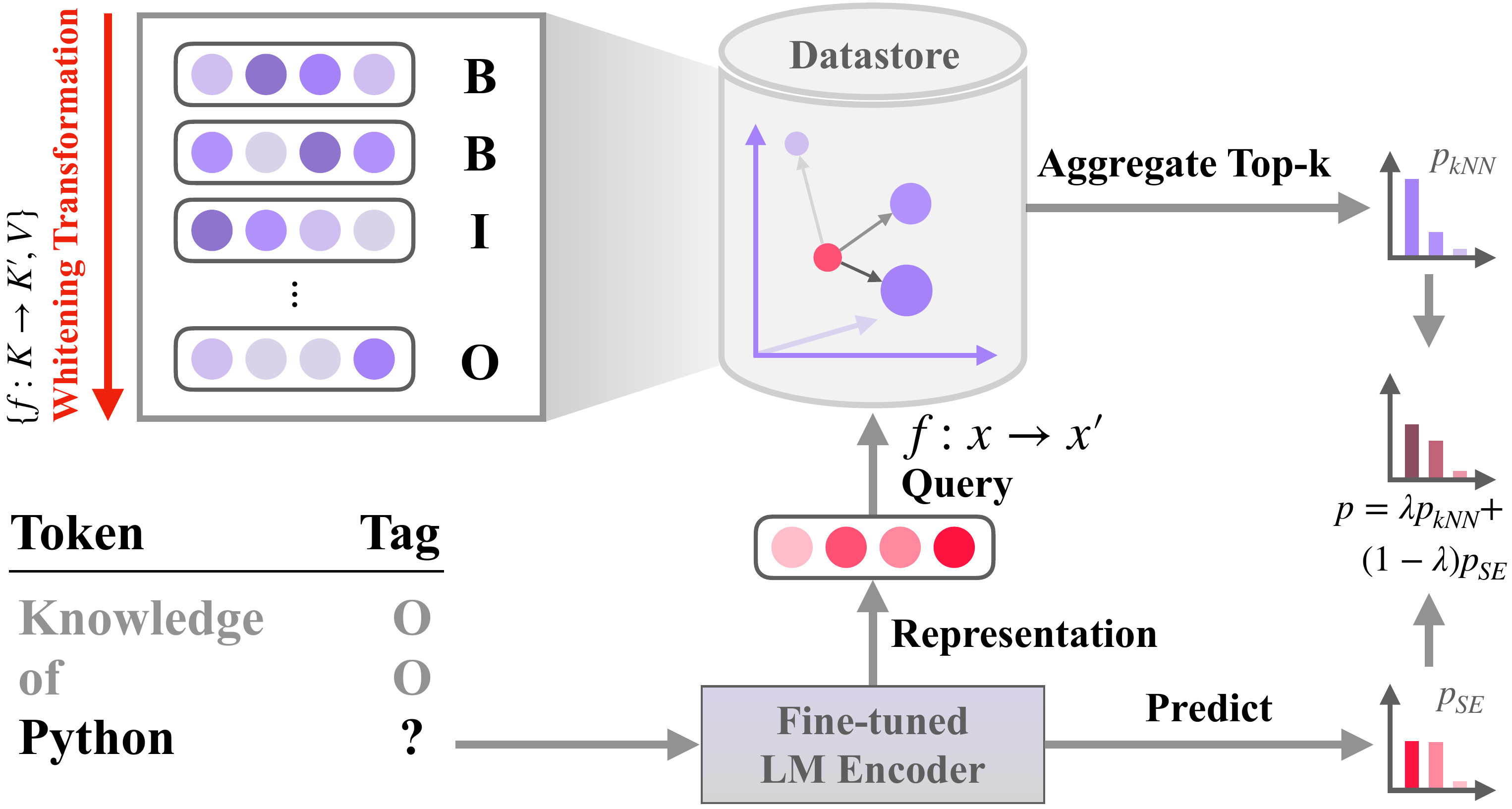}
    \caption{\textbf{Setup of \nnose{}.} The datastore consists of paired contextual token representations obtained from a fine-tuned encoder and the corresponding \texttt{BIO} tag. 
    We use a whitening transformation to enhance the isotropy of token representations. 
    During inference, i.e., retrieving tokens, we use the same whitening transformation on the test token’s representation to retrieve the $k$-nearest neighbors from the datastore. 
    We interpolate the encoder and \knn{} distributions with a hyperparameter $\lambda$ as the final distribution.}
    \label{fig:fig1-7}
\end{figure}

Therefore, we seek to answer the question:\ \emph{\textbf{How effective are nearest neighbors retrieval methods for occupational skill extraction?}} 
Our contributions are as follows:
\begin{itemize}
    \item To the best of our knowledge, we are the first to investigate encoder-based \knn{} retrieval by leveraging \emph{multiple} datasets.
    \item Furthermore, we present a novel domain-specific \roberta{}-based language model, JobBERTa, tailored to the job market domain.
    \item We conduct an extensive analysis to show the advantages of \knn{} retrieval, in contrast to prior work that primarily focuses on hyperparameter-specific analysis.\footnote{Code: \url{https://github.com/jjzha/nnose}.}
\end{itemize}

\section{Nearest Neighbor Skill Extraction}
\subsection{Skill Extraction.} The task of SE is formulated as a sequence labeling problem. 
We define a set of job description sentences $\mathcal{X}$, where each $d \in \mathcal{X}$ represents a set of sequences with the $j^{\text{th}}$ input sequence $\mathcal{X}^j_d= \{x_1, x_2, ..., x_i\}$, with a corresponding target sequence of \texttt{BIO}-labels $\mathcal{Y}^j_d = \{y_1, y_2, ..., y_i\}$. The labels include ``\texttt{B}'' (beginning of a skill token), ``\texttt{I}'' (inside skill token), and ``\texttt{O}'' (any outside token). 
The objective is to use $\mathcal{D}$ in training a labeling algorithm that accurately predicts entity spans by assigning an output label $y_i$ to each token $x_i$.

\subsection{\nnose{}}\label{subsec:knnse}
\looseness=-1
The core idea of \nnose{} is that we augment the extraction of skills during inference with a \knn{} retrieval component and a datastore consisting of context--token pairs. 
\cref{fig:fig1-7} outlines our two-step approach. 
First, we extract skills by getting token representation $\boldsymbol{h}_i$ from $x_i$ and assign a probability distribution $p_{\mathrm{SE}}$ for each $\boldsymbol{h}_i$ in the input sentence. 
Second, we use each $\boldsymbol{h}_i$ to find the most similar token representations in the datastore and get the probability distribution $p_{\mathrm{kNN}}$, aggregated from the $k$-nearest context--token pairs. 
Last, we obtain the final probability distribution $p$ by interpolating between the two distributions. 
In addition to formalizing \nnose{}, we apply the Whitening Transformation (\cref{subsec:zca}) to the embeddings, an important process for \knn{} approaches as used in previous work~\citep{su2021whitening, yin2022efficient}.

\subsection{Datastore} 
The datastore $\mathcal{D}$ comprises key--value pairs $(\boldsymbol{h}_i, y_i)$, where each $\boldsymbol{h}_i$ represents the contextualized token embedding computed by a \emph{fine-tuned} SE encoder, and $y_i \in \{\text{\tt{B}, \tt{I}, \tt{O}}\}$ denotes the corresponding gold label. 
Typically, the datastore consists of all tokens from the training set. 
In contrast to the approach employed by~\citet{wang2022k} for \knn{}--NER, where they only store \texttt{B} and \texttt{I} tags in the datastore (only named entities), we also include the \texttt{O}-tag in the datastore. This allows us to retrieve non-named entities, which is more intuitive than assigning non-entity probability mass to the \texttt{B} and \texttt{I} tokens.

\subsection{Inference} 
During inference, the \nnose{} model aims to predict $y_i$ based on the contextual representation of $x_i$ (i.e., $\boldsymbol{h}_i$). 
This representation is used to query the datastore for \knn{} using an $L^2$ distance measure (following~\citealp{Khandelwal2020Generalization}), denoted as $d(\cdot, \cdot)$. 
Once the neighbors are retrieved, the model computes a distribution over the neighbors by applying a softmax function with a temperature parameter $T$ to their negative distances (i.e., similarities). 
This aggregation of probability mass for each label (\text{\tt{B}, \tt{I}, \tt{O}}) across all occurrences in the retrieved targets is represented as:

\begin{equation}
\small
p_{\mathrm{kNN}}(y_i \mid x_i) \propto \sum_{(k_i, v_i) \in \mathcal{D}} \mathbbm{1}_{y=v_i} \exp\left(\frac{-d(\boldsymbol{h}_i, \boldsymbol{k})}{T}\right).
\end{equation}

Items that do not appear in the retrieved targets have zero probability. 
Finally, we interpolate the nearest neighbors distribution $p_{\mathrm{kNN}}$ with the fine-tuned model distribution $p_{\mathrm{SE}}$ using a tuned parameter $\lambda$ to produce the final \nnose{} distribution $p$:

\begin{equation}
\begin{aligned}
p(y_i \mid x_i) = \hspace{1mm} & \lambda \times p_{\mathrm{kNN}}\left(y_i \mid x_i\right) + (1-\lambda) \times p_{\mathrm{SE}}\left(y_i \mid x_i\right).
\end{aligned}
\end{equation}

\subsection{Whitening Transformation}\label{subsec:zca}
Several works (\citealp{li-etal-2020-sentence, su2021whitening, huang-etal-2021-whiteningbert-easy}) note that if a set of vectors are isotropic, we can assume it is derived from the Standard Orthogonal Basis, which also indicates that we can properly calculate the similarity between embeddings. 
Otherwise, if it is anisotropic, we need to transform the original sentence embedding to enforce isotrophorism, and then measure similarity. \citet{su2021whitening, huang-etal-2021-whiteningbert-easy} applies the vector whitening approach~\citep{koivunen1999feasibility} on BERT~\citep{devlin2019bert}. 
The Whitening Transformation (\texttt{WT}), initially employed in data preprocessing, aims to eliminate correlations among the input data features for a model.
In turn, this can improve the performance of certain models that rely on uncorrelated features.
Other works (\citealp{gao2018representation, ethayarajh2019contextual, li2020sentence, yan2021consert, jiang2022promptbert}, among others) found that (frequency) biased \emph{token} embeddings hurt final sentence representations. 
These works often link token embedding bias to the token embedding anisotropy and argue it is the main reason for the bias. 
We apply \texttt{WT} to the token embeddings like previous work for nearest neighbor retrieval~\citep{yin2022efficient}. 
In short, \texttt{WT} transforms the mean value of the embeddings into 0 and the covariance matrix into the identity matrix, and these transformations are then applied to the original embeddings. 
We apply \texttt{WT} to the embeddings before putting them in the datastore and before querying the datastore.
The workflow of \texttt{WT} is detailed in \cref{app:wt}.

\section{Experimental Setup}\label{sec:exp}

\begin{table}[t]
\centering
\resizebox{\linewidth}{!}{
\begin{tabular}[t]{lllrrrr}
\toprule
\textbf{Dataset}       &    \textbf{Loc.}    & \textbf{License}    &   \textbf{Train}   & \textbf{Dev.}  &   \textbf{Test}  & $\mathcal{D}$ (tokens) \\\midrule
\textsc{SkillSpan}     &    *        & CC-BY-4.0  &   5,866   & 3,992 &   4,680 & 86.5K  \\
\textsc{Sayfullina}    &    UK       & Unknown    &   3,706   & 1,854 &   1,853 & 53.1K   \\
\textsc{Green}         &    UK       & CC-BY-4.0  &   8,670   & 963   &   336   & 209.5K  \\\midrule
\textsc{Total}         & \multicolumn{5}{c}{}                                                          & 349.2K \\
\bottomrule
\end{tabular}}
\caption{\textbf{Dataset Statistics.} We provide statistics for all three datasets, including the location and license. Input granularity is at the token level, with performance measured in span-F1. The size of the datastore $\mathcal{D}$ is in tokens and determined by embedding tokens and their context from the training sets, resulting in approximately 350K keys. See~\cref{data:examples7} for examples.
}
\label{tab:num_post7}
\end{table}

\subsection{Data}

All datasets are in English and have different label spaces. 
We transform all skills to the same label space and give each token a generic tag (i.e., {\texttt{B}, \texttt{I}, \texttt{O}}).
We give a brief description of each dataset below and \cref{tab:num_post7} summarizes them:

\paragraph{\textsc{SkillSpan}~\citep{zhang-etal-2022-skillspan}.}
This job posting dataset includes annotations for skills and knowledge derived from the ESCO taxonomy.
To fit our approach, we flatten the two label layers into one layer (i.e., \texttt{BIO}).
The baseline is the JobBERT model, which was continuously pre-trained on a dataset of 3.2 million job posting sentences. 
The industries represented in the data range from tech to more labor-intensive sectors.

\paragraph{\textsc{Sayfullina}~\citep{sayfullina2018learning}}
is used for soft skill sequence labeling. 
Soft skills are personal qualities that contribute to success, such as teamwork, dynamism, and independence. 
Data originated from the UK.
This is the smallest dataset among the three, with no specified industries.

\paragraph{\textsc{Green}~\citep{green-maynard-lin:2022:LREC}.}
A dataset for extracting skills, qualifications, job domain, experience, and occupation labels. 
The dataset consists of jobs from the UK, and the industries represented include IT, finance, healthcare, and sales. 
This is the largest dataset among the three.

\subsection{Models}

We use 3 English-based LMs: 1 general-purpose and 2 domain-specific models. 
Implementation details for fine-tuning and \nnose{} are in~\cref{sec:params}.

\paragraph{JobBERT~\citep{zhang-etal-2022-skillspan}} is a 110M parameter BERT-based model continuously pre-trained~\citep{gururangan2020don} on 3.2M English job posting sentences. 
It outperforms \bertb{} on several skill-specific tasks.

\paragraph{RoBERTa~\citep{liu2019roberta}.} 
We also use \roberta{} (123M parameters). 
It showed to outperform JobBERT in our initial experiments and we therefore include this model as a baseline.

\paragraph{JobBERTa (Ours).} 
Given that RoBERTa outperformed JobBERT, we create another baseline and release a model named JobBERTa. 
This is a \roberta{} model continuously pre-trained~\citep{gururangan2020don} on the same 3.2M JD sentences as JobBERT.

\begin{table}[t!]
\centering
\setlength{\arrayrulewidth}{0.25mm}
\resizebox{\linewidth}{!}{
\begin{tabular}[t]{lllll|l}%
\toprule
                                  & \textbf{Setting}                         & \textsc{\textbf{SkillSpan}}        & \textsc{\textbf{Sayfullina}}        &  \textsc{\textbf{Green}}           & \textbf{avg.\ span-F1}    \\\midrule
JobBERT~\citep{zhang-etal-2022-skillspan}  &                                 &  60.47                    &  88.16                     &  42.55                    & 63.73            \\         
\hspace{0.2em} + \knn{}                   & \dwt{}                          &  61.06 \up{0.59}          &  88.25 \up{0.09}           &  43.56 \up{1.01}          & 64.29 \up{0.56}  \\   
\hspace{0.2em} + \knn{}                   & \adwt{}                         &  60.93 \up{0.48}          &  88.26 \up{0.10}           &  44.44 \up{1.89}          & 64.54 \up{0.81} \\\midrule

RoBERTa~\citep{liu2019roberta}             &                                 &  63.88                    &  91.97                     &  44.49                    & 66.78            \\          
\hspace{0.2em} + \knn{}                   & \dwt{}                          &  63.57 \down{0.31}        &  91.97 \none{0.00}         &  45.02    \up{0.53}       & 66.85 \up{0.07}  \\ 
\hspace{0.2em} + \knn{}                   & \adwt{}                         &  63.98 \up{0.10}          &  91.97 \none{0.00}         &  44.86 \up{0.37}          & 66.94 \up{0.16}     \\\midrule    

JobBERTa (This work)                      &                                 &  63.74                    &  92.06                     &  49.61                    & 68.47  \\          
\hspace{0.2em} + \knn{}                   & \dwt{}                          &  64.14 \up{0.40}          &  91.89 \down{0.17}         &  50.35 \up{0.74}          & 68.79 \up{0.32}   \\  
\hspace{0.2em} + \knn{}                   & \adwt{}                         &  \textbf{64.24} \up{0.50}\textsuperscript{$\dagger$}   &  \textbf{92.15} \up{0.09} &  \textbf{50.78} \up{1.17}\textsuperscript{$\dagger$} & \textbf{69.06} \up{0.59}\\          
\bottomrule            
\end{tabular}}
\caption{
\textbf{Test Set Results.} Two settings are considered for each model based on dev.\ set results in~\cref{dev:results}: \dd{} refers to the in-dataset datastore, containing keys from the specific training data, while \ad{} represents a datastore with keys from all available training sets. The notation \texttt{+WT} indicates the application of Whitening Transformation to the keys before adding them to and querying the datastore. The performance impact of using \knn{} is indicated as \up{} (increase), \down{} (decrease), or \textbf{--} (no change). The best-performing setup for each dataset is highlighted. For the top-performing model (JobBERTa), \textsuperscript{$\dagger$} signifies statistical significance over the baseline using a token-level McNemar test~\citep{mcnemar1947note}. The avg.\ span-F1 performance of each model across the three datasets is displayed.
}
\label{tab:res}
\end{table}

\section{Results}\label{sec:results7}

We evaluate the performance of fine-tuning models enhanced with \nnose{}. 
We consider different setups: First, we compare using the Whitening Transformation (\texttt{+WT}) or without. 
Second, we explore two datastore setups: One using an in-dataset datastore (\dd{}), where each respective training set is stored separately, and another where all datasets are stored in the datastore (\ad{}). 
In the latter setup, we encode all three datasets with each fine-tuned model, and each model has its own \texttt{WT} matrix. 
For example, we fine-tune a model on \textsc{SkillSpan} and encode the training set tokens of \textsc{SkillSpan}, \textsc{Sayfullina}, and \textsc{Green} to populate the datastore. 
From the results on the development set (\cref{tab:dev}, \cref{dev:results}), we observe that adding \texttt{WT} consistently improves performance. 
Therefore, we only report the span-F1 scores on each test set (\cref{tab:res}) \emph{with} \texttt{WT} and the average over all three datasets. 

\subsection{Best Model Performance} In~\cref{tab:res}, we show that the best-performing baseline model is JobBERTa, achieving more than 4 points span-F1 improvement over JobBERT and 2 points higher than RoBERTa on average. 
This confirms the effectiveness of DAPT in improving language models~\citep{han-eisenstein-2019-unsupervised,alsentzer2019publicly,gururangan2020don,lee2020biobert,nguyen2020bertweet,zhang-etal-2022-skillspan}.

\subsection{Best \nnose{} Setting} 
We confirm the trends from dev.\ on test: The largest improvements come from using the setup with \texttt{WT}, especially in the \adwt{} setting. 
All models seem to benefit from the \nnose{} setup, JobBERT and JobBERTa shows the largest improvements, with the largest gains observed in the \adwt{} datastore setup. 
In summary, \adwt{} consistently demonstrates performance enhancements across all experimental setups.%

\begin{figure*}[t]
    \centering
    \includegraphics[width=\linewidth]{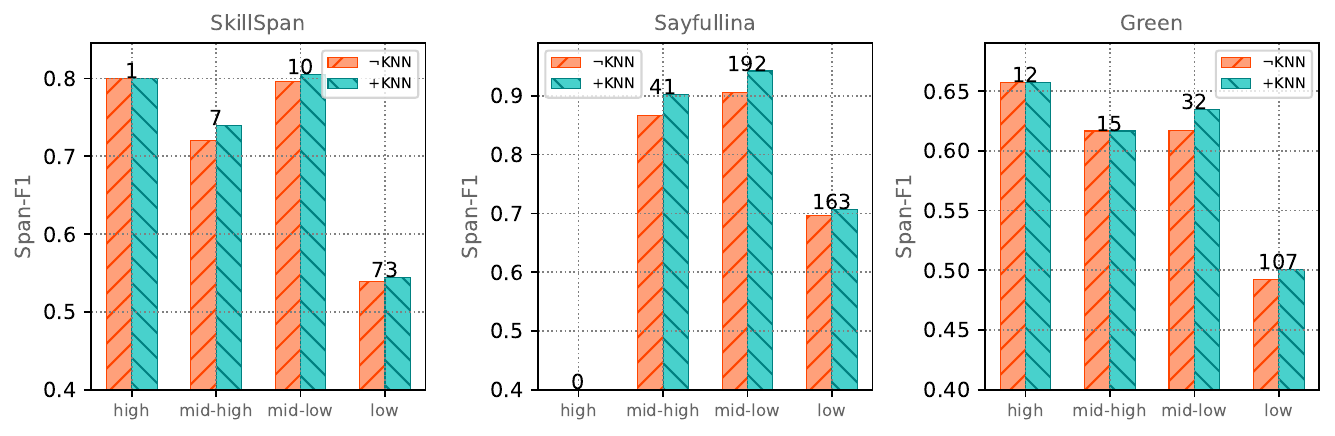}
    \caption{
    \textbf{Long-tail Prediction Performance.} \knn{} is based on the datastore with all the datasets. We categorize the occurrences of a skill in the test set with respect to the training set. For example, a skill in the test set occurs two times in the training set, we put this in the ``low'' bin. There are three frequency ranges: \emph{high}: 10--15, \emph{mid--high}: 7--10, \emph{mid--low}: 4--6, \emph{low}: 0--3. \textsc{Sayfullina} does not have any test set skills that occur more than 10 times in the training set. On top of the bars is the number of predicted skills for the test set in each bucket.
    }
    \label{fig:longtail}
\end{figure*}

\section{Analysis}
As we store training tokens from all datasets in the datastore, we expect the model to recall a greater number of skills based on the current context during inference. In turn, this would lead to improved downstream model performance. We want to address the challenges of SE datasets by predicting long-tail patterns, and if we observe improvements in detecting unseen skills in a cross-dataset setting. 

To investigate in which situations our model improves, we are analyzing the following:
\circled{1} The predictive capability of \nnose{} in relation to rarely occurring skills compared to regular fine-tuning (\cref{subsec:longtail}). Skills exhibit varying frequencies across datasets, we categorize the skill frequencies into buckets and compare the performance between vanilla fine-tuning and the inclusion of \knn{}.
\circled{2} If \nnose{} actually retrieves from other datasets when they are combined (\cref{subsec:all}), and if there is a sign of leveraging multiple datasets, then;
\circled{3} How much does \nnose{} enhance performance in a cross-dataset setting (\cref{subsec:unseen})? Our results indicate a large performance drop when a fine-tuned SE model, trained on one dataset, is applied to another dataset, highlighting the sparsity across datasets. We demonstrate that \nnose{} helps alleviate this, both from an empirical perspective and by inspecting the prediction errors (\cref{subsec:errors}).

\subsection{Long-tail Skills Prediction}\label{subsec:longtail}
\citet{Khandelwal2020Generalization} observed that due to explicitly memorizing the training data, NNLMs effectively predict rare patterns. 
We analyze whether the performance of ``long-tail skills'' improves using \nnose{}. 
A visualization of the long-tail distribution of skills is in \cref{fig:skilldistribution} (\cref{app:skilldistribution}).

We present the results in \cref{fig:longtail}. 
We investigate the performance of JobBERTa with and without \knn{} based on the occurrences of skills in the evaluation set relative to the train set. 
We count the skills in the evaluation set that occur a number of times in the training set, ranging from 0--15 occurrences and is grouped into low, mid--low, mid--high, and high--frequency bins (0--3, 4--6, 7--10, 10--15, respectively). 
This approach estimates the number of skills the LM recalls from the training stage.

Our findings reveal that skills with low-frequent skills are the most difficult and make up the largest bucket, and our approach is able to improve on them on all three datasets.
For \textsc{SkillSpan}, we observe an improvement in the low-frequency bin, from 53.9$\rightarrow$54.5 span-F1. 
Similarly, \textsc{Green} exhibits a similar trend with an improvement in the low-frequency bin (49.2$\rightarrow$50.1). 
Interestingly, it also shows gains in most other frequency bins. 
Last, for \textsc{Sayfullina}, there is also an improvement (69.7$\rightarrow$70.7 in the low bin). 
It is worth pointing out that there are many skills that fall in the low bin in \textsc{SkillSpan} and \textsc{Green}. 
This is exactly where \nnose{} improves most for these datasets. For \textsc{Sayfullina}, we notice the largest number of predicted skills is in the mid--low bin. This is where we also see improvements for \nnose{}.

\subsection{Retrieving From All Datasets}\label{subsec:all}
\begin{figure*}[t]
    \centering
    \includegraphics[width=\linewidth]{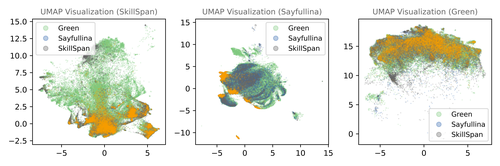}
    \caption{
    \textbf{UMAP Visualization of Nearest Neighbors Retrieval.} The datastore consists of the training set (\texttt{+WT}) of all three datasets used in this work. Each colored dot represents a non-\texttt{O} token from the training set. The embeddings are generated using JobBERTa. The orange shade represents the retrieved neighbors with $k=4$ for each token that is a skill (i.e., not an \texttt{O} token). Note that for the middle plot, the orange shade covers the blue clusters \textsc{Sayfullina}. \textsc{Green} has the green shade and \textsc{SkillSpan} are the darkcyan colors.
    }
    \label{fig:ds-viz}
\end{figure*}

We presented the best improvements of \nnose{} in the \adwt{} datastore in \cref{sec:results7}.
An important question remains: Does the \adwt{} setting retrieve from all datasets? 
Qualitatively,~\cref{fig:ds-viz} shows the UMAP visualization \citep{mcinnes2018umap-software} of representations stored in each \adwt{} datastore. 
We mark the retrieved neighbors with orange for each downstream dev.\ set. 
In all plots, we observe that \textsc{Green} is prominent in the representation space (green), while \textsc{SkillSpan} (darkcyan) and \textsc{Sayfullina} (blue) form distinct clusters. 
Each plot has its own pattern: \textsc{SkillSpan} and \textsc{Sayfullina} have well-shaped clusters, while \textsc{Green} consists of one large cluster. 
\textsc{SkillSpan} and \textsc{Sayfullina} mostly retrieve from their own clusters. 
In contrast, \textsc{Green} retrieves from the entire representation space, which could explain the largest span-F1 performance gains (\cref{tab:res}).
This suggests that \knn{} effectively leverages multiple datasets in most cases (qualitative analysis see \cref{app:qualitative}).

\begin{figure*}[t!]
    \centering
    \includegraphics[width=\linewidth]{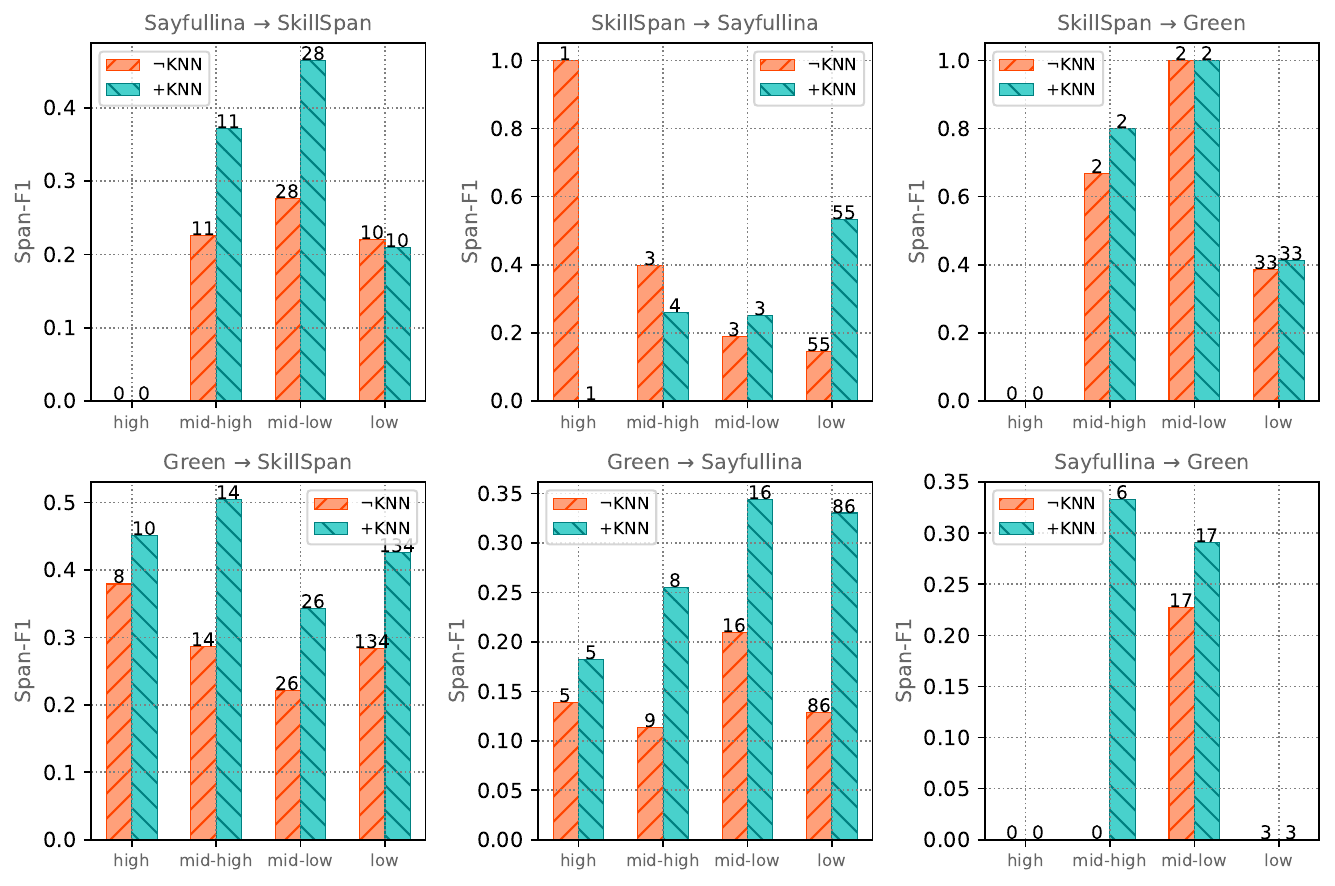}
    \caption{
    \textbf{Cross-dataset Long-tail Performance.} Similar to~\cref{fig:longtail}, we plot the cross-dataset long-tail performance. \nnose{} uses the datastore with all datasets. Training and evaluation data (test) are indicated in graph titles. Frequency bins are based on the training data span frequency; there are three frequency ranges: \emph{high}: 10--15, \emph{mid--high}: 7--10, \emph{mid--low}: 4--6, \emph{low}: 0--3.
    }
    \label{fig:crossdataset}
\end{figure*}

\subsection{Prediction of Unseen Skills}\label{subsec:unseen}
The UMAP plots in \cref{fig:ds-viz} suggest that some datasets are closer to each other than others. 
To quantify this, we investigate the overlap of annotated skills between datasets and assess cross-dataset performance of \nnose{} on unseen skills.

\paragraph{Overlap of Datasets.}
We calculate the exact span overlap of skills between the training sets of the datasets using the Jaccard similarity coefficient \citep{jaccard1901distribution}: $J(A,B) = \frac{|A \cap B|}{|A \cup B|}$, where $A$ and $B$ are sets of multi-token spans (e.g., ``manage a team'') from two separate training sets.
The Jaccard similarity coefficients are as follows: $J$(\textsc{SkillSpan}, \textsc{Sayfullina}) = 0.35, $J$(\textsc{Sayfullina}, \textsc{Green}) = 0.10, and $J$(\textsc{SkillSpan}, \textsc{Green}) = 0.29. 
These Jaccard coefficients indicate overlap between unique skill spans across datasets, suggesting that \nnose{} can introduce the model to new and unseen skills.

\begin{table}[t]
\centering
\setlength{\arrayrulewidth}{0.25mm}
\begin{tabular}{lllll}
\toprule
& $\downarrow$Trained on     & \textsc{SkillSpan}        & \textsc{Sayfullina} &  \textsc{Green}      \\\midrule       
\vanilla& \textsc{SkillSpan} &  \cellcolor{gray}         &  18.05              &  43.17               \\          
& \textsc{Sayfullina}        &  9.44                     &  \cellcolor{gray}   &  11.79                \\          
& \textsc{Green}             &  29.67                    &  15.93              &  \cellcolor{gray}    \\\midrule   
& \textsc{All}               &  59.33                    &  90.16              &  44.59               \\\midrule     

\withknn& \textsc{SkillSpan} &  \cellcolor{gray}         &  45.86 \up{27.81}   &  45.44 \up{2.27}     \\          

& \textsc{Sayfullina}        &  26.16 \up{16.72}         &  \cellcolor{gray}   &  25.38 \up{13.59}     \\          
& \textsc{Green}             &  41.22 \up{11.55}         &  46.58 \up{30.65}   &  \cellcolor{gray}     \\\midrule   
& \textsc{All}               &  59.51 \up{0.31}          &  90.33 \up{0.17}    &  45.63 \up{1.04}     \\\bottomrule
\end{tabular}
\caption{
\textbf{Results of Unseen Skills based on JobBERTa (\adwt{}).} In the vanilla setting, models trained on one skill dataset are applied to another on test, showing varied performance. However, applying \knn{} improves the detection of unseen skills. Diagonal results can be found in~\cref{tab:res}. Refer to~\cref{tab:unseenparams} for tuned hyperparameters.
}
\label{tab:unseen}
\end{table}

\paragraph{Results.}
\cref{tab:unseen} presents the performance of JobBERTa across datasets. 
For completeness, we include a baseline where JobBERTa is fine-tuned on a union of all datasets (\textsc{All}). 
We notice training on the union of the data never leads to the best target dataset performance.
Generally, we observe that in-domain data is best, both in vanilla and NNOSE setups (diagonal in \cref{tab:unseen}). Performance drops when a model is applied to a dataset other than the one it was trained on (off-diagonal). 
Using \nnose{} leads to substantial improvements across the challenging off-diagonal (cross-dataset) settings, while performance remains stable within datasets. 
We observe the largest improvements when applied to \textsc{Sayfullina}, with \emph{up to a 30\%} increase in span-F1. 
This is likely due to \textsc{Sayfullina} consisting mostly of soft skills, which are less prevalent in \textsc{SkillSpan} and \textsc{Green}, making it beneficial to introduce soft skills. 
Conversely, when the model is trained on \textsc{Sayfullina}, the absolute improvement on \textsc{SkillSpan} is lower, indicating that skill datasets can benefit each other to different extents.
 
\subsection{Cross-dataset Long-tail Analysis}
\cref{tab:unseen} shows improvements when \nnose{} is used in favor of vanilla fine-tuning. 
\cref{fig:crossdataset} presents the long-tail performance analysis in the cross-dataset scenario, similar to \cref{fig:longtail}. 
We observe the largest gains with \nnose{} in the low or mid--low frequency bins. 
However, exceptions are \textsc{SkillSpan}$\rightarrow$\textsc{Green} and \textsc{Sayfullina}$\rightarrow$\textsc{Green}, where most gains occur in the mid--high bin. 
Notably, \textsc{Sayfullina}$\rightarrow$\textsc{Green} demonstrates higher performance with \nnose{}, where all 6 skills are incorrectly predicted in the mid--high bin.
An analysis of precision and recall in \cref{tab:pr} (\cref{app:cr}) substantiates that the improvements are both precision and recall-based, with gains of up to 40 recall points and 35.4 precision points in \textsc{Green}$\rightarrow$\textsc{Sayfullina}. There is also an improvement up to 35.5 recall points and 34.1 precision points for \textsc{SkillSpan}$\rightarrow$\textsc{Sayfullina}.
This further solidifies that memorizing tokens (i.e., storing all skills in the datastore) helps recall as mentioned in~\citet{Khandelwal2020Generalization}, and more importantly, highlighting the benefits of \nnose{} in cross-dataset scenarios for SE.

\begin{table*}[t]
    \centering
    \resizebox{\linewidth}{!}{
    \begin{tabular}{lll}
    \toprule
                           & \textbf{False Positives}                           & \textbf{False Negatives}  \\\midrule
                           & cleaning                                           & GCP \\
    \textsc{SkillSpan}     & decisive                                           & IBM MQ \\
                           & Apache Camel                                       & AWS \\
                           & building consumer demand for sustainable products  & budget responsible\\\midrule
                           & empathy                                            & leadership      \\
      \textsc{Sayfullina}  & leadership management                              &                 \\
                           & communication                                      &                 \\
                           & ability to manage and prioritise multiple assignments and tasks &    \\\midrule
                           & SQL scripting languages                            & software engineering \\
          \textsc{Green}   & Manage a team                                      & development \\
                           & troubleshooting activities                         & DevOps \\
                           & dealing with tenants                               & Cisco network administration \\\bottomrule
    \end{tabular}
    }
    \caption{\textbf{FPs \& FNs of \nnose{}.} We show several examples of false positives and false negatives in each dataset. We only show the predictions of \nnose{} that are \emph{not} in the vanilla model predictions.}
    \label{tab:fpfn}
\end{table*}

\subsection{Qualitative Check on Prediction Errors}\label{subsec:errors}
We perform a qualitative analysis on the false positives (fp) and false negatives (fn) of \nnose{} predictions compared to vanilla fine-tuning for each dataset. 
This analysis tells us whether a prediction corresponds to an actual skill, even if it does not contribute positively to the span-F1 metric. 
We observe that \nnose{} %
produces a significant number of false positives that are ``similar'' to genuine skills. 
In~\cref{tab:fpfn}, for each dataset, we picked five fps and fns that represent hard, soft, and personal skills well (if applicable). We show the fps and fns for JobBERTa with \nnose{}, we only show predictions that are \emph{not} in the vainlla model predictions. 
In \textsc{Sayfullina}, there is only one fn. We notice from the errors, and especially the fps, that these are definitely skills, indicating the benefit of \nnose{} helping to predict new skills or missed annotations.

\section{Related Work}
\subsection{Skill Extraction}
The dynamic nature of labor markets has led to an increase in tasks related to JD, including skill extraction~\citep{kivimaki-etal-2013-graph,zhao2015skill,sayfullina2018learning,smith2019syntax,tamburri2020dataops,shi2020salience, chernova2020occupational,bhola-etal-2020-retrieving,gugnani2020implicit,fareri2021skillner,konstantinidis2022knowledge, zhang-etal-2022-skillspan,zhang-jensen-plank:2022:LREC,zhang2022skill,green-maynard-lin:2022:LREC,gnehm-bhlmann-clematide:2022:LREC,beauchemin2022fijo,decorte2022design, ao2023skill, goyal-etal-2023-jobxmlc, zhang2023escoxlmr}. 
These works employ methods such as sequence labeling \citep{sayfullina2018learning, smith2019syntax, chernova2020occupational, zhang-etal-2022-skillspan, zhang2022skill}, multi-label classification \citep{bhola-etal-2020-retrieving}, and graph-based methods \citep{shi2020salience, goyal-etal-2023-jobxmlc}. 
Recent methodologies include domain-specific models where LMs are continuously pre-trained on unlabeled JD \citep{zhang-etal-2022-skillspan, gnehm-bhlmann-clematide:2022:LREC}. 
However, none of these methodologies introduce a retrieval-augmented model like \nnose{}.

\subsection{General Retrieval-augmentation}
In retrieval augmentation, LMs can utilize external modules to enhance their context-processing ability.
Two approaches are commonly used: First, using a separately trained model to retrieve relevant documents from a collection. 
This approach is employed in open-domain question answering tasks~\citep{petroni-etal-2021-kilt} and with specific models such as ORQA~\citep{lee-etal-2019-latent}, REALM~\citep{guu2020retrieval}, RAG~\citep{lewis2020retrieval}, FiD~\citep{izacard2021distilling}, and ATLAS~\citep{izacard2022few}.

Second, previous work on explicit memorization showed promising results with a cache~\citep{grave2017improving}, which serves as a type of datastore. 
The cache contains past hidden states of the model as keys and the next word as tokens in key--value pairs. Memorization of hidden states in a datastore, involves using the \knn{} algorithm as the retriever. 
The first work of the \knn{} algorithm as the retrieval component was by \citet{Khandelwal2020Generalization}, leading to several LM decoder-based works.

\subsection{Decoder-based Nearest Neighbor Approaches}
Decoder-based nearest neighbors approaches are primarily focused on language modeling~\citep{Khandelwal2020Generalization, he-etal-2021-efficient, yogatama2021adaptive, ton2022regularized, shi2022nearest, jin2022plug, bhardwaj2022adaptation, xu2023nearest} and machine translation~\citep{khandelwal2021nearest, zheng2021adaptive, jiang2021learning, jiang2022towards, wang2022efficient, martins2022efficient, martins2022chunk, zhu2022knowledge, du2023federated, zhu2023knn, min-etal-2023-nonparametric, min2023silo}. 
These approaches often prioritize efficiency and storage space reduction, as the datastores for these tasks can contain billions of tokens.

\subsection{Encoder-based Nearest Neighbor Approaches.}
Encoder-based nearest neighbor approaches have been explored in tasks such as named entity recognition~\citep{wang2022k} and emotion classification~\citep{yin2022efficient}. 
Here, the datastores are limited to single datasets with the sentence (or token) gold label pairs. 
Instead, we show the potential of adding multiple datasets in the datastore.

\section{Conclusion}
We introduce \nnose{}, an LM that incorporates and leverages a non-parametric datastore for nearest neighbor retrieval of skill tokens. 
To the best of our knowledge, we are the first to introduce the nearest neighbors retrieval component for the extraction of occupational skills. 
We evaluated \nnose{} on three relevant skill datasets with a wide range of skills and show that \nnose{} enhances the performance of all LMs used in this work \emph{without} additionally tuning the LM parameters. 
Through the combination of train sets in the datastore, our analysis reveals that \nnose{} effectively leverages all the datasets by retrieving tokens from each. 
Moreover, \nnose{} not only performs well on rare skills but also enhances the performance of more frequent patterns. 
Lastly, we observe that our baseline models exhibit poor performance when applied in a cross-dataset setting. 
However, with the introduction of \nnose{}, the models improve across all settings. 
Overall, our findings indicate that \nnose{} is a promising approach for application-specific skill extraction setups and potentially helps discover skills that were missed in manual annotations.

\section*{Limitations}
We consider several limitations: One is the limited diversity of the datasets used in this work. 
Our study was constrained by the use of only three English datasets. 
By focusing solely on English data, we might have overlooked insights that exist in other languages. 
While these datasets were carefully selected to ensure relevance and quality, the limited scope of the data may restrict the generalizability of our findings to other SE datasets. 
Future research includes incorporating a wider range of datasets from diverse sources to obtain a more comprehensive understanding of the topic. 
Potential interesting future work should include validation on whether \nnose{} works in a multilingual setting.

Another limitation is that we do skill detection and not specific labeling of the extracted spans, i.e., extracting generic \texttt{B}, \texttt{I}, \texttt{O} tags.
This was to ensure that the datasets could be used in unison in the datastore. 
Interesting future work could extending \nnose{} to include labeled skills in the datastore.

\section*{Ethics Statement}

The subject of job-related language models is a highly contentious topic, often sparking intense debates surrounding the issue of bias. 
We acknowledge that LMs such as JobBERTa and \nnose{} possess the potential for inadvertent consequences, such as unconscious bias and dual-use when employed in the candidate selection process for specific job positions. 
There are research efforts to develop fairer recommender systems in the field of human resources, focusing on mitigating biases (e.g., \citealp{mujtaba2019ethical, raghavan2020mitigating, deshpande2020mitigating, kochling2020discriminated, sanchez2020does, wilson2021building, vanimproving, arafan2022end}). 
Nevertheless, one potential approach to alleviating such biases involves the retrieval of sparse skills for recall (e.g., this work). 
It is important to note, however, that we have not conducted an analysis to ascertain whether this particular method exacerbates any pre-existing forms of bias.

\clearpage
\section{Appendix}

\begin{figure*}[t]
\begin{minipage}{0.3\textwidth}
\footnotesize
\begin{minted}[frame=single,
               framesep=2mm,
               linenos=true,
               xleftmargin=15pt,
               tabsize=1]{xml}
Experience	O
in	        O
working     B
on          I
a           I
cloud-based I
application I
running     O
on          O
Docker      B
.           O

A           O
degree      B
in          I
Computer    I
Science     I
or          O
related     O
fields      O
.           O
\end{minted}
\caption{\textbf{Data Example for SkillSpan.} In \textsc{SkillSpan}, we would like to note the long skills.} 
\label{json-skillspan}
\end{minipage}\hfill
\begin{minipage}{0.3\textwidth}
\footnotesize
\begin{minted}[frame=single,
               framesep=2mm,
               linenos=true,
               xleftmargin=15pt,
               tabsize=1]{xml}
ability     O
to          O
work        B
under       I
stress      I
condition   O

due         O
to          O
the         O
dynamic     B
nature      O
of          O
the         O
group       O
environment O
,           O
the         O
ideal       O
candidate   O
will        O
\end{minted}
\caption{\textbf{Data Example for Sayfullina.} In \textsc{Sayfullina}, the skills are usually soft-like skills.} 
\label{json-sayfullina}
\end{minipage}\hfill
\begin{minipage}{0.3\textwidth}
\footnotesize
\begin{minted}[frame=single,
               framesep=2mm,
               linenos=true,
               xleftmargin=15pt,
               tabsize=1]{xml}
A               O
sound           O
understanding   O
of              O
the             O
Care            B
Standards       I
together        O
with            O
a               O
Nursing         B
qualification   I
and             O
current         O
NMC             B
registration    I
are             O
essential       O
for             O
this            O
role            O
\end{minted}
\caption{\textbf{Data Example for Green.} There are many qualification skills (e.g., certificates).} 
\label{json-green}
\end{minipage}
\end{figure*}

\subsection{Whitening Transformation Algorithm}\label{app:wt}

\begin{algorithm}[htbp]
\SetAlgoLined
 \textbf{input:} Embeddings $\left\{x_i\right\}_{i=1}^N$;\\

 Compute $\mu=\frac{1}{N} \sum_{i=1}^N x_i$ and $\Sigma$ of $\left\{x_i\right\}_{i=1}^N$

 Compute $U, \Lambda, U^\top = \text{SVD}(\Sigma)$

 Compute $W = U\sqrt{\Lambda^{-1}}$

 \For{$i = 1, 2, ..., n$}{
    $\widetilde{x}_i=\left(x_i-\mu\right) W$
 
 }
 \textbf{return} $\left\{\widetilde{x}_i\right\}_{i=1}^N$;
 \caption{Whitening Transformation Workflow}
 \label{algo:white}
\end{algorithm}

We apply the whitening transformation to the query embedding and the embeddings in the datastore. 
We can write a set of token embeddings as a set of row vectors: $\left\{x_i\right\}_{i=1}^N$. Additionally, a linear transformation $\widetilde{x}_i=\left(x_i-\mu\right) W$ is applied, where $\mu=\frac{1}{N} \sum_{i=1}^N x_i$. 
To obtain the matrix $W$, the following steps are conducted: First, we obtain the original covariance matrix

\begin{equation}
    \Sigma=\frac{1}{N} \sum_{i=1}^N\left(x_i-\mu\right)^\top\left(x_i-\mu\right).
\end{equation}

Afterwards, we obtain the transformed covariance matrix $\widetilde{\Sigma}=W^\top \Sigma W$, where we specify $\widetilde{\Sigma}=I$. 
Therefore, $\Sigma =\left(W^\top\right)^{-1} W^{-1} = \left(W^{-1}\right)^\top W^{-1}$. 
Here, $\Sigma$ is a positive definite symmetric matrix that satisfies the following singular value decomposition (SVD;~\citealp{golub1971singular}) as indicated by \citet{su2021whitening}:
$\Sigma= U\Lambda U^\top.$
$U$ is an orthogonal matrix, $\Lambda$ is a diagonal matrix, and the diagonal elements are all positive. 
Therefore, let $W^{-1}=\sqrt{\Lambda}U^\top$, we obtain the solution: $W = U\sqrt{\Lambda^{-1}}$. Putting it all together, as input, we have the set of embeddings $\left\{x_i\right\}_{i=1}^N$. We compute $\mu$ and $\Sigma$ of $\left\{x_i\right\}_{i=1}^N$. 
Then, we perform SVD on $\Sigma$ to obtain matrices $U$, $\Lambda$, and $U^\top$. Using these matrices, we calculate the transformation matrix $W$. 
Finally, we apply the transformation to each embedding in the set by subtracting $\mu$ and multiplying by $W$. We are left with $\widetilde{x}_i=\left(x_i-\mu\right) W$. Note that we do \texttt{WT} \emph{before} we store the embedding in the datastore, and apply \texttt{WT} to the token embedding before we query the datastore.

We show the Whitening Transformation procedure in~\cref{algo:white}. Note that \citet{li-etal-2020-sentence, su2021whitening} introduced a dimensionality reduction factor $k$ on $W$ ($W[:, :k]$). 
he diagonal elements in the matrix $\Lambda$ obtained from the SVD algorithm are in descending order. 
One can decide to keep the first $k$ columns of $W$ in line 6. 
This is similar to PCA \citep{abdi2010principal}. 
However, empirically, we found that reducing dimensionality had a negative effect on downstream performance, thus we omit that in this implementation.

\subsection{Data Examples}\label{data:examples7}

\begin{table}[ht]
    \centering
    \begin{tabular}{ll}
    \toprule
    \textsc{SkillSpan}      & \cref{json-skillspan} \\
    \textsc{Sayfullina}     & \cref{json-sayfullina} \\
    \textsc{Green}          & \cref{json-green} \\
    \bottomrule
    \end{tabular}
    \caption{Data example references for each dataset.}
    \label{tab:dataexamples}
\end{table}

In \cref{tab:dataexamples}, we refer to several listings of examples of the datasets. Notably in \textsc{SkillSpan}, the original samples contain two columns of labels. These refer to skills and knowledge. To accommodate for the approach of \nnose{}, we merge the labels together and thus removing the possible nesting of skills.~\citet{zhang-etal-2022-skillspan} mentions that there is not a lot of nesting of skills. Following~\citet{zhang-etal-2022-skillspan}, we prioritize the skills column when merging the labels. When there is nesting, we keep the labels of skills and remove the knowledge labels.

\begin{table*}[!ht]
\begin{minipage}{0.48\textwidth}
\centering
\resizebox{\linewidth}{!}{
\begin{tabular}[t]{rrlll}
\toprule
 Dataset $\rightarrow$          &                                    & \textsc{SkillSpan}   & \textsc{Sayfullina} &  \textsc{Green}                       \\\midrule
JobBERT                         & $k$                                & 4                    & 4                   &  16                   \\
                                & $\lambda$                          & 0.3                  & 0.3                 &  0.15                 \\
                                & $T$                                & 0.1                  & 2.0                 &  10.0                 \\\midrule
RoBERTa                         & $k$                                & 32                   & 4                   & 64                    \\
                                & $\lambda$                          & 0.3                  & 0.3                 & 0.25                  \\
                                & $T$                                & 10.0                 & 0.1                 & 10.0                  \\\midrule
JobBERTa                        &  $k$                               & 16                   & 4                   & 8                     \\
                                &  $\lambda$                         & 0.2                  & 0.1                 & 0.1                   \\
                                &  $T$                               & 5.0                  & 10.0                & 10.0                  \\
\midrule
                                &  $k$                               & \multicolumn{3}{c}{\{4, 8, 16, 32, 64, 128\}}                     \\
Search Space                    &  $\lambda$                         & \multicolumn{3}{c}{\{0.1, 0.15, 0.2, 0.25, ..., 0.9\}}            \\
                                &  $T$                               & \multicolumn{3}{c}{\{0.1, 0.5, 1.0, 2.0, 3.0, 5.0, 10.0\}}       \\\bottomrule
\end{tabular}}
\caption{\textbf{Tuned Hyperparameters on Dev.} These are for \dd{}.}
\label{tab:params1}
\end{minipage}\hfill
\begin{minipage}{0.48\textwidth}
\centering
\resizebox{\linewidth}{!}{
\begin{tabular}[t]{rrlll}
\toprule
 Dataset $\rightarrow$          &                                    & \textsc{SkillSpan}   & \textsc{Sayfullina} &  \textsc{Green}                       \\\midrule
JobBERT                         & $k$                                & 4                    & 4                   &  64                   \\
                                & $\lambda$                          & 0.35                 & 0.35                &  0.4                  \\
                                & $T$                                & 2.0                  & 0.1                 &  5.0                  \\\midrule
RoBERTa                         & $k$                                & 32                   & 4                   & 16                    \\
                                & $\lambda$                          & 0.35                 & 0.45                & 0.25                  \\
                                & $T$                                & 0.1                  & 0.1                 & 1.0                   \\\midrule
JobBERTa                        &  $k$                               & 64                   & 128                 & 128                   \\
                                &  $\lambda$                         & 0.25                 & 0.35                & 0.45                   \\
                                &  $T$                               & 10.0                 & 0.5                 & 10.0                  \\
\midrule
                                &  $k$                               & \multicolumn{3}{c}{\{4, 8, 16, 32, 64, 128\}}                     \\
Search Space                    &  $\lambda$                         & \multicolumn{3}{c}{\{0.1, 0.15, 0.2, 0.25, ..., 0.9\}}            \\
                                &  $T$                               & \multicolumn{3}{c}{\{0.1, 0.5, 1.0, 2.0, 3.0, 5.0, 10.0\}}       \\\bottomrule
\end{tabular}}
\caption{\textbf{Tuned Hyperparameters on Dev.} These are for \dwt{}.}
\label{tab:params2}
\end{minipage}\hfill
\begin{minipage}{0.48\textwidth}
\centering
\resizebox{\linewidth}{!}{
\begin{tabular}[t]{rrlll}
\toprule
 Dataset $\rightarrow$          &                                    & \textsc{SkillSpan}   & \textsc{Sayfullina} &  \textsc{Green}                       \\\midrule
JobBERT                         & $k$                                & 4                    & 16                  &  32                   \\
                                & $\lambda$                          & 0.3                  & 0.25                &  0.15                 \\
                                & $T$                                & 10.0                 & 5.0                 &  10.0                 \\\midrule
RoBERTa                         & $k$                                & 16                   & 8                   & 8                     \\
                                & $\lambda$                          & 0.15                 & 0.1                 & 0.1                   \\
                                & $T$                                & 10.0                 & 10.0                & 10.0                  \\\midrule
JobBERTa                        &  $k$                               & 8                    & 4                   & 8                     \\
                                &  $\lambda$                         & 0.2                  & 0.15                & 0.1                   \\
                                &  $T$                               & 0.5                  & 0.1                 & 10.0                  \\
\midrule
                                &  $k$                               & \multicolumn{3}{c}{\{4, 8, 16, 32, 64, 128\}}                     \\
Search Space                    &  $\lambda$                         & \multicolumn{3}{c}{\{0.1, 0.15, 0.2, 0.25, ..., 0.9\}}            \\
                                &  $T$                               & \multicolumn{3}{c}{\{0.1, 0.5, 1.0, 2.0, 3.0, 5.0, 10.0\}}       \\\bottomrule
\end{tabular}}
\caption{\textbf{Tuned Hyperparameters on Dev.} These are for \ad{}.}
\label{tab:params3}
\end{minipage}\hfill
\begin{minipage}{0.48\textwidth}
\centering
\resizebox{\linewidth}{!}{
\begin{tabular}[t]{rrlll}
\toprule
 Dataset $\rightarrow$          &                                    & \textsc{SkillSpan}   & \textsc{Sayfullina} &  \textsc{Green}                       \\\midrule
JobBERT                         & $k$                                & 32                   & 4                   &  128                  \\
                                & $\lambda$                          & 0.3                  & 0.3                 &  0.4                  \\
                                & $T$                                & 1.0                  & 0.5                 &  2.0                  \\\midrule
RoBERTa                         & $k$                                & 128                  & 128                 & 64                    \\
                                & $\lambda$                          & 0.35                 & 0.1                 & 0.25                   \\
                                & $T$                                & 0.1                  & 0.5                 & 0.1                   \\\midrule
JobBERTa                        &  $k$                               & 32                   & 8                   & 128                   \\
                                &  $\lambda$                         & 0.15                 & 0.3                 & 0.2                   \\
                                &  $T$                               & 0.1                  & 0.1                 & 2.0                   \\
\midrule
                                &  $k$                               & \multicolumn{3}{c}{\{4, 8, 16, 32, 64, 128\}}                     \\
Search Space                    &  $\lambda$                         & \multicolumn{3}{c}{\{0.1, 0.15, 0.2, 0.25, ..., 0.9\}}            \\
                                &  $T$                               & \multicolumn{3}{c}{\{0.1, 0.5, 1.0, 2.0, 3.0, 5.0, 10.0)\}}       \\\bottomrule
\end{tabular}}
\caption{\textbf{Tuned Hyperparameters on Dev.} These are for \adwt{}.}
\label{tab:params4}
\end{minipage}\hfill
\end{table*}

\begin{table}
    \centering
\begin{tabular}{lllll}
\toprule
             $\downarrow$Trained on                & Hyperparams.\                      & \textsc{SkillSpan}   & \textsc{Sayfullina} &  \textsc{Green}        \\\midrule          
             \textsc{SkillSpan}         &  $k$                               & \cellcolor{gray!25}  & 16                  & 32                   \\
                                        &  $\lambda$                         & \cellcolor{gray!25}  & 0.9                 & 0.7                   \\
                                        &  $T$                               & \cellcolor{gray!25}  & 0.1                 & 0.5                   \\\midrule 
             \textsc{Sayfullina}        &  $k$                               & 64                   & \cellcolor{gray!25} & 32                   \\
                                        &  $\lambda$                         & 0.9                  & \cellcolor{gray!25} & 0.8                   \\
                                        &  $T$                               & 0.1                  & \cellcolor{gray!25} & 0.1                   \\\midrule
             \textsc{Green}             &  $k$                               & 32                   & 32                  & \cellcolor{gray!25}   \\
                                        &  $\lambda$                         & 0.85                 & 0.9                 & \cellcolor{gray!25}   \\
                                        &  $T$                               & 0.5                  & 0.1                 & \cellcolor{gray!25}   \\\midrule
             \textsc{All}               &  $k$                               & 4                    & 128                 & 32                  \\
                                        &  $\lambda$                         & 0.25                 & 0.6                 & 0.65                  \\
                                        &  $T$                               & 1.0                  & 1.0                 & 0.5                  \\
                                        \midrule
                                        &  $k$                               & \multicolumn{3}{c}{\{4, 8, 16, 32, 64, 128\}}                     \\
             Search Space               &  $\lambda$                         & \multicolumn{3}{c}{\{0.1, 0.15, 0.2, 0.25, ..., 0.9\}}            \\
                                        &  $T$                               & \multicolumn{3}{c}{\{0.1, 0.5, 1.0, 2.0, 3.0, 5.0, 10.0\}}       \\\bottomrule
\end{tabular}
\caption{
\textbf{Results of Unseen Skills (Development Set) based on JobBERTa.}
}
\label{tab:unseenparams}
\end{table}

\subsection{Implementation Details}
\label{sec:params}

\subsection{General Implementation} We obtain all LMs from the Transformers library~\citep{wolf2020transformers} and implement JobBERTa using the same library. All learning rates for fine-tuning are $5 \times 10^{-5}$ using the AdamW optimizer~\citep{loshchilov2017decoupled}. We use a batch size of 16 and a maximum sequence length of 128 with dynamic padding. The models are trained for 20 epochs with early stopping using a patience of 5. We implement the retrieval component using the FAISS library~\citep{johnson2019billion}, which is a standard for nearest neighbors retrieval-augmented methods.\footnote{\url{https://faiss.ai/}} %

\subsection{JobBERTa} We apply domain-adaptive pre-training~\citep{gururangan2020don}, which involves continued self-supervised pre-training of a large LM on domain-specific text. This approach enhances the modeling of text for downstream tasks within the domain. We continue pre-training on a \texttt{roberta-base} checkpoint with 3.2M job posting sentences from~\citet{zhang-etal-2022-skillspan}. We use a batch size of 8 and run MLM for a single epoch following~\citet{gururangan2020don}. The rest of the hyperparameters are set to the defaults in the Transformer library.\footnote{\url{https://github.com/huggingface/transformers/blob/main/examples/pytorch/language-modeling/run_mlm.py}}

\subsection{\nnose{} Setup}
Following previous work, the keys used in \nnose{} are the 768-dimensional representation logits obtained from the final layer of the LM (input to the softmax). We perform a single forward pass over the training set of each dataset to save the keys and values, i.e., the hidden representation and the corresponding gold BIO tag. The FAISS index is created using all the keys to learn 4096 cluster centroids. During inference, we retrieve $k$ neighbors. The index looks up 32 cluster centroids while searching for the nearest neighbors. For all experiments, we compute the squared Euclidean ($L^2$) distances with full precision keys. The difference in inference speed is almost negligible, with the \knn{} module taking a few extra seconds compared to regular inference. For the exact hyperparameter values, we indicate them in the next paragraph.

\begin{table*}[t]
\centering
\resizebox{\linewidth}{!}{
\begin{tabular}{llllll}%
\toprule
Dataset (Dev.) $\rightarrow$                 & Setting                                 & \textsc{SkillSpan}   & \textsc{Sayfullina}    &  \textsc{Green}           & avg.\ Span-F1\\\midrule          
JobBERT~\citep{zhang-etal-2022-skillspan}     &                                         &  61.08               &  89.26                 &  37.27                    & 62.54           \\         
\hspace{0.2em} + \knn{}                      & \dd{}                                   &  61.56 \up{0.48}     &  89.69 \up{0.43}       &  37.48 \up{0.21}          & 62.91 \up{0.37} \\ 
\hspace{0.2em} + \knn{}                      & \dwt{}                                  &  61.77 \up{0.69}     &  89.78 \up{0.52}       &  38.07 \up{0.80}          & 63.21 \up{0.67} \\   
\hspace{0.2em} + \knn{}                      & \ad{}                                   &  61.58 \up{0.50}     &  89.50 \up{0.24}       &  37.27 \none{0.00}        & 62.78 \up{0.24} \\
\hspace{0.2em} + \knn{}                      & \adwt{}                                 &  61.50 \up{0.42}     &  89.37 \up{0.11}       &  38.19 \up{0.92}          & 63.02 \up{0.48} \\\midrule 

RoBERTa~\citep{liu2019roberta}                &                                         &  65.02               &  92.91                 &  40.33                    & 66.09  \\          
\hspace{0.2em} + \knn{}                      & \dd{}                                   &  65.36 \up{0.34}     &  92.76 \down{0.15}     &  40.53 \up{0.20}          & 66.22 \up{0.13} \\ 
\hspace{0.2em} + \knn{}                      & \dwt{}                                  &  65.34 \up{0.32}     &  93.07 \up{0.16}       &  41.22 \up{0.89}          & 66.54 \up{0.45} \\         
\hspace{0.2em} + \knn{}                      & \ad{}                                   &  64.98 \down{0.04}   &  92.78 \down{0.13}     &  40.60 \up{0.27}          & 66.12 \up{0.03} \\  
\hspace{0.2em} + \knn{}                      & \adwt{}                                 &  65.38 \up{0.36}     &  92.92 \up{0.01}       &  41.11 \up{0.77}          & 66.47 \up{0.38} \\\midrule    

JobBERTa (This work)                         &                                         &  65.15               &  92.09                 &  40.59                    & 65.94  \\          
\hspace{0.2em} + \knn{}                      & \dd{}                                   &  65.25 \up{0.10}     &  91.99 \down{0.10}     &  41.31 \up{0.72}          & 66.18 \up{0.24} \\ 
\hspace{0.2em} + \knn{}                      & \dwt{}                                  &  65.21 \up{0.06}     &  92.10 \up{0.01}       &  41.41 \up{0.82}          & 66.24 \up{0.30} \\        
\hspace{0.2em} + \knn{}                      & \ad{}                                   &  65.15 \none{0.00}   &  92.04 \down{0.05}     &  40.83 \up{0.24}          & 66.01 \up{0.07} \\   
\hspace{0.2em} + \knn{}                      & \adwt{}                                 &  65.22 \up{0.07}     &  92.13 \up{0.04}       &  41.45 \up{0.86}          & 66.26 \up{0.32} \\          
\bottomrule              
\end{tabular}}
\caption{\textbf{Development Set Results.} There are four settings for each model. \dd{}: in-dataset datastore (i.e., the datastore only contains the keys from the specific training data it is applied on). \ad{}: The datastore contains the keys from all available training datasets. $+W$: Whitening Transformation is applied to the keys before adding them to the datastore or querying the datastore. We indicate the performance increase (\up{}), decrease (\down{}), or no change (\textbf{--}) when using \knn{} compared to not using \knn{}. Additionally, we show the average span-F1 performance of each model across the three datasets. In the development set, it seems that an in-dataset datastore works best.}
\label{tab:dev}
\end{table*}

\subsection{Hyperparameters \nnose{}} The best-performing hyperparameters and search space can be found in~\cref{tab:params1}, \cref{tab:params2}, \cref{tab:params3}, and \cref{tab:params4}. We report the $k$-nearest neighbors, $\lambda$ value, and softmax temperature $T$ for each dataset and model.

In \cref{tab:unseenparams}, we show the hyperparameters for the cross-dataset analysis. In the vanilla setting, we apply the models trained on a particular skill dataset to another skill dataset, similar to transfer learning. We observe a significant discrepancy in performances cross-dataset, indicating a wide range of skills. However, when \knn{} is applied, it improves the detection of unseen skills. The datastore contains tokens from all datasets.

\begin{figure*}[ht]
    \centering
    \includegraphics[width=\linewidth]{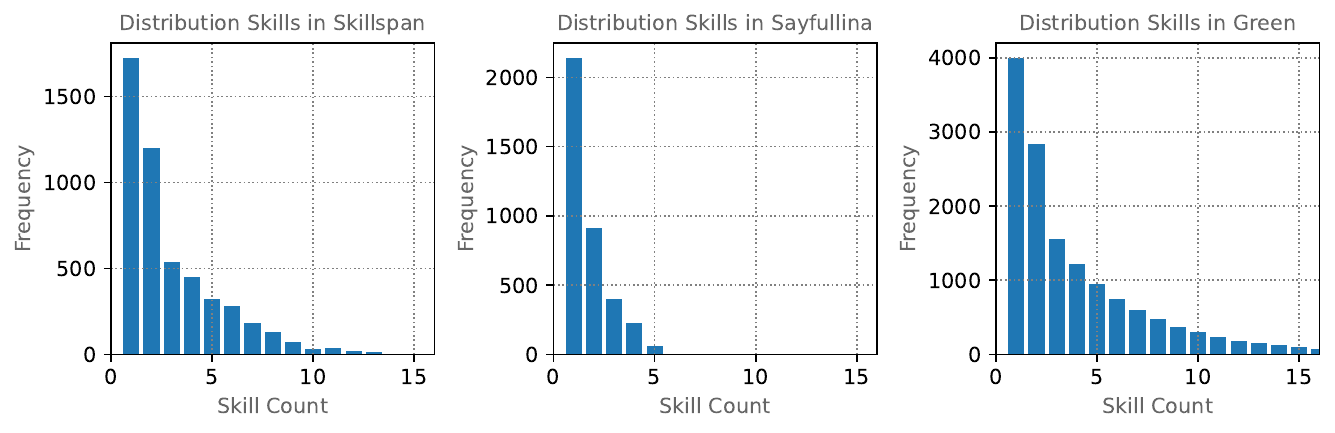}
    \caption{
    \textbf{Frequency Distribution of Skill Occurrences in the Train Set.} We display the frequency distribution of skill occurrences in each train set. \emph{How to read}: For instance, in the case of Sayfullina, there are over 2,000 skills that occur only \textbf{once} in the training set. We demonstrate that all skill datasets exhibit an inherent long-tail pattern.
    }
    \label{fig:skilldistribution}
\end{figure*}

\begin{table*}[t]
    \centering
    \resizebox{\linewidth}{!}{
    \begin{tabular}{lllll}
    \toprule
                                                           & \multicolumn{2}{c}{Vanilla}                   & \multicolumn{2}{c}{+\knn{}} \\
    Setup$\downarrow$                                      & Precision            & Recall                  & Precision             & Recall    \\\midrule
    \textsc{Sayfullina}$\rightarrow$\textsc{SkillSpan}     & 10.20                & 10.50                   & 37.67\up{27.47}       & 29.62\up{19.12}     \\
    \textsc{Green}$\rightarrow$\textsc{SkillSpan}          & 28.40                & 33.56                   & 46.00\up{11.60}       & 46.29\up{12.73}     \\\midrule
    \textsc{SkillSpan}$\rightarrow$\textsc{Sayfullina}     & 15.19                & 23.42                   & 49.25\up{34.06}       & 58.95\up{35.53}     \\
    \textsc{Green}$\rightarrow$\textsc{Sayfullina}         & 12.80                & 21.58                   & 48.21\up{35.41}       & 61.87\up{40.29}     \\\midrule
    \textsc{SkillSpan}$\rightarrow$\textsc{Green}          & 52.01                & 37.42                   & 55.37\up{3.36}        & 38.74\up{1.32}     \\
    \textsc{Sayfullina}$\rightarrow$\textsc{Green}         & 17.79                & 7.64                    & 39.83\up{22.04}       & 18.31\up{10.67}     \\\bottomrule
    \end{tabular}}
    \caption{\textbf{Precision \& Recall Numbers Cross-dataset on Test.} We show the precision and recall numbers in the cross-dataset setup. We use the \adwt{} setup here, with JobBERTa as the backbone model.}
    \label{tab:pr}
\end{table*}

\subsection{Development Set Results}\label{dev:results}
We show the dev.\ set results in~\cref{tab:dev}. Overall, the patterns of improvements hold across datasets and models. We base the test set result on the best-performing setups in the development set, i.e., \dwt{} and \adwt{}.

\subsection{Frequency Distribution of Skills}\label{app:skilldistribution}
We show the skill frequency distribution of the datasets in~\cref{fig:skilldistribution}, as mentioned in~\cref{subsec:longtail}. Here, we show evidence of the long-tail pattern in skills for each dataset. There is a cut-off at count 15 for \textsc{Green}, indicating that there are skills in the development set that occur more than 15 times.

\begin{table}
\centering
\resizebox{\linewidth}{!}{
    \begin{tabular}{|l|c|}
    \hline
    \multicolumn{2}{|c|}{JobBERTa $\rightarrow$ \textsc{SkillSpan}}\\\hline
    \hline
    Current token               & \texttt{IT}                    \\ \hline
    Gold label                  & \texttt{O} \\\hline
    LM prediction probs         & \texttt{[0.277, 0.404, 0.319]} \\
    \hline\hline
    Nearest neighbors ($k=8$)   & \texttt{['IT', 'Software', 'Software', 'Cloud', }\\ 
                                & \texttt{'Cloud', 'Database', 'Ag', 'software']} \\\hline
    Aggregated \knn{} scores    & \texttt{[0.000, 0.132, 0.868]}        \\\hline\hline

    Final predicted probs       & \texttt{[0.221, 0.350, 0.429]}\\\hline
    
    \end{tabular}}
    \caption{\textbf{Cherry Picked Qualitative Sample \nnose{} of Higher Precision.} We show a qualitative sample of using JobBERTa on \textsc{SkillSpan}. In this case, we see more weight being put on a specific tag, resulting in higher precision.}
    \label{tab:qualitative1}
\end{table}

\begin{table}
    \centering
    \resizebox{\linewidth}{!}{
    \begin{tabular}{|l|c|}
    \hline
    \multicolumn{2}{|c|}{JobBERTa $\rightarrow$ \textsc{SkillSpan}}\\\hline
    \hline
    Current token               & \texttt{coding}                    \\ \hline
    Gold label                  & \texttt{B} \\\hline
    LM prediction probs         & \texttt{[0.988, 0.000, 0.012]} \\
    \hline\hline
    Nearest neighbors ($k=8$)   & \texttt{['programming', 'coding', 'programming', 'debugging', }\\ 
                                & \texttt{'scripting', 'writing', 'coding', 'programming']} \\\hline
    Aggregated \knn{} scores    & \texttt{[1.000, 0.000, 0.000]}        \\\hline\hline

    Final predicted probs       & \texttt{[0.991, 0.000, 0.009]}\\\hline\hline
    \cellcolor{gray!25}         &  \cellcolor{gray!25}        \\\hline\hline
    Current token               & \texttt{skills}                    \\ \hline
    Gold label                  & \texttt{I} \\\hline
    LM prediction probs         & \texttt{[0.000, 0.990, 0.010]} \\
    \hline\hline
    Nearest neighbors ($k=8$)   & \texttt{['skills', 'skills', 'skills', 'skills', 'skills', }\\ 
                                & \texttt{'skills', 'skills', 'skills']} \\\hline
    Aggregated \knn{} scores    & \texttt{[0.000, 1.000, 0.000]}        \\\hline\hline

    Final predicted probs       & \texttt{[0.000, 0.992, 0.008]}\\\hline
    
    \end{tabular}
    }
    \caption{\textbf{Cherry Picked Qualitative Sample \nnose{} of Multiple Tokens.} We show a qualitative sample of using JobBERTa on \textsc{SkillSpan} with multi-token annotations and how this behaves.}
    \label{tab:qualitative2}
\end{table}

\begin{table}
    \centering
    \resizebox{\linewidth}{!}{
    \begin{tabular}{|l|c|}
    \hline
    \multicolumn{2}{|c|}{JobBERTa $\rightarrow$ \textsc{Green}}\\\hline
    \hline
    Current token               & \texttt{tools}                    \\ \hline
    Gold label                  & \texttt{I} \\\hline
    LM prediction probs         & \texttt{[0.250, 0.374, 0.379]} \\
    \hline\hline
    Nearest neighbors ($k=8$)   & \texttt{['tools', 'tools', 'transport', 'transport', }\\ 
                                & \texttt{'transport', 'transport', 'car', 'transport']} \\\hline
    Aggregated \knn{} scores    & \texttt{[0.124, 0.626, 0.250]}        \\\hline\hline

    Final predicted probs       & \texttt{[0.234, 0.399, 0.366]}\\\hline
    
    \end{tabular}
    }
    \caption{\textbf{Cherry Picked Qualitative Sample \nnose{} of Randomness.} We show a qualitative sample of using JobBERTa on \textsc{SkillSpan}.The language model puts high confidence on the tag \texttt{I}, which is the correct tag. Here the retrieved neighbors do not seem too relevant, which in this case is mostly random chance that it got it correctly.}
    \label{tab:qualitative3}
\end{table}

\begin{table}
    \centering
    \resizebox{\linewidth}{!}{
    \begin{tabular}{|l|c|}
    \hline
    \multicolumn{2}{|c|}{JobBERTa $\rightarrow$ \textsc{SkillSpan}}\\\hline
    \hline
    Current token               & \texttt{optimistic}                    \\ \hline
    Gold label                  & \texttt{B} \\\hline
    LM prediction probs         & \texttt{[0.998, 0.000, 0.002]} \\
    \hline\hline
    Nearest neighbors ($k=8$)   & \texttt{['proactive', 'responsible', 'holistic', 'operational', }\\ 
                                & \texttt{'positive', 'open', 'professional', 'agile']} \\\hline
    Aggregated \knn{} scores    & \texttt{[1.000, 0.000, 0.000]}        \\\hline\hline

    Final predicted probs       & \texttt{[0.999, 0.000, 0.001]}\\\hline
    
    \end{tabular}
    }
    \caption{\textbf{Cherry Picked Qualitative Sample \nnose{} of Variety.} We show a qualitative sample of using JobBERTa on \textsc{SkillSpan}. The language model puts high confidence in the tag \texttt{B}, which is the correct tag. The retrieved neighbors are frequently relevant.}
    \label{tab:qualitative4}
\end{table}

\subsection{Qualitative Results \nnose{}}\label{app:qualitative}

We show several qualitative results of \nnose{}. In \cref{tab:qualitative1}, we show a qualitative sample of using JobBERTa on \textsc{SkillSpan}. The current token is ``IT'' with gold label \texttt{O}. The language model puts 0.4 softmax probability on the tag \texttt{I}. By retrieving the nearest neighbors, the final probability mass gets shifted towards \texttt{O} with probability 0.43, which is the correct tag.

In \cref{tab:qualitative2}, we show a qualitative sample of using JobBERTa on \textsc{SkillSpan} with multi-token annotations and how this behaves. The current skill is ``coding skills'' with gold labels \texttt{B} and \texttt{I} respectively. Both the model and \knn{} puts high confidence in the correct label. Note that the nearest neighbors of ``coding'' are quite varied, which shows the benefit of \nnose{}. Note that all the retrieved ``skills'' tokens are from different contexts.

In \cref{tab:qualitative3}, we show a qualitative sample of using JobBERTa on \textsc{SkillSpan}. The current token is ``optimistic'' with gold label \texttt{B}. This is a so-called ``soft skill''. The language model puts high confidence in the tag \texttt{B}, which is the correct tag. The retrieved neighbors are frequently relevant, but sometimes less. This indicates that the retrieved neighbors (all soft skills) occur in similar contexts.

In \cref{tab:qualitative4}, we show a qualitative sample of using JobBERTa on \textsc{SkillSpan}. The current token is ``optimistic'' with gold label \texttt{B}. This is a so-called ``soft skill''. The language model puts high confidence in the tag \texttt{B}, which is the correct tag. The retrieved neighbors are frequently relevant, but sometimes less. This indicates that the retrieved neighbors (all soft skills) occur in similar contexts.

\section{Further Cross-dataset Analysis}\label{app:cr}
In~\cref{tab:pr}, we checked the precision and recall numbers for the cross-dataset setup with \adwt{} and JobBERTa as the backbone model. When using \nnose{}, we generally notice an increase in precision, with the largest when applied to \textsc{Sayfullina}. The largest gains are with respect to recall, we notice a significant gain in all setups, where the recall and precision increase is mixed. This indicates that \nnose{} is a useful method for both precision-focused and recall-focused applications, as we are storing skills in the datastore to be retrieved.

\newpage
\part{Linking Skills to Existing Sources}
\newpage

\chapter{{E}ntity {L}inking in the {J}ob {M}arket {D}omain}
\label{chap:chap8}
The work presented in this chapter is based on a paper that has been accepted as: \bibentry{zhang2024eljob}.

\newcommand{\bartb}{BART$_\text{base}$}
\newcommand{\bartl}{BART$_\text{large}$}

\newpage

\section*{Abstract}
In Natural Language Processing, entity linking (EL) has centered around Wikipedia, but yet remains underexplored for the job market domain.
Disambiguating skill mentions can help us get insight into the current labor market demands. In this work, we are the first to explore EL in this domain, specifically targeting the linkage of occupational skills to the ESCO taxonomy~\citep{le2014esco}. Previous efforts linked coarse-grained (full) sentences to a corresponding ESCO skill. In this work, we link more fine-grained span-level mentions of skills. We tune two high-performing neural EL models, a bi-encoder~\citep{wu-etal-2020-scalable} and an autoregressive model~\citep{cao2021autoregressive}, on a synthetically generated mention--skill pair dataset and evaluate them on a human-annotated skill-linking benchmark. Our findings reveal that both models are capable of linking implicit mentions of skills to their correct taxonomy counterparts. Empirically, BLINK outperforms GENRE in strict evaluation, but GENRE performs better in loose evaluation (accuracy@k).\footnote{The source code can be found at \url{https://github.com/jjzha/el_esco}.}

\section{Introduction}
Labor market dynamics, influenced by technological changes, migration, and digitization, have led to the availability of job descriptions (JD) on platforms to attract qualified candidates~\citep{brynjolfsson2011race,brynjolfsson2014second,balog2012expertise}. 
It is important to extract and link surface form skills to a unique taxonomy entry, allowing us to quantify the current labor market dynamics and determine the demands and needs. We attempt to tackle the problem of \emph{entity linking} (EL) in the job market domain, specifically the linking of fine-grained span-level skill mentions to a specific taxonomy entry. 

Generally, EL is the task of linking mentions of entities in unstructured text documents to their respective unique entities in a knowledge base (KB), most commonly Wikipedia~\citep{he-etal-2013-learning}. Recent models address this problem by producing entity representations from a (sub)set of KB information, e.g., entity descriptions~\citep{logeswaran-etal-2019-zero,wu-etal-2020-scalable}, fine-grained entity types~\citep{raiman2018deeptype,onoe2020fine,ayoola-etal-2022-refined}, or generation of the input text autoregressively~\citep{cao2021autoregressive,de2022multilingual}.

For skill linking specifically, we use the European Skills, Competences, Qualifications and Occupations (ESCO;~\citealp{le2014esco}) taxonomy due to its comprehensiveness. Previous work classified spans to its taxonomy code via multi-class classification~\citep{zhang-etal-2022-kompetencer} without surrounding context and neither the full breadth of ESCO.~\citet{gnehm-etal-2022-fine} approaches it as a sequence labeling task, but only uses more coarse-grained ESCO concepts, and not the full taxonomy. Last, others attempt to match the full sentence to their respective taxonomy title~\citep{decorte2022design,decorte2023extreme,clavie2023large}. 

The latter comes with a limitation: The taxonomy title does not indicate which subspan in the sentence it points to, without an exact match. We define this as an \emph{implicit} skill, where mentions (spans) in the sentence do not have an exact string match with a skill in the ESCO taxonomy. The differences can range from single tokens to entire phrases. For example, we can link ``being able to work together'' to ``plan teamwork''.\footnote{\url{https://t.ly/3VUJG}} If we know the exact span, this implicit skill can be added to the taxonomy as an alternative choice for the surface skill. As a result, this gives us a more nuanced view of the labor market skill demands. Therefore, we attempt to train models to the linking of both implicit and explicit skill mentions.

\paragraph{Contributions.} Our findings can be summarized as follows: \circled{1} We pose the task of skill linking as an entity linking problem, showing promising results of successful linking with two entity linking systems.
\circled{2} We present a qualitative analysis showing that the model successfully links implicit skills to their respective skill entry in ESCO.

\begin{table}[t]
    \centering
    \begin{tabular}{lrrr}
    \toprule
             & Instances & Unique Titles & \texttt{UNK}\\
    \midrule
    Train    & 123,619   & 12,984   & 14,641\\
    Dev.     & 480       & 149      & 233 \\
    Test     & 1,824     & 455      & 813\\
    \bottomrule
    \end{tabular}
    \caption{\textbf{Data Statistics.} Data distribution of train, dev, and test splits. \texttt{UNK} indicates skills mentions that are not linked to a corresponding taxonomy title.
    }
    \label{tab:data}
\end{table}

\section{Methodology}

\subsection{Definition}
In EL, we process the input document $\mathcal{D} = \{w_1, \ldots, w_r\}$, a collection of entity mentions denoted as $\mathcal{MD} = \{m_1, \ldots, m_n\}$, and a KB, ESCO in our case: $\mathcal{E} = \{e_1, \ldots, e_{13890}, \text{\texttt{UNK}}\}$. The objective of an EL model is to generate a list of mention-entity pairs $\{(m_i, e_i)\}_{i=1}^{n}$, where each entity $e$ corresponds to an entry in a KB.
We assume that both the titles and descriptions of the entities are available, which is a common scenario in EL research~\citep{ganea-hofmann-2017-deep,logeswaran-etal-2019-zero,wu-etal-2020-scalable}. We also assume that each mention in the document has a corresponding valid gold entity present in the knowledge base, including \texttt{UNK}. This scenario is typically referred to as ``in-KB evaluation''. Similar to prior research efforts~\citep{logeswaran-etal-2019-zero,wu-etal-2020-scalable}, we also presuppose that the mentions within the document have already been tagged. %

\subsection{Data}
\looseness=-1
We use ESCO titles as ground truth labels, containing 13,890 skills.\footnote{Per version 1.1.1, accessed on 01 August 2023.}
\cref{tab:data} presents the train, dev, and test data in our experiments. We leverage the train set introduced by~\citet{decorte2023extreme}\footnote{\url{https://t.ly/edqkp}} along with the dev and test sets provided in~\citet{decorte2022design}.\footnote{\url{https://t.ly/LcqQ7}} 
The train set is synthetically generated by~\citet{decorte2023extreme} with the \texttt{gpt-3.5-turbo-0301} model~\citep{openai-chat}. Specifically, this involves taking each skill from ESCO and prompting the model to generate sentences resembling JD sentences that require that particular skill. The dev and test splits, conversely, are derived from actual job advertisements sourced from the study by~\citet{zhang-etal-2022-skillspan}. These JDs are annotated with spans corresponding to specific skills, and these spans have subsequently been manually linked to ESCO, as described in the work of~\citet{decorte2022design}. In cases where skills cannot be linked, two labels are used, namely \texttt{UNDERSPECIFIED} and \texttt{LABEL NOT PRESENT}. For the sake of uniformity, we map both of these labels to a generic \texttt{UNK} tag. We used several heuristics based on Levenshtein distance and sentence similarity to find the exact subspans if it exceeds certain thresholds, otherwise, it is \texttt{UNK}. This process is outlined in~\cref{app:preprocess}. In addition, some data examples can be found in~\cref{app:examples}. The number of \texttt{UNK}s in the data is also in~\cref{tab:data}. During inference, the \texttt{UNK} title is a prediction option for the models. 

\subsection{Models}
We use two EL models, selected for their robust performance in EL on Wikipedia.
\footnote{For the hyperparameter setups, we refer to~\cref{app:hyperparams}.}

\paragraph{BLINK~\citep{wu-etal-2020-scalable}.}
BLINK uses a bi-encoder architecture based on BERT~\citep{devlin-etal-2019-bert}, for modeling pairs of mentions and entities. The model processes two inputs:
\begin{align*}
&\text{\texttt{[CLS]}} \hspace{0.5mm} \text{ctxt}_{\text{l}} \hspace{0.5mm}\text{\texttt{[S]}} \hspace{0.5mm} \text{mention} \hspace{0.5mm} \text{\texttt{[E]}} \hspace{0.5mm} \text{ctxt}_{\text{r}}  \hspace{0.5mm}\text{\texttt{[SEP]}}
\end{align*}
Where ``mention'', ``ctxt$_\text{l}$'', and ``ctxt$_\text{r}$'' corresponds to the wordpiece tokens of the mention, the left context, and the right context. The mention is denoted by special tokens \texttt{[S]} and \texttt{[E]}. The entity and its description are structured as follows:
\begin{align*}
&\text{\texttt{[CLS]}} \hspace{0.5mm} \text{title} \hspace{0.5mm} \text{\texttt{[ENT]}} \hspace{0.5mm} \text{description} \hspace{0.5mm}\text{\texttt{[SEP]}}
\end{align*}
Here, ``title'' and ``description'' represent the wordpiece tokens of the skills's title and description, respectively. \texttt{[ENT]} is a special token to separate the two representations. We train the model to maximize the dot product of the \texttt{[CLS]} representation of the two inputs, for the correct skill in comparison to skills within the same batch. For each training pair $(m_i, e_i)$, the loss is computed as
\begin{equation}
    \mathcal{L}\left(m_i, e_i\right) = -\operatorname{s}\left(m_i, e_i\right) + \log \sum_{j=1}^B \exp \left(\operatorname{s}\left(m_i, e_j\right)\right),
\end{equation}
where the objective is to minimize the distance between $m_i$ and $e_i$ while encouraging the model to assign a higher score to the correct pair and lower scores to randomly sampled incorrect pairs. Hard negatives are also used during training, these are obtained by finding the top 10 predicted skills for each training example. These extra hard negatives are added to the random in-batch negatives.

\paragraph{GENRE~\citep{cao2021autoregressive}.} GENRE formulates EL as a retrieval problem using a sequence-to-sequence model based on BART~\citep{lewis-etal-2020-bart}. This model generates textual entity identifiers (i.e., skill titles) and ranks each entity $e \in \mathcal{E}$ using an autoregressive approach:
\begin{equation}
\operatorname{s}(e \mid x) = p_\theta(y \mid x)=\prod_{i=1}^N p_\theta\left(y_i \mid y_{<i}, x\right),
\end{equation}
where $y$ represents the set of $N$ tokens in the identifier of entity $e$ (i.e., entity tile), and $\theta$ denotes the model parameters. During decoding, the model uses a constrained beam search to ensure the generation of valid identifiers (i.e., only producing valid titles that exist within the KB, including \texttt{UNK}). 

\subsection{Setup} We train a total of six models: for BLINK, these are \bertb{} and \bertl{} (uncased;~\citealp{devlin-etal-2019-bert}) trained on ESCO, and another large version trained on Wikipedia and ESCO sequentially.
GENRE has the same setup, but then with BART~\citep{lewis-etal-2020-bart}. Additionally, we apply the released models from both BLINK and GENRE (large, trained on Wikipedia) in a zero-shot manner and evaluate their performance. The reason we use Wikipedia-based models is that we hypothesize this is due to many skills in ESCO also having corresponding Wikipedia pages (e.g., Python\footnote{\url{https://en.wikipedia.org/wiki/Python_(programming_language)}} or teamwork\footnote{\url{https://en.wikipedia.org/wiki/Teamwork}}), thus could potentially help linking. Next, to address unknown entities (\texttt{UNK}), we include them as possible label outputs. 
Last, our evaluation metric is Accuracy@$k$, following prior research~\citep{logeswaran-etal-2019-zero,wu-etal-2020-scalable, zaporojets2022tempel}.

\begin{table*}[t]
    \centering
    \resizebox{\textwidth}{!}{%
    \begin{tabular}{lrrrrrrr}
    \toprule
         & \textbf{Train Source} & \textbf{Acc@1}  & \textbf{Acc@4} & \textbf{Acc@8} & \textbf{Acc@16} & \textbf{Acc@32} \\
    \midrule
Random                      & & \std{0.22}{0.00} & \std{0.88}{0.00} & \std{1.76}{0.00} & \std{3.52}{0.00} & \std{7.04}{0.00} \\
TF-IDF                      & & \std{2.25}{0.00} & \cellcolor{lightgray} & \cellcolor{lightgray} & \cellcolor{lightgray} & \cellcolor{lightgray}\\
\midrule
BLINK (\texttt{bert-base})   &  ESCO             & \std{12.74}{0.49} & \std{22.81}{0.79} & \std{27.70}{0.82} & \std{32.44}{1.33} & \std{36.46}{1.07}\\
BLINK (\texttt{bert-large})  &  ESCO             & \std{12.77}{0.94} & \std{22.58}{1.47} & \std{27.24}{1.23} & \std{31.75}{0.89} & \std{36.10}{1.28}\\
BLINK (\texttt{bert-large})  &  Wiki (0-shot)    & \std{23.30}{0.00} & \textbf{\std{32.89}{0.00}} & \textbf{\std{38.16}{0.00}} & \std{42.60}{0.00} & \std{45.56}{0.00}\\
BLINK (\texttt{bert-large})  &  Wiki + ESCO      & \textbf{\std{23.55}{0.14}} & \std{32.63}{0.16} & \std{37.38}{0.09} & \textbf{\std{43.25}{0.13}} & \textbf{\std{48.98}{0.21}} \\\midrule
GENRE (\texttt{bart-base})           &  ESCO             & \std{1.47}{0.05} & \std{4.84}{1.74} & \std{10.46}{6.81} & \std{11.30}{4.18} & \std{15.51}{4.62}\\
GENRE (\texttt{bart-large})          &  ESCO             & \std{2.33}{0.44} & \std{5.74}{1.43} & \std{8.18}{2.21} & \std{11.13}{2.42} & \std{15.26}{2.66}\\
GENRE (\texttt{bart-large})          &  Wiki (0-shot)    & \std{6.91}{0.00} & \std{12.34}{0.00} & \std{15.52}{0.00} & \std{21.60}{0.00} & \std{33.17}{0.00}\\
GENRE (\texttt{bart-large})          &  Wiki + ESCO      & \textbf{\std{11.48}{0.41}} & \textbf{\std{21.26}{0.43}} & \textbf{\std{27.40}{0.78}} & \textbf{\std{37.21}{0.69}}  & \textbf{\std{49.78}{1.05}} \\
\bottomrule
    \end{tabular}%
    }    
    \caption{\textbf{Skill Linking Results.} We show the results of the various models used. There are two \texttt{base} and four \texttt{large} models. Training sources are either ESCO or a combination of Wikipedia and ESCO. The results are the average and standard deviation over five seeds. For the 0-shot setup, we apply the fine-tuned models from the work of~\citet{wu-etal-2020-scalable} and~\citet{cao2021autoregressive} to the ESCO test set once. We have a random and TF-IDF-based baseline.}
    \label{tab:results}
\end{table*}

\section{Results}
\cref{tab:results} presents the results. Each model is trained for five seeds, and we report the average with standard deviation. We make use of a random and TF-IDF-based baseline. Firstly, we observe that the strict linking performance (i.e., Acc@1) is rather modest for both BLINK and GENRE. But most models outperform the baselines. Notably, the top-performing models in this context are the \bertl{} and \bartl{} models, which were further fine-tuned from Wikipedia EL with ESCO. As expected, scores improve considerably as we increase the value of $k$.  Secondly, for both BLINK and GENRE, model size seems not to have a substantial impact when trained only on ESCO. Specifically for BLINK, the performance remains consistent for Acc@1 and exhibits only a slight decline as we relax the number of candidates for performance evaluation. For GENRE, the observed trend remains largely unchanged, even with a larger $k$.

Remarkably, the zero-shot setup performance of both BLINK and GENRE, when trained on Wikipedia, surpasses that of models trained solely on ESCO. For Wikipedia-based evaluation, GENRE usually outperforms BLINK. We notice the opposite in this case. For BLINK, this improvement is approximately 11 accuracy points for $k = 1$. Meanwhile, for GENRE, we observe an increase of roughly 9 accuracy points when trained on both Wikipedia and ESCO. This trend persists for a larger $k$, reaching up to a 12.5 accuracy point improvement for BLINK and a 34 accuracy point improvement for GENRE in the case of Acc@32. Furthermore, we show that further fine-tuning the Wikipedia-trained models on ESCO contributes to an improved EL performance at $k = \{1, 16, 32\}$ for both models. For \texttt{UNK}-specific results, we refer to~\cref{app:unk}. We confirm our hypothesis that Wikipedia has concepts that are also in ESCO, this gives the model strong prior knowledge of skills.

\begin{table}[t!]
\newcolumntype{b}{X}
\newcolumntype{s}{>{\hsize=.6\hsize}X}
    \centering
    \small
    \begin{tabularx}{\linewidth}{bss}
    \toprule
    \textbf{Mention}                                                                                                                       &    \textbf{BLINK}                                                        &   \textbf{GENRE}                                    \\
    \midrule
    \circled{1} Work in a way that is \textcolor{purple}{\textbf{patient-centred}} and inclusive.                                             &    \textcolor{teal}{person centred care} (K0913)                    & \textcolor{red}{work in an organised manner} (T) \\    \midrule
    \circled{2}    You can \textcolor{purple}{\textbf{ride a bike}}.                                                                             &    \textcolor{red}{sell bicycles} (S1.6.1)                      & \textcolor{teal}{drive two-wheeled vehicles} (S8.2.2)              \\    \midrule
    \circled{3} It is expected that you are a super user of the \textcolor{purple}{\textbf{MS office tools}}.                                 &    \textcolor{teal}{use Microsoft Office} (S5.6.1)                       & \textcolor{red}{tools for software configuration management} (0613) \\    \midrule
    \circled{4} \textcolor{purple}{\textbf{Picking and packing}}.                                                                             &    \textcolor{red}{carry out specialised packing for customers} (S6.1.3) & \textcolor{teal}{perform loading and unloading operations} (S6.2.1)\\    \midrule
    \circled{5} You are expected to be able to further \textcolor{purple}{\textbf{develop your team}} - both personally and professionally. \textcolor{orange}{GOLD: \textbf{manage a team} (S4.8.1)}  &    \textcolor{red}{manage personal professional development} (S1.14.1)   & \textcolor{red}{shape organisational teams based on competencies} (S4.6.0) \\ \midrule
    \circled{6} Our games are developed using Unity so we expect all our programmers to have solid knowledge of mobile game development in Unity3D and \textcolor{purple}{\textbf{C\#}}.  & \textcolor{teal}{C\#} (K0613)        & \textcolor{teal}{C\#} (K0613) \\
    \bottomrule
    \end{tabularx}%
    \caption{
    We show six qualitative examples. The mention is indicated with purple and we show the predictions ($\mathbf{k = 1}$) of BLINK and GENRE. Green predictions mean correct, and red indicates wrong linking with respect to the ground truth. We also show the ESCO ID, indicating the differences in concepts. The results show successful linking of implicit mentions of skills. In example (5), we show how the linked results are still valid while being different concepts. However, evaluation does not count it as a correct hit.}
    \label{tab:quali}
\end{table}

\section{Discussion}\label{sec:discussion}

\subsection{Qualitative Analysis}
We manually inspected a subset of the predictions. We present qualitative examples in~\cref{tab:quali}. We found the following trends upon inspection: 
\begin{itemize}
    \itemsep0em
    \item The EL models exhibit success in linking implicit and explicit mentions to their respective taxonomy titles (e.g., \circled{1}, \circled{2}, \circled{4}, \circled{6}).
    \item In cases of hard skills (\circled{3}, \circled{6}), BLINK correctly matches ``MS office tools'' to ``using Microsoft Office'', which is not an exact match. Both models predict the explicit mention ``C\#'' correctly to the C\# taxonomy title.
    \item We found that the models predict paraphrased versions of skills that could also be considered correct (\circled{4}, \circled{5}), even being entirely different concepts (i.e., different ESCO IDs).
\end{itemize}

\subsection{Evaluation Limitation}
We qualitatively demonstrate the linking of skills that are implicit and/or valid.
Empirically, we observe that the strict linking of skills leads to an underestimation of model performance. We believe this limitation is rooted in evaluation. In train, dev, and test, there is only \emph{one} correct gold label. We reciprocate the findings by~\citet{li-etal-2020-efficient}, where they found that a large number of predictions are ``technically correct'' but limitations in Wikipedia-based evaluation falsely penalized their model (i.e., a more or less precise version of the same entity).
Especially \circled{5} in~\cref{tab:quali} shows this challenge for ESCO, we can consider multiple links to be correct for a mention given a particular context. This highlights the need for appropriate EL evaluation sets, not only for ESCO, but for EL in general.

\section{Conclusion}
We present entity linking in the job market domain, using two existing high-performing neural models. We demonstrate that the bi-encoder architecture of BLINK is more suited to the job market domain compared to the autoregressive GENRE model. While strict linking results favor BLINK over GENRE, if we relax the number of candidates, we observe that GENRE performs slightly better. From a qualitative perspective, the performance of strict linking results is modest due to limitations in the evaluation set, which considers only one skill correct per mention. However, upon examining the predictions, we identify valid links, suggesting the possibility of multiple correct links for a particular mention, highlighting the need for more comprehensive evaluation. We hope this work sparks interest in entity linking within the job market domain.

\section*{Limitations}
In the context of EL for ESCO, our approach has several limitations. Firstly, it only supports English, and might not generalize to other languages. Secondly, our EL model is trained on synthetic training data, which may not fully capture the intricacies and variations present in real-world documents. The use of synthetic data could limit its performance on actual, real JD texts. Nevertheless, we have human-annotated evaluation data.
Moreover, in our evaluation process, we use only one gold-standard ESCO title as the correct answer. This approach may not adequately represent a real-world scenario, where multiple ESCO titles could be correct as shown in~\cref{tab:quali}.

\clearpage

\section{Appendix}

\begin{algorithm*}[t]

\KwData{%
  $sentence$: The input sentence \\
  $target\_subspan$: The target subspan \\
  $threshold$: The Levenshtein distance similarity threshold \\
}

\KwResult{%
  $most\_similar\_ngram$: The most similar n-gram \\
}

\BlankLine

$all\_ngrams \leftarrow$ GenerateAllNgrams($sentence$)

$filtered\_ngrams \leftarrow$ FilterNgrams($all\_ngrams$, $target\_subspan$, $threshold$)

$most\_similar\_ngram \leftarrow$ None \\
$max\_similarity \leftarrow 0$

\For{$ngram$ in $filtered\_ngrams$}{
    $subspan\_embedding \leftarrow$ EncodeWithSBERT($target\_subspan$) \\
    $ngram\_embedding \leftarrow$ EncodeWithSBERT($ngram$)
    
    $similarity \leftarrow$ CosineSimilarity($subspan\_embedding$, $ngram\_embedding$)
    
    \If{$similarity > max\_similarity$ \textbf{and} $similarity > 0.5$}{
        $max\_similarity \leftarrow similarity$ \\
        $most\_similar\_ngram \leftarrow ngram$
    }
    \Else{$most\_similar\_ngram$ = UNK}
}

\Return $most\_similar\_ngram$

\caption{Find the most similar n-gram to a target subspan}
\label{alg:find-similar-ngram}
\end{algorithm*}

\subsection{Data Preprocessing}\label{app:preprocess}

We outline the preprocessing steps for the training set. In~\citet{decorte2023extreme}, there are sentence--ESCO skill title pairs. The data is synthetically generated by GPT-3.5. Where for each ESCO skill title a set of 10 sentences is generated. A crucial limitation for entity linkers is that the generated sentence does not have the ESCO skill title as an exact match in the sentence, but at most slightly paraphrased. To find the most similar subspan in the sentence to the target skill, we have to apply some heuristics. In Algorithm~\ref{alg:find-similar-ngram}, we denote our algorithm to find the most similar subspan. Our method is a brute force approach, where we create all possible n-grams until the maximum length of the sentence, and compare the target subspan against each n-gram. Based on Levenshtein distance, we filter the results, where we only take the top 80\% n-grams. Then, we encode both target subspan and n-gram with SentenceBERT~\citep{reimers-gurevych-2019-sentence}, the similarity is based on cosine similarity. If the similarity does not exceed 0.5, the candidate subspan is \texttt{UNK} and the ESCO title will also be \texttt{UNK}, otherwise, we take the most similar n-gram. Empirically, we found that these thresholds worked best. Note that this method is not error-prone, but allows us to generate implicit and negative examples to train entity linkers. We show two qualitative examples in~\cref{lst:example_train} and discuss the quality in~\cref{app:examples}.

\begin{table*}[ht]
    \centering
    \resizebox{\textwidth}{!}{%
    \begin{tabular}{lrrrrrrr}
    \toprule
         & \textbf{Train Source} & \textbf{Acc@1}  & \textbf{Acc@4} & \textbf{Acc@8} & \textbf{Acc@16} & \textbf{Acc@32} \\
    \midrule
BLINK (\texttt{bert-large}) \texttt{UNK} & Wiki + ESCO & \std{1.38}{0.12}  & \std{3.32}{0.22} & \std{4.67}{0.33} & \std{7.68}{0.42} & \std{10.70}{0.58} \\\midrule
GENRE (\texttt{bart-large}) \texttt{UNK} & Wiki + ESCO & \std{1.65}{0.20}  & \std{4.99}{0.50} & \std{9.23}{0.58} & \std{16.01}{0.48} & \std{24.70}{2.52} \\    \bottomrule
    \end{tabular}%
    }
    \caption{\textbf{\texttt{UNK} Linking Results.} We show the results of BLINK and GENRE predicting \texttt{UNK}. We use the best-performing models, based on~\cref{tab:results}.}
    \label{tab:unk}
\end{table*}

\begin{figure*}[t]
\begin{minipage}{\textwidth}
\begin{minted}[frame=single,
               framesep=3mm,
               linenos=true,
               xleftmargin=15pt,
               tabsize=2]{json}
{
    "context_left": "we're looking for someone who is 
    passionate about", 
    "context_right": "and eager to share their 
    knowledge with others.", 
    "mention": "young horse training", 
    label_title": "young horses training", 
    "label": "Principles & techniques of educating 
    young horses important simple body control 
    exercises.", 
    "label_id": 2198
}
{
    "context_left": "Hands-on experience with", 
    "context_right": "is a must-have qualification 
    for this job.", 
    "mention": "various hand-operated printing 
    devices", 
    "label_title": "types of hand-operated printing 
    devices", 
    "label": "Process of creating various types 
    hand-operated printing devices, such as stamps, 
    seals, embossing labels or inked pads and their 
    applications.", 
    "label_id": 10972
}
\end{minted}
\caption{\textbf{Two Training Examples.} The training examples are in the format for BLINK, there is the left context, right context, and the mention. The label title is the ESCO skill, and the label is the description of the label title. The label ID is the ID that refers to the label title.}
\label{lst:example_train}
\end{minipage}
\end{figure*}

\begin{figure*}[t]
\begin{minipage}{\textwidth}
\begin{minted}[frame=single,
               framesep=3mm,
               linenos=true,
               xleftmargin=15pt,
               tabsize=2]{json}
{
    "context_left": "You must have an", 
    "context_right": "with a high-quality mindset.", 
    "mention": "analytical proactive and structured 
    workstyle", 
    "label_title": "work in an organised manner", 
    "label": "Stay focused on the project at hand, 
    at any time. Organise, manage time, plan, 
    schedule and meet deadlines.", 
    "label_id": 3884
}
\end{minted}
\end{minipage}
\caption{\textbf{One Evaluation Example.} The evaluation example is in the format for BLINK, there is the left context, right context, and the mention. The label title is the ESCO skill, and the label is the description of the label title. The label ID is the ID that refers to the label title.}
\label{lst:example_dev}
\end{figure*}

\subsection{Data Examples}\label{app:examples}
We show a couple of data examples from the training (\cref{lst:example_train}) and development set (\cref{lst:example_dev}). In the training examples, we show an example with a mention that is the same as the original ESCO title (``young horse training''). In addition, we have an example where there is an ``implicit'' mention (i.e., the mention does not exactly match with the label title). This shows that our algorithm works to an extent. For the development example, this is another implicit mention. However, these samples are human annotated. 
There are also quite some \texttt{UNKs} given the training data. We show that this is helping the model predict \texttt{UNK}.

\subsection{Implementation Details}\label{app:hyperparams}
For training both BLINK\footnote{\url{https://github.com/facebookresearch/BLINK}} and GENRE,\footnote{\url{https://github.com/facebookresearch/genre}} we use their respective repositories. All models are trained for 10 epochs, for a batch size of 32 for training and 8 for evaluation. For both BLINK and GENRE we use 5\% warmup. For the base models we use learning rate $2 \times 10^{-5}$ and for the large models we use $2 \times 10^{-6}$. The maximum context and candidate length is 128 for both models. Each model is trained on an NVIDIA A100 GPU with 40GBs of VRAM and an AMD Epyc 7662 CPU. The seed numbers the models are initialized with are 276800, 381552, 497646, 624189, 884832. We run all models with the maximum number of epochs (10) and select the best-performing one based on validation set performance for accuracy@1.

\subsection{\texttt{UNK} Evaluation}\label{app:unk}
In~\cref{tab:unk}, we show the performance of both BLINK and GENRE on the \texttt{UNK} label. We use the best-performing models based on~\cref{tab:results}. Generally, we observe that GENRE is better in predicting \texttt{UNKs} than BLINK. However, the exact linking results (i.e., Acc@1) are low. This can potentially be alleviated by actively training for predicting \texttt{UNKs}~\citep{zhu-etal-2023-learn}.

\newpage
\part{Conclusions}
\newpage
\label{chap:conc}
\chapter{Summary and Conclusions}
 
The landscape of recent technological advances is transforming online talent platforms, including job boards and social media. 
Job seekers and employers are increasingly leveraging digital tools to post and apply for job opportunities, as well as to discover or provide training opportunities~\citep{spottl20214th}. 
To effectively manage talent and bridge skills gaps in education and training programs, companies, education institutions, and training providers require up-to-date and transparent information on skills and qualifications, which can be derived from large-scale job vacancy data.

In this thesis, we introduce \emph{Computational Job Market Analysis} and make contributions to three areas in this domain: 1) Annotation, datasets, and domain-specific models (i.e., resources) in \cref{chap:chap1}, \cref{chap:chap2}, \cref{chap:chap3}, and \cref{chap:chap4} (\textbf{Part II}). 2) Several modeling methodologies for skill extraction based on domain adaptation, weak supervision, and retrieval-augmentation in~\cref{chap:chap5}, \cref{chap:chap6}, and \cref{chap:chap7} (\textbf{Part III}). 3) Last, exploring how to link extracted spans to an existing taxonomy in~\cref{chap:chap8} (\textbf{Part IV}). In the following part, we conclude the thesis with answers to the proposed research questions (\cref{chap:intro}) and suggestions for future work (\cref{sec:futurework}).

\begin{quote}
    \textbf{RQ1: \rqone{}}
\end{quote}

\noindent
 In~\cref{chap:chap1}, we introduced \jobstack{}, a dataset for de-identification of privacy-related entities in job postings. We posed this task as a sequence labeling task and used both BiLSTM-based~\citep{plank-etal-2016} and transformer-based models~\citep{devlin2019bert} and compared them to each other on this task. We found that transformer-based models consistently outperform BiLSTM-based models that have been standard for de-identification in the medical domain, around 6\% span-F1 on the development set and 1\% span-F1 on the test set of \jobstack{}.

\begin{quote}
    \textbf{RQ2: \rqtwo{}}
\end{quote}

\noindent
For the task of de-identification, we explored multi-task learning, where we made use of varied auxiliary data: Generic named entity recognition (CoNLL2003;~\citealp{tjong2003introduction}) and de-identification in medical data (I2B2/UTHealth;~\citealp{stubbs_annotating_2015}) to improve upon the task of de-identification of job postings. We showed that, indeed, making use of a close downstream task was beneficial, where I2B2/UTHealth was the most helpful for overall span-F1 performance. In our analysis, we found that I2B2/UTHealth is particularly helpful for predicting profession entities and with CoNLL2003, the model predicts locations slightly better. A combination of both CoNLL2003 and I2B2/UTHealth resulted in the highest recall scores on the test set. This shows the usefulness of multi-task learning for this task (\cref{chap:chap1}).

\begin{quote}
    \textbf{RQ3: \rqthree{}}
\end{quote}

\noindent
In~\cref{chap:chap2}, we proposed a novel active learning algorithm: Cartography Active Learning, as an example for a speed up of annotation and more efficient model training.
We optimize for selecting the most informative data concerning a model by leveraging insights from data maps~\citep{swayamdipta-etal-2020-dataset}.
We show empirically that our method is on par or significantly outperforms various popular active learning methods based on uncertainty sampling~\citep{lewis1994sequential, lewis1994heterogeneous} or diverse batching~\citep{geifman2017deep, sener2017active, gissin2019discriminative, zhdanov2019diverse}. We show in our analysis that our algorithm chooses more informative samples, has less overlap of instances with other active learning algorithms, and performs equally or better than full-dataset performance using $<$14\% training data.

\begin{quote}
    \textbf{RQ4: \rqfour{}}
\end{quote}

\noindent
In~\cref{chap:chap3}, we start by developing annotation guidelines for skill extraction. We draw inspiration from the ESCO taxonomy and annotate for skill and knowledge components on the token level with three annotators. We define several annotation rules for knowledge and skill annotation, such as: ``knowledge is something that one possesses, and cannot (usually) physically execute''. For example, knowledge of the ``Python'' programming language. For a skill, we expect it to start with a verb, but if it includes a modal verb, the modal verb will not be tagged. For instance, in the sentence ``you will train new staff'', the skill will be ``train new staff''. After six rounds of annotation, we calculated the agreement over tokens and the span surface form. We obtain a Fleiss' $\kappa$ of 0.70--0.75, which indicates a substantial agreement (i.e., accuracy) over the span annotations. In the end, we introduce \textsc{SkillSpan}, a job posting dataset of more than 10,000 sentences annotated for skill and knowledge components.

\begin{quote}
    \textbf{RQ5: \rqfive{}}
\end{quote}

\noindent
We release domain-adapted BERT and SpanBERT~\citep{joshi2020spanbert} models---JobBERT and JobSpanBERT. In this scenario, domain-adapted means further training of the language model using the pre-training objectives they have been trained on (e.g., masked language modeling). We train BERT and SpanBERT further on 3.2M English job posting sentences. Our analysis shows that domain-adaptive pre-training helps to improve downstream performance on the task of skill extraction by around 2 span-F1 on the \textsc{SkillSpan} test set. %

\begin{quote}
\textbf{RQ6: \rqsix{}}
\end{quote}

\noindent
We showed that domain adaptive pre-training helps improve performance on the task of skill extraction for English and expect it to work as well on a mid-resource language like Danish. We contribute \textsc{DaJobBERT}, our domain-adapted language model for the Danish job market (\cref{chap:chap4}) continuously pre-trained on 24.5M Danish job posting sentences. It improves upon regular Danish BERT (DaBERT) by 19.4\% weighted macro-F1 on the task of skill classification. 

\begin{quote}
    \textbf{RQ7: \rqseven{}}
\end{quote}

\noindent
Additionally, in~\cref{chap:chap4}, we present a novel skill classification dataset for competences in Danish: \textsc{Kompetencer}. We made use of the ESCO taxonomy~\citep{le2014esco} to categorize extracted spans in Danish job postings into their respective taxonomy code using distant supervision (defined in~\cref{def:weaksupervision}). Via manual inspection, we found that 70.4\% of distantly supervised labels were correct for the Danish development split and 14.1\% correct for the Danish test split. In the test split, there were frequently unknown spans. Finally, we manually correct the distantly supervised labels.

\begin{quote}
    \textbf{RQ8: \rqeight{}}
\end{quote}

\noindent
In~\cref{chap:chap5}, we explore the compatibility of the ESCO skill taxonomy as a weak supervision signal for skill extraction. Employing various skill representation methods established in prior research, we demonstrate the usefulness of ESCO skill representations in this context. Our results show mostly gains in loose-F1 score (\cref{sec:results_chap5}), suggesting a substantial overlap between predicted and gold spans. However, there is a need for refinement in offset methods to precisely extract the correct span, which can be achieved through techniques like human post-editing or automated methods such as candidate filtering. 

\begin{quote}
    \textbf{RQ9: \rqnine{}}
\end{quote}

\noindent
In~\cref{chap:chap6}, we introduce \escolmr{}: A domain-adapted, multilingual language model further pre-trained on the ESCO taxonomy via a link-aware pre-training approach~\citep{yasunaga2022linkbert}. We evaluated across an extensive set of datasets in the domain across four languages, \escolmr{} outperformed \xlmr{} in six out of nine job-related tasks, consisting of skill extraction and classification. 
We attribute the enhanced performance of \escolmr{} to its effectiveness on shorter span lengths, emphasizing the value of pre-training on the ESCO dataset. Notably, \escolmr{} exhibited improved performance on both frequent surface spans and different span lengths. 

\begin{quote}
    \textbf{RQ10: \rqten{}}
\end{quote}

\noindent
Skills can be mentioned in different frequencies across multiple datasets. In~\cref{chap:chap7}, we showed a dataset-unifying method for combining several skill extraction datasets to address the challenge of the long-tail distribution of skills (\cref{sub:challenges}). We present \nnose{}, an LM that integrates a non-parametric datastore for nearest neighbor retrieval of skill tokens of three combined datasets for skill extraction. \nnose{} was assessed on three diverse skill datasets, showcasing its ability to enhance the performance of all language models used without additional parameter tuning with a significant performance increase on the in-dataset setting. Our analysis indicates that \nnose{} effectively leverages all datasets by retrieving tokens from each, demonstrating proficiency not only with rare, but also improving the performance of more common skills. Notably, models' performance without an external datastore deteriorates in cross-dataset settings, but the introduction of \nnose{} leads to overall improvement across all cross-dataset scenarios, with the largest improvement of 30.7\% span-F1. These findings suggest that \nnose{} holds promise for application-specific skill extraction setups and further qualitative analysis may uncover skills overlooked in manual annotations in other datasets.

\begin{quote}
    \textbf{RQ11: \rqeleven{}}
\end{quote}

\noindent
Lastly, in~\cref{chap:chap8}, we explore entity linking in the job market domain to finally match the extracted skills with a taxonomical counterpart instead of multi-class classification as used in~\cref{chap:chap4}. We used two established neural models. Our findings indicate that the bi-encoder architecture of BLINK outperforms the autoregressive GENRE model in strict linking results. Strict linking performance is modest due to evaluation set limitations, considering only one correct skill per mention. However, upon closer examination of predictions, valid links emerge, suggesting the potential for multiple correct links per mention. This underscores the need for a more comprehensive evaluation approach.

\section{Future Directions}\label{sec:futurework}

\paragraph{End-to-End Datasets} As categorizing skills can give us a snapshot of the current labor market, it is important to develop capable language technology systems that can do this succinctly. Strictly speaking,~\cref{chap:chap8} shows the most comprehensive way to do the categorization. However, this is not the only approach, as one could directly go from sentence or skill to taxonomy title as shown in~\cref{chap:chap4} and ~\citep{decorte2023extreme,clavie2023large}. A possible approach to building a large-scale dataset for matching skills with taxonomy codes in the most efficient way is to consider it as a multi-label problem: One sentence could have multiple possible skill codes. This not only allows for training a model to do this task but also to evaluate fairly, where multiple taxonomy codes are valid predictions. 

If there is a scenario where it is interesting to see the matched span in a sentence with a grounded skill, one can also decide to still think of the task as an entity linking problem, with multiple valid skills.

\paragraph{Large Language Models} Assuming that we want to extract skill spans from the sentences or documents, another challenge is discontinuous spans as indicated in~\cref{chap:chap0}. Given the skill ``\textit{critical, independent, and open-minded thinking skills}'', one has to make modeling choices here: Are we keeping this as a \emph{multiple skill span} or do we separate this conjunction in ``critical thinking skills'', ``independent thinking skills'', ``open-minded thinking skills''? One hypothesis is that generative large language models can solve this task, by automatically linking the skill span at hand to a taxonomy entry. In addition, there needs to be some effort on how the gold labels are represented and to evaluate this correctly.

We are currently in exciting times, the capabilities of large language models are showing signs of general task solvers~\citep{mccann2018natural, sanh2021multitask, wei2022finetuned, muennighoff2022crosslingual}. In the context of this thesis, one interesting direction is to investigate to what extent these systems can do end-to-end extraction and matching. Now, in these models, context length is not a prevailing issue anymore. Thus, full job descriptions can be used as input to the model. As the output is generated, one needs to consider constraining it to a specific subset of data (e.g., ESCO).

On a broader note, skill extraction is one of the many applications within Computational Job Market Analysis. Skills can be extracted from resumes and other types of documents. One can use the extracted skills as features for recommending jobs or courses to a specific individual or other types of talent acquisition. Additionally, we can create systems for career growth guidance. Concurrently, the integration of automated methodologies in such applications introduces potential challenges and uncertainties related to fairness, privacy, reproducibility, controllability, and transparency. To address these concerns, it is evident that research in NLP plays a pivotal role, offering insights and solutions to tackle the aforementioned risks.

\newpage

\addcontentsline{toc}{chapter}{Bibliography}
\bibliography{references,anthology}

\end{document}